\documentclass[journal]{IEEEtran}

\pdfoutput=1
\usepackage[utf8]{inputenc}
\usepackage{amssymb}
\usepackage{supertabular}

\usepackage{bbm}
\usepackage{rotating}
\usepackage{dsfont}

\usepackage{bm}
\usepackage{makecell}
\usepackage{array}
\usepackage{eucal}
\usepackage{graphicx}
\usepackage{epstopdf}
\usepackage{amsfonts,amsthm}
\usepackage{setspace}
\usepackage{cite}
\usepackage[colorlinks,linkcolor=blue,anchorcolor=blue,citecolor=blue]{hyperref}
\usepackage{subfig}
\usepackage{subfloat}
\usepackage{multirow,booktabs}

\usepackage{amsmath,mathrsfs}
\usepackage[misc]{ifsym}

\usepackage{amsmath,graphicx}
\usepackage{epstopdf}
\usepackage{amsfonts,amsthm}
\usepackage{algorithmic}
\usepackage[ruled,linesnumbered,boxed]{algorithm2e}
\usepackage{setspace}
\usepackage{cite}
\usepackage[colorlinks,linkcolor=blue,anchorcolor=blue,citecolor=blue]{hyperref}
\usepackage{subfig}
\usepackage{subfloat}
\usepackage{multirow,booktabs}

\usepackage{amsmath}
\usepackage{amssymb}
\usepackage[misc]{ifsym}
\usepackage{url}

\usepackage{color,soul}
\definecolor{myred}{RGB}{255,0,0}
\definecolor{myblue}{RGB}{0,0,255}
\usepackage{threeparttable}

\usepackage{hyperref}
\hypersetup{hypertex=true,
            colorlinks=true,
            linkcolor=red,
            anchorcolor=green,
            citecolor=blue}

\newtheorem{proposition}{Proposition}[section]

\newtheorem{Definition}{Definition}[section]
\newtheorem{Theorem}{Theorem}[section]
\newtheorem{Remark}{Remark}[section]
\newtheorem{Lemma}{Lemma}[section]

\newcommand{\tabincell}[2]{\begin{tabular}{@{}#1@{}}#2\end{tabular}}

\if CLASSOPTIONcompsoc
  \usepackage[nocompress]{cite}
\else
  \usepackage{cite}
\fi

\ifCLASSINFOpdf
\else
\fi
\begin{document}

\title{
Nonconvex
Robust 
High-Order Tensor Completion
Using 
Randomized Low-Rank Approximation
}

\author{Wenjin~Qin,~Hailin~Wang,~\IEEEmembership{Student Member,~IEEE,}~Feng~Zhang,~Weijun~Ma,~Jianjun~Wang,~\IEEEmembership{Member,~IEEE,}\\
%
~and~Tingwen~Huang,~\IEEEmembership{Fellow,~IEEE}
%
\thanks{
This research was supported in part by the National Natural Science Foundation of China under
 Grant 12071380, Grant 12101512;
in part by
the National Key Research and Development Program of China under Grant  2021YFB3101500;
 in part by
 the   China Postdoctoral Science Foundation  under Grant  2021M692681;
 in part by
 the Natural Science Foundation of Chongqing, China,  under Grant cstc2021jcyj-bshX0155;
 and in part by
the  Fundamental Research Funds for the Central Universities under Grant SWU120078.
(Corresponding author: Jianjun Wang.)
}
\thanks{Wenjin Qin and Feng Zhang are with the School of Mathematics and Statistics, Southwest University, Chongqing 400715, China (e-mail:
qinwenjin2021@163.com,
zfmath@swu.edu.cn
 ).
Hailin Wang is with the School of Mathematics and Statistics, Xi'an
Jiaotong University, Xi'an 710049, China (e-mail: wanghailin97@163.com).}
\thanks{Weijun Ma is with the School of Information Engineering, Ningxia University, Yinchuan 750021, China
(e-mail: Weijunma\_2008@sina.com).
}
\thanks{Jianjun Wang is with the School of Mathematics and Statistics, Southwest
University, Chongqing 400715, China, and also with the Research Institute of
Intelligent Finance and Digital Economics, Southwest University, Chongqing
400715, China (e-mail: wjj@swu.edu.cn).}
\thanks{
Tingwen Huang is with the Department of Mathematics, Texas A\&M University at Qatar, Doha 23874, Qatar (e-mail: tingwen.huang@qatar.tamu.edu).
}
}
\maketitle
\begin{abstract}
Within the tensor singular value decomposition (T-SVD) framework, 
existing robust low-rank tensor completion 
approaches
have made great achievements in various areas of science and engineering.
Nevertheless, these  methods 
involve the 
T-SVD based low-rank approximation,
which 
suffers from high computational costs
when dealing with large-scale tensor data.
Moreover, most of them are  only applicable to third-order tensors.
Against these issues,
in this article,
two
 efficient low-rank 
 tensor approximation approaches 
 fusing
 randomized techniques are 
 first devised
under the order-$d$ ($d\geq3$) T-SVD  framework.
On this basis, we then further investigate the robust high-order tensor completion (RHTC) problem,
in which a double nonconvex 
model 
along with
its corresponding 
fast
optimization  algorithms with convergence guarantees are developed. 
To the best of our knowledge,
this is the first study to incorporate the 
randomized low-rank approximation 
 into the RHTC problem. 
Empirical
 studies on large-scale
 synthetic
and real  tensor data illustrate that
the proposed method outperforms other state-of-the-art 
approaches
in terms of
both computational efficiency and estimated precision.
%
\end{abstract}

\IEEEpeerreviewmaketitle
\vspace{-0.3cm}
\section{\textbf{Introduction}}
Multidimensional data including
medical images,
remote sensing images,  light field images, 
color videos,
and beyond, are becoming increasingly prevailing 
in 
various domains
such as
neuroscience \cite{
beckmann2005tensorial
},
chemometrics \cite{schutt2017quantum},
data mining \cite{
papalexakis2016tensors11
},
machine learning \cite{
sidiropoulos2017tensor11
},
image processing
\cite{
marquez2020compressive,
lin2022robust,
long2021bayesian
},
and
computer vision \cite{
bibi2017high,
zhang2019robust,
hou2021robust55
}.
Compared to
vectors and matrices,
tensors 
possess a more powerful capability to 
characterize the 
inherent
structural information
underlying these 
data from a higher-order perspective.
Nevertheless,
due
to various 
factors such as 
 occlusions, abnormalities,
software glitches, 
or sensor failures,
the tensorial data 
faced
in practical applications
can often suffer from elements loss and noise/outliers corruption.
Hence,
robust low-rank tensor completion (RLRTC) has been widely concerned by a large number of scholars
\cite{
goldfarb2014robust1,
huang2015provable22,
zhao2015bayesian12,
chen2019nonconvex5555,
huang2020robust1,
chen2021auto,
liu2021simulated,
li2021robust,
he2022coarse,
jiang2019robust,wang2020robust, wang2019robust1,lou2019robust1,
song2020robust,ng2020patched,he2020robust,
chen2020robust1,zhao2020nonconvex1,qiu2021nonlocal,zhao2022robust1  
}.

RLRTC
belongs to a 
canonical inverse problem,
which aims to 
reconstruct
the underlying  low-rank tensor 
from 
partial observations of 
target tensor corrupted by noise/outliers. 
{Mathematically, the  RLRTC model 
can be formulated as follows: 
}
\begin{align}
\label{intro3}
\min_{{\boldsymbol{\mathcal{L}}},{\boldsymbol{\mathcal{E}}} }\;\;
 \Psi({\boldsymbol{\mathcal{L}}})
+ \lambda\Upsilon({\boldsymbol{\mathcal{E}}}),
\;\; \text{s.t.}\;\;
\boldsymbol{\bm{P}}_{{{\Omega}}}({\boldsymbol{\mathcal{L}}}+{\boldsymbol{\mathcal{E}}})=
\boldsymbol{\bm{P}}_{{{\Omega}}}({\boldsymbol{\mathcal{M}}}),
\end{align}
where 
$\Psi({\boldsymbol{\mathcal{L}}})$
represents
the regularizer measuring 
tensor low-rankness
%
%
(also called
the
certain relaxation 
of  tensor rank),
$\Upsilon({\boldsymbol{\mathcal{E}}})$
\footnote{
In specific problems, if we
assume that the noise/outliers  follows the
Laplacian distribution or the Gaussian distribution, then $\Upsilon(\boldsymbol{\mathcal{E}})$ can be chosen as $\|\bm{\mathcal{E}}\|_1$ or $\|\bm{\mathcal{E}}\|_F^{2}$,
respectively.
}
denotes the  noise/outliers 
 regularization,
$\lambda > 0 
$ 
is a 
trade-off parameter that balances these two
terms,
${\boldsymbol{\mathcal{M}}}$ is the observed tensor,
and
$\boldsymbol{\bm{P}}_{{{\Omega}}}(\cdot 
)$
 is the projection operator
onto the observed index set ${{{\Omega}}}$ 
such that
\begin{eqnarray*}
\label{intro2}
\big(\boldsymbol{\bm{P}}_{{{\Omega}}}(\boldsymbol{\mathcal{M}})\big)_{i_1,\cdots ,i_d}=
\begin{cases}
{\boldsymbol{\mathcal{M}}}_{i_1,\cdots ,i_d},\;\text{if}\;(i_1,\cdots ,i_d) \in \Omega,\\
0,\;\;\;\;\;\;\;\;\;\;\;\;\;
\text{otherwise}.
\end{cases}
\end{eqnarray*}
If the index set ${{{\Omega}}}$ is the whole set, i.e., no elements are missing, then the  model (\ref{intro3}) reduces
to the Tensor Robust Principal Component Analysis (TRPCA)
problem \cite{
lu2019tensor,zhang2020low1,   
gao2020enhanced,li2021nonconvex11,qiu2022fast,wang2022tensor55,
zheng2020tensor,    shi2021robust1, yang2022nonconvex11
}.
If there is  no corruption, i.e., ${\boldsymbol{\mathcal{E}}}=0$, then the model (\ref{intro3}) 
 is equivalent   to
the
Low-Rank Tensor Completion (LRTC)
problem \cite{
zhang2016exact,
kong2018t,    
 zhang2021low55,    
%
 xu2021fast22,
  wang2021generalized11,
 liu2013tensor,bengua2017efficient, 
 zhao2021tensor,
 xue2022laplacian,
 qiu2022noisy
}.
Therefore,
RLRTC can be viewed as
a generalized form of both LRTC and TRPCA.

%
Nevertheless,
there exist different definitions 
of  tensor rank
and its 
corresponding
relaxation 
within  different tensor decomposition frameworks,
which  makes the optimization  problem (\ref{intro3})
 extremely complicated. 
 The  commonly-used
 frameworks
 contain 
  CANDECOMP/PARAFAC (CP) 
\cite{
kolda2009tensor
},
Tucker 
\cite{tucker1966some},
 tensor train (TT) \cite{oseledets2011tensor}, tensor ring (TR)  
\cite{zhao2016tensor},
 and tensor singular value decomposition (T-SVD)
\cite{kilmer2011factorization,
kernfeld2015tensor
}.
Among them, T-SVD presents the first closed multiplication operation 
 called tensor-tensor product (t-product), 
 and 
 derives 
 a
 novel  tensor tubal rank \cite{kilmer2013third} 
  that 
  well characterizes the intrinsic low-rank structure of a tensor.
In particular,
 the recent work \cite{kilmer2021tensor} revolutionarily proved the
best
representation and compression theories of T-SVD,
making it more notable in 
capturing the ``spatial-shifting" correlation and  the global structure  information
underlying tensors.
 With
  these advantages, the
  robust low-tubal-rank tensor completion
  \cite{jiang2019robust,wang2020robust, wang2019robust1,lou2019robust1,
song2020robust,ng2020patched, he2020robust,
chen2020robust1,zhao2020nonconvex1,qiu2021nonlocal,zhao2022robust1}
 modeled by (\ref{intro3})
  and its  variants
\cite{
lu2019tensor,zhang2020low1,   
gao2020enhanced,li2021nonconvex11,qiu2022fast,wang2022tensor55,
zheng2020tensor,    shi2021robust1, yang2022nonconvex11,
zhang2016exact, kong2018t, 
 zhang2021low55,  
 xu2021fast22,
  wang2021generalized11
}
have recently caught 
many scholars' attention. 
However,
we observe that these approaches  
are only relevant to third-order tensors. %
To fix this problem,
Qin et al. established 
an 
order-$d$ ($d\geq3$)  T-SVD 
algebraic framework \cite{qin2022low}
based on 
a family of
invertible  linear transforms,
and then preliminary investigated the 
model,  algorithm, and theory
for
the
robust high-order tensor completion (RHTC)
\cite{qin2022low,wang2023guaranteed55, qin2022robust , qin2021robust
}.
The RHTC methods 
generated by
other tensor
factorization 
frameworks
can be found in \cite{
goldfarb2014robust1,
huang2015provable22,
zhao2015bayesian12,
chen2019nonconvex5555,
huang2020robust1,
chen2021auto,
liu2021simulated,
li2021robust,
he2022coarse}.

%
Although the 
above deterministic
RLRTC investigations
have already made some 
achievements
in 
real-world applications,
%
they
encounter
enormous challenges 
in dealing with 
those    tensorial  data characterized  by 
large volumes, high dimensions, 
 complex  structures, etc.
This is mainly attributed to the fact that
they 
require 
 to perform 
 multiple low-rank approximations
based on  a specific
tensor factorization. 
%
%
Calculating 
such an approximation 
generally involves the
singular value decomposition (SVD) or  T-SVD,
which is very time-consuming and inefficient  
when the data size scales up.
Motivated by the  reason that  randomized  algorithms   %
can 
accelerate the computational 
speed
of  their  conventional counterparts at the expense of
slight accuracy loss
\cite{halko2011finding11,mahoney2011randomized,woodruff2014sketching,martinsson2020randomized55 ,buluc2021randomized55
},
effective
low-rank tensor approximation  approaches using 
randomized   sketching techniques (e.g.,
\cite{
zhang2018randomized,tarzanagh2018fast,che2022fast, wang2015fast,battaglino2018practical,
malik2018low,che2019randomized,minster2020randomized,ahmadi2020randomized})
have attracted more and more attention in recent years.
Among these
methods,
we obviously find that 
the ones
 based on T-SVD framework are only  appropriate 
 for 
third-order tensors.
%
In addition,
RHTC researches
in combination with randomized low-rank approximation are relatively lacking. 
%
With the  rapid development of
information technologies, such as  Internet of Things, and big data,
large-scale high-order tensors 
encountered
in real 
scenarios are growing explosively, 
 like
order-four color videos  and
multi-temporal
remote sensing images,
order-five light filed images,
order-six bidirectional texture function images. Therefore,
 in this work,
we consider developing 
fast and
efficient 
randomness-based large-scale
 high-order tensor
 representation and recovery
methods under the  T-SVD framework.

\vspace{-0.4cm}
\subsection{\textbf{Our Contributions}}
Main
contributions of this work are summarized as follows:

1) Firstly, within the  order-$d$  T-SVD   algebraic framework 
\cite{qin2022low}, 
two 
efficient 
randomized  algorithms for low-rank approximation of high-order
tensor
are devised 
considering their 
adaptability
to large-scale  tensor data. 
 The developed
 approximation methods
  obtain a significant advantage in computational speed against  
 the  optimal  $k$-term approximation \cite{qin2022low}
 (also called truncated T-SVD)
 with a slight loss of precision. 

2)
Secondly, an effective 
and scalable 
model for RHTC 
is proposed 
in virtue of 
nonconvex
low-rank and noise/outliers 
regularizers.
Based on
the 
proposed randomized low-rank approximation schemes,
we then design 
     two
alternating direction method of multipliers (ADMM) framework based
fast  algorithms  with convergence guarantees
to solve the formulated model,
through which
any
low T-SVD rank high-order  tensors
with simultaneous elements loss and noise/outliers corruption
can be 
reconstructed 
efficiently  and  accurately.

3)
Thirdly,
our proposed RHTC
algorithm can be applied to 
a series of 
large-scale 
reconstruction
tasks,
such as
the
restoration of fourth-order color videos and  multi-temporal remote sensing images, and 
fifth-order light-field images.
Experimental results  demonstrate that the proposed method 
 %
 achieves 
 competitive performance in 
estimation accuracy 
and CPU
running time  than other state-of-the-art ones. 
Strikingly,
in the case of sacrificing 
a little precision,
our 
versions    
combined with
 randomization ideas 
decrease the CPU time by about $50\%$$\sim$$70\%$ 
compared with 
the 
deterministic version.

\vspace{-0.4cm}
\subsection{\textbf{Organization}}
\textcolor[rgb]{0.00,0.00,0.00}{
The remainder of the paper is organized as follows.
Section \ref{related} gives a brief summary of 
related work.
The 
main notations and preliminaries 
are
introduced in Section \ref{nota},
and then 
we develop two  efficient randomized algorithms for  low-rank approximation of high-order tensor
 in Section \ref{approximation}.
Section \ref{model} proposes 
effective
nonconvex model
 and 
 optimization
 algorithm for RHTC. 
%
In Section \ref{experiments}, extensive experiments
 are conducted to evaluate the effectiveness  of  the proposed method. 
Finally, we 
conclude our work
in Section \ref{conclusion}.}

What is noteworthy is that
 this article can be regarded as
an expanded version of 
our  previous
conference  paper  \cite{qin2022robust}.
Built off
the conference  version,
this paper makes 
the following changes:
1)
the 
high-order tensor approximation algorithm that fuses the randomized blocked strategy
is added;
2) the original
low-rank and noise/outliers regularizers 
are further enhanced
with more flexible regularizers;
%
3)
two accelerated 
algorithms
for solving the newly formulated non-convex model are designed  via the proposed low-rank approximation strategies;
4)
a large number of
experiments concerning with 
high-order tensor approximation and restoration are added.

\renewcommand{\arraystretch}{0.0} 
\setlength\tabcolsep{3.0pt}
\begin{table*}[!htbp]
  \centering
  \caption{The main notions and preliminaries for order-$d$ tensor.}
  \label{notation_part1}
  \small
  \scriptsize
  \footnotesize
  \begin{threeparttable}
    \begin{tabular}{l|l|l|l}
    \hline
    Notations&Descriptions&Notations&Descriptions\cr
    \hline
    \hline


    ${{\bm{A}}} \in \mathbb{R} ^{n_{1}\times n_{2}}$ & matrix &
     $\bm{I}_{n} \in \mathbb{R} ^{n\times n}$  &  $n\times n $  identity matrix
    \cr

    $\operatorname{trace}(\bm{A})$ & matrix trace &
    $ {{\bm{A}}}^{\mit{H}} ({{\bm{A}}}^{\mit{T}})$ & conjugate transpose (transpose)
     \cr

     $ \langle {{\bm{A}}},{{\bm{B}}} \rangle=\operatorname{trace} ({{\bm{A}}}^{\mit{H}} \cdot {{\bm{B}}})$& matrix inner product&
    ${\|{{\bm{A}}}\|}_{\bm{w}, {\mathrm{S}}_{p}}= {\big(\sum_{i} {\bm{w}}_{i} \; \big|\sigma_{i}({{\bm{A}}})\big| ^{p}\big)}^\frac{1}{p}$
    &matrix weighted  Schatten-$p$   norm
    \cr

     ${\boldsymbol{\mathcal{A}}} \in \mathbb{R}^{n_1\times \cdots \times n_d}$ & order-$d$ tensor&
  ${\boldsymbol{\mathcal{A}}}_{i_{1}\cdots i_{d}}$ or ${\boldsymbol{\mathcal{A}}}{(i_{1},\cdots, i_{d})}$	& $(i_1,\cdots,i_d)$-th  entry
    \cr

     ${{\bm{A}}_{(k)}} \in \mathbb{R}^{n_k\times
   \prod_{j\neq k } {n_j}
    } $
     & mode-$k$ unfolding of $\boldsymbol{\mathcal{A}}$ &
    ${\|{\boldsymbol{\mathcal{A}}}\|}_{\infty}= \max_{i_{1} \cdots i_d} |{\boldsymbol{\mathcal{A}}_{i_{1} \cdots i_d}}|$ &tensor infinity norm
    \cr

    ${\|{\boldsymbol{\mathcal{A}}}\|}_{{\boldsymbol{\mathcal{W}}},\ell_q}= ({\sum 
    }
    {\boldsymbol{\mathcal{W}}_{i_{1} \cdots i_d}}
    |{\boldsymbol{\mathcal{A}}_{i_{1} \cdots i_d}}|^{q})^{\frac{1}{q}}$ &tensor weighted $\ell_q$-norm &
    ${\|{\boldsymbol{\mathcal{A}}}\|}_{\mathnormal{F}}= {(\sum_{i_{1} \cdots i_d} |{\boldsymbol{\mathcal{A}}_{i_{1} \cdots i_d}}|^{2})^\frac{1}{2}}$ & tensor Frobenius norm
     \cr



    $
    \boldsymbol{\mathcal{C}}={\boldsymbol{\mathcal{A}}}{*}_{\mathfrak{L}} {\boldsymbol{\mathcal{B}}}
    $
    & order-$d$
    t-product under linear transform $\mathfrak{L}$
    &
   $ {{\boldsymbol{\mathcal{A}}}}^{\mit{H}} ({{\boldsymbol{\mathcal{A}}}}^{\mit{T}} )$
    &transpose (conjugate transpose)
    \cr

    \hline

$\langle {\boldsymbol{\mathcal{A}}},{\boldsymbol{\mathcal{B}}} \rangle  $&
\multicolumn{3} {l} {
  the inner product between order-$d$ tensors ${\boldsymbol{\mathcal{A}}}$ and ${\boldsymbol{\mathcal{B}}}$, i.e.,
$\langle {\boldsymbol{\mathcal{A}}},{\boldsymbol{\mathcal{B}}} \rangle  = {\sum}_{j=1}^{n_3 n_4 \cdots n_d} \langle {\bm{\mathcal{A}}}^{<j>},{\bm{\mathcal{B}}}^{<j>} \rangle$.
  }  \cr

${\boldsymbol{\mathcal{A}}}^{<j>} \in \mathbb{R}^{n_1 \times n_{2}}$ &
    \multicolumn{3} {l}{
    the matrix frontal slice of ${\boldsymbol{\mathcal{A}}}$,
     $
    {\boldsymbol{\mathcal{A}}}^{<j>}={\boldsymbol{\mathcal{A}}}{(:,:,i_3,\cdots, i_{d})},\;
    j={\sum_{a=4}^{d} }   {(i_a-1){\Pi}_{b=3}^{a-1}n_b}+i_3$
    .}\cr

      $\boldsymbol{\mathcal{A}} \;{\times}_{n}\; \bm{M}$ &
   \multicolumn{3} {l}{ the mode-$n$ product of tensor $\boldsymbol{\mathcal{A}}$ with matrix ${\bm{M}},\;
   \boldsymbol{\mathcal{B}} =\boldsymbol{\mathcal{A}}{\times}_{n} {\bm{M} } $
   $\Longleftrightarrow$
   ${\bm{B}}_{(n)} = {\bm{M}} \cdot {\bm{A}}_{(n)}$.} \cr

%
${\operatorname{bdiag}} ({{\boldsymbol{\mathcal{A}}}}) \in \mathbb{R}^{n_1n_3\cdots n_d \times n_2n_3\cdots n_d}$&
\multicolumn{3}{l}
{
$ {\operatorname{bdiag}} ({{\boldsymbol{\mathcal{A}}}})$
 is a block diagonal matrix whose $i$-th block equals to ${\boldsymbol{\mathcal{A}}}^{<i>}$,
 $\forall
i \in \{1,2,\cdots,n_3\cdots n_d\}$.
}
\cr

f-diagonal/f-upper triangular tensor
${\boldsymbol{\mathcal{A}}}$&
\multicolumn{3} {l} {
frontal slice ${\boldsymbol{\mathcal{A}}}^{<j>}$
 of ${\boldsymbol{\mathcal{A}}}$
is {a diagonal matrix (an upper triangular matrix)},
 $\forall
j \in \{1,2,\cdots,n_3\cdots n_d\}$.
}
\cr

 identity tensor ${\boldsymbol{\mathcal{I}}} \in \mathbb{R}^{n \times n \times n_3\times \cdots \times n_{d}}$  &
 \multicolumn{3} {l} {
identity tensor ${\boldsymbol{\mathcal{I}}} $ is defined to be
a tensor
such that
$ {{\mathfrak{L}} ({\boldsymbol{\mathcal{I}}}) }^{<j>} =\bm{I}_{n},
\forall
j \in \{1,2, \cdots, n_3\cdots n_d\}$.}\cr

 Gaussian random tensor ${\boldsymbol{\mathcal{G}}}$&
 \multicolumn{3} {l} {
the entries of
${{\mathfrak{L}} ({\boldsymbol{\mathcal{G}}})}^{<j>}$
follow
the standard normal distribution,
$
\forall
j \in \{1,2,\cdots,n_3 \cdots n_d\}$.
}
  \cr

  orthogonal tensor ${\boldsymbol{\mathcal{Q}}} $&
  \multicolumn{3} {l} {
  orthogonal tensor
  satisfies: ${\boldsymbol{\mathcal{Q}}}^{\mit{T}}{*}_{\mathfrak{L}}{\boldsymbol{\mathcal{Q}}} =
 {\boldsymbol{\mathcal{Q}}}{*}_{\mathfrak{L}} {\boldsymbol{\mathcal{Q}}}^{\mit{T}}={\boldsymbol{\mathcal{I}}}$,
 while
 partially orthogonal
tensor satisfies:  ${\boldsymbol{\mathcal{Q}}}^{\mit{T}}{*}_{\mathfrak{L}}{\boldsymbol{\mathcal{Q}}} =
{\boldsymbol{\mathcal{I}}}$.
  }
  \cr

  $\textrm{H-TSVD}({\boldsymbol{\mathcal{A} }}, \mathfrak{L})$&
  \multicolumn{3} {l} { order-$d$ T-SVD  factorization, i.e.,
  ${\boldsymbol{\mathcal{A}}}={\boldsymbol{\mathcal{U}}} {*}_{\mathfrak{L}}   {\boldsymbol{\mathcal{S}}} {*}_{\mathfrak{L}}
{\boldsymbol{\mathcal{V}}}^{\mit{T}}$,
where ${\boldsymbol{\mathcal{U}}} $ and
${\boldsymbol{\mathcal{V}}} $ are  orthogonal,
 ${\boldsymbol{\mathcal{S}}} $ is  f-diagonal.
  }
  \cr

$\textrm{H-TQR}({\boldsymbol{\mathcal{A} }}, \mathfrak{L})$&
\multicolumn{3} {l} {order-$d$ tensor QR-type factorization, i.e.,
${\boldsymbol{\mathcal{A}}}={\boldsymbol{\mathcal{Q}}} {*}_{\mathfrak{L}}   {\boldsymbol{\mathcal{R}}}$,
where ${\boldsymbol{\mathcal{Q}}} $ is
  orthogonal
while
${\boldsymbol{\mathcal{R}}} $
 is  f-upper triangular.
}
\cr

${\operatorname{rank}}_{tsvd} ({{\boldsymbol{\mathcal{A}}}}) $&
\multicolumn{3} {l} {
 ${\operatorname{rank}}_{tsvd}({\boldsymbol{\mathcal{A}}})
= \sum_{i} {\mathbbm{1}
\big[
{\boldsymbol{\mathcal{S}}}(i,i,:,\cdots,:) \neq \boldsymbol{0}
\big]}$, where
${\boldsymbol{\mathcal{S}}}$ originates from 
the middle component of
${\boldsymbol{\mathcal{A}}} = {\boldsymbol{\mathcal{U}}}{{*}_{\mathfrak{L}}}{\boldsymbol{\mathcal{S}}} {*}_{\mathfrak{L}}
 {\boldsymbol{\mathcal{V}}}^{\mit{T}}$.
}
\cr
    \hline
    \end{tabular}
    \end{threeparttable}
   \vspace{-0.5cm}
\end{table*}

\vspace{-0.3cm}

\section{\textbf{Related Work}}\label{related}

Based on 
different 
 factorization 
schemes, 
the 
representative
RLRTC methods can be broadly summarized as follows.



%
%

\subsubsection{\textbf{RLRTC 
Based on \text{T-SVD}  Factorization}}
%
Lu et al. \cite{lu2019tensor}
rigorously deduced
 a novel tensor nuclear norm (TNN)  corresponding to 
T-SVD
that is proved to be the convex envelope of the tensor average rank.
Sequentially,
Jiang et al. \cite{jiang2019robust}
 conducted a rigourous study for the RLRTC problem,
which is  modeled by 
the TNN and 
 $\ell_{1}$-norm penalty terms.
Besides,
Wang et al.    also adopted this novel TNN or 
slice-weighted TNN
plus
 a sparsity measure inducing $\ell_1$-norm to develop the RLRTC methods \cite{wang2020robust,wang2019robust1}.
Theoretically, the deterministic and 
non-asymptotic upper bounds on the estimation error
are established from a statistical standpoint.
However, the previous methods may suffer from disadvantage due to the limitation of Fourier transform \cite{qiu2021nonlocal}.
Aiming at this issue, by utilizing the  generalized
transformed TNN (TTNN) and $\ell_1$-norm regularizers,
 Song et al. \cite{song2020robust}  proposed an unitary transform method for  RLRTC
and also analyzed its  recovery guarantee.
Continuing along this vein,
a patched-tubes unitary transform approach for  RLRTC
was  proposed  by Ng et al. \cite{ng2020patched}.
Nevertheless,
the TTNN 
 is  a loose approximation of the tensor tubal rank, which usually leads to the
over-penalization of  the
optimization problem
and hence 
causes some unavoidable biases in real applications. 
 In addition,
 as indicated by \cite{fan2001variable},
 the $\ell_{1}$-norm
might not be statistically optimal in more challenging scenarios.
Recently,
to break the shortcomings existing in the 
TNN 
 and $\ell_{1}$-norm regularization terms, 
%
some researchers \cite{
chen2020robust1,zhao2020nonconvex1,qiu2021nonlocal,zhao2022robust1
}
designed
new 
nonconvex low-rank 
and noise/outliers  regularization  terms
to study the RLRTC problem from the 
model, algorithm, and theory.
%
But these 
methods are only limited to the case of third-order tensors
 and face the 
  high computational expense  of T-SVD.

\subsubsection{\textbf{RLRTC 
Based on Other  Factorization Schemes}}
%
Liu et al. \cite{liu2013tensor} primitively developed a new  Tucker
nuclear norm, i.e., Sum-of-Nuclear-Norms of  unfolding matrices of a tensor (SNN),
as convex relaxation of the tensor tucker rank.
 Then, the
 RLRTC approach 
within the Tucker format
 was investigated
 in  \cite{goldfarb2014robust1,huang2015provable22}
 via 
   combining the SNN regularization 
 with $\ell_{1}$-norm loss function.
Zhao et al.  \cite{zhao2015bayesian12}  proposed a variational Bayesian inference framework for
CP rank determination and applied it to the RLRTC problem.
Within the   TT  factorization, 
  Bengua et al.  \cite{bengua2017efficient} proposed  a novel   TT  nuclear norm
as the convex surrogate of the TT rank.
Furthermore,
in virtue of
an auto-weighted mechanism,
Chen et al. \cite{chen2021auto}
studied a 
new RLRTC method 
modeled by
the TT nuclear norm  and  $\ell_{1}$-norm  regularizers.
%
Under  the TR  decomposition, 
 by utilizing
the TR
nuclear norm and $\ell_{1}$-norm regularizers,
the   model,  algorithm, and theoretical analysis for RLRTC
 were developed by
Huang et al. \cite{huang2020robust1}.
 To be more robust against both missing entries and noise/outliers,
an effective iterative $\ell_p$-regression ($0<p\leq2$) TT completion method   was  developed in \cite{liu2021simulated}.
In parallel,
 integrating 
 TR rank
with $\ell_{p,\epsilon}$-norm ($0<p\leq1$), Li et al. \cite{li2021robust} proposed  a new RLRTC formulation.
Besides,
He et al. \cite{he2022coarse}  put forward a novel two-stage coarse-to-fine framework 
for RLRTC of visual 
 data
 in the TR factorization. 
However, 
these 
deterministic RLRTC
methods mostly
involve 
 multiple SVDs 
of unfolding matrices,
 which 
 experiences  high 
 computational costs
 when dealing with 
 large-scale tensor data.

\begin{figure*}[!htbp]
\renewcommand{\arraystretch}{0.01}
\setlength\tabcolsep{0.8pt}
\centering
\begin{tabular}{cccc}
\centering
%
\scriptsize{\footnotesize(CPU-Time, Relative-Error, PSNR)}
&\scriptsize {\footnotesize(1522.18s, 0.0691, 33.19db)}
& \scriptsize{\footnotesize(191.55s,  0.0735, 32.65db)}
&\scriptsize {\footnotesize(180.46s,  0.0737, 32.63db)}
%
\\
\includegraphics[width=1.6in, height=0.80in]{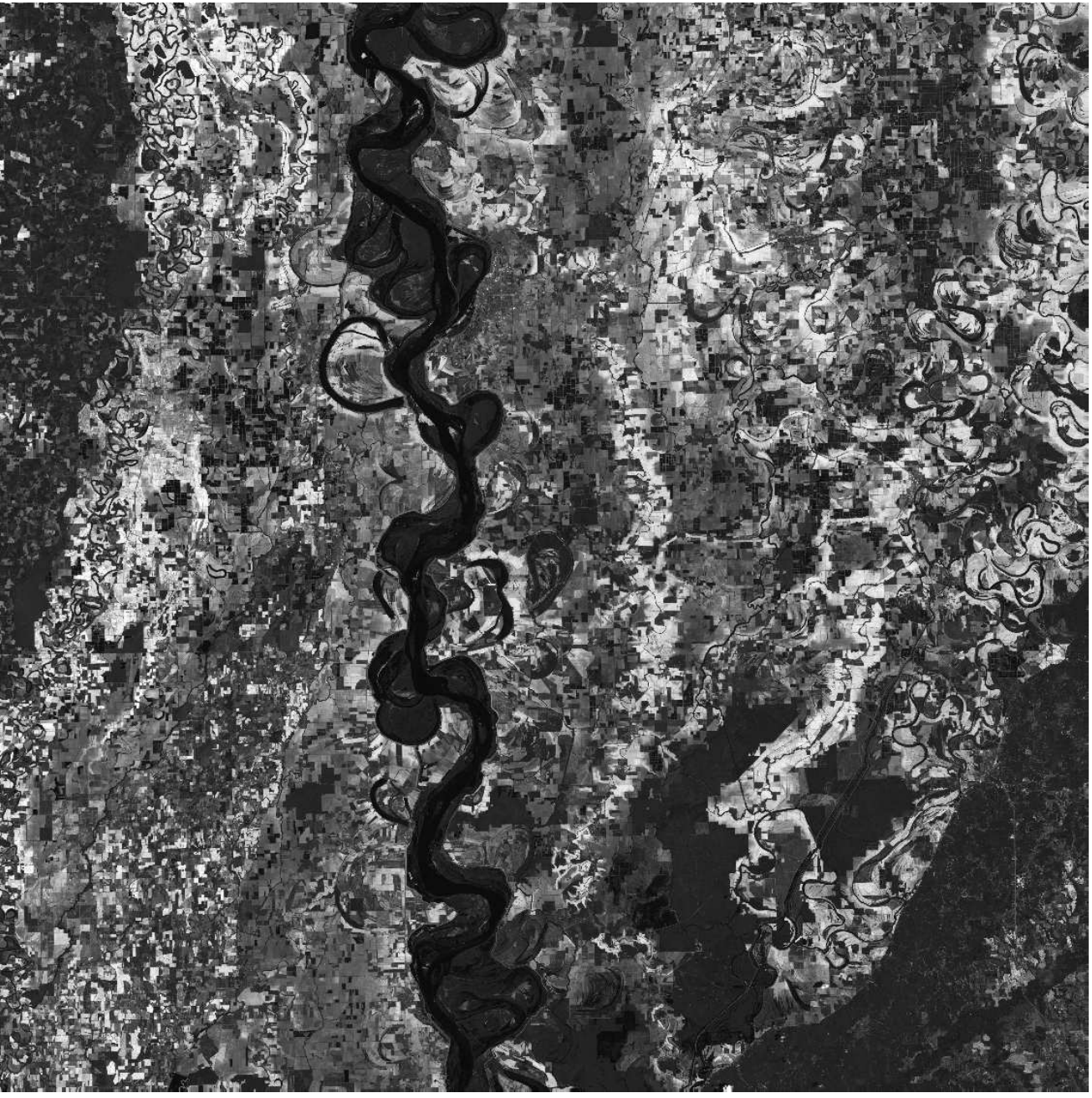}&
\includegraphics[width=1.6in, height=0.80in]{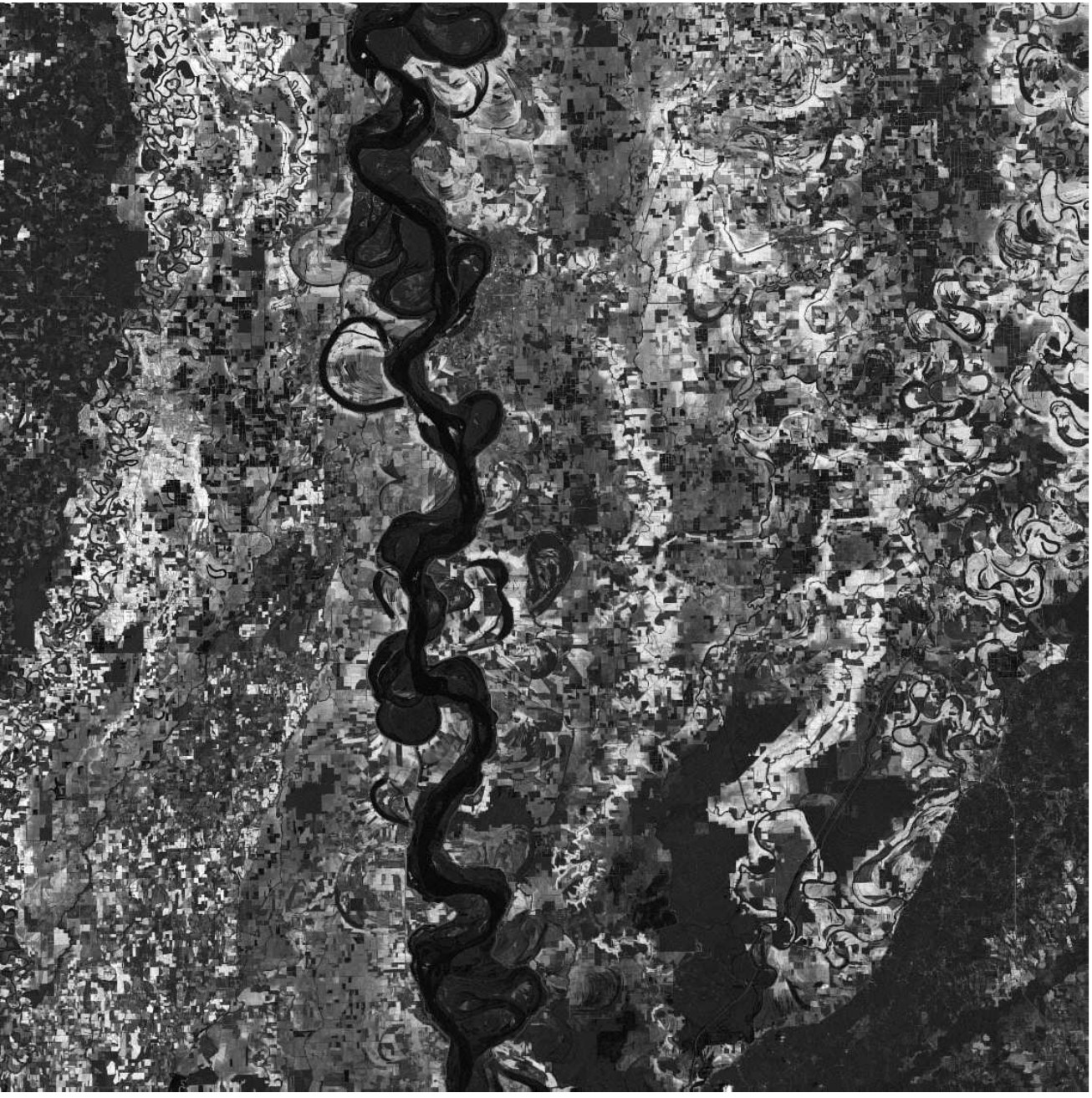}&
\includegraphics[width=1.6in, height=0.80in]{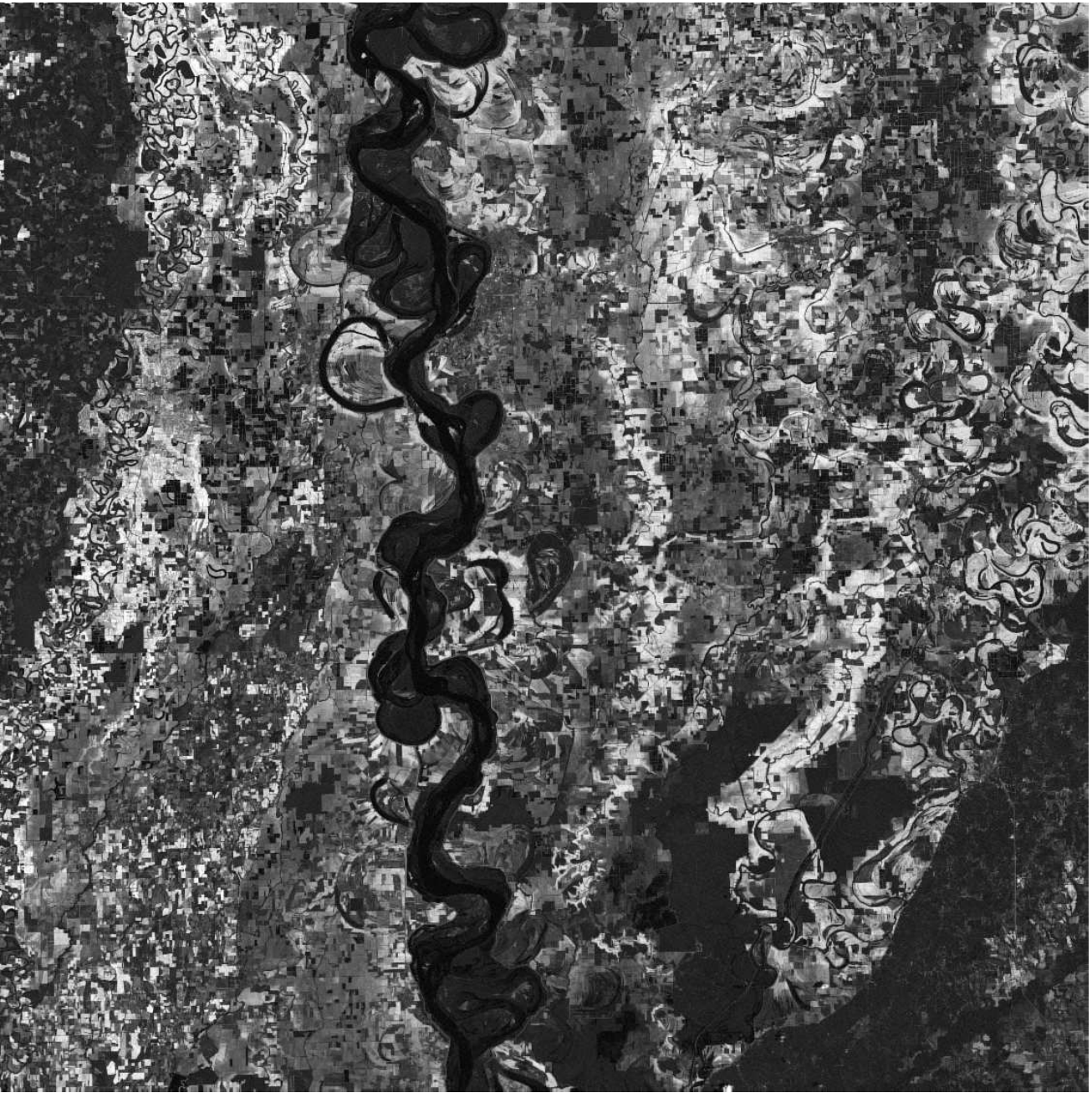}&
\includegraphics[width=1.6in, height=0.80in]{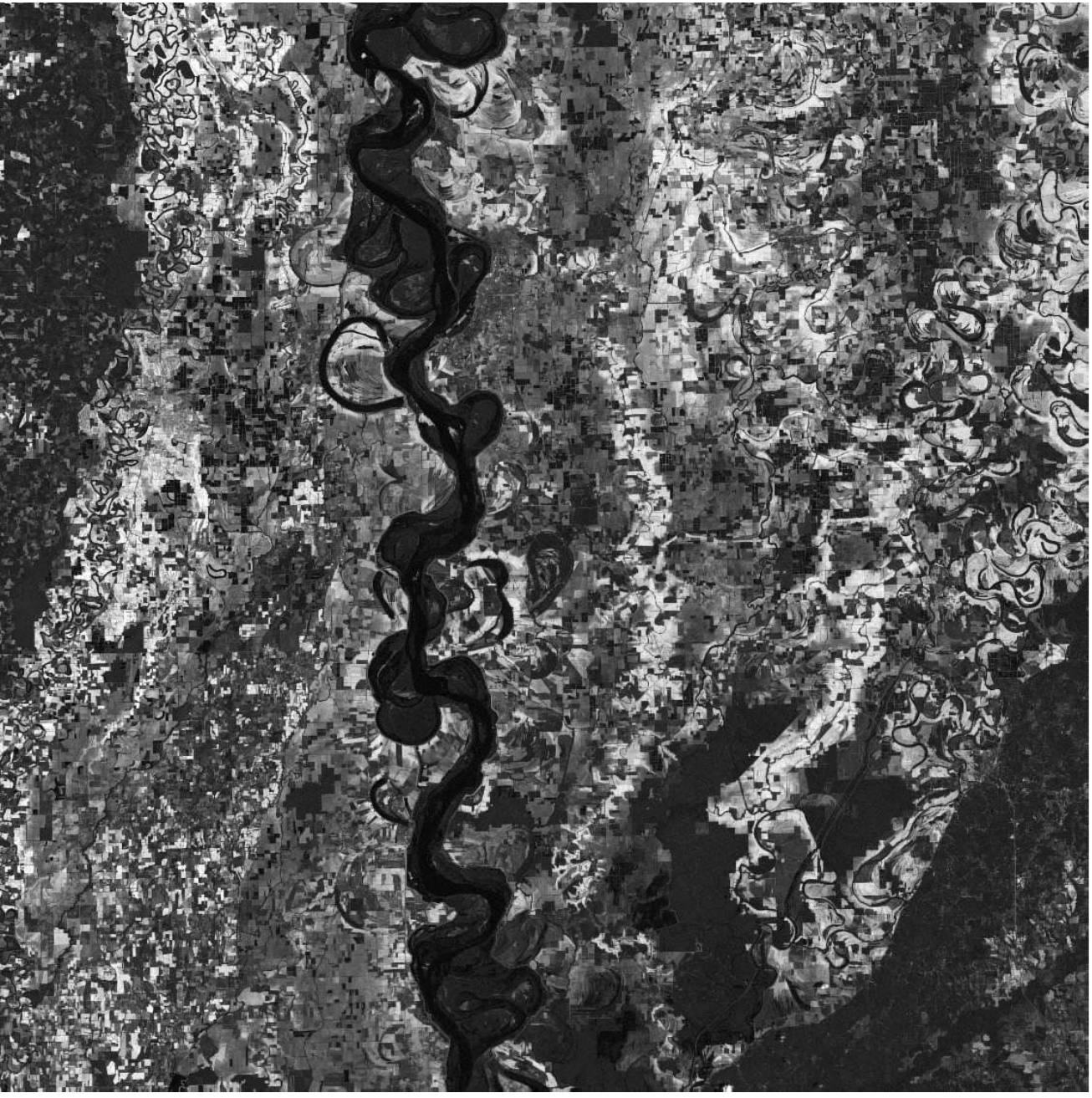}
\\
\scriptsize{\footnotesize(CPU-Time, Relative-Error, PSNR)}
%
&\scriptsize {\footnotesize(1136.81s, 0.0358, 37.85db)}
& \scriptsize{\footnotesize(260.76s,  0.0374, 37.47db)}
&\scriptsize {\footnotesize(246.17s,  0.0380, 37.32db)}
\\
\includegraphics[width=1.6in, height=0.80in]{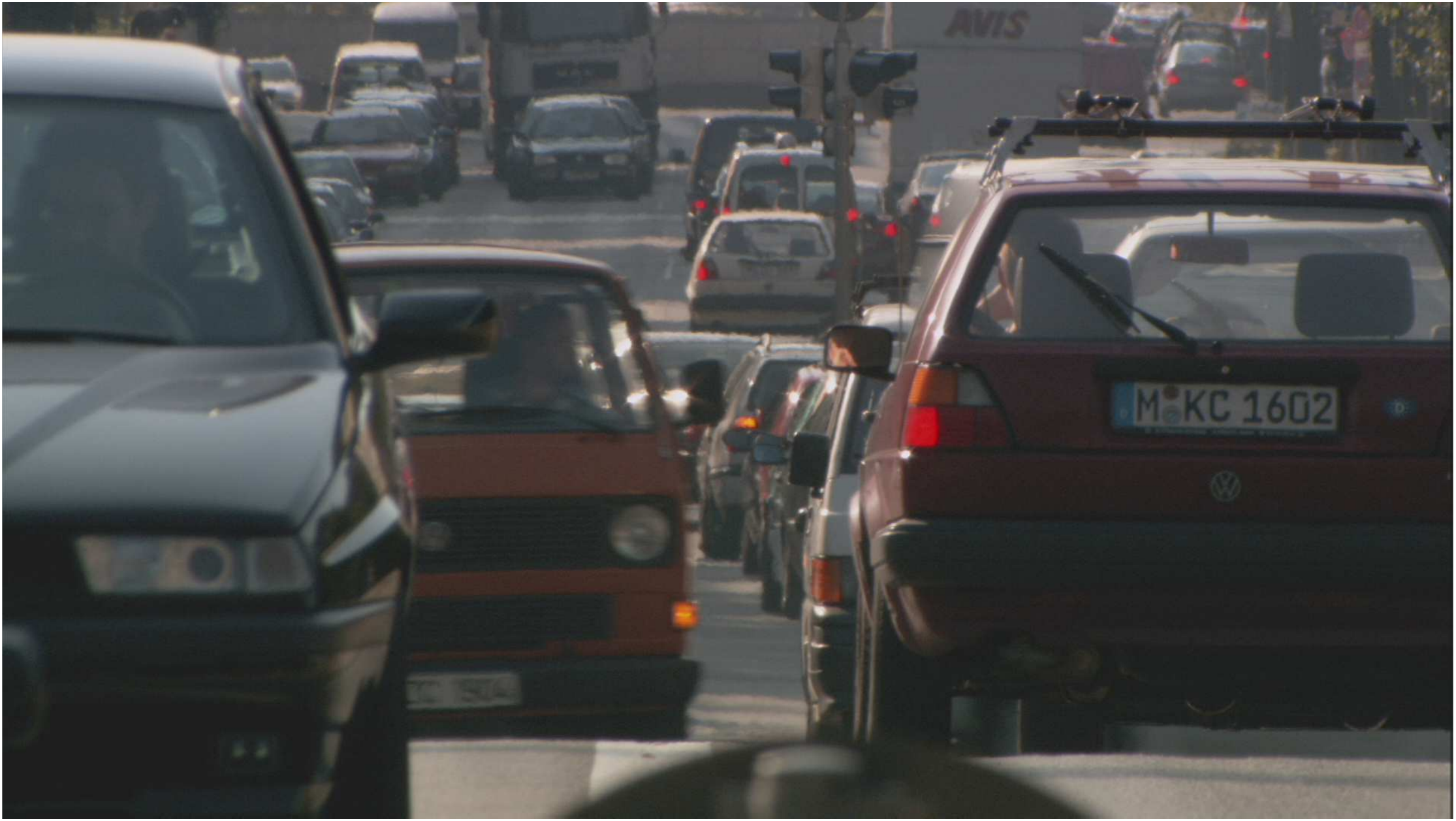}&
\includegraphics[width=1.6in, height=0.80in]{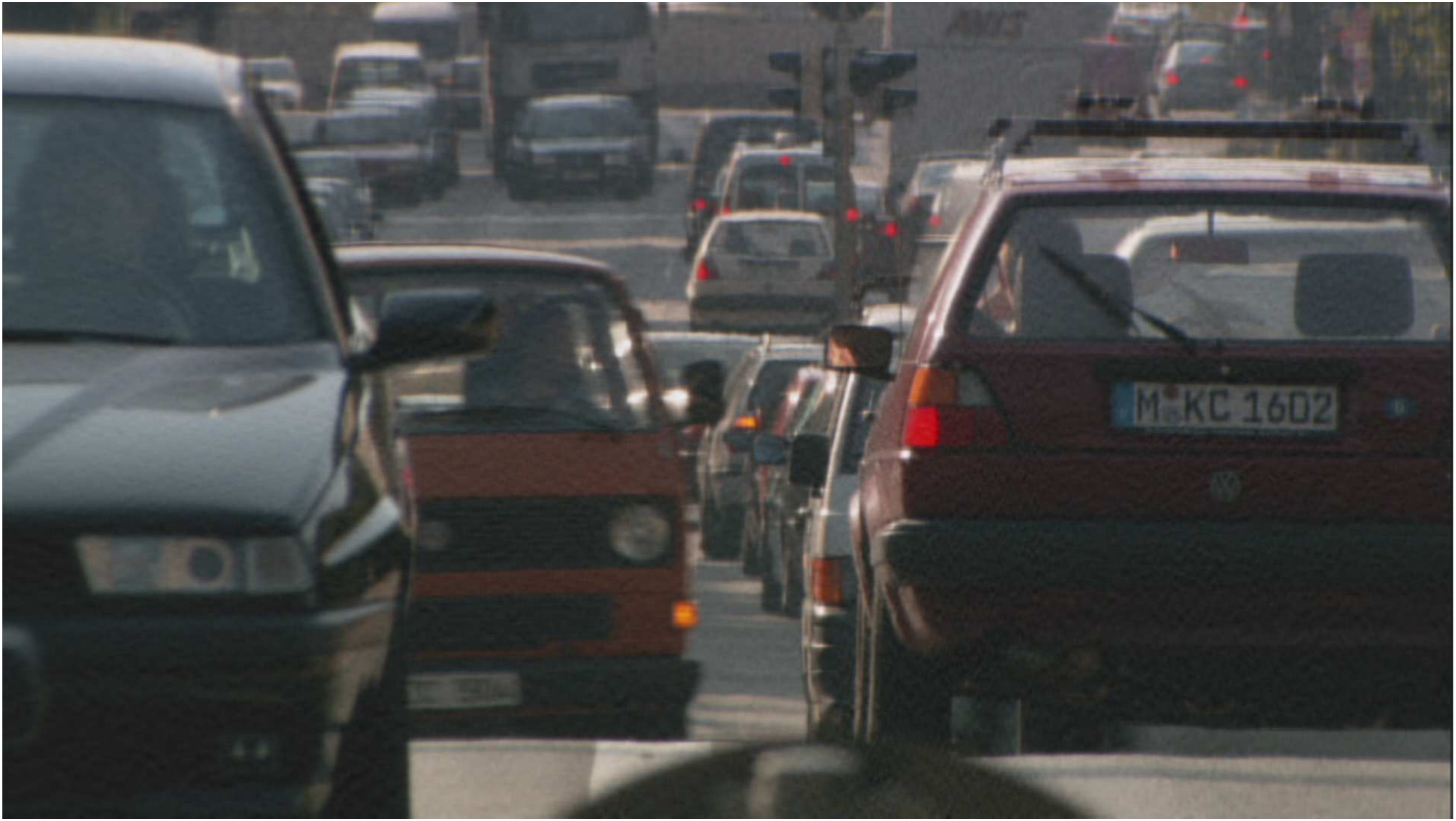}&
\includegraphics[width=1.6in, height=0.80in]{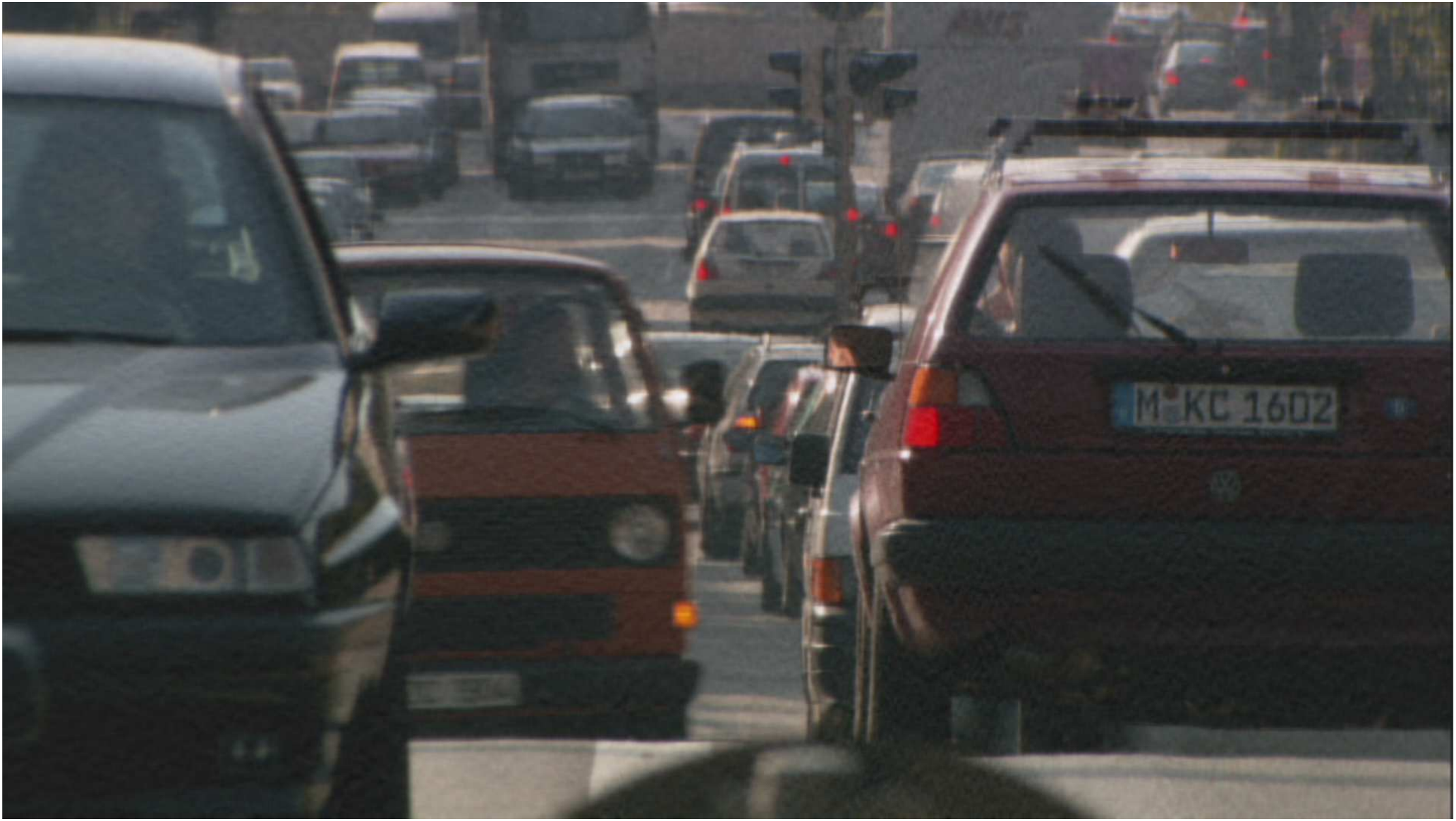}&
\includegraphics[width=1.6in, height=0.80in]{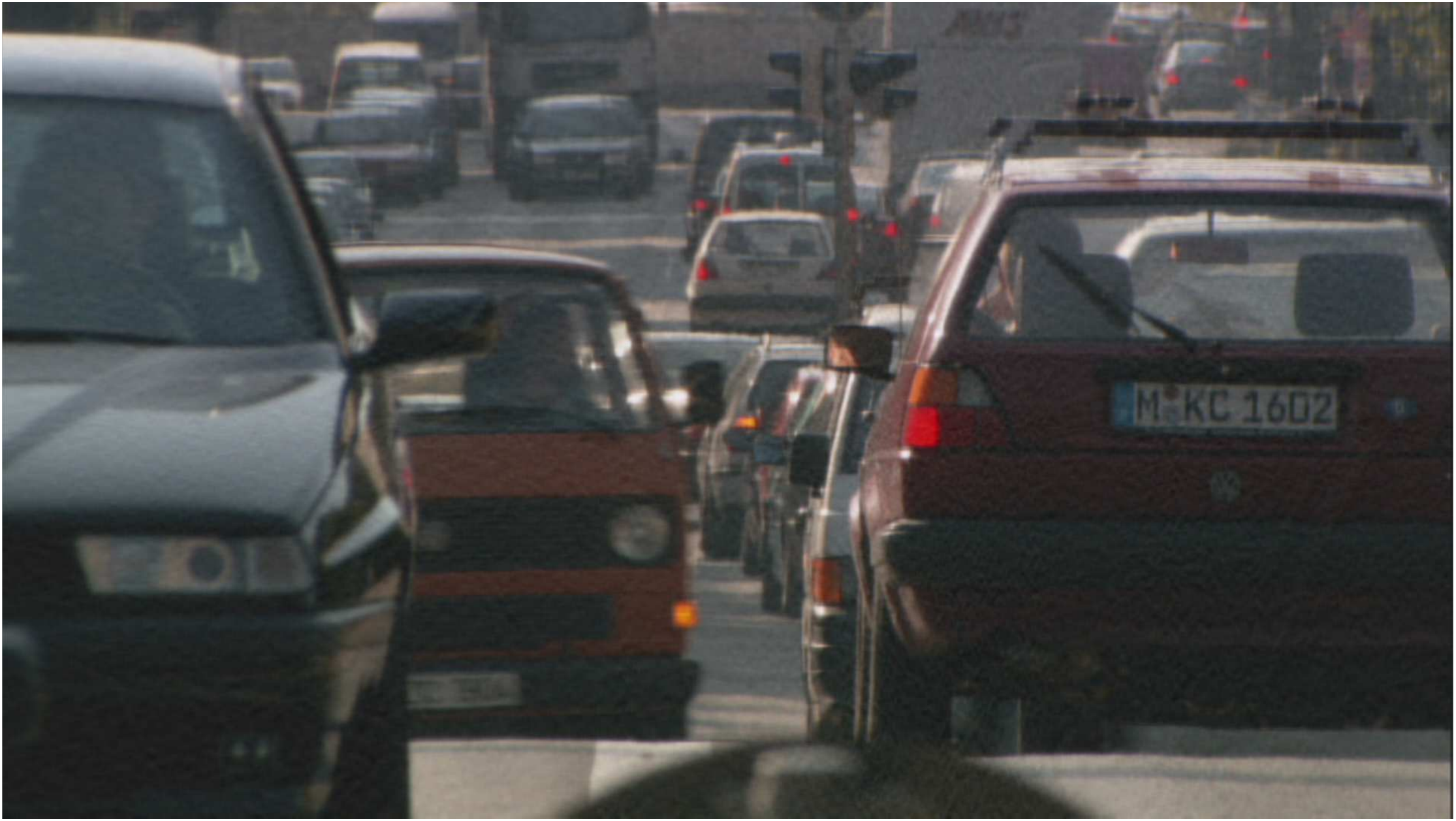}
\\
%
\scriptsize{\footnotesize(CPU-Time, Relative-Error, PSNR)}
&\scriptsize {\footnotesize(586.05s,  0.0436, 36.27db)}
&\scriptsize {\footnotesize(142.57s,  0.0455, 35.88db)}
& \scriptsize{\footnotesize(122.92s,  0.0463, 35.74db)}
%
\\
\includegraphics[width=1.6in, height=0.80in]{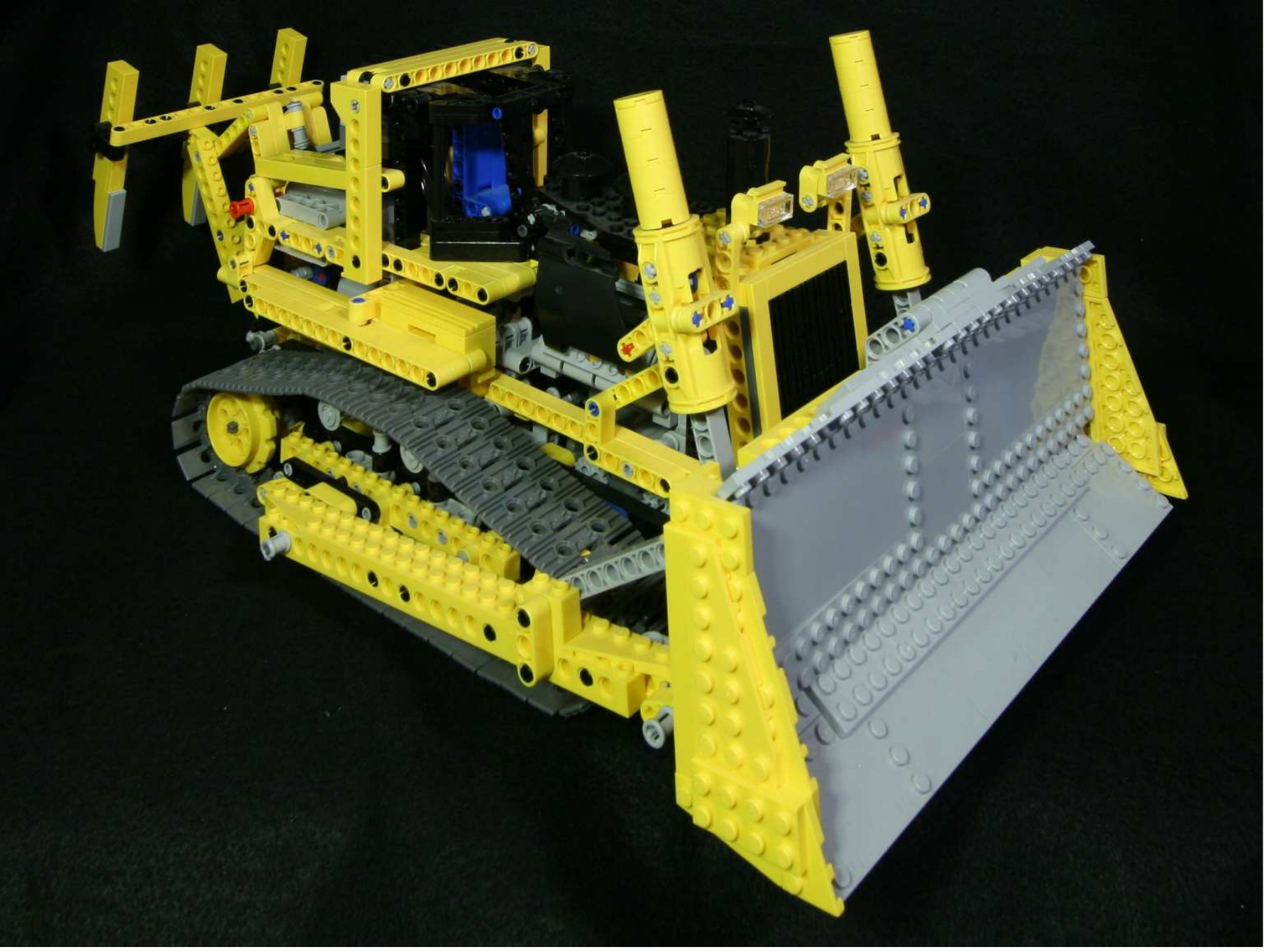}&
\includegraphics[width=1.6in, height=0.80in]{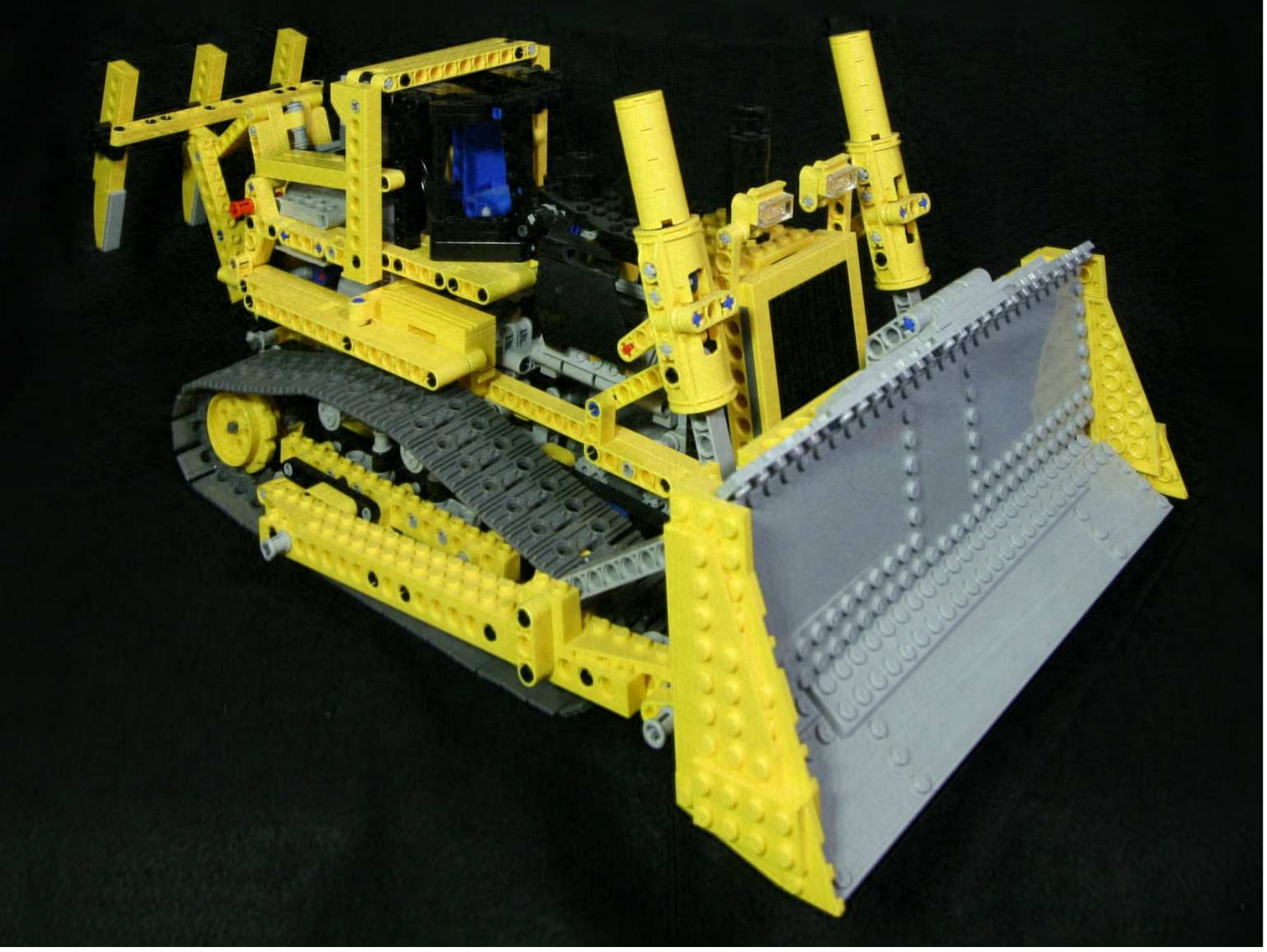}&
\includegraphics[width=1.6in, height=0.80in]{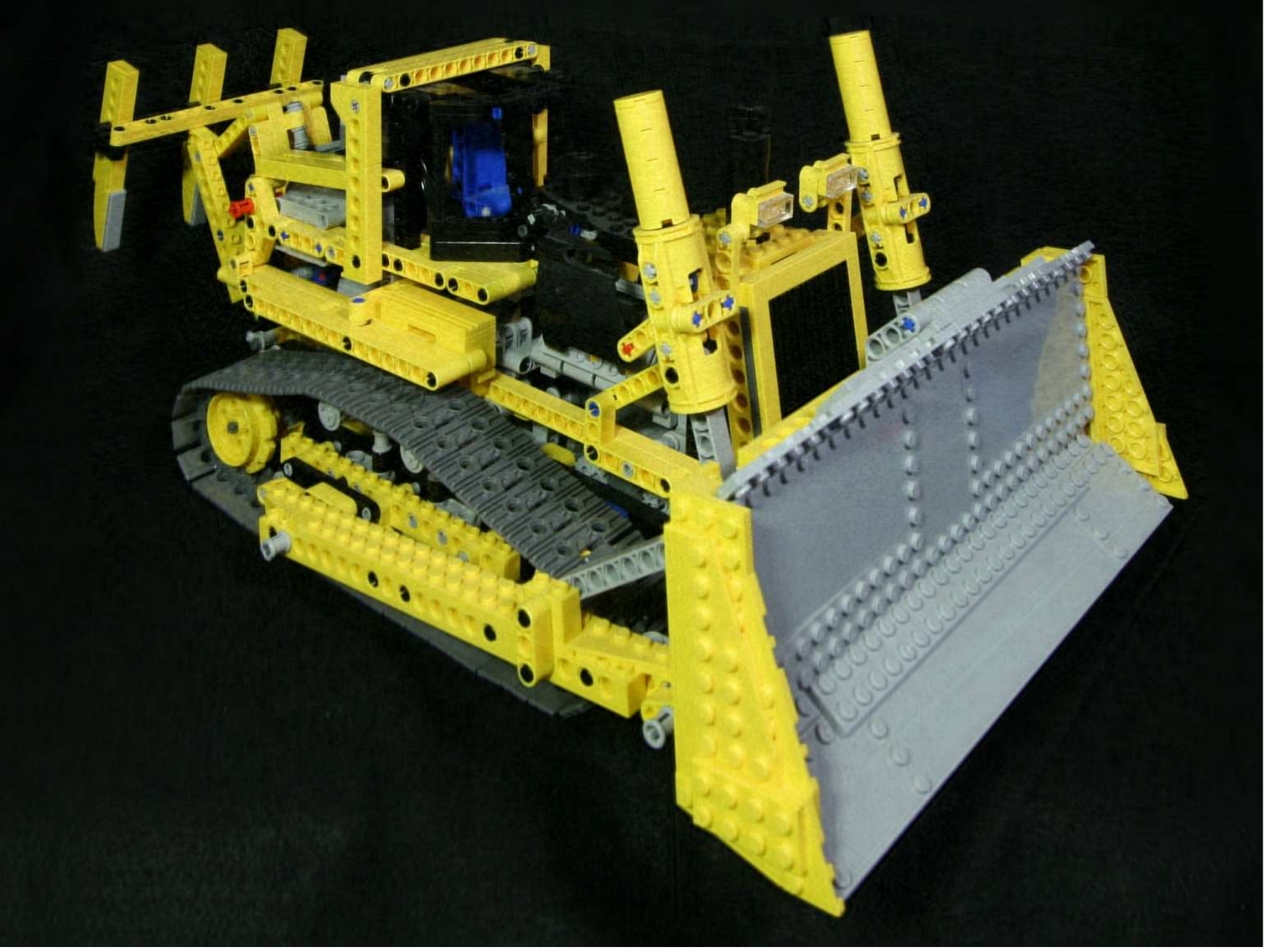}&
\includegraphics[width=1.6in, height=0.80in]{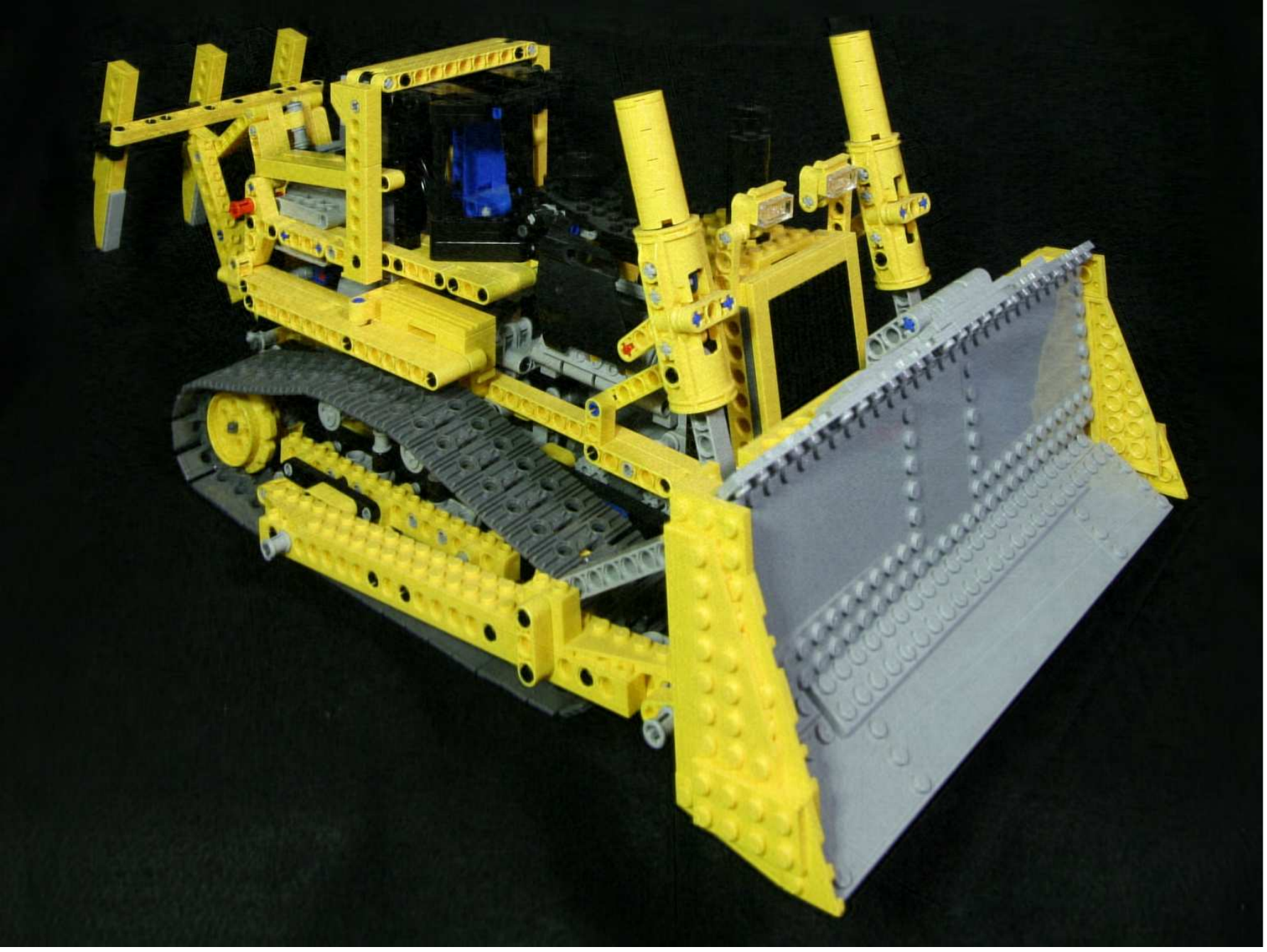}
\\
\small{(I) {{Original}}}
&\small {(II) {Truncated T-SVD \cite{qin2022low}
}}
&\small{(III) {Proposed R-TSVD}}
&\small{(IV) {Proposed RB-TSVD
 }}
\\
\end{tabular}
\caption
{
 Examples of large-scale high-order  
 tensor approximation.
 From top
to bottom
are 
multi-temporal remote sensing images
($4500\times 4500 \times 6 \times 11$),
color video 
($1080\times 1920 \times 3 \times 500$),
and 
light field images 
($1152\times 1536 \times 3 \times 17\times 17$),
respectively.
(I) 
Original versions; 
(II)-(IV) The  approximated version 
obtained by truncated T-SVD, 
 R-TSVD,  RB-TSVD, respectively.
}
\vspace{-0.6cm}
\label{intro_image22}
\end{figure*}

 \vspace{-0.19cm}

\section{\textbf{
notations and preliminaries
}}\label{nota}
For brevity, 
the main notations and preliminaries utilized 
in 
the whole paper
are summarized
in Table \ref{notation_part1},
most of which originate form  the literature \cite{qin2022low}.

In this work,
 we let
 ${\mathfrak{L}}(\boldsymbol{\mathcal{A}})$ represent 
 the result of invertible linear transforms $\mathfrak{L}$ on $\boldsymbol{\mathcal{A}} \in \mathbb{R}^{n_1\times \cdots \times  n_d}$, i.e.,
 \begin{align}\label{trans}
 \mathfrak{L}(\boldsymbol{\mathcal{A}}) =\boldsymbol{\mathcal{A}} \;{\times}_{3} \;\bm{U}_{n_3} \;{\times}_{4} \;\bm{U}_{n_4} \cdots
    {\times}_{d} \;\bm{U}_{n_d},
\end{align}
 where   the 
 transform matrices $\bm{U}_{n_i} \in \mathbb{C}^{n_i \times n_i}$ of $\mathfrak{L}$ satisfies:
   \begin{align}\label{orth}
  {{\bm{U}}}_{n_i} \cdot {{\bm{U}}}^{\mit{H}}_{n_i}={{\bm{U}}}^{\mit{H}}_{n_i} \cdot {{\bm{U}}}_{n_i}=\alpha_i {{\bm{I}}}_{n_i},  \forall i\in \{3,\cdots,d\},
   \end{align}
   in which $\alpha_i>0$  is a  constant.
  The inverse operator of $\mathfrak{L}(\boldsymbol{\mathcal{A}})$
   is defined as
   $ \mathfrak{L}^{-1} (\boldsymbol{\mathcal{A}}) =\boldsymbol{\mathcal{A}} \;{\times}_{d} \;\bm{U}^{-1}_{n_d} \;{\times}_{{d-1}} \;{{\bm{U}}_{n}}_{d-1}^{-1} \cdots
    {\times}_{3} \;\bm{U}^{-1}_{n_3}$,
    and  $\mathfrak{L}^{-1} (\mathfrak{L}(\boldsymbol{\mathcal{A}}))=\boldsymbol{\mathcal{A}}$.

\begin{Definition}\label{def10}
\textbf{(Order-$d$ WTSN)}
Let $\mathfrak{L}$ be any invertible linear transform in (\ref{trans}) and it satisfies (\ref{orth}),
${\boldsymbol{\mathcal{S}}}$ be from the middle 
component of
${\boldsymbol{\mathcal{A}}} = {\boldsymbol{\mathcal{U}}}{{*}_{\mathfrak{L}}}{\boldsymbol{\mathcal{S}}} {*}_{\mathfrak{L}}
 {\boldsymbol{\mathcal{V}}}^{\mit{T}}$.
Then,
the weighted  tensor Schatten-$p$ norm (WTSN) of ${\boldsymbol{\mathcal{A}}} \in \mathbb{R}^{ n_1\times \cdots \times n_d} $
  is defined as
%
%
\begin{align*}\label{WTSN}
\| {\boldsymbol{\mathcal{A}}}\|_{ {\boldsymbol{\mathcal{W}}}, {\boldsymbol{\mathcal{S}}}_{p} }
&:= {\Big(\frac{1}{\rho} \big\|\operatorname{bdiag}\big( {\mathfrak{L}}(\boldsymbol{\mathcal{A}} )\big)\big\|_
{
{\bm{w}}  
,
 {\mathrm{S}}_{p}
}^{p}\Big)}^{1/p}
\\
&=\Big(
\frac{1}{\rho}
\sum_{j=1}^{n_3n_4\cdots n_d}
\big\|
{{\mathfrak{L}} (\boldsymbol{\mathcal{A}})}^{< j>}
\big\|_
{{ 
{\bm{w}}^{(j)},
 {\mathrm{S}}_{p}}}^{p}
\Big)^{ 
{1}/{p}
}
\\
&=\Big(
\frac{1}{\rho} \sum_{j=1}^{n_3n_4\cdots n_d}
\operatorname{trace}\big(
{\boldsymbol{\mathcal{W}}}^{<j>}
\cdot  \big| { {\mathfrak{L}}({{\boldsymbol{\mathcal{S}}}})} ^ {<j>}
\big|^{p}\big)
\Big)^{
{1}/{p}
}, 
\end{align*}
where 
$\boldsymbol{\mathcal{W}}$ is the nonnegative  weight 
composed of an order-$d$ f-diagonal tensor,
$
\bm{w}
=
\operatorname{diag} 
(   \operatorname{bdiag}(\boldsymbol{\mathcal{W}})      
)$,
$ 
{\bm{w}^{(j)}}
=\operatorname{diag} 
(  \boldsymbol{\mathcal{W}}^{<j>}  )$,
and $\rho=\alpha_3\alpha_4\cdots \alpha_d>0$ is a
constant determined by the  invertible linear transform
$\mathfrak{L}$.
\end{Definition}

\begin{Remark}
The
high-order WTSN (HWTSN)
assigns different weight values to different singular values in the 
transform domain:
the larger one
is multiplied by a smaller weight while the smaller one 
 is multiplied by a larger weight.
 That is,
 the  weight values should be inversely proportional to the singular values in the transform domain.
In particular, the HWTSN \textbf{1)}
is equivalent to
the high-order  tensor
Schatten-$p$ norm (HTSN) when 
weighting is not 
taken into account, \textbf{2)}
reduces to  
the high-order weighted TNN 
(HWTNN)\cite{qin2021robust} 
when $p=1$,
and
\textbf{3)}
simplifies to
the high-order TNN (HTNN)\cite{qin2022low} 
when $p=1$, and 
$\boldsymbol{\mathcal{W}}$ is not considered.
\end{Remark}

\begin{Theorem}\label{th_app}
\textcolor[rgb]{0.00,0.00,0.00}{\textbf{(
Optimal
 $k$-term approximation 
\cite{qin2022low})}
Let the \text{T-SVD} of
${\boldsymbol{\mathcal{A}}} \in \mathbb{R}^{n_1\times \cdots \times  n_d}$
be
%
${\boldsymbol{\mathcal{A}}}={\boldsymbol{\mathcal{U}}}{*}_{\mathfrak{L}}{\boldsymbol{\mathcal{S}}}{*}_{\mathfrak{L}}{\boldsymbol{\mathcal{V}}}^{\mit{T}}$
and define
${\boldsymbol{\mathcal{A}}}_{k}={\sum}_{i=1}^{k}
 {\boldsymbol{\mathcal{U}}}(:,i,:,\cdots,:) {*}_{\mathfrak{L}} {\boldsymbol{\mathcal{S}}}(i,i,:,\cdots,:){*}_{\mathfrak{L}}
 {\boldsymbol{\mathcal{V}}}(:,i,:,\cdots,:)^{\mit{T}}$
for some $k<\min(n_1,n_2)$.
Then, ${\boldsymbol{\mathcal{A}}}_{k}=\arg\min_{
{\tilde{\boldsymbol{{\mathcal{A}}}}} \in 
\bm{\Theta}
}
\|{\boldsymbol{\mathcal{A}}}-{\tilde{\boldsymbol{{\mathcal{A}}}}}\|_{\mathnormal{F}}$,
where $
\bm{\Theta}
=\{ {\boldsymbol{\mathcal{X}}} {*}_{\mathfrak{L}} {\boldsymbol{\mathcal{Y}}} |
{\boldsymbol{\mathcal{X}}} \in \mathbb{R}^{n_1\times k \times n_3 \times \cdots  \times  n_d},
{\boldsymbol{\mathcal{Y}}} \in \mathbb{R}^{k\times n_2  \times n_3 \times \cdots  \times  n_d}\}$.
This
implies that ${\boldsymbol{\mathcal{A}}}_{k}$ is 
the
approximation
of ${\boldsymbol{\mathcal{A}}}$ with the T-SVD rank at most $k$.
}
\end{Theorem}

\vspace{-0.4cm}
\section{\textbf{
Randomized Techniques Based High-Order Tensor Approximation
}}\label{approximation}

%

The optimal $k$-term approximation presented in Theorem \ref{th_app}
  is
  time-consuming 
 for large-scale tensors.
To tackle this issue, 
an efficient
QB approximation for high-order tensor is developed
in virtue of randomized projection
techniques.
%
On this basis,
we put forward an effective
randomized 
algorithm for
calculating
the high-order T-SVD (\text{abbreviated as \textbf{R-TSVD}}).
To be  slightly more specific, 
the  calculation   
 of R-TSVD can be subdivided into the  following two steps:

\renewcommand{\arraystretch}{0.1}
 \begin{algorithm}[!htbp]
\setstretch{0.0}
     \caption
     {
    The Basic \textbf{randQB} Approximation. 
    }
     \label{RTSVD-RANK777}
      \KwIn{
      $\bm{\mathcal{A}}\in\mathbb{R}^{n_1\times \cdots\times n_d}$,  invertible linear transform:
       $\mathfrak{L}$, target T-SVD rank:  $k$, 
        oversampling parameter: $s$. 
       }
      {\color{black}\KwOut{
      $\bm{\mathcal{Q}},\bm{\mathcal{B}}$.
      }}
      	Set $\hat{l} = k + s$, and generate 
      a Gaussian random tensor $\bm{\mathcal{G}}\in\mathbb{R}^{n_2\times \hat{l}\times n_3 \times \cdots \times  n_d}$\;

      Construct
      a random projection of 
      $\bm{\mathcal{A}}$ as
      ${\boldsymbol{\mathcal{Y}}}={\boldsymbol{\mathcal{A}}}{*}_{\mathfrak{L}} {\boldsymbol{\mathcal{G}}}$\;

Form
the 
partially orthogonal
tensor ${{\boldsymbol{\mathcal{Q}}}}$ by computing the T-QR factorization of $\bm{\mathcal{Y}}$\;

%

%

     $ {{\boldsymbol{\mathcal{B}}}}= {\boldsymbol{\mathcal{Q}}}^{\mit{T}}{*}_{\mathfrak{L}} {{\boldsymbol{\mathcal{A}}}}$
    .
   \end{algorithm}

 \begin{itemize}
   \item \textbf{Step I (Randomized Step):} Compute an approximate basis for the range of the target tensor $\bm{\mathcal{A}}
   \in \mathbb{R}^{n_1 \times n_2 \times n_3 \times  \cdots \times n_d}
   $ via  randomized projection
   techniques.
   That is to say,
    we require an orthogonal subspace basis 
   tensor $\bm{\mathcal{Q}} \in \mathbb{R}^{n_1 \times l \times n_3 \times  \cdots \times n_d}
   $ which satisfies
   \begin{equation}\label{tqb11}
{\boldsymbol{\mathcal{A}}}\approx
{\boldsymbol{\mathcal{Q}}} {*}_{\mathfrak{L}}  {\boldsymbol{\mathcal{B}}}=
{\boldsymbol{\mathcal{Q}}}{*}_{\mathfrak{L}}  {{\boldsymbol{\mathcal{Q}}}}^{\mit{T}} {*}_{\mathfrak{L}}  {\boldsymbol{\mathcal{A}}}.
\end{equation}
%
The approximation presented by (\ref{tqb11}) can also be regarded as a kind of low-rank factorization/approximation of ${\boldsymbol{\mathcal{A}}}$,
called QB factorization or QB approximation in our work.
A basic randomized 
technique for computing the
 QB factorization 
 is 
 shown in  Algorithm \ref{RTSVD-RANK777},
 which is denoted as  the basic 
 randomized QB (randQB)
 approximation. 
   \item \textbf{Step II (Deterministic Step):} Perform the deterministic T-QR factorization on the reduced tensor ${\boldsymbol{\mathcal{B}}}^{\mit{T}}$,
   i.e., ${\boldsymbol{\mathcal{B}}}^{\mit{T}}={\boldsymbol{\mathcal{Q}}}_{1} {*}_{\mathfrak{L}}  {\boldsymbol{\mathcal{R}}}$.
   Then, execute the deterministic
   T-SVD on the smaller tensor ${\boldsymbol{\mathcal{R}}}$, i.e.,
   ${\boldsymbol{\mathcal{R}}}=\hat{{\boldsymbol{\mathcal{U}}}} {*}_{\mathfrak{L}}  {\boldsymbol{\mathcal{S}}} {*}_{\mathfrak{L}}
   {\hat{{\boldsymbol{\mathcal{V}}}}}^{\mit{T}}$. Thus,
    \begin{align}\label{tqb}
{\boldsymbol{\mathcal{A}}}
&\approx
{\boldsymbol{\mathcal{Q}}} {*}_{\mathfrak{L}}  \hat{{\boldsymbol{\mathcal{V}}}}  {*}_{\mathfrak{L}}  {\boldsymbol{\mathcal{S}}}
{*}_{\mathfrak{L}}
{\hat{\boldsymbol{\mathcal{U}}}}^{\mit{T}} {*}_{\mathfrak{L}}  {{\boldsymbol{\mathcal{Q}}}_{1}}^{\mit{T}}
%
\notag \\
&
=
{\boldsymbol{\mathcal{U}}} {*}_{\mathfrak{L}}  {{\boldsymbol{\mathcal{S}}}}  {*}_{\mathfrak{L}}
{{\boldsymbol{\mathcal{V}}}}^{\mit{T}}
(\textit{Let}
\;
{{\boldsymbol{\mathcal{U}}}}={{\boldsymbol{\mathcal{Q}}}}{*}_{\mathfrak{L}}
\hat{{\boldsymbol{\mathcal{V}}}},
{{\boldsymbol{\mathcal{V}}}}={{\boldsymbol{\mathcal{Q}}}_{1}}{*}_{\mathfrak{L}}
\hat{{\boldsymbol{\mathcal{U}}}}
)
.
\end{align}
 \end{itemize}
\begin{algorithm}[!htbp]
\setstretch{0.01}
     \caption{
     \text{Transform Domain Version}: 
    \textbf{R-TSVD}.
     }
     \label{transform-version-fp11}
      \KwIn{$\bm{\mathcal{A}}\in\mathbb{R}^{n_1\times \cdots\times n_d}$,  
      transform:
       $\mathfrak{L}$,
       target T-SVD rank:  $k$, 
        oversampling parameter: $s$,  power iteration: $t$.
       }
      {\color{black}\KwOut{
$\bm{\mathcal{U}}$, $\bm{\mathcal{S}}$, $\bm{\mathcal{V}}$.
      }}

Set $\hat{l} = k + s$ and initialize a Gaussian random tensor $\bm{\mathcal{G}}\in\mathbb{R}^{n_2\times \hat{l} \times n_3 \times \cdots \times  n_d}$\;

       Compute the results of $\mathfrak{L}$ on $\boldsymbol{\mathcal{A} }$ and $\boldsymbol{\mathcal{G} }$, i.e.,
       $\mathfrak{L}(\boldsymbol{\mathcal{A}}), \mathfrak{L}(\boldsymbol{\mathcal{G}})$\;

      \For{$v=1,2,\cdots, n_3\cdots n_d$}
      {

 $[ \mathfrak{L}(\boldsymbol{\mathcal{Q}})^{<v>},\sim]=\operatorname{qr}(
     \mathfrak{L}(\boldsymbol{\mathcal{A}})^{<v>}\cdot \mathfrak{L}(\boldsymbol{\mathcal{G}})^{<v>}
     )$\;

       \For{$ j=1,2,\ldots,t$}
    {
     $[ \mathfrak{L}(\boldsymbol{\mathcal{Q}}_{1})^{<v>},\sim]=\operatorname{qr}\big(
     {( \mathfrak{L}(\boldsymbol{\mathcal{A}})^{<v>})}^{\mit{T}}
     \cdot \mathfrak{L}(\boldsymbol{\mathcal{Q}})^{<v>}
    \big )$\;
   $
   [ \mathfrak{L}(\boldsymbol{\mathcal{Q}})^{<v>},
   \sim]=\operatorname{qr}\big(
   \mathfrak{L}(\boldsymbol{\mathcal{A}})^{<v>}
  \cdot \mathfrak{L}(\boldsymbol{\mathcal{Q}}_{1})^{<v>}
   \big)$\;
   }



 $[\mathfrak{L}(\boldsymbol{\mathcal{Q}}_{1})^{<v>},\mathfrak{L}(\boldsymbol{\mathcal{R}})^{<v>}]= \textrm{qr} \big(
 (\mathfrak{L}(\boldsymbol{\mathcal{A}})^{<v>})^{\mit{T}}\cdot
 \mathfrak{L}(\boldsymbol{\mathcal{Q}})^{<v>}
 \big) $;\\

$[\mathfrak{L}(\hat{\boldsymbol{\mathcal{U}}})^{<v>},\mathfrak{L}(\boldsymbol{\mathcal{S}})^{<v>},\mathfrak{L}(\hat{\boldsymbol{\mathcal{V}}})^{<v>}]= \textrm{svd} \big(\mathfrak{L}(\boldsymbol{\mathcal{R}})^{<v>}\big) $;\\

$
\mathfrak{L}(\boldsymbol{\mathcal{V}})^{<v>}
= \mathfrak{L}(\boldsymbol{\mathcal{Q}}_{1})^{<v>}\cdot \mathfrak{L}(\hat{\boldsymbol{\mathcal{U}}})^{<v>}
$;\\

$
\mathfrak{L}(\boldsymbol{\mathcal{U}})^{<v>}
= \mathfrak{L}(\boldsymbol{\mathcal{Q}})^{<v>}\cdot \mathfrak{L}(\hat{\boldsymbol{\mathcal{V}}})^{<v>}
$;\\


}
${\boldsymbol{\mathcal{U}}} \leftarrow {\mathfrak{L}}^{-1}(
       {\mathfrak{L}}({\boldsymbol{\mathcal{U}}})
        )$,
        ${\boldsymbol{\mathcal{S}}} \leftarrow {\mathfrak{L}}^{-1}(
       {\mathfrak{L}}({\boldsymbol{\mathcal{S}}})
        )$,
        ${\boldsymbol{\mathcal{V}}} \leftarrow {\mathfrak{L}}^{-1}(
       {\mathfrak{L}}({\boldsymbol{\mathcal{V}}})
        )$.
    \end{algorithm}


\begin{Remark}
\textbf{(Power Iteration Strategy)}
To further improve the accuracy of randQB approximation
of   ${{\boldsymbol{\mathcal{A}}}}$,
 we can additionally apply the power iteration  scheme, which  multiplies alternately with
 ${{\boldsymbol{\mathcal{A}}}}$ and ${{{\boldsymbol{\mathcal{A}}}}}^{\mit{T}}$, i.e.,
 ${({{\boldsymbol{\mathcal{A}}}} {*}_{\mathfrak{L}} {{{\boldsymbol{\mathcal{A}}}}}^{\mit{T}})}^{t}
 {*}_{\mathfrak{L}} {{\boldsymbol{\mathcal{A}}}}$,
 where $t$ is a  nonnegative integer.
 Besides, to avoid the rounding error of float point arithmetic obtained from performing the power iteration,
  the reorthogonalization step
  is required.
 Thus,
the randQB approximation algorithm
incorporating power iteration   strategy 
 can be obtained by adding the following steps after the third step of Algorithm \ref{RTSVD-RANK777}, i.e.,

    \For{$ j=1,2,\ldots,t$}
    {
     $[ {{\boldsymbol{\mathcal{Q}}}_{1}},\sim]=\operatorname{H-TQR}(
     {\boldsymbol{\mathcal{A}}}^{\mit{T}} {*}_{\mathfrak{L}} {{\boldsymbol{\mathcal{Q}}}},
     \mathfrak{L})$;\;



   $[ {{\boldsymbol{\mathcal{Q}}}},\sim]=\operatorname{H-TQR}(
   {\boldsymbol{\mathcal{A}}}{*}_{\mathfrak{L}} {{\boldsymbol{\mathcal{Q}}}_{1}},
   \mathfrak{L})$;\;
   }
\end{Remark}

\begin{algorithm}[!htbp]
\setstretch{0.01}
     \caption{
  Transform Domain Version: 
 \textbf{RB-TSVD}.
     }
     \label{transform-version-fp}
      \KwIn{$\bm{\mathcal{A}}\in\mathbb{R}^{n_1\times \cdots\times n_d}$,  
      transform:
       $\mathfrak{L}$,
       target T-SVD Rank:  $k$,
        block size: $b$,
        power iteration: $t$.
       }
      {\color{black}\KwOut{
$\bm{\mathcal{U}}$, $\bm{\mathcal{S}}$, $\bm{\mathcal{V}}$.
      }}

      	Let  $\hat{l}$   be  a number slightly larger than
     $k$, and generate
      a Gaussian random tensor $\bm{\mathcal{G}}\in\mathbb{R}^{n_2\times \hat{l}\times n_3 \times \cdots \times  n_d}$\;

        Compute the results of $\mathfrak{L}$ on $\boldsymbol{\mathcal{A} }$ and $\boldsymbol{\mathcal{G} }$, i.e.,
       $\mathfrak{L}(\boldsymbol{\mathcal{A}}), \mathfrak{L}(\boldsymbol{\mathcal{G}})$\;

      \For{$v=1,2,\cdots, n_3\cdots n_d$}
      {

  $ {\mathfrak{L}(\boldsymbol{\mathcal{Q}}) }^{<v>}=[\;\;];\;\;
       {\mathfrak{L}(\boldsymbol{\mathcal{B}})}^{<v>}=[\;\;]$\;

       \For{$ j=1,2,\ldots,t$}
    {
     $[ \mathfrak{L}(\boldsymbol{\mathcal{W}})^{<v>},\sim]=\operatorname{qr}(
     \mathfrak{L}(\boldsymbol{\mathcal{A}})^{<v>}\cdot \mathfrak{L}(\boldsymbol{\mathcal{G}})^{<v>}
     )$\;
   $
   [ \mathfrak{L}(\boldsymbol{\mathcal{G}})^{<v>},
   \sim]=\operatorname{qr}\big(
  {( \mathfrak{L}(\boldsymbol{\mathcal{A}})^{<v>})}^{\mit{T}}\cdot \mathfrak{L}(\boldsymbol{\mathcal{W}})^{<v>}
   \big)$\;
   }

  $ \mathfrak{L}(\boldsymbol{\mathcal{W}})^{<v>}=
   \mathfrak{L}(\boldsymbol{\mathcal{A}})^{<v>}\cdot \mathfrak{L}(\boldsymbol{\mathcal{G}})^{<v>}$
   \;

  $\mathfrak{L}(\boldsymbol{\mathcal{H}})^{<v>}=
   {( \mathfrak{L}(\boldsymbol{\mathcal{A}})^{<v>})}^{\mit{T}}\cdot \mathfrak{L}(\boldsymbol{\mathcal{W}})^{<v>}$\;

\For{$i=1,2,\cdots,\lfloor\frac{\hat{l}}{ b}\rfloor$}
              {
           $ \mathfrak{L}\big({\boldsymbol{\mathcal{G}}}^{(i)}\big)^{<v>}=
             \mathfrak{L}(\boldsymbol{\mathcal{G}})^{<v>} \big(:,(i-1)b+1:ib \big)$\;

$
\mathfrak{L}\big({\boldsymbol{\mathcal{Y}}}^{(i)}\big)^{<v>}
=
-
\mathfrak{L}(\boldsymbol{\mathcal{Q}})^{<v>} \cdot \mathfrak{L}(\boldsymbol{\mathcal{B}})^{<v>} \cdot
\mathfrak{L}({\boldsymbol{\mathcal{G}}}^{(i)})^{<v>}
+
\mathfrak{L}(\boldsymbol{\mathcal{W}})^{<v>} \big(:,(i-1)b+1:ib \big)
$
\;

$[ {\mathfrak{L}({\boldsymbol{\mathcal{Q}}}^{(i)})}^{<v>},\mathfrak{L}({\boldsymbol{\mathcal{R}}}^{(i)})^{<v>}]
=\operatorname{qr}\big(\mathfrak{L}({\boldsymbol{\mathcal{Y}}}^{(i)})^{<v>}\big)$\;

$[ {\mathfrak{L}({\boldsymbol{\mathcal{Q}}}^{(i)})}^{<v>}, {( \mathfrak{L}({\hat{\boldsymbol{\mathcal{R}}}}^{(i)} ))}^{<v>}  ]=\operatorname{qr}(
{\mathfrak{L}({\boldsymbol{\mathcal{Q}}}^{(i)})}^{<v>}
-
{\mathfrak{L}({\boldsymbol{\mathcal{Q}}})}^{<v>}\cdot
{( \mathfrak{L}(\boldsymbol{\mathcal{Q}})^{<v>})}^{\mit{T}}\cdot
{\mathfrak{L}({\boldsymbol{\mathcal{Q}}}^{(i)})}^{<v>}
)$\;

$
{\mathfrak{L}({\boldsymbol{\mathcal{R}}}^{(i)})}^{<v>}=
{( \mathfrak{L}({\hat{\boldsymbol{\mathcal{R}}}}^{(i)} ))}^{<v>}\cdot
{\mathfrak{L}({\boldsymbol{\mathcal{R}}}^{(i)})}^{<v>}
$\;

$
{\mathfrak{L}({\boldsymbol{\mathcal{B}}}^{(i)})}^{<v>}=
{\big( \mathfrak{L}( {{\boldsymbol{\mathcal{R}}}^{(i)}})^{<v>}\big)}^{\mit{-T}}
\cdot
\big[
{\big( {\mathfrak{L}({\boldsymbol{\mathcal{H}}})}^{<v>}  (:,(i-1)b+1:ib)\big)}^{\mit{T}}
-
{ ({\mathfrak{L}({\boldsymbol{\mathcal{Y}}}^{(i)})}^{<v>})
}^{\mit{T}} \cdot
{\mathfrak{L}({\boldsymbol{\mathcal{Q}}})}^{<v>}\cdot{\mathfrak{L}({\boldsymbol{\mathcal{B}}})}^{<v>}
-
{ ({\mathfrak{L}({\boldsymbol{\mathcal{G}}}^{(i)})}^{<v>})
}^{\mit{T}} \cdot
 {( \mathfrak{L}(\boldsymbol{\mathcal{B}})^{<v>})}^{\mit{T}}
\cdot
{\mathfrak{L}({\boldsymbol{\mathcal{B}}})}^{<v>} \big]
$\;

$\mathfrak{L}(\boldsymbol{\mathcal{Q}})^{<v>} =[ \mathfrak{L}(\boldsymbol{\mathcal{Q}})^{<v>}, {\mathfrak{L}({\boldsymbol{\mathcal{Q}}}^{(i)})}^{<v>}]$\;
$\mathfrak{L}(\boldsymbol{\mathcal{B}})^{<v>} =
[ \mathfrak{L}(\boldsymbol{\mathcal{B}})^{<v>}, {\mathfrak{L}({\boldsymbol{\mathcal{B}}}^{(i)})}^{<v>}]^{\mit{T}}$\;

%
%
 }

 $[\mathfrak{L}(\boldsymbol{\mathcal{Q}}_{1})^{<v>},\mathfrak{L}(\boldsymbol{\mathcal{R}}_{1})^{<v>}]= \textrm{qr} \big(
 (\mathfrak{L}(\boldsymbol{\mathcal{B}})^{<v>})^{\mit{T}}
 \big) $;\\

$[\mathfrak{L}(\hat{\boldsymbol{\mathcal{U}}})^{<v>},\mathfrak{L}(\boldsymbol{\mathcal{S}})^{<v>},\mathfrak{L}(\hat{\boldsymbol{\mathcal{V}}})^{<v>}]= \textrm{svd} \big(\mathfrak{L}(\boldsymbol{\mathcal{R}}_{1})^{<v>}\big) $;\\

$
\mathfrak{L}(\boldsymbol{\mathcal{V}})^{<v>}
= \mathfrak{L}(\boldsymbol{\mathcal{Q}}_{1})^{<v>}\cdot \mathfrak{L}(\hat{\boldsymbol{\mathcal{U}}})^{<v>}
$;\\

$
\mathfrak{L}(\boldsymbol{\mathcal{U}})^{<v>}
= \mathfrak{L}(\boldsymbol{\mathcal{Q}})^{<v>}\cdot \mathfrak{L}(\hat{\boldsymbol{\mathcal{V}}})^{<v>}
$;\\


}
${\boldsymbol{\mathcal{U}}} \leftarrow {\mathfrak{L}}^{-1}(
       {\mathfrak{L}}({\boldsymbol{\mathcal{U}}})
        )$,
        ${\boldsymbol{\mathcal{S}}} \leftarrow {\mathfrak{L}}^{-1}(
       {\mathfrak{L}}({\boldsymbol{\mathcal{S}}})
        )$,
        ${\boldsymbol{\mathcal{V}}} \leftarrow {\mathfrak{L}}^{-1}(
       {\mathfrak{L}}({\boldsymbol{\mathcal{V}}})
        )$.
        \vspace{-0.0cm}
    \end{algorithm}

According to the above analysis,
the computational procedure of R-TSVD is shown in  Algorithm  \ref{transform-version-fp11}.
To
obtain high performance of linear algebraic computation,  
we further
investigate  the blocked version of
basic randQB approximation, 
and then
derive a randomized blocked  algorithm
for computing high-order T-SVD (abbreviated as  \textbf{RB-TSVD}, see Algorithm \ref{transform-version-fp}). 
 \textcolor[rgb]{0.00,0.00,0.00}{\textbf{Owing to the space limitations of this paper,
 the  detailed derivation of 
blocked randQB approximation and
its induced
RB-TSVD are 
given in the supplementary material.}}
%
 The experimental results shown
 in Figure \ref{intro_image22}  
 indicate that
 for various real-world large-scale tensors,
  the proposed
   RB-TSVD and R-TSVD algorithms achieve up to
    $4$$\sim$$8$X
 faster running time than 
 the previous truncated T-SVD
 while maintaining 
 similar accuracy.
 %
 Besides, there is no significant difference between RB-TSVD and R-TSVD in  
 calculation time.  
 Only for very large-scale tensors (i.e., the spatial dimensions up to about $10^4 \times 10^4$)
 does the advantage of RB-TSVD  show up,
 which is verified in the supplementary materials.

\vspace{-0.3cm}

\section{\textbf{Rubust High-Order Tensor Completion}}\label{model}

\subsection{\textbf{Proposed Model}}

In this subsection, we formally introduce
the  double nonconvex
 model for RHTC,
in which  the low-rank 
component
 is constrained by the  HWTSN 
  ({see Definition \ref{def10}}),
while the 
 noise/outlier component 
 is regularized by its weighted  
 $\ell_{q}$-norm ({see Table \ref{notation_part1}}).
 Specifically,
 suppose that we are given  a low T-SVD rank 
 tensor ${\boldsymbol{\mathcal{L}}} \in \mathbb{R}^{n_1 \times  \cdots \times  n_d}$
corrupted by the 
noise or outliers.
The corrupted part can be 
represented by the 
tensor ${\boldsymbol{\mathcal{E}}} \in \mathbb{R}^{n_1\times  \cdots \times  n_d}$.
Here, both ${\boldsymbol{\mathcal{L}}}$ and ${\boldsymbol{\mathcal{E}}}$ are of arbitrary magnitude.
We do not know the T-SVD rank of ${\boldsymbol{\mathcal{L}}}$.
Moreover,
 we have no idea about the locations of the nonzero entries of ${\boldsymbol{\mathcal{E}}}$,
not even how many there are.
Then, the goal of the RHTC problem  is to
achieve 
the
reconstruction (either exactly  or approximately)
of low-rank component
${\boldsymbol{\mathcal{L}}}$
from an observed subset of 
corrupted tensor 
${\boldsymbol{\mathcal{M}}}={\boldsymbol{\mathcal{L}}}+{\boldsymbol{\mathcal{E}}}$.
Mathematically, 
the  proposed RHTC model can be formulated
as follows:
\begin{equation} \label{orin_nonconvex}
\min_{{\boldsymbol{\mathcal{L}}},{\boldsymbol{\mathcal{E}}}}
\|  {\boldsymbol{\mathcal{L}}} \|_{{ {\boldsymbol{\mathcal{W}}}_{1}, {\boldsymbol{\mathcal{S}}}_{p} }}  ^{p}
 + \lambda\|{\boldsymbol{\mathcal{E}}}\|_{{\boldsymbol{\mathcal{W}}}_{2},\ell_q}^{q},
\boldsymbol{\bm{P}}_{{{\Omega}}}({\boldsymbol{\mathcal{L}}}+{\boldsymbol{\mathcal{E}}})=
\boldsymbol{\bm{P}}_{{{\Omega}}}({\boldsymbol{\mathcal{M}}}),
\end{equation}
where $0<p,q<1$,
$
\lambda > 0
$ is the penalty parameter,
$\|  {\boldsymbol{\mathcal{L}}} \|_{{ {\boldsymbol{\mathcal{W}}}_{1}, {\boldsymbol{\mathcal{S}}}_{p} }} $
denotes the HWTSN of  ${\boldsymbol{\mathcal{L}}}$ while
$\|{\boldsymbol{\mathcal{E}}}\|_{{\boldsymbol{\mathcal{W}}}_{2},\ell_q}$
represents  the weighted $\ell_q$-norm  of  ${\boldsymbol{\mathcal{E}}}$,
and ${\boldsymbol{\mathcal{W}}}_{1}$ and ${\boldsymbol{\mathcal{W}}}_{2}$ are
the weight tensors, which
 will be updated automatically in the subsequent ADMM optimization,
 \text{see  \ref{optalg} for details.} 

\begin{Remark}
In the 
model (\ref{orin_nonconvex}),
the HWTSN not only gives better approximation
to the original low-rank assumption, but also considers
the importance of different 
singular components.
 Comparing with the previous  
 regularizers, e.g., HWTNN and
 HTNN/HTSN (treat the different 
 rank components equally),
 %
 the proposed one is  tighter and more  feasible. 
 Besides, the  weighted $\ell_q$-norm
 has a superior potential to be sparsity-promoting 
 in comparison with the $\ell_1$-norm and $\ell_q$-norm.
Therefore, the joint  HWTSN and weighted $\ell_q$-norm
enables the underlying 
low-rank structure 
in the observed tensor ${\boldsymbol{\mathcal{M}}}$
to be well captured, and  the 
robustness against noise/outliers  to  be well 
enhanced.
The  proposed two
nonconvex 
regularizers
mainly
involve several key
ingredients: 
1) flexible
  linear transforms $\mathfrak{L}$;
 2)
 adjustable parameters $p$ and $q$;
3) 
automatically updated
weight tensors ${\boldsymbol{\mathcal{W}}}_{1}$ and ${\boldsymbol{\mathcal{W}}}_{2}$.
Their various combinations can degenerate to  
many existing 
RLRTC  models.
\end{Remark}

\subsection{\textbf{HWTSN Minimization Problem}}
In this subsection, we mainly present the solution method of HWTSN minimization problem, that is, the method of solving
\begin{align}\label{wtsn_prox1}
\arg\min_{ {\boldsymbol{\mathcal{X}}}}
\tau \| {\boldsymbol{\mathcal{X}}}\|_{ { {\boldsymbol{\mathcal{W}}}, {\boldsymbol{\mathcal{S}}}_{p} }}^{p}
+
\frac{1}{2}\|{\boldsymbol{\mathcal{X}}}-{\boldsymbol{\mathcal{Z}}}\|^{2}_{\mathnormal{F}}.
\end{align}
%
%
%
%
\noindent
Before  providing 
the solution to problem (\ref{wtsn_prox1}),
we first introduce the key lemma and definition.

\begin{Lemma} \label{lemm_lp}
\cite{zuo2013generalized}
For 
the 
given $p$ ($0 < p < 1$) and $w > 0$, the 
optimal solution of the following
optimization problem
\begin{equation}\label{equ_gst1}
\min_{x} w
|x|^{p}
 + \frac{1}{2} {(x-s)}^2
\end{equation}
is given 
by the
generalized soft-thresholding (GST)
 operator: 
\begin{align*}\label{lq_shrink}
\hat{x}=\operatorname{ {{GST}} }\big(s,w,p\big)
=
\begin{cases}
0,\;\;\;\;\;\;\;\;\;\;\;\;\;\;\;\;\;\;\;\;\;\;\;\;\;\;\;\text{if}\;|s|\leq\delta,\\
\operatorname{sign}(s) {\hat{\alpha}}^{*} \;\;\;\;\;\;\;\;\;\;\;\;\;\;\;\;\text{if}\;|s|>\delta,
\end{cases}
\end{align*}
where
$\delta={[2w(1-p)]}^{\frac{1}{2-p}}+w p {[2w(1-p)]}^{\frac{p-1}{2-p}}$ is a threshold value,
$\operatorname{sign}(s)$ denotes the signum function,
and
${\hat{\alpha}}^{*}$  can be obtained by solving 
 ${\alpha}+w p {\alpha}^{p-1}-|s|=0 \; (\alpha>0)$.
\end{Lemma}

\begin{Definition}\label{gtsvt}
\textbf{(GTSVT operator)}
Let
${\boldsymbol{\mathcal{A}}}={\boldsymbol{\mathcal{U}}}{*}_{\mathfrak{L}}{\boldsymbol{\mathcal{S}}}{*}_{\mathfrak{L}}
{\boldsymbol{\mathcal{V}}}^{\mit{T}}$
be the  \text{T-SVD} of
 ${\boldsymbol{\mathcal{A}}} \in \mathbb{R}^{n_1\times  \cdots \times n_d}$.
For any $\tau>0$, $0<p<1$, then
the generalized Tensor Singular Value Thresholding (GTSVT) operator of ${\boldsymbol{\mathcal{A}}}$   is defined as follows
\begin{equation}\label{tsvt_operator}
{\boldsymbol{\mathcal{D}}}_{{\boldsymbol{\mathcal{W}}},p,\tau}({\boldsymbol{\mathcal{A}}})
={\boldsymbol{\mathcal{U}}}{*}_{\mathfrak{L}}
{\boldsymbol{\mathcal{S}}}_{{\boldsymbol{\mathcal{W}}},p,\tau}
{*}_{\mathfrak{L}}{\boldsymbol{\mathcal{V}}}^{\mit{T}},
\end{equation}
where
${\boldsymbol{\mathcal{S}}}_{{\boldsymbol{\mathcal{W}}},p,\tau}={\mathfrak{L}^{-1}} \big (GST(
{{\mathfrak{L}}({\boldsymbol{\mathcal{S}}})},
\tau {\boldsymbol{\mathcal{W}}},p)\big)$,
 ${\boldsymbol{\mathcal{W}}} \in \mathbb{R}^{n_1\times  \cdots \times n_d}$ is 
 the weight parameter
composed of an
order-$d$ f-diagonal tensor,
and  $\operatorname{GST}$ 
denotes the element-wise shrinkage
operator.
\end{Definition}

\begin{Remark}
Since the larger singular values usually carry more important information than the smaller ones,
 the GTSVT operator  
 requires to
 satisfy:
 the larger singular values
 in the transform domain
 should be shrunk less, while the
smaller ones should be shrunk 
more.
  In other words,
 the  weights are selected 
 inversely 
 to the singular values in the transform domain.
Thus,
 the original components corresponding to the larger singular values will be less affected.
  %
  %
 The GTSVT operator is more flexible than the T-SVT operator proposed in \cite{qin2022low} (shrinks all singular values with the same threshold)
 and  the WTSVT
 operator proposed in \cite{qin2021robust}, 
and provides more degree of freedom for the approximation to the original problem.
\end{Remark}

\begin{Theorem}\label{theorem_svt}
  Let $\mathfrak{L}$ be any invertible linear transform in (\ref{trans}) and it satisfies (\ref{orth}),
$m=\min{(n_1,n_2)}$.
For any $\tau>0$ and ${\boldsymbol{\mathcal{Z}}} \in \mathbb{R}^{n_1\times  \cdots \times n_d}$,
if the weight parameter satisfies
$$
0\leq {\boldsymbol{\mathcal{W}}}^{<j>}(1,1) \leq
\cdots
\leq {\boldsymbol{\mathcal{W}}}^{<j>}(m,m),\;
\forall
j \in 
[n_3  \cdots n_d],
$$
then
the GTSVT operator    (\ref{tsvt_operator}) obeys 
\begin{align}\label{eq_svt}
{\boldsymbol{\mathcal{D}}}_{{\boldsymbol{\mathcal{W}}},p,\tau}({\boldsymbol{\mathcal{Z}}})= 
\arg\min_{ {\boldsymbol{\mathcal{X}}}}
\tau \| {\boldsymbol{\mathcal{X}}}\|_{ { {\boldsymbol{\mathcal{W}}}, {\boldsymbol{\mathcal{S}}}_{p} }}^{p}
+
\frac{1}{2}\|{\boldsymbol{\mathcal{X}}}-{\boldsymbol{\mathcal{Z}}}\|^{2}_{\mathnormal{F}}.
\end{align}
\end{Theorem}

\begin{proposition}\label{prop11}
Let ${{\boldsymbol{\mathcal{A}}}}= {{\boldsymbol{\mathcal{Q}}}}
{*}_{\mathfrak{L}}  {\boldsymbol{\mathcal{B}}} \in \mathbb{R}^{n_1 \times n_2 \times \cdots\times n_d}$, where
${{\boldsymbol{\mathcal{Q}}}}
\in \mathbb{R}^{n_1 \times k\times  \cdots\times n_d}$ is partially orthogonal
 and
${{\boldsymbol{\mathcal{B}}}}
\in \mathbb{R}^{k \times n_2\times  \cdots\times n_d}$.
Then, we have
$$
{\boldsymbol{\mathcal{D}}}_{{\boldsymbol{\mathcal{W}}},p,\tau}({\boldsymbol{\mathcal{A}}})={{\boldsymbol{\mathcal{Q}}}}
{*}_{\mathfrak{L}} {\boldsymbol{\mathcal{D}}}_{{\boldsymbol{\mathcal{W}}},p,\tau}({\boldsymbol{\mathcal{B}}}).
$$
%
\end{proposition}

\begin{proposition}\label{ppppp6666}
Let ${{\boldsymbol{\mathcal{A}}}}= {{\boldsymbol{\mathcal{Q}}}_{1}}
{*}_{\mathfrak{L}}  {\boldsymbol{\mathcal{B}}} {*}_{\mathfrak{L}}
{{\boldsymbol{\mathcal{Q}}}_{2}^{\mit{T}}}
\in \mathbb{R}^{n_1 \times n_2 \times \cdots\times n_d}$, where
${{\boldsymbol{\mathcal{Q}}}_{1}}
\in \mathbb{R}^{n_1 \times k\times  \cdots\times n_d}$ and ${{\boldsymbol{\mathcal{Q}}}_{2}}
\in \mathbb{R}^{n_2 \times k\times  \cdots\times n_d}$ are partially orthogonal,
${{\boldsymbol{\mathcal{B}}}}
\in \mathbb{R}^{k \times k \times  \cdots\times n_d}$.
Then, we have
\begin{equation*}
{\boldsymbol{\mathcal{D}}}_{{\boldsymbol{\mathcal{W}}},p,\tau}({\boldsymbol{\mathcal{A}}})={{\boldsymbol{\mathcal{Q}}}_{1}}
{*}_{\mathfrak{L}} {\boldsymbol{\mathcal{D}}}_{{\boldsymbol{\mathcal{W}}},p,\tau}({\boldsymbol{\mathcal{B}}})
{*}_{\mathfrak{L}}
{{\boldsymbol{\mathcal{Q}}}_{2}^{\mit{T}}}.
\end{equation*}
\end{proposition}

\begin{Remark}
From the Definition \ref{gtsvt} and Theorem \ref{theorem_svt},
 we can find that the major computational bottleneck of HWTSN minimization problem (\ref{wtsn_prox1})
 is to execute 
 the GTSVT operator  involving T-SVD
 multiple times.
 Based on the previous
 Proposition \ref{prop11},\ref{ppppp6666},  
we can 
avoid expensive computation
by instead calculating GTSVT on a smaller tensor ${{\boldsymbol{\mathcal{B}}}}$.
In other words, we can efficiently calculate GTSVT operator according to the following two steps:
1) compute two orthogonal subspace basis 
   tensor $\bm{\mathcal{Q}}_{1}, \bm{\mathcal{Q}}_{2}$ via random
   projection techniques;
   2) perform the GTSVT operator on a smaller tensor ${{\boldsymbol{\mathcal{B}}}}$.
 The computational procedure of GTSVT is shown in  Algorithm  \ref{RTSVD-RANK12}.
 \textbf{What is particularly noteworthy is that the Algorithm  \ref{RTSVD-RANK12}
is highly parallelizable 
because  the operations across frontal slices can be readily distributed across different 
processors.
Therefore, additional computational gains can be achieved in virtue of the 
parallel computing framework.
}
\end{Remark}

\begin{algorithm}[!htbp]
\setstretch{0.0} 
\caption{
GST
algorithm
                \cite{zuo2013generalized}. 
}\label{gst}
  \KwIn{
 $s,w,p,J=3$ or $4$.}
  {\color{black}\KwOut{
  $\operatorname{ {{GST}} }\big(s,w,p\big)$.}}

${{\delta}}^{GST}_{p}(w)={[2w(1-p)]}^{\frac{1}{2-p}} + wp {{[2w(1-p)]}^{\frac{p-1}{2-p}}}$\;
 \eIf{$|s|\leq  {{\delta}}^{GST}_{p}(w)$}
{
$\operatorname{ {{GST}} }\big(s,w,p\big)=0$
\;
}
{$j=0, x^{(j)}=|s|$\;
   \For{$ j=0,1,\cdots, J$}
     {
     $x^{(j+1)}=|s|-wp(x^{(j)})^{p-1}$;\\
     $j=j+1$;
    }
      $
      \operatorname{ {{GST}} }\big(s,w,p\big)
      =\operatorname{sgn}(s)x^{(j)}$;
}
\end{algorithm}

\vspace{-0.6cm}

 \begin{algorithm}[!htbp]
\setstretch{0.0}
     \caption{
     High-Order GTSVT,
     ${\boldsymbol{\mathcal{D}}}_{{\boldsymbol{\mathcal{W}}},p,\tau}({\boldsymbol{\mathcal{A}}},\mathfrak{L})$.
     }
     \label{RTSVD-RANK12}
      \KwIn{$\bm{\mathcal{A}}\in\mathbb{R}^{n_1\times \cdots\times n_d}$,
       transform:  $\mathfrak{L}$, target T-SVD Rank: $k$,
        weight tensor: 
        ${\boldsymbol{\mathcal{W}}} \in \mathbb{R}^{n_{1}\times  \cdots \times n_d}$,
          block size: $b$,
$0<p<1$,
 $\tau> 0$,
 power iteration: $t$.
       }

  	Let  $\hat{l}$   be  a number slightly larger than $k$ 
      and generate
      a Gaussian random tensor $\bm{\mathcal{G}}\in\mathbb{R}^{n_2\times \hat{l}\times n_3 \times \cdots \times  n_d}$\;

        Compute the results of $\mathfrak{L}$ on $\boldsymbol{\mathcal{A} }$ and $\boldsymbol{\mathcal{G} }$, i.e.,
       $\mathfrak{L}(\boldsymbol{\mathcal{A}}), \mathfrak{L}(\boldsymbol{\mathcal{G}})$\;

{
 \For{$v=1,2,\cdots, n_3 n_4 \cdots n_d$}
      {

\If{ \text{utilize the unblocked randomized technique} 
}
{
    Execute
             Lines $\textbf{4-12}$ of Algorithm \ref{transform-version-fp11}
             to obtain $ {\mathfrak{L}}({\boldsymbol{\mathcal{U}}})^{<v>}$,
             $ {\mathfrak{L}}({\boldsymbol{\mathcal{S}}})^{<v>}$, and $ {\mathfrak{L}}({\boldsymbol{\mathcal{V}}})^{<v>}$
              \;
}

\ElseIf{ \text{utilize the blocked randomized technique} 
}
{
    Execute
             Lines $\textbf{4-24}$ of Algorithm \ref{transform-version-fp}
             to obtain $ {\mathfrak{L}}({\boldsymbol{\mathcal{U}}})^{<v>}$,
             $ {\mathfrak{L}}({\boldsymbol{\mathcal{S}}})^{<v>}$, and $ {\mathfrak{L}}({\boldsymbol{\mathcal{V}}})^{<v>}$
              \;
}
\ElseIf{
\text{not utilize the randomized technique}
}
{
$[{\mathfrak{L}}({\boldsymbol{\mathcal{U}}})^{<v>},{\mathfrak{L}}({\boldsymbol{\mathcal{S}}})^{<v>},{\mathfrak{L}}({\boldsymbol{\mathcal{V}}})^{<v>}]= \textrm{svd} \big({\boldsymbol{\mathcal{A}}}_{\mathfrak{L}}^{<v>} \big) $
              \;
}



$
\hat{\bm{S}}
= 
\operatorname{ {{GST}} }\big\{
{\operatorname{diag} ( 
\mathfrak{L}(\boldsymbol{\mathcal{S}})^{<v>}
)},
\tau\cdot {\operatorname{diag} (  {{\boldsymbol{\mathcal{W}}}}^{<v>} )},
p
\big\}
$;\\

${  \mathfrak{L}(\boldsymbol{\mathcal{C}})   }^{<v>}= 
{\mathfrak{L}(\boldsymbol{\mathcal{U}})}^{<v>}\cdot
%
\operatorname{diag} ( \hat{\bm{S}})
\cdot
{   (\mathfrak{L}(\boldsymbol{\mathcal{V}})^{<v>})   }  ^{\mit{T}}$
;\\
}
}

{

}

    		 {\color{black}\KwOut{
${\boldsymbol{\mathcal{D}}}_{{\boldsymbol{\mathcal{W}}},p,\tau}({\boldsymbol{\mathcal{A}}},\mathfrak{L})
\leftarrow {\mathfrak{L}}^{-1}(
      {\mathfrak{L}}({\boldsymbol{\mathcal{C}}}))
$
.}}

    \end{algorithm}

\vspace{-0.3cm}

\subsection{\textbf{Optimization Algorithm}}\label{optalg}
In this subsection,
the 
ADMM
framework  \cite{boyd2011distributed
}
is adopted to solve the proposed  model (\ref{orin_nonconvex}).
The nonconvex model (\ref{orin_nonconvex})  can be equivalently reformulated as follows:
\begin{equation}\label{equ_nonconvex}
\min_{{\boldsymbol{\mathcal{L}}},{\boldsymbol{\mathcal{E}}}}
\|  {\boldsymbol{\mathcal{L}}} \|_{{ {\boldsymbol{\mathcal{W}}}_{1}, {\boldsymbol{\mathcal{S}}}_{p} }}  ^{p}
 + \lambda\|\boldsymbol{\bm{P}}_{{{\Omega}}}({\boldsymbol{\mathcal{E}}})\|_{{\boldsymbol{\mathcal{W}}}_{2},\ell_q}^{q},
\; \;
\text{s.t.} \;\;
{\boldsymbol{\mathcal{L}}}+{\boldsymbol{\mathcal{E}}}=
{\boldsymbol{\mathcal{M}}}.
\end{equation}
The  augmented Lagrangian function of (\ref{equ_nonconvex}) is
\begin{align}
\mathcal{F}(
{\boldsymbol{\mathcal{L}}},{\boldsymbol{\mathcal{E}}},{\boldsymbol{\mathcal{Y}}},\beta)=
\|  {\boldsymbol{\mathcal{L}}} \|_{{ {\boldsymbol{\mathcal{W}}}_{1}, {\boldsymbol{\mathcal{S}}}_{p} }}  ^{p}
 + \lambda\|\boldsymbol{\bm{P}}_{{{\Omega}}}({\boldsymbol{\mathcal{E}}})\|_{{\boldsymbol{\mathcal{W}}}_{2},\ell_q}^{q}+
\notag \\
\label{wtsn_admm}
\langle {\boldsymbol{\mathcal{Y}}}, {
{\boldsymbol{\mathcal{L}}}+{\boldsymbol{\mathcal{E}}}-\boldsymbol{\mathcal{M}}} \rangle +
{\beta}/{2}
\|  {\boldsymbol{\mathcal{L}}}+{\boldsymbol{\mathcal{E}}} -{\boldsymbol{\mathcal{M}}}\|^{2}_{{{\mathnormal{F}}}},
\end{align}
where ${\boldsymbol{\mathcal{Y}}}$ is the dual variable and $\beta$ is the regularization  parameter.
   The ADMM framework alternately updates each optimization variable until convergence.
The iteration template of the ADMM at the ($k+1$)-th iteration is described as follows:
\begin{align}
\label{L_hat}
{\boldsymbol{\mathcal{L}}}^{k+1}
& = \arg \min_{\boldsymbol{\mathcal{L}}}  \big\{
\mathcal{F}({\boldsymbol{\mathcal{L}}},{\boldsymbol{\mathcal{E}}}^{k},{\boldsymbol{\mathcal{Y}}}^{k},
\beta^k) \big\}, \\
\label{E_hat}
{\boldsymbol{\mathcal{E}}}^{k+1}
& = \arg \min_{\boldsymbol{\mathcal{E}}  }     \;\big\{
\mathcal{F}({\boldsymbol{\mathcal{L}}}^{k+1},{\boldsymbol{\mathcal{E}}},{\boldsymbol{\mathcal{Y}}}^{k},
\beta^k)\big\}, \\
\label{Y_hat}
{\boldsymbol{\mathcal{Y}}}^{k+1}
&= {\boldsymbol{\mathcal{Y}}}^{k}+\beta^{k}({\boldsymbol{\mathcal{L}}}^{k+1}+{\boldsymbol{\mathcal{E}}}^{k+1}-{\boldsymbol{\mathcal{M}}}),\\
\label{vartheta}
\beta^{k+1}
&= \min \left( \beta^{\operatorname{max}},\vartheta \beta^k \right),
\end{align}
\noindent where $\vartheta >1$  is a control constant. 
Now we solve the 
subproblem (\ref{L_hat})  and (\ref{E_hat})
 explicitly in the ADMM, respectively.

\noindent {\textbf{Update ${\boldsymbol{\mathcal{L}}^{k+1}}$ (low-rank component)}}
The  optimization subproblem  (\ref{L_hat}) concerning 
${\boldsymbol{\mathcal{L}}^{k+1}}$ can  be written as
\begin{align}\label{L_prox}
%
\min_{\boldsymbol{\mathcal{L}}}
\big\|  {\boldsymbol{\mathcal{L}}} \big\|_{ { {  { {\boldsymbol{\mathcal{W}}}_{1}}^{k}}, 
{\boldsymbol{\mathcal{S}}}_{p}}}^{p}
+
{\beta^k}/{2}
\big\|{\boldsymbol{\mathcal{L}}}-{\boldsymbol{\mathcal{M}}}+{\boldsymbol{\mathcal{E}}}^{k}
+
{{\boldsymbol{\mathcal{Y}}}^{k}}/{\beta^k}
\big\|^2_{\mathnormal{F}}.
\end{align}
Let ${\boldsymbol{\mathcal{G}}}^{k}=
{\boldsymbol{\mathcal{M}}}-{\boldsymbol{\mathcal{E}}}^{k}
-
{{\boldsymbol{\mathcal{Y}}}^{k}}/{\beta^k}$.
Using  the 
GTSVT algorithm  
that incorporates the randomized schemes, 
the  subproblem (\ref{L_prox})
can be efficiently solved, 
i.e.,
${\boldsymbol{\mathcal{L}}}^{k+1}=
{\boldsymbol{\mathcal{D}}}_{
{{\boldsymbol{\mathcal{W}}}_{1}}^{k},p,\frac{1}{\beta^k}
}({\boldsymbol{\mathcal{G}}}^{k},\mathfrak{L})$.
%

\begin{Remark}
(\textbf{\textbf{Update \texorpdfstring{${{\boldsymbol{\mathcal{W}}}_{1}}$} \; via  reweighting strategy}})
The weight tensor ${\boldsymbol{\mathcal{W}}_{1}}$
can be  adaptively tuned at each iteration,
and its formula in  the $k$-th iteration is given by
\begin{equation*}
{({\boldsymbol{\mathcal{W}}_{1}}^{k})} ^{<j>}(i,i)= 
{\frac{c_{1}} {
{
 {({  {\boldsymbol{\mathcal{K}}}^{k}})} ^{<j>}(i,i)
+\epsilon_{1}
}}} \cdot\frac{1}{\beta^{k}}
,\;
\end{equation*}
where $j \in \{1,\cdots,n_3\cdots n_d\}$, $i\in \{1,\cdots,\min( n_1,n_2)\}$, $c_{1} > 0$ is a  constant,
$\epsilon_{1}$ is a small non-negative  constant  to avoid division by zero,
and the entries  on the diagonal of
${({{  {\boldsymbol{\mathcal{K}}} }^{k}})}^{<j>}$
represent
the singular values of
${
{
{\mathfrak{L}}
(
{\boldsymbol{\mathcal{G}}} ^{k}
)
}}
^{<j>}$.
In such a reweighted technique,
the sparsity performance is enhanced after each iteration
and
the 
updated ${{\boldsymbol{\mathcal{W}}}_{1}}^{k}$
satisfy:
\begin{equation*}
{({\boldsymbol{\mathcal{W}}_{1}}^{k})} ^{<j>}(m,m) 
 \geq 
 {({\boldsymbol{\mathcal{W}}_{1}}^{k})} ^{<j>} (n,n) 
 \geq 0,
 \; \forall m\geq n.
 \end{equation*}
\end{Remark}
%
%
\noindent {\textbf{Update ${\boldsymbol{\mathcal{E}}^{k+1}}$ (noise/outliers component)}}
The optimization subproblem (\ref{E_hat})
with respect to $\boldsymbol{\mathcal{E}}^{k+1}$ 
can  be written as
\begin{align*}
\min_{
\boldsymbol{\mathcal{E}}} 
\lambda\|\boldsymbol{\bm{P}}_{{{\Omega}}}({\boldsymbol{\mathcal{E}}})\|_{{{\boldsymbol{\mathcal{W}}}_{2}}^{k},\ell_q}^{q}
 +
{\beta^k}/{2}
\big\|
{\boldsymbol{\mathcal{E}}}-{\boldsymbol{\mathcal{M}}}+{\boldsymbol{\mathcal{L}}}^{k+1}
+
{{\boldsymbol{\mathcal{Y}}}^{k}} / {\beta^k}
\big\|^2_{\mathnormal{F}}.
\end{align*}
%
Let ${\boldsymbol{\mathcal{H}}}^{k}=
{\boldsymbol{\mathcal{M}}}-{\boldsymbol{\mathcal{L}}}^{k+1}
-
{{\boldsymbol{\mathcal{Y}}}^{k}}/{\beta^k}$.
The above problem 
can be solved by the following two subproblems
 with respect to $\boldsymbol{\bm{P}}_{{{\Omega}}}({\boldsymbol{\mathcal{E}}}^{k+1})$ and $\boldsymbol{\bm{P}}_{{{\Omega}}_{\bot}}({\boldsymbol{\mathcal{E}}}^{k+1})$
, respectively.
Note that the weight tensor ${\boldsymbol{\mathcal{W}}_{2}}$ is updated at each iteration,
 and its form at the $k$-th  iteration is set as follows:
\begin{equation*}
{({\boldsymbol{\mathcal{W}}_{2}}^{k})}(i_1,\cdots,i_d)= 
{\frac{c_2} {
{
\big |{({  {\boldsymbol{\mathcal{H}}}^{k}})} (i_1,\cdots,i_d)\big|
+\epsilon_2
}}} \cdot\frac{1}{\beta^{k}}
,\;
\end{equation*}
in which $c_{2} > 0$ is a  constant, $\epsilon_2>0$ is a small constant  to avoid division by zero.

%
%
\noindent
 \textbf{
 Regarding $\boldsymbol{\bm{P}}_{{{\Omega}}}({\boldsymbol{\mathcal{E}}}^{k+1})$
 }:
the optimization subproblem
 with respect to $\boldsymbol{\bm{P}}_{{{\Omega}}}({\boldsymbol{\mathcal{E}}}^{k+1})$
 is formulated as following
%
\begin{align}\label{E_prox_ome}
\min_{
\boldsymbol{\bm{P}}_{{{\Omega}}}(\boldsymbol{\mathcal{E}})
} 
\lambda\|\boldsymbol{\bm{P}}_{{{\Omega}}}({\boldsymbol{\mathcal{E}}})\|_{{{\boldsymbol{\mathcal{W}}}_{2}}^{k},\ell_q}^{q}
 +
{\beta^k}/{2}
\big\| \boldsymbol{\bm{P}}_{{{\Omega}}}(
{\boldsymbol{\mathcal{E}}}-{\boldsymbol{\mathcal{H}}}^{k}
)
\big\|^2_{\mathnormal{F}}.
\end{align}
%
\noindent
The  closed-form solution for  subproblem
(\ref{E_prox_ome}) can be computed by generalized element-wise shrinkage operator, i.e.,
\begin{align*}
\boldsymbol{\bm{P}}_{{{\Omega}}}( {\boldsymbol{\mathcal{E}}}^{k+1})=
\operatorname{{GST}}
(
\boldsymbol{\bm{P}}_{{{\Omega}}}(
{\boldsymbol{\mathcal{H}}}^{k}),   
{\lambda}
\cdot
({\beta^{k}} )^{-1}
\cdot \boldsymbol{\bm{P}}_{{{\Omega}}}(
{{\boldsymbol{\mathcal{W}}}_{2}}^{k})
,q
).
%
\end{align*}
\noindent
  \textbf{
   Regarding
$\boldsymbol{\bm{P}}_{{{\Omega}}_{\bot}}({\boldsymbol{\mathcal{E}}}^{k+1})$}:
the optimization subproblem
 with respect to $\boldsymbol{\bm{P}}_{{{\Omega}}_{\bot}}({\boldsymbol{\mathcal{E}}}^{k+1})$
 is formulated as following
%
 \begin{align}\label{E_prox_ome1}
\boldsymbol{\mathrm{P}}_{{{\Omega}}_{\bot}}({\boldsymbol{\mathcal{E}}}^{k+1})=
\min_{
{\boldsymbol{\bm{P}}}_{{{\Omega}}_{\bot}} (\boldsymbol{\mathcal{E}})
}
{\beta^k/2}
\big\| \boldsymbol{\bm{P}}_{{{\Omega}}_{\bot}}(
{\boldsymbol{\mathcal{E}}}-{\boldsymbol{\mathcal{H}}}^{k}
)
\big\|^2_{\mathnormal{F}}.
\end{align}
\noindent
The  closed-form solution for  subproblem
(\ref{E_prox_ome1})  can be obtained through the standard least square regression method.
%
%
%
\begin{algorithm}[!htbp]
\setstretch{0.0} 
\caption{
Solve the proposed model (\ref{orin_nonconvex}) by ADMM.
}\label{algorithm1}

 \KwIn{
$\boldsymbol{\bm{P}}_{{{\Omega}}}({\boldsymbol{\mathcal{M}}}) \in \mathbb{R}^{n_1 \times \cdots \times n_d}$,
$\mathfrak{L}$,
 $\lambda$,
  target T-SVD Rank: $k$,
          block size: $b$,
 power iteration: $t$,
  $0<p,q<1$,
  $\tau> 0$,
  $c_1, c_2, {\epsilon}_1, {\epsilon}_2$.
  }

 \textbf{Initialize:} ${\boldsymbol{\mathcal{L}}}^{0}={\boldsymbol{\mathcal{E}}}^{0}={\boldsymbol{\mathcal{Y}}}^{0}=\boldsymbol{0}$,
 $\vartheta$,
$\beta^0$,
$\beta^{\max}$,
$\varpi$,
$k=0$\;

\While{\text{not converged}}
{
 Update ${\boldsymbol{\mathcal{L}}}^{k+1}$ by computing 
 Algorithm \ref{RTSVD-RANK12};
 \\
%
  Update $\boldsymbol{\bm{P}}_{{{\Omega}}}({\boldsymbol{\mathcal{E}}}^{k+1})$ by computing (\ref{E_prox_ome});\\
     Update $\boldsymbol{\bm{P}}_{{{\Omega}}_{\bot}}({\boldsymbol{\mathcal{E}}}^{k+1})$ by computing (\ref{E_prox_ome1});\\

 Update   ${\boldsymbol{\mathcal{Y}}}^{k+1}$ by computing (\ref{Y_hat});\\

 Update $\beta^{k+1}$ by computing (\ref{vartheta});\\

 Check the convergence conditions
\begin{align*}
%
&\|{\boldsymbol{\mathcal{L}}}^{k+1}-{\boldsymbol{\mathcal{L}}}^{k}\|{_{\infty}}
 \leq \varpi,
\;
\|{\boldsymbol{\mathcal{E}}}^{k+1}-{\boldsymbol{\mathcal{E}}}^{k}\|{_{\infty}}  \leq \varpi,\\
&\|
 \boldsymbol{\boldsymbol{\mathcal{L}}}^{k+1}+
{\boldsymbol{\mathcal{E}}}^{k+1}- {\boldsymbol{\mathcal{M}}}\|{_{\infty}}
   \leq \varpi.
\end{align*}
}
{\color{black}\KwOut{
  ${\boldsymbol{\mathcal{L}}}  \in \mathbb{R}^{n_1 \times \cdots \times n_d}$.
  }}
\end{algorithm}
\vspace{-0.5cm}

\subsection{\textbf{Convergence Analysis}}
In this subsection,
we provide a theoretical guarantees for the convergence of the proposed
Algorithm \ref{algorithm1},
the detailed proof of which is given in the supplementary material.

\begin{Theorem}\label{conver}
 Let $\mathfrak{L}$ be any invertible linear transform in (\ref{trans}) and it satisfies (\ref{orth}),
$m=\min{(n_1,n_2)}$.
 If the diagonal elements of all matrix frontal slices 
 on weighted tensor ${{\boldsymbol{\mathcal{W}}}_{1}}^{k}$ are sorted in a non-descending
order, i.e.,
$$
 {({\boldsymbol{\mathcal{W}}_{1}}^{k})} ^{<j>}(1,1)
\cdots
\leq {({\boldsymbol{\mathcal{W}}_{1}}^{k})} ^{<j>}(m,m),\;
\forall
j \in 
[n_3  \cdots n_d],
$$
then the sequences $\{{\boldsymbol{\mathcal{L}}}^{k+1}\}$, $\{{\boldsymbol{\mathcal{E}}}^{k+1}\}$ and $\{{\boldsymbol{\mathcal{Y}}}^{k+1}\}$
generated by Algorithm \ref{algorithm1} satisfy:
\begin{align*}
&1)\lim_{k\rightarrow \infty} {\|{\boldsymbol{\mathcal{L}}}^{k+1}-{\boldsymbol{\mathcal{L}}}^{k}\|{_{\mathnormal{F}}}}=0;
\;\;
2)\lim_{k\rightarrow \infty} {\|{\boldsymbol{\mathcal{E}}}^{k+1}-{\boldsymbol{\mathcal{E}}}^{k}\|{_{\mathnormal{F}}}}=0;\\
&3)\lim_{k\rightarrow \infty}{
\|{\boldsymbol{\mathcal{M}}}-{\boldsymbol{\mathcal{L}}}^{k+1}-{\boldsymbol{\mathcal{E}}}^{k+1}\|{_{\mathnormal{F}}}}=0.
\end{align*}
\end{Theorem}

\vspace{-0.45cm}

\subsection{\textbf{Complexity Analysis}}
Given an input tensor
$\boldsymbol{\bm{P}}_{{{\Omega}}}({\boldsymbol{\mathcal{M}}}) \in \mathbb{R}^{n_1 \times \cdots \times n_d}$,
we analyze the 
per-iteration
complexity of Algorithm \ref{algorithm1} with/without randomized techniques.
The  per-iteration  of Algorithm \ref{algorithm1} needs  to update ${\boldsymbol{\mathcal{L}}}$,
$\boldsymbol{\bm{P}}_{{{\Omega}}}({\boldsymbol{\mathcal{E}}})$,
$\boldsymbol{\bm{P}}_{{{\Omega}}_{\bot}}({\boldsymbol{\mathcal{E}}})$,
${\boldsymbol{\mathcal{Y}}}$, respectively.
Upadating
$\boldsymbol{\bm{P}}_{{{\Omega}}}({\boldsymbol{\mathcal{E}}})$ requires to perform GST operation
with a complexity of
$\boldsymbol{\mathcal{O}}\big( |\Omega| \big)$, 
where $|\Omega|$ denotes 
the cardinality of $\Omega$.
$\boldsymbol{\bm{P}}_{{{\Omega}}_{\bot}}({\boldsymbol{\mathcal{E}}})$ and
${\boldsymbol{\mathcal{Y}}}$ can be updated by a low consumed algebraic computation.
The update of  ${\boldsymbol{\mathcal{L}}}$ mainly involves
matrix-matrix product,
economic QR/SVD decomposition,
       linear transforms $\mathfrak{L}(\cdot)$
      and  its inverse operator  $ \mathfrak{L}^{-1}(\cdot)$. Specifically,
%
for any  invertible linear transforms $\mathfrak{L}$, 
 the per-iteration complexity of  ${\boldsymbol{\mathcal{L}}}$
is 
\begin{enumerate}
 \item
$\boldsymbol{\mathcal{O}}\big(\prod_{i=1}^{d}{n_i} \cdot \sum_{j=3}^{d}{n_j} + \hat{l}   \cdot \prod_{k=1}^{d}{n_k} \big)$,\;
with randomized technique;

\item
$\boldsymbol{\mathcal{O}}\big(\prod_{i=1}^{d}{n_i} \cdot \sum_{j=3}^{d}{n_j} + \min \{n_1,n_2\} \cdot \prod_{k=1}^{d}{n_k}\big )$,\;
without randomized technique.
\end{enumerate}
%
For some special invertible linear transforms $\mathfrak{L}$, e.g., FFT,
the per-iteration complexity of  ${\boldsymbol{\mathcal{L}}}$
is 
\begin{enumerate}
  \item $\boldsymbol{\mathcal{O}}\big(\prod_{i=1}^{d}{n_i} \cdot  \sum_{j=3}^{d} \log({n_j})+ \hat{l} \cdot \prod_{k=1}^{d}{n_k}\big)$,\;
  with randomized technique;
  \item $\boldsymbol{\mathcal{O}}\big(\prod_{i=1}^{d}{n_i} \cdot  \sum_{j=3}^{d} \log({n_j})+ 
  \min \{n_1,n_2\} \cdot \prod_{k=1}^{d}{n_k}\big)$,\;
  without randomized technique.
\end{enumerate}
%
It is obvious that
the versions
using randomized technique can be advantageous when $\hat{l} \ll \min \{n_1,n_2\}$.

\vspace{-0.3cm}

\section{\textbf{EXPERIMENTAL RESULTS}}\label{experiments}

In this section, we perform extensive experiments on both 
synthetic 
and   real-world tensor data 
to substantiate the superiority and effectiveness of the proposed approach.
%
%
%
 %
All the experiments  are implemented on  the platform of Windows 10 and Matlab (R2016a) with an
Intel(R) Xeon(R) Gold-5122 3.60GHz CPU and 192GB memory.

\vspace{-0.346cm}

\subsection {\textbf{Synthetic Experiments}}
In this subsection, we
mainly perform the
efficiency/precision validation  and convergence study 
on the synthetic high-order
 tensors,
and also compare the obtained
results of the proposed method (\textbf{HWTSN+w$\ell_q$})
 with the baseline 
RLRTC method induced by high-order T-SVD framework,
i.e.,
\textbf{HWTNN+$\ell_1$} \cite{qin2021robust}. 
Two fast  versions
(i.e., they 
 incorporate the unblocked and blocked randomized
strategies, respectively) of ``\textbf{HWTSN+w$\ell_q$}"
 are called \textbf{HWTSN+w$\ell_q$(UR)} and \textbf{HWTSN+w$\ell_q$(BR)}, respectively.

\begin{table*}
  \caption{
  The CPU time  and RelError values 
  obtained by fourth-order 
  synthetic tensors restoration. 
  }
  \label{sys_fourth1}
  \centering
  \renewcommand\arraystretch{0.79}
\setlength\tabcolsep{6pt}
\scriptsize
\begin{tabular}{c cc cc  cc cc }
    %
     \hline
\multirow{2}{*}{
   \text{Algorithm Parameters}
     }&
    \multicolumn{2}{c }{
    HWTNN+$\ell_1$ \cite{qin2021robust}
    }&
    \multicolumn{2}{c } {HWTSN+w$\ell_q$
    }&
    \multicolumn{2}{c } {HWTSN+w$\ell_q$(UR)
    }&
    \multicolumn{2}{c } {HWTSN+w$\ell_q$(BR)
    }\\
     \cmidrule(rl){2-3}  \cmidrule(rl){4-5} \cmidrule(rl){6-7} \cmidrule(rl){8-9} 
  \qquad&   
  Time (s) & RelError&Time (s)& RelError&Time (s)& RelError&Time (s)& RelError 
 \\
     \hline

  \specialrule{0em}{1pt}{1pt}
\hline

$\mathfrak{L}=\text{FFT}, k=50, t=1,p=q=0.9$
& \multirow{2}{*}{1119}   &\multirow{2}{*}{9.893e$-$8}
&1670&4.933e$-$9
&722&4.528e$-$9
 &728&4.522e$-$9
  \\

$\mathfrak{L}=\text{FFT}, k=50,  t=1,p=q=0.7$
&   \qquad &\qquad
 &1640&9.179e$-$9
&699&9.325e$-$9
 &712&  1.039e$-$8  \\


$\mathfrak{L}=\text{DCT}, k=50,  t=1,p=q=0.9$
&   \multirow{2}{*}{1230}  & \multirow{2}{*}{9.925e$-$8}
 &1802&4.949e$-$9
&710&5.189e$-$9
 &727&  4.869e$-$9 \\
$\mathfrak{L}=\text{DCT}, k=50, t=1,p=q=0.7$
&   \qquad  &\qquad
 &1731&9.218e$-$9
&691&1.042e$-$8
 &707& 9.622e$-$9  \\
$\mathfrak{L}=\text{ROT}, k=50, t=1,p=q=0.9$
&   \multirow{2}{*}{1229}  & \multirow{2}{*}{9.085e$-$8}
&1784&4.927e$-$9
&706&5.131e$-$9
 &720&   5.121e$-$9   \\

$\mathfrak{L}=\text{ROT}, k=50,  t=1,p=q=0.7$
&   \qquad  &\qquad
  &1749&9.520e$-$9
&700&9.711e$-$9
 &716&9.798e$-$9
      \\

\hline

 \multicolumn{9}{c}{
  $N_1=N_2=1000 , N_3=N_4=5$, $R=50, sr=0.5, \tau =0.4$
     }

      \\
       \specialrule{0em}{2pt}{2pt}
\hline

$\mathfrak{L}=\text{FFT}, k=50,  t=1,p=q=0.9$
&   \multirow{2}{*}{844} & \multirow{2}{*}{7.885e$-$8}
&  1121&7.208e$-$9
&  462 &  7.395e$-$9
&   456&    7.656e$-$9       \\

$\mathfrak{L}=\text{FFT}, k=50, t=1,p=q=0.7$
&  \qquad  &\qquad
 & 1129  & 7.602e$-$9
&  475&     7.843e$-$9
  &  454 &    6.756e$-$9   \\


$\mathfrak{L}=\text{DCT}, k=50,  t=1,p=q=0.9$
&   \multirow{2}{*}{920} & \multirow{2}{*}{7.916e$-$8}
   & 1210  &7.259e$-$9
& 452  &   7.785e$-$9
   &460   & 7.923e$-$9     \\
$\mathfrak{L}=\text{DCT}, k=50,  t=1,p=q=0.7$
&    \qquad &\qquad
 &  1229 &7.419e$-$9
& 474&  6.622e$-$9
    & 467   &  6.901e$-$9     \\

$\mathfrak{L}=\text{ROT}, k=50,  t=1,p=q=0.9$
&  \multirow{2}{*}{921}  &\multirow{2}{*}{8.035e$-$8}
 &1227  &6.823e$-$9
&  461&  7.978e$-$9
   & 452   &   6.965e$-$9   \\
$\mathfrak{L}=\text{ROT}, k=50,  t=1,p=q=0.7$
&     \qquad&\qquad
 &  1126 &8.872e$-$9
&  484 &    6.844e$-$9
&469  &    6.124e$-$9      \\

\hline

 \multicolumn{9}{c}{
   $N_1=N_2=1000, N_3=N_4=5$, $R=50, sr=0.5, \tau =0.2$
     }
    \\

  \specialrule{0em}{2pt}{2pt}
\hline

$\mathfrak{L}=\text{FFT}, k=100, t=1,p=q=0.9$
&  \multirow{2}{*}{5141} &\multirow{2}{*}{6.474e$-$8}
& 6939  &   3.704e$-$9
& 2234  &  4.051e$-$9
& 2178 &    3.769e$-$9      \\

$\mathfrak{L}=\text{FFT}, k=100,  t=1,p=q=0.7$
&  \qquad  &\qquad
&  6696 &   6.926e$-$9
  &2191   & 7.619e$-$9
    & 2091  &  7.033e$-$9    \\


$\mathfrak{L}=\text{DCT}, k=100,  t=1,p=q=0.9$
&  \multirow{2}{*}{5871} &\multirow{2}{*}{6.489e$-$8}
&  8031  &   3.701e$-$9
&   2853&   4.049e$-$9
  & 2775  &      3.766e$-$9\\
$\mathfrak{L}=\text{DCT}, k=100, t=1,p=q=0.7$
&   \qquad  &\qquad
&  7608 &     6.114e$-$9
&  2673 &   6.909e$-$9
  &  2650  &     6.619e$-$9 \\
$\mathfrak{L}=\text{ROT}, k=100, t=1,p=q=0.9$
&  \multirow{2}{*}{6137} &\multirow{2}{*}{6.489e$-$8}
&  7863  &   3.091e$-$9
&  2731 & 3.672e$-$9
& 2707  &   3.464e$-$9           \\

$\mathfrak{L}=\text{ROT}, k=100,  t=1,p=q=0.7$
&   \qquad  & \qquad
 &   7470&6.264e$-$9
&  2610  &  6.933e$-$9
&  2599 &6.672e$-$9

      \\

\hline

 \multicolumn{9}{c}{
  $N_1=N_2=2000 , N_3=N_4=5$, $R=100, sr=0.5, \tau =0.4$
     }

      \\
       \specialrule{0em}{2pt}{2pt}
\hline

$\mathfrak{L}=\text{FFT}, k=100,  t=1,p=q=0.9$
&  \multirow{2}{*}{4520} &\multirow{2}{*}{5.588e$-$8}
&  4745 &4.872e$-$9
& 1494  &  5.236e$-$9
  & 1468  &    5.027e$-$9    \\

$\mathfrak{L}=\text{FFT}, k=100, t=1,p=q=0.7$
&   \qquad &\qquad
&  4685 & 5.022e$-$9
&  1477 &     5.739e$-$9
 &  1461 &      5.488e$-$9   \\


$\mathfrak{L}=\text{DCT}, k=100,  t=1,p=q=0.9$
&  \multirow{2}{*}{4178} &\multirow{2}{*}{5.396e$-$8}
& 4509   & 5.358e$-$9  
&   1506&  5.959e$-$9   
  &  1491  &    5.749e$-$9 
  \\
$\mathfrak{L}=\text{DCT}, k=100,  t=1,p=q=0.7$
&  \qquad   &\qquad
& 4412   &5.121e$-$9
&  1488 &     5.749e$-$9
 &  1481  &    5.591e$-$9    \\

$\mathfrak{L}=\text{ROT}, k=100,  t=1,p=q=0.9$
&  \multirow{2}{*}{4143} &\multirow{2}{*}{5.387e$-$8}
& 4399   &5.138e$-$9
& 1486  &     6.017e$-$9
 &   1457 &   5.932e$-$9   \\
$\mathfrak{L}=\text{ROT}, k=100,  t=1,p=q=0.7$
&    \qquad &\qquad
&   4385 & 4.971e$-$9
&   1483&    5.704e$-$9
   & 1470   &    5.369e$-$9   \\

\hline

 \multicolumn{9}{c}{
   $N_1=N_2=2000, N_3=N_4=5$, $R=100, sr=0.5, \tau =0.2$
     }
    \\
\end{tabular}
\end{table*}

\begin{table*}
  \caption{
  The CPU time and RelError values 
  obtained by fifth-order 
  synthetic tensors restoration. 
  }
  \label{sys_fifth1}
  \centering
  \renewcommand\arraystretch{0.79}
\setlength\tabcolsep{6.0pt}
\footnotesize
\scriptsize
\begin{tabular}{c cc cc  cc cc  }
    %
     \hline
\multirow{2}{*}{
   \text{Algorithm Parameters}
     }&
    \multicolumn{2}{c }{
    HWTNN+$\ell_1$ \cite{qin2021robust}
    }&
    \multicolumn{2}{c } {HWTSN+w$\ell_q$
    }&
    \multicolumn{2}{c } {HWTSN+w$\ell_q$(UR)
    }
    &
    \multicolumn{2}{c } {HWTSN+w$\ell_q$(BR)
    }
    \\
     \cmidrule(rl){2-3}  \cmidrule(rl){4-5} \cmidrule(rl){6-7} \cmidrule(rl){8-9}
  \qquad&   
  Time (s)& RelError&Time (s)& RelError&Time (s)& RelError&Time (s)& RelError
 \\
     \hline

  \specialrule{0em}{1pt}{1pt}
\hline

$\mathfrak{L}=\text{FFT}, k=50, t=1,p=q=0.9$
& \multirow{2}{*}{1286}   &\multirow{2}{*}{9.368e$-$8}
&  1902 &4.329e$-$9
& 866  &    4.699e$-$9
  & 859  &   4.673e$-$9    \\

$\mathfrak{L}=\text{FFT}, k=50,  t=1,p=q=0.7$
&   \qquad &\qquad
&  1871 &       8.855e$-$9
&  856 &      1.001e$-$8
 &  841 &     9.612e$-$9 \\


$\mathfrak{L}=\text{DCT}, k=50,  t=1,p=q=0.9$
& \multirow{2}{*}{1424}   &\multirow{2}{*}{9.465e$-$8}
&  2057  &     5.085e$-$9
&  899 &    4.674e$-$9
  &  889  &   4.349e$-$9   \\
$\mathfrak{L}=\text{DCT}, k=50, t=1,p=q=0.7$
&   \qquad  &\qquad
&  1980  &     8.539e$-$9
&  886 &  1.032e$-$8
    &  866  &    9.428e$-$9  \\
$\mathfrak{L}=\text{ROT}, k=50, t=1,p=q=0.9$
& \multirow{2}{*}{1422}   &\multirow{2}{*}{9.454e$-$8}
&  2069 &4.731e$-$9
&   911 &    4.641e$-$9
&  883 &      4.890e$-$9      \\

$\mathfrak{L}=\text{ROT}, k=50,  t=1,p=q=0.7$
&    \qquad & \qquad
   &   2006&9.273e$-$9
&  982  &     6.675e$-$9
&  947 &4.467e$-$9

      \\

\hline

 \multicolumn{9}{c}{
  $N_1=N_2=1000 , N_3=N_4=N_5=3$, $R=50, sr=0.5, \tau =0.4$
     }

      \\
       \specialrule{0em}{2pt}{2pt}
\hline

$\mathfrak{L}=\text{FFT}, k=50,  t=1,p=q=0.9$
& \multirow{2}{*}{979}   &\multirow{2}{*}{7.865e$-$8}
& 1300  &     6.655e$-$9
&  561 &    7.111e$-$9
& 558  &   7.405e$-$9    \\

$\mathfrak{L}=\text{FFT}, k=50, t=1,p=q=0.7$
 & \qquad  & \qquad
&   1301 &6.623e$-$9
& 586  &       7.587e$-$9
&  583 &         7.526e$-$9   \\


$\mathfrak{L}=\text{DCT}, k=50,  t=1,p=q=0.9$
& \multirow{2}{*}{1082}   &\multirow{2}{*}{7.926e$-$8}
&   1411 &     7.174e$-$9
&  582 &      6.688e$-$9
&   577 &   6.746e$-$9   \\
$\mathfrak{L}=\text{DCT}, k=50,  t=1,p=q=0.7$
&   \qquad  &\qquad
&   1397 &     8.348e$-$9
& 601  &   7.113e$-$9
   &  594  &   7.852e$-$9   \\

$\mathfrak{L}=\text{ROT}, k=50,  t=1,p=q=0.9$
 & \multirow{2}{*}{1088}   &\multirow{2}{*}{7.516e$-$8}
&   1415 &     7.319e$-$9
& 582  &    6.856e$-$9
  &  572  &   6.929e$-$9   \\
$\mathfrak{L}=\text{ROT}, k=50,  t=1,p=q=0.7$
&   \qquad  &\qquad
&   1398 &      7.638e$-$9
&  603 &   7.577e$-$9
   & 592   &   6.734e$-$9   \\

\hline

 \multicolumn{9}{c}{
   $N_1=N_2=1000, N_3=N_4=N_5=3$, $R=50, sr=0.5, \tau =0.2$
     }
    \\

  \specialrule{0em}{2pt}{2pt}
\hline

$\mathfrak{L}=\text{FFT}, k=100, t=1,p=q=0.9$
& \multirow{2}{*}{5787}   &\multirow{2}{*}{6.266e$-$8}
&  7789 &      3.062e$-$9
&  2712 &    3.648e$-$9
    &  2635 & 3.487e$-$9      \\

$\mathfrak{L}=\text{FFT}, k=100,  t=1,p=q=0.7$
&  \qquad  &\qquad
& 7402  &       5.899e$-$9
&  2564 &  6.492e$-$9
     &2525   &  6.217e$-$9     \\


$\mathfrak{L}=\text{DCT}, k=100,  t=1,p=q=0.9$
& \multirow{2}{*}{5626}   &\multirow{2}{*}{6.248e$-$8}
& 7603   &       3.041e$-$9
&  2801 &     3.572e$-$9
 &  2770  &  3.241e$-$9    \\
$\mathfrak{L}=\text{DCT}, k=100, t=1,p=q=0.7$
&    \qquad &\qquad
&   7194 &     6.504e$-$9
&  2647 &     7.001e$-$9
 & 2621   &    6.898e$-$9  \\
$\mathfrak{L}=\text{ROT}, k=100, t=1,p=q=0.9$
& \multirow{2}{*}{5615}   &\multirow{2}{*}{6.395e$-$8}
   & 7596  &3.256e$-$9
&   2788 &       3.811e$-$9
&   2764&       3.533e$-$9      \\

$\mathfrak{L}=\text{ROT}, k=100,  t=1,p=q=0.7$
&    \qquad  &   \qquad
 & 7301  &5.219e$-$9
&  2691  &     5.864e$-$9
&  2669 &5.790e$-$9

      \\

\hline

 \multicolumn{9}{c}{
  $N_1=N_2=2000 , N_3=N_4=N_5=3$, $R=100, sr=0.5, \tau =0.4$
     }

      \\
       \specialrule{0em}{2pt}{2pt}
\hline

$\mathfrak{L}=\text{FFT}, k=100,  t=1,p=q=0.9$
& \multirow{2}{*}{4797}   &\multirow{2}{*}{5.243e$-$8}
& 5210  &       4.995e$-$9
&  1740 &      5.518e$-$9
&  1715 &    5.305e$-$9  \\

$\mathfrak{L}=\text{FFT}, k=100, t=1,p=q=0.7$
&   \qquad &\qquad
&  5117 &       5.236e$-$9
& 1726  &     5.769e$-$9
  & 1683  &    5.458e$-$9   \\


$\mathfrak{L}=\text{DCT}, k=100,  t=1,p=q=0.9$
& \multirow{2}{*}{4636}   &\multirow{2}{*}{5.237e$-$8}
&  4669  &     5.388e$-$9
&  1646 &    6.104e$-$9
 &   1619 &  5.636e$-$9   \\
$\mathfrak{L}=\text{DCT}, k=100,  t=1,p=q=0.7$
&    \qquad &\qquad
&   4596&      5.019e$-$9
&  1642 &     5.887e$-$9
 &  1615  &   5.451e$-$9   \\

$\mathfrak{L}=\text{ROT}, k=100,  t=1,p=q=0.9$
& \multirow{2}{*}{4342}   &\multirow{2}{*}{4.893e$-$8}
&    4658&      5.007e$-$9
& 1669  &    5.705e$-$9
  &  1665  &     5.362e$-$9 \\
$\mathfrak{L}=\text{ROT}, k=100,  t=1,p=q=0.7$
&     \qquad&\qquad
&  4637  &     4.899e$-$9
&  1683 &     5.310e$-$9
 & 1657   &    5.154e$-$9  \\

\hline

 \multicolumn{9}{c}{
   $N_1=N_2=2000, N_3=N_4=N_5=3$, $R=100, sr=0.5, \tau =0.2$
     }
    \\
    \vspace{-0.4cm}
\end{tabular}
\end{table*}

 In our synthetic experiments,
the  ground-truth low T-SVD rank
tensor ${{\boldsymbol{\mathcal{L}}}}$ with ${\operatorname{rank}}_{tsvd} ({{\boldsymbol{\mathcal{L}}}}) = R$
is generated by performing the order-$d$ t-product
${{\boldsymbol{\mathcal{L}}}}={{\boldsymbol{\mathcal{L}}}_{1}}{*}_{\mathfrak{L}}{{\boldsymbol{\mathcal{L}}}}_{2}$,
where the entries of  ${{\boldsymbol{\mathcal{L}}}}_{1} \in \mathbb{R}^{N_1\times R \times N_3\times \cdots \times  N_d}$
and  ${{\boldsymbol{\mathcal{L}}}}_{2} \in \mathbb{R}^{R\times N_2 \times  N_3 \times  \cdots \times N_d}$
are independently sampled from the normal distribution ${\mathcal{N}}(0,1)$.
\textcolor[rgb]{0.00,0.00,0.00}{Three  invertible linear transforms $\mathfrak{L}$ are adopted to the t-product:
(a) Fast Fourier Transform (FFT);
(b) Discrete Cosine Transform (DCT);
(c) Random Orthogonal Transform  (ROT)}.
Suppose that ${\Omega}$ is the observed index set
that is generated uniformly at random
while  ${{\Omega}_{\bot}}$ is the unobserved index set,
in which $|\Omega| =sr \cdot 
\prod_{a=1}^{d} N_a 
$, $sr$ denotes the sampling ratio,
$|\Omega|$ represents the cardinality of $\Omega$.
Then,
we construct
the noise/outliers
tensor ${{\boldsymbol{\mathcal{E}}}} \in \mathbb{R}^{N_1 \times N_2 \times N_3 \times \cdots \times N_d}$ 
as follows:
\textbf{1)}  
all the elements 
in 
${{\Omega}_{\bot}}$
are all equal to $0$;
\textbf{2)}
the $\tau \cdot |\Omega|$  elements randomly selected in ${\Omega}$ are each valued as
$\{\pm1\}$
with   equal probability   $\frac{1}{2}$,
and the remaining elements in ${\Omega}$ are set to $0$.
Finally, we 
form the observed tensor as
 $\boldsymbol{\bm{P}}_{{{\Omega}}}({\boldsymbol{\mathcal{M}}})=
\boldsymbol{\bm{P}}_{{{\Omega}}}({\boldsymbol{\mathcal{L}}}+{\boldsymbol{\mathcal{E}}})$.
We evaluate the restoration 
performance by 
CPU time and 
Relative Error (RelError) defined as
$$
\operatorname{RelError
}:=
{\|{\boldsymbol{\mathcal{L}}}-
\hat{{\boldsymbol{\mathcal{L}}}}
\|_{{{\mathnormal{F}}}}} / {\|{\boldsymbol{\mathcal{L}}}\|_{{{\mathnormal{F}}}}},
$$
where $\hat{{\boldsymbol{\mathcal{L}}}}$ 
denote the estimated result of the ground-truth ${{\boldsymbol{\mathcal{L}}}}$. 
%

\subsubsection {\textbf{Efficiency/Precision   Validation}} \label{fourth_sys}
%
%
Firstly, we verify the accuracy/effectiveness  
of the proposed algorithm as well as the compared 
ones on 
the following two types of
 synthetic tensors:
 \textbf{(I)}
      $N_1=N_2=N, N \in \{1000,2000\}, N_3=N_4=5$;
      \textbf{(II)}
      $N_1=N_2=N, N \in \{1000,2000\}, 
      N_3=N_4=N_5=3$.
In our experiments,
 we set
 $R=0.05 \cdot\min(N_1,N_2), sr=0.5, \tau \in \{0.4,0.2\}$,
  $p,q \in \{0.9,0.7\}$, $\mathfrak{L} \in \{\text{FFT}, \text{DCT}, \text{ROT}\}$,
 $
 b=\lfloor\frac{ k+5} 
 { 3}\rfloor,
 t=1$,
$\vartheta=1.1, 
\beta^0=10^{-3},
\beta^{\max}=10^{8},
\varpi=10^{-6},\epsilon_{1}=\epsilon_{2}=10^{-16}$,
$c_1=\alpha \cdot\min(N_1,N_2)$,
$\alpha \in
\{5,10,15,20,25\}
$,
$c_2=1$,
 and  $\lambda \in \{0.02, 0.03, 0.05,0.08\}$.
The experimental results are presented
in Table \ref{sys_fourth1} and Table \ref{sys_fifth1}, 
from which we can observed that
the RelError 
 values obtained 
 from the proposed method 
 are  relatively small in all case,
 which indicates that
the 
proposed algorithm
can accurately complete the latent low T-SVD rank tensor ${{\boldsymbol{\mathcal{L}}}}$
while removing the noise/outliers. 
 Besides,
 the versions integrated with 
 randomized techniques can greatly 
shorten the computational time
 under different linear transforms $\mathfrak{L}$.

 %
\begin{figure}[!htbp]
\centering
\renewcommand{\arraystretch}{0.1}
\setlength\tabcolsep{0.0pt}
\begin{tabular}
{ccc}
\centering
\includegraphics[width=1.18in, height=0.964in]{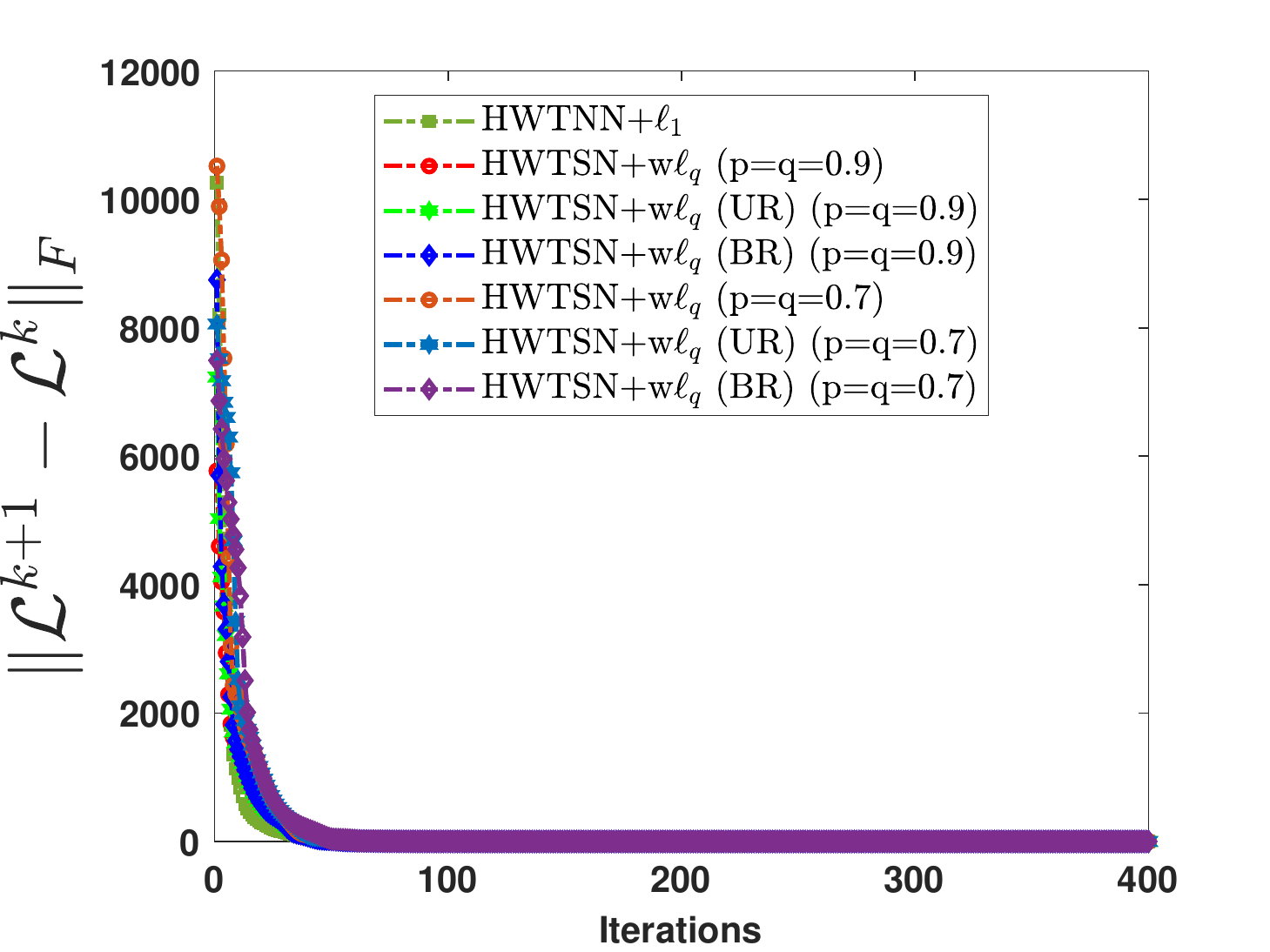}&
\includegraphics[width=1.18in, height=0.964in]{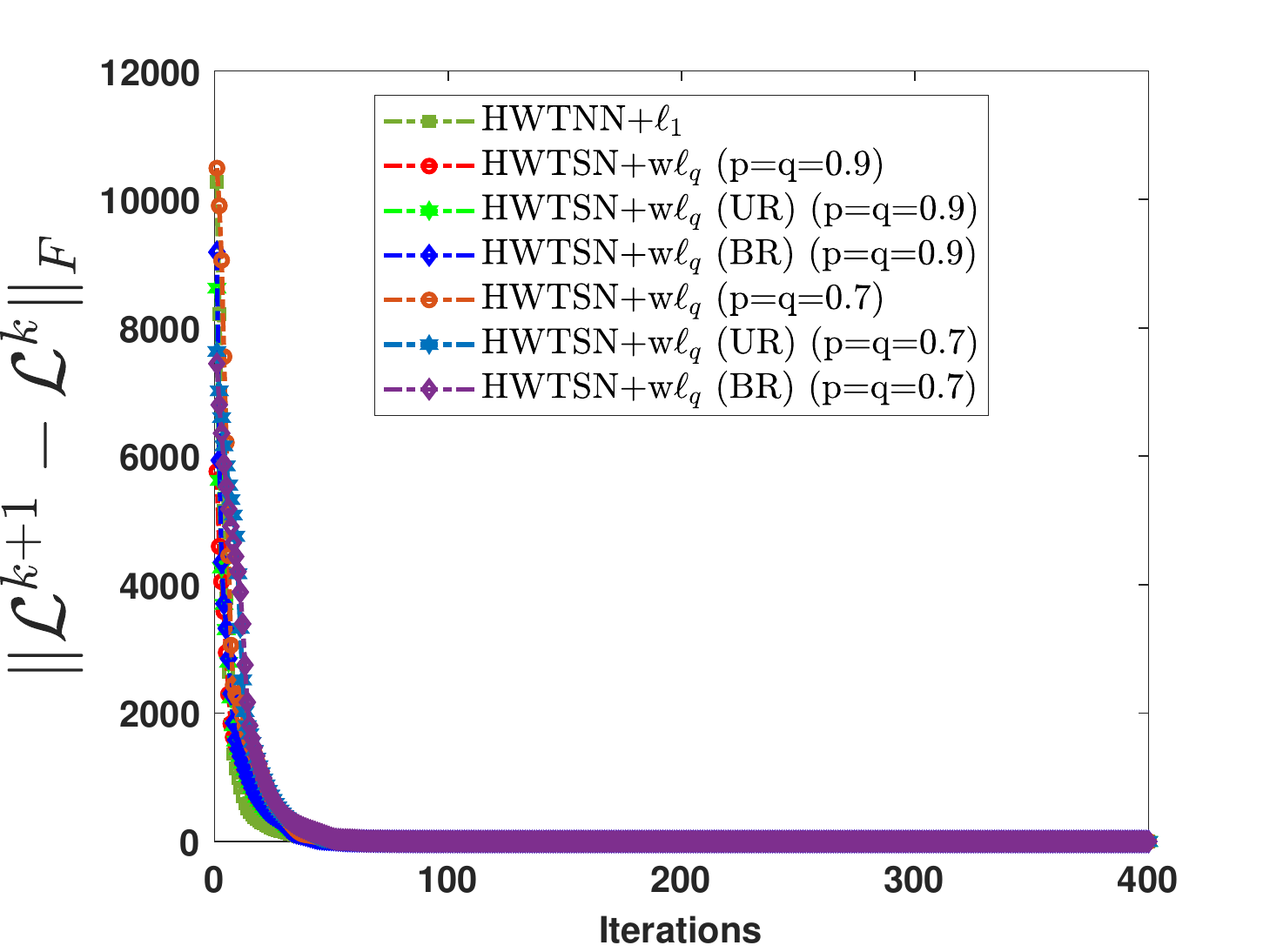}&
\includegraphics[width=1.18in, height=0.964in]{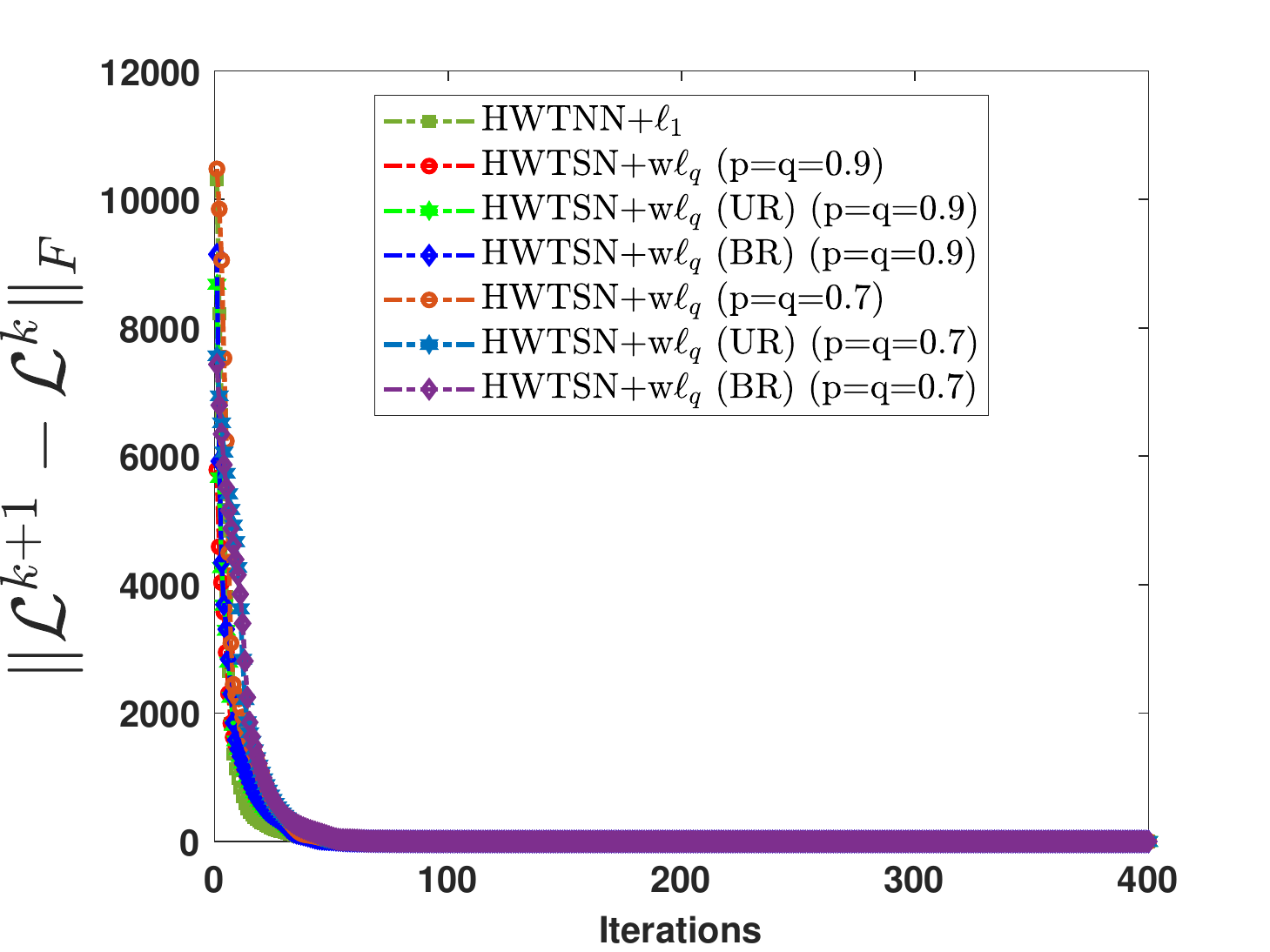}
\\
\includegraphics[width=1.18in, height=0.964in]{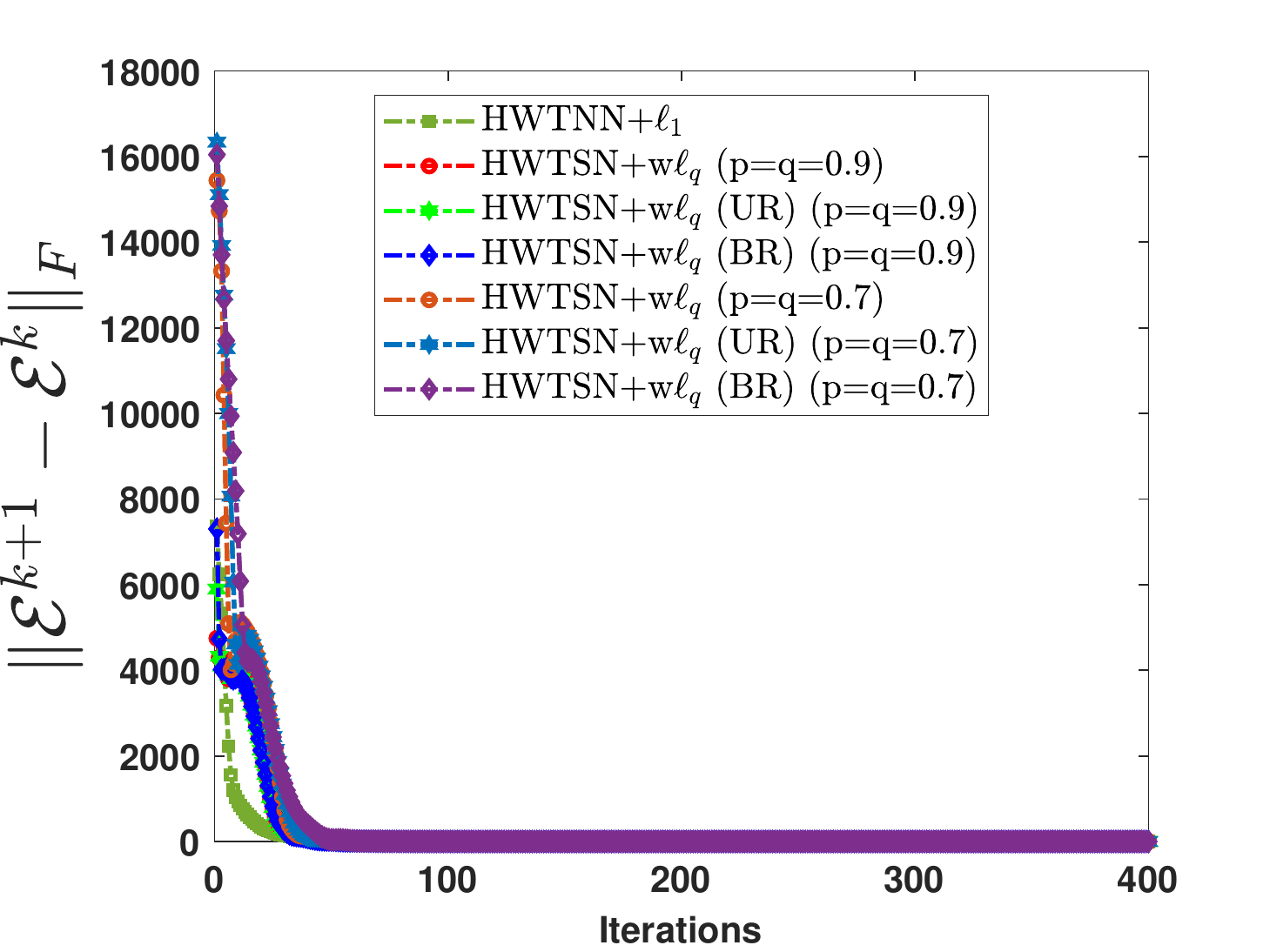}&
\includegraphics[width=1.18in, height=0.964in]{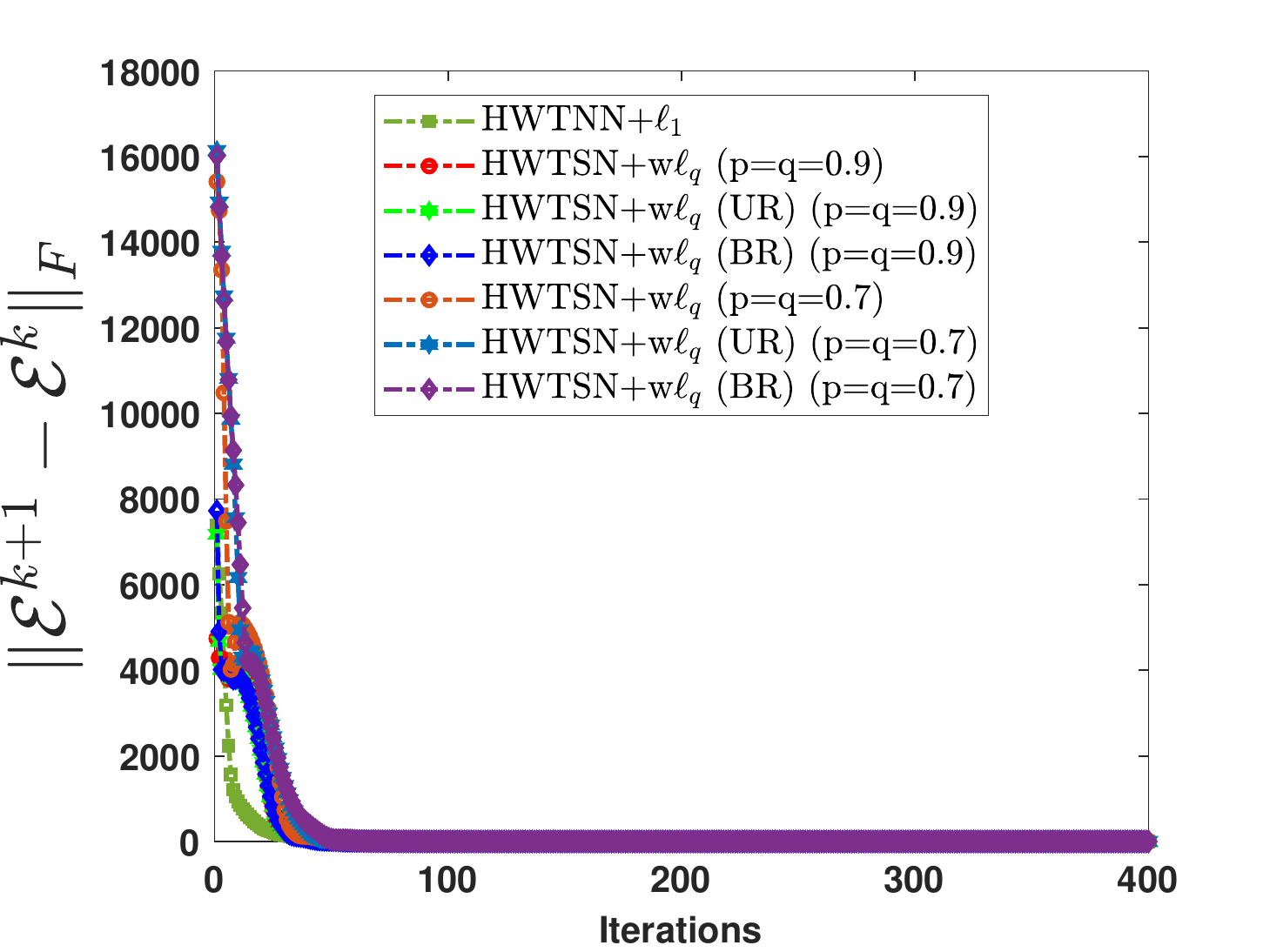}&
\includegraphics[width=1.18in, height=0.964in]{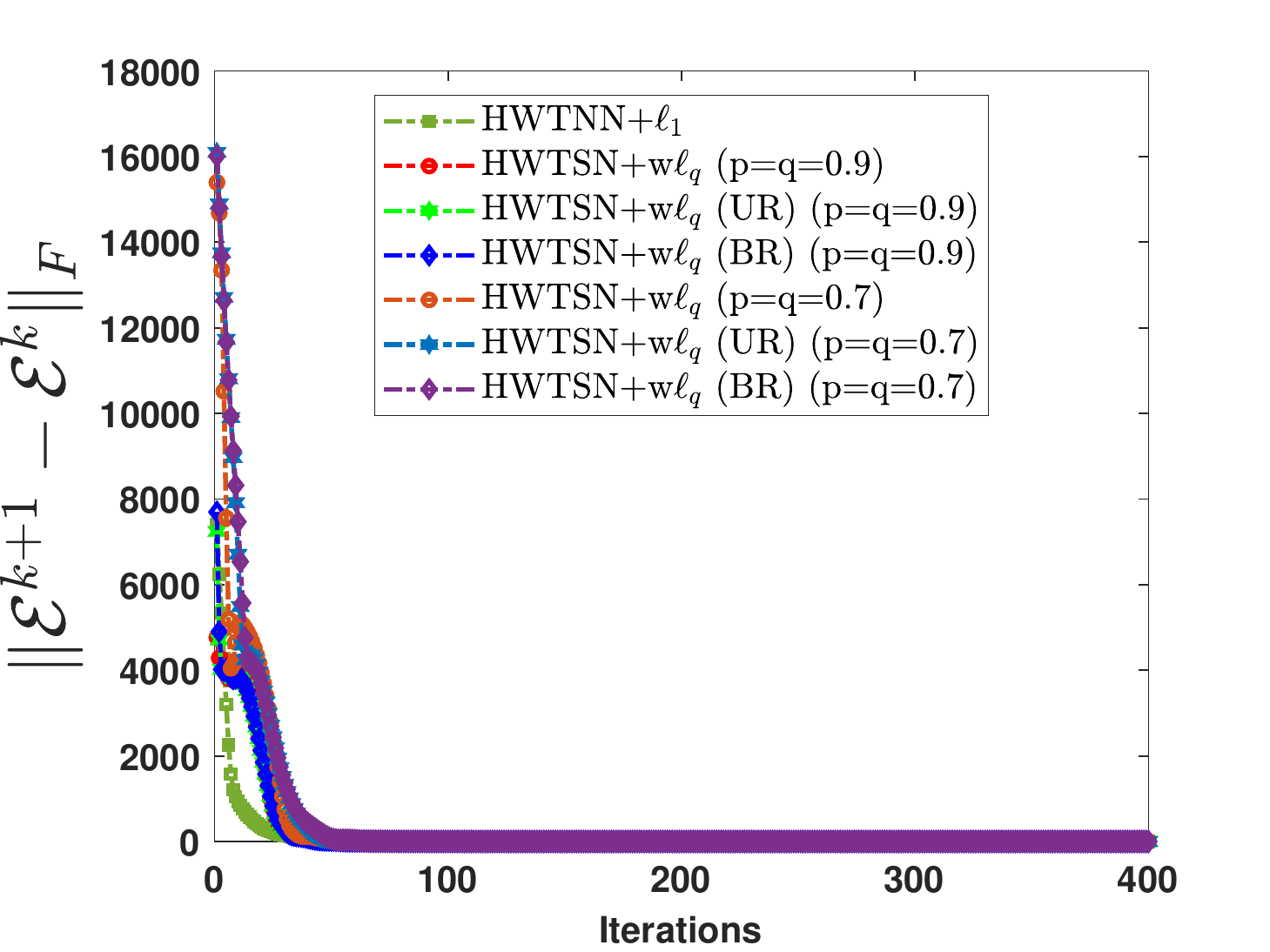}
\\
\includegraphics[width=1.18in, height=0.964in]{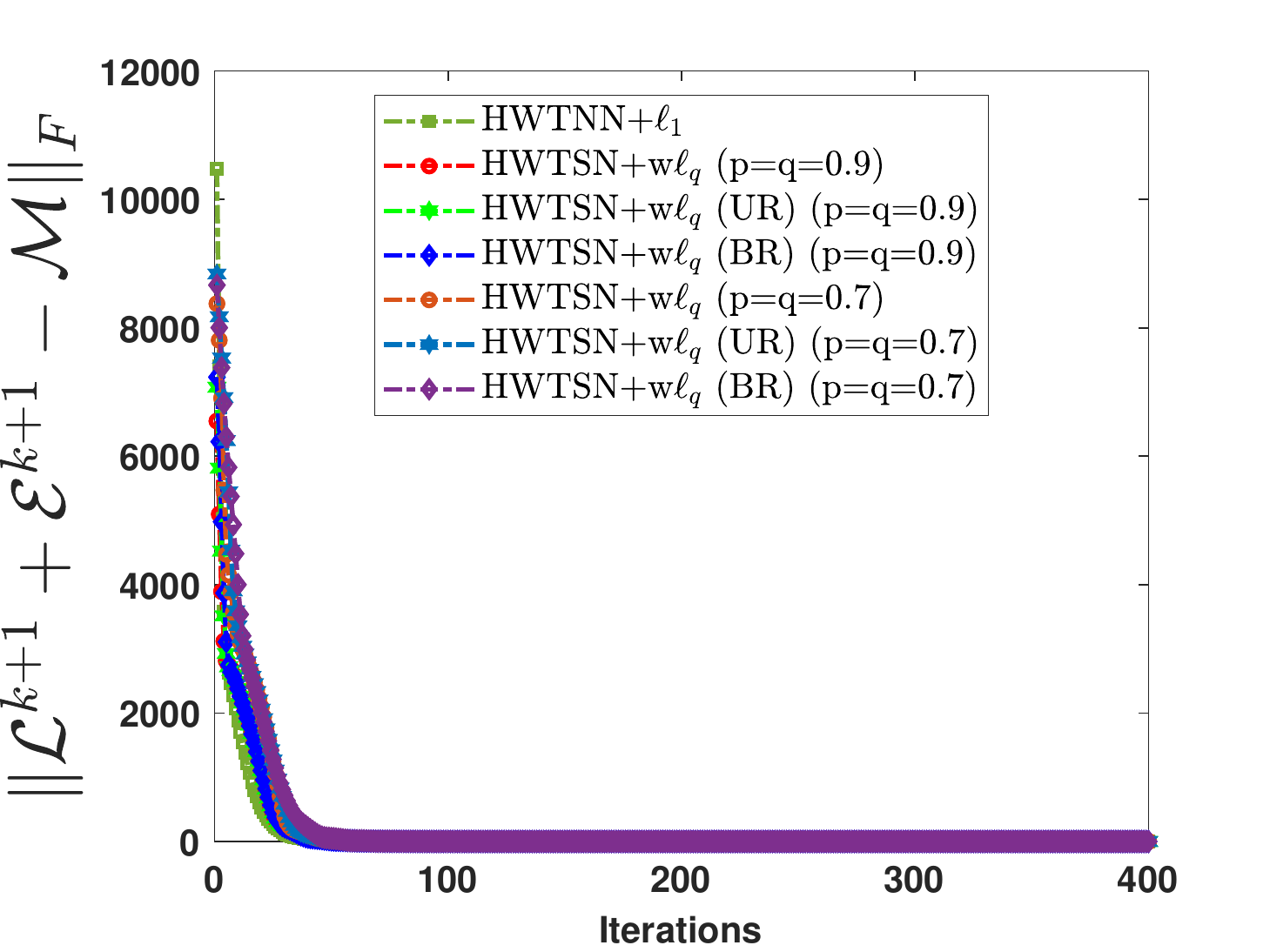}&
\includegraphics[width=1.18in, height=0.964in]{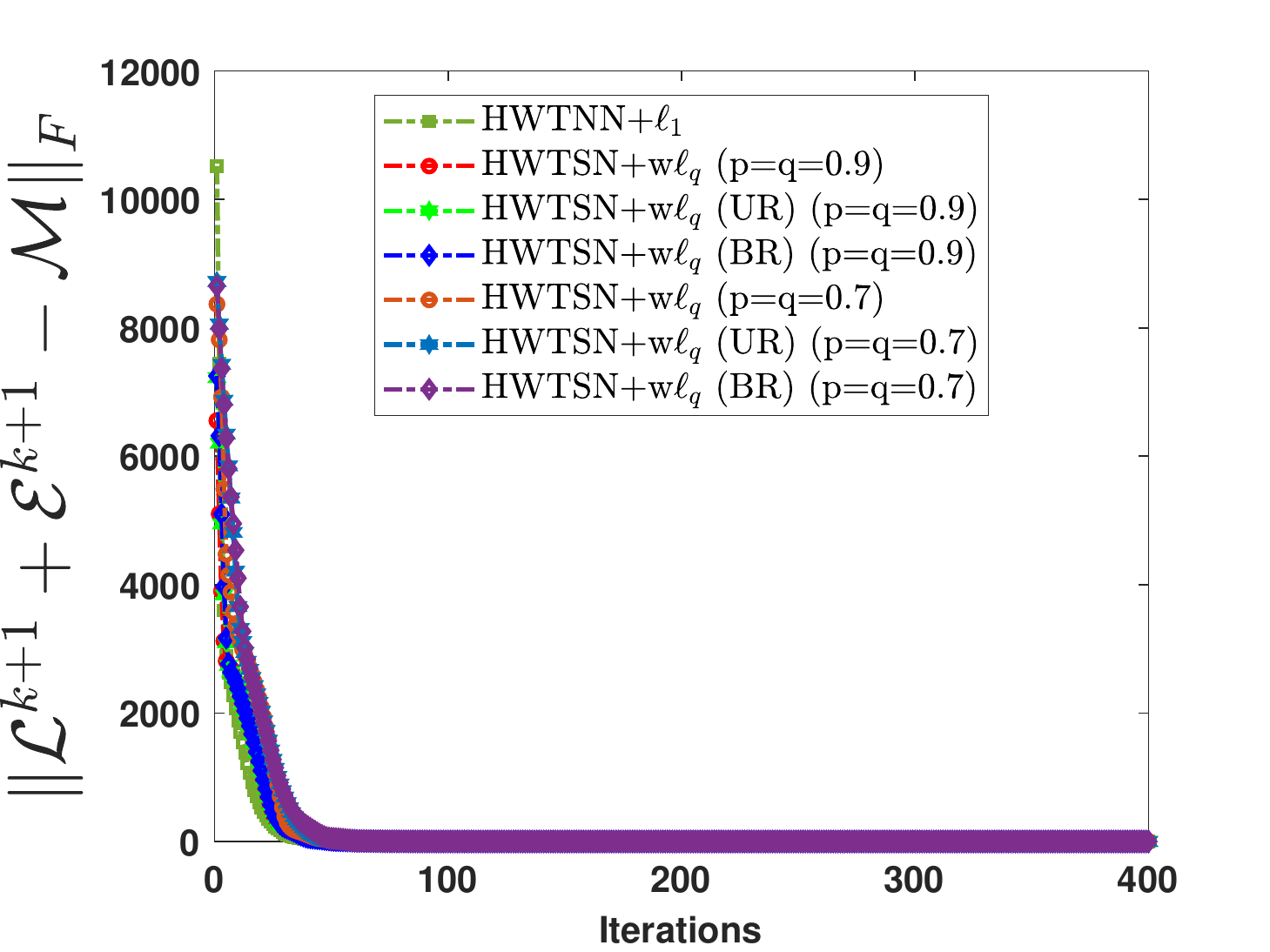}&
\includegraphics[width=1.18in, height=0.964in]{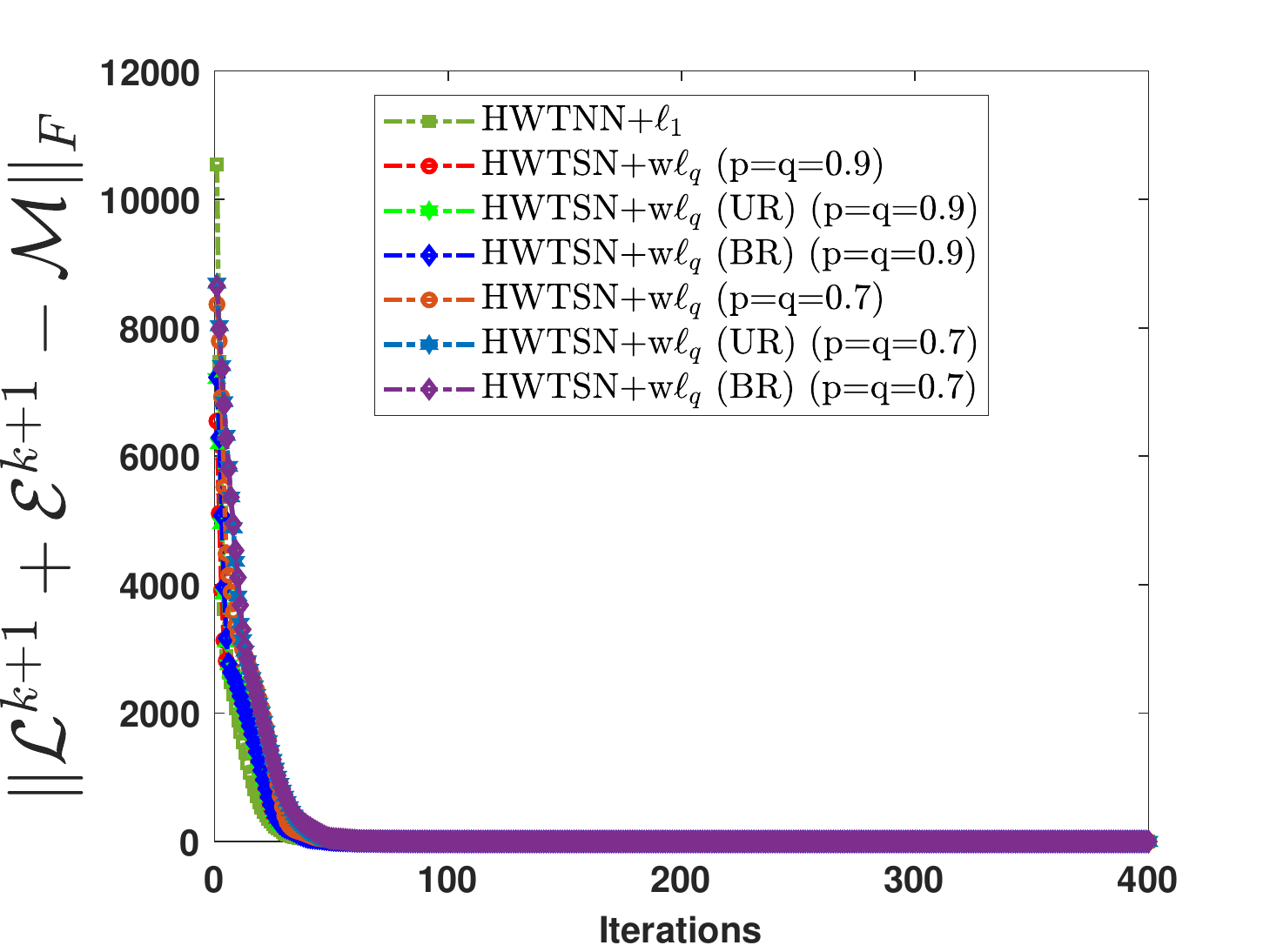}
\\
{(I) $\mathfrak{L}=\small{\text{FFT}}$}& {(II) $\mathfrak{L}=\small{\text{DCT}}$} & {(III) $\mathfrak{L}=\small{\text{ROT}}$}
\end{tabular}
\caption{
The 
convergence behavior
of  the
proposed and competitive 
RHTC
 algorithms.
 The x-coordinate is 
 the number of iterations,
the y-coordinates are the sequence 
Chg$1$-Chg$3$.
%
}
\label{conver_robust11}
\end{figure}

\subsubsection {\textbf{Convergence  Study}}
Secondly,
we mainly 
analyze the convergence behavior  
of the proposed and competitive  algorithms
on the following synthetic tensor:
$N_1=N_2=1000, N_3=N_4=N_5=3, R=50, sr = 0.5, \tau=0.4$.
The
parameter settings of the proposed algorithm are the same as those utilized 
in the previous experiments.
Then,
we record  three type values, i.e.,
$\operatorname{Chg1}:={\|{\boldsymbol{\mathcal{L}}}^{k+1}-{\boldsymbol{\mathcal{L}}}^{k}\|{_{\mathnormal{F}}}}$,
$\operatorname{Chg2}:={\|{\boldsymbol{\mathcal{E}}}^{k+1}-{\boldsymbol{\mathcal{E}}}^{k}\|{_{\mathnormal{F}}}}$,
$\operatorname{Chg3}:={\|{\boldsymbol{\mathcal{M}}}-{\boldsymbol{\mathcal{L}}}^{k+1}-{\boldsymbol{\mathcal{E}}}^{k+1}\|{_{\mathnormal{F}}}}$,
obtained by various RHTC algorithms  at the $k$-th iteration, respectively.
The recorded results are plotted in Figure \ref{conver_robust11},
which is exactly consistent with the Theorem \ref{conver},   
i.e.,
 the obtained $\operatorname{Chg1}$,
$\operatorname{Chg2}$, and
$\operatorname{Chg3}$
  gradually approach $0$
  when the proposed %
  algorithm
  iterates to a certain number of times.
  %

\vspace{-0.3cm}
 \subsection{\textbf{Real-World Applications}}
 \textbf{Experiment Settings:} 
 In this subsection,
 we apply the proposed method (\textbf{HWTSN+w$\ell_q$}) and its two accelerated versions
 to several real-world applications,
 and also compare it 
 with  other RLRTC approaches:
SNN+$\ell_1$\cite{
goldfarb2014robust1
},
TRNN+$\ell_1$ \cite{huang2020robust1},
%
%
%
TTNN+$\ell_1$ \cite{song2020robust},
TSP-$k$+$\ell_1$ \cite{lou2019robust1},
LNOP
\cite{chen2020robust1},
NRTRM \cite{qiu2021nonlocal},
and
HWTNN+$\ell_1$ \cite{qin2021robust}.
In our experiments, we 
 normalize the gray-scale
 value of the
 tested tensors to the interval
 $[0, 1]$.
 For the RLRTC methods based on 
third-order T-SVD, we reshape the last two or three modes of tested tensors 
into one mode.
 The observed tensor is constructed as follows:
 the random-valued impulse noise with ratio $\tau$ is uniformly and randomly   %
 added to each frontal slice of the ground-truth 
 tensor,
 and then we 
 sample ($sr \cdot \prod_{i=1}^{d} n_i$)
pixels from the noisy tensor to form the observed tensor 
 $
\boldsymbol{\bm{P}}_{{{\Omega}}}({\boldsymbol{\mathcal{M}}})$ at random.
Unless otherwise stated, all parameters involved in the competing methods were optimally assigned or selected as suggested in the reference papers.
The Peak Signal-to-Noise Ratio (PSNR),
%
the structural similarity (SSIM), and the CPU time 
are employed to evaluate the recovery performance.
%

\begin{figure*}[!htbp]
\renewcommand{\arraystretch}{0.528}
\setlength\tabcolsep{0.4pt}
\centering
\begin{tabular}{ccc ccc  cccccc}
\centering

\includegraphics[width=0.585in, height=0.48in]{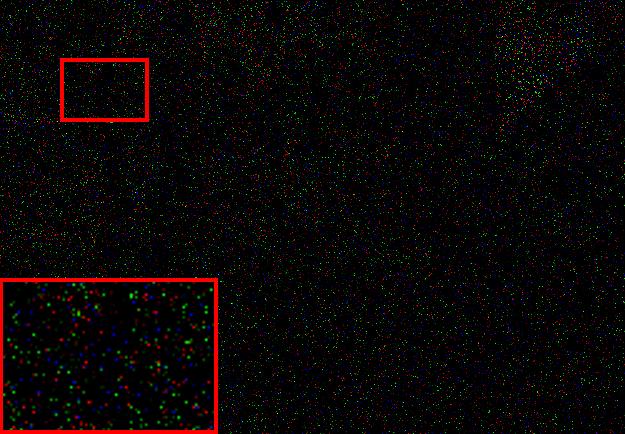}&
\includegraphics[width=0.585in, height=0.48in]{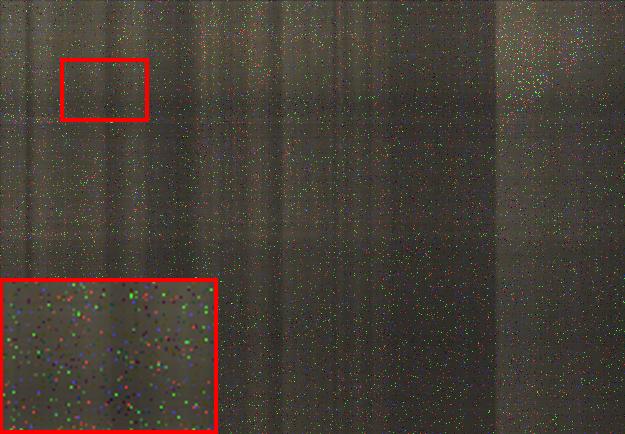}&
\includegraphics[width=0.585in, height=0.48in]{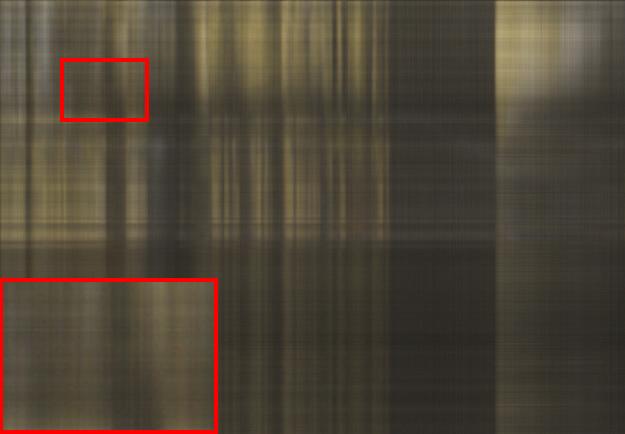}&
\includegraphics[width=0.585in, height=0.48in]{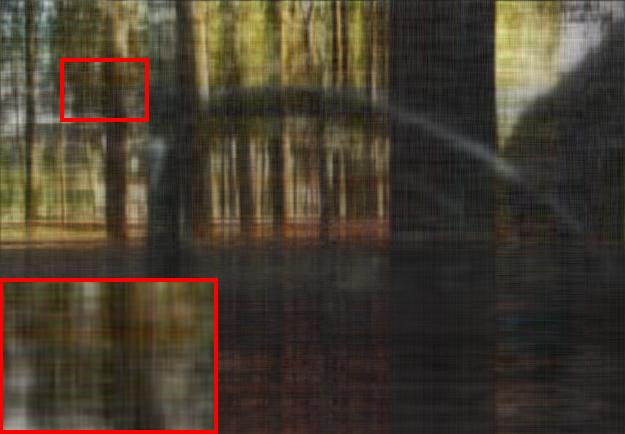}&
\includegraphics[width=0.585in, height=0.48in]{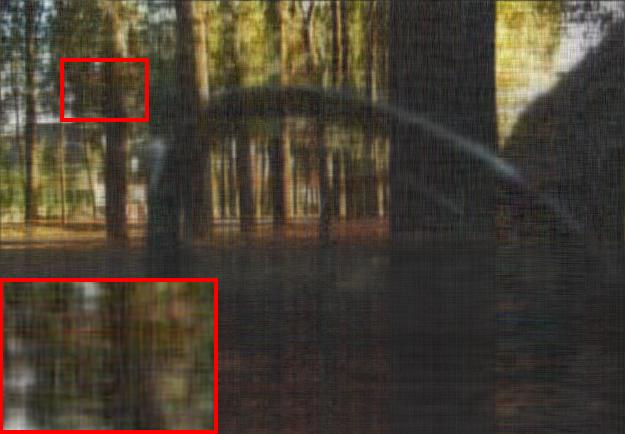}&
\includegraphics[width=0.585in, height=0.48in]{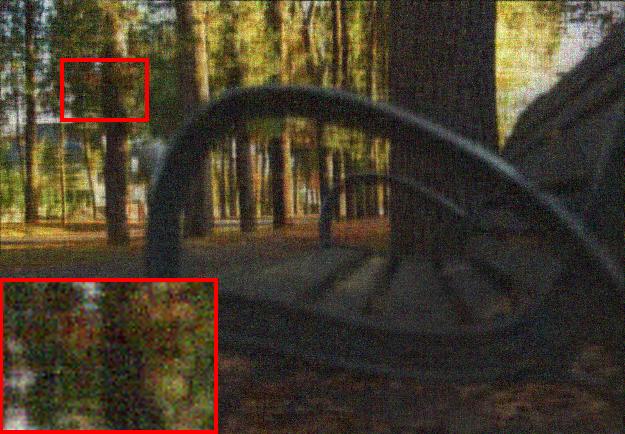}&
\includegraphics[width=0.585in, height=0.48in]{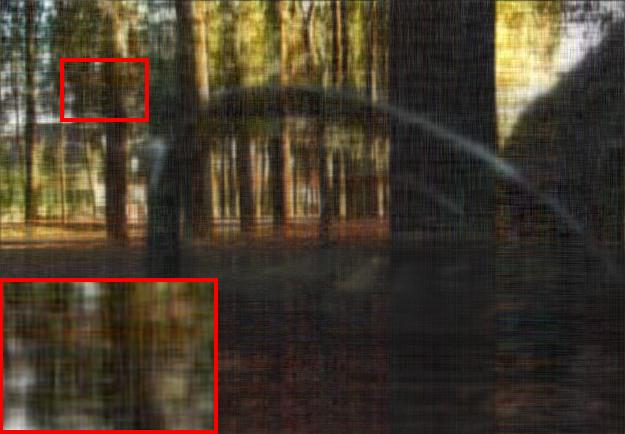}&
\includegraphics[width=0.585in, height=0.48in]{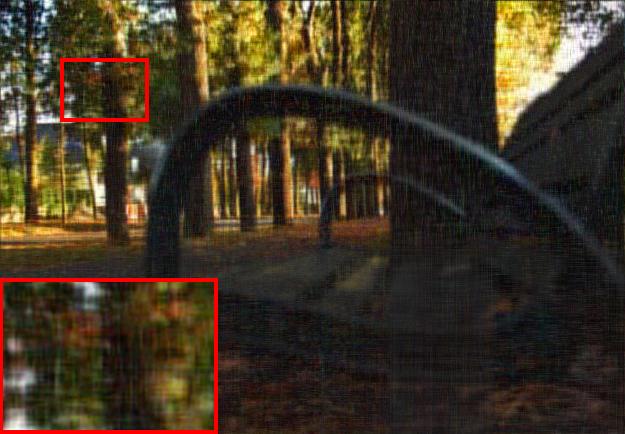}&
\includegraphics[width=0.585in, height=0.48in]{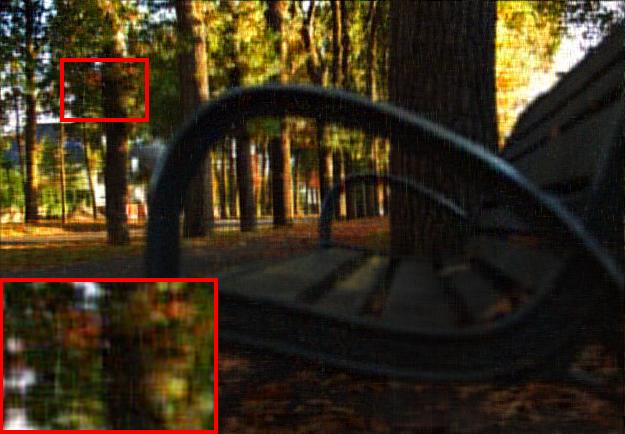}&
\includegraphics[width=0.585in, height=0.48in]{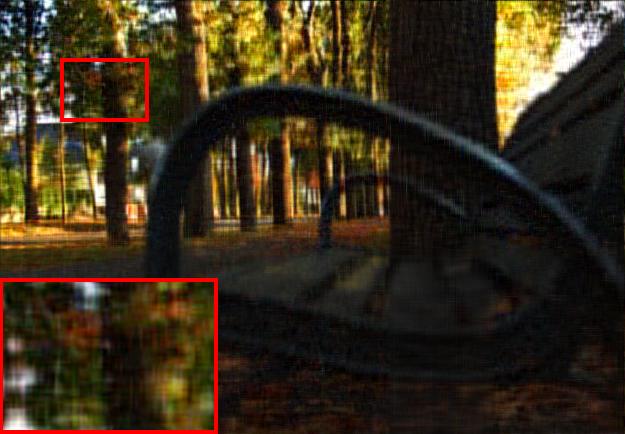}&
\includegraphics[width=0.585in, height=0.48in]{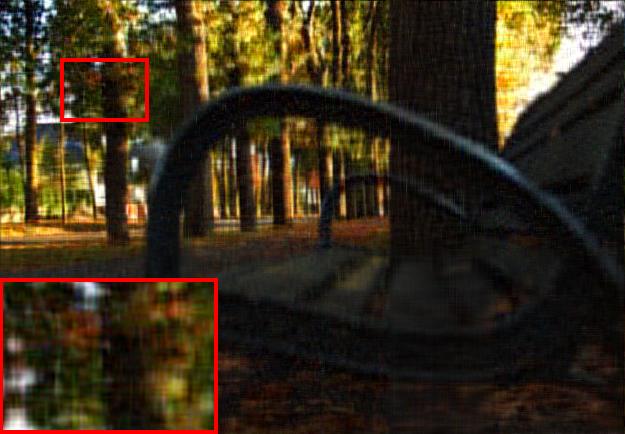}
&
\includegraphics[width=0.585in, height=0.48in]{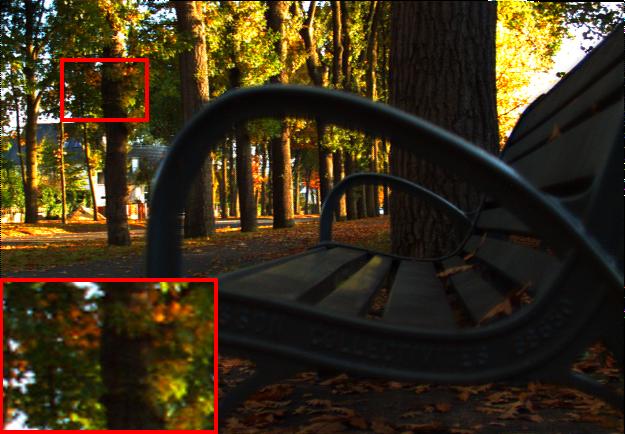}

\\


\includegraphics[width=0.585in, height=0.48in,angle=0]{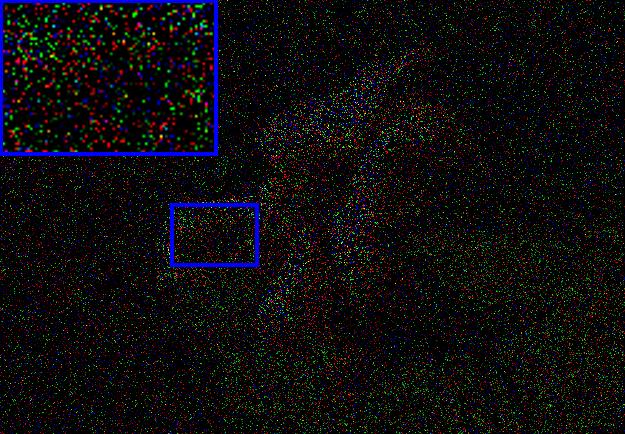}&
\includegraphics[width=0.585in, height=0.48in]{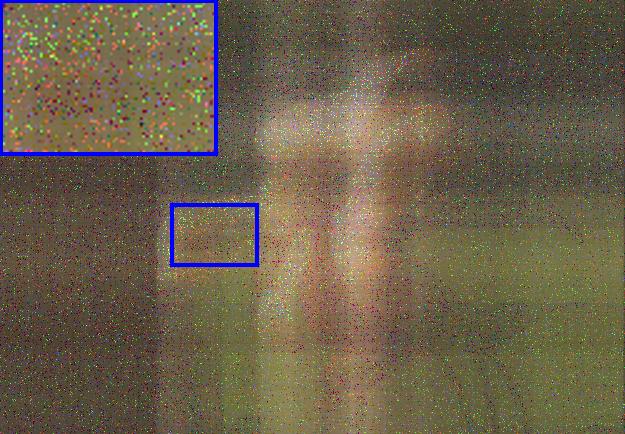}&
\includegraphics[width=0.585in, height=0.48in]{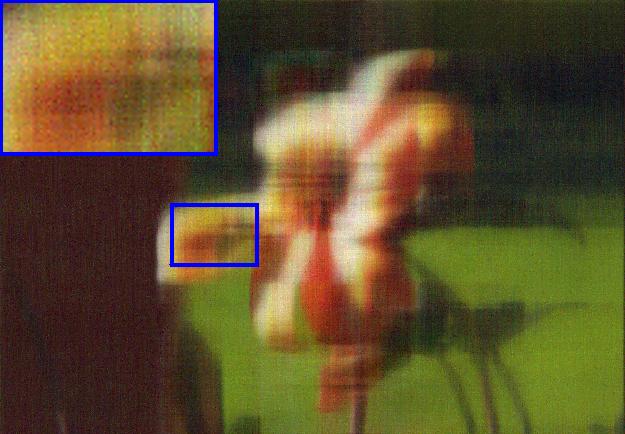}&
\includegraphics[width=0.585in, height=0.48in]{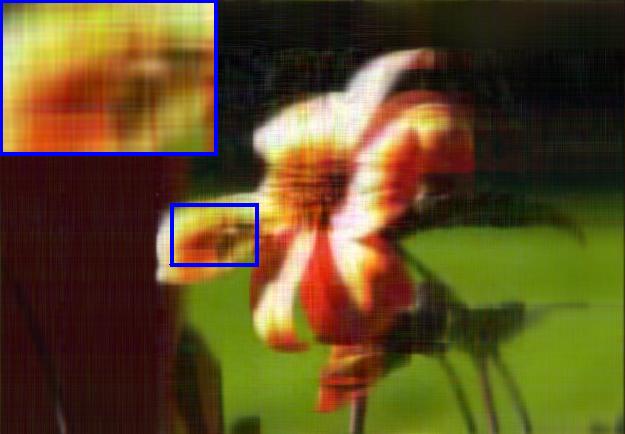}&
\includegraphics[width=0.585in, height=0.48in]{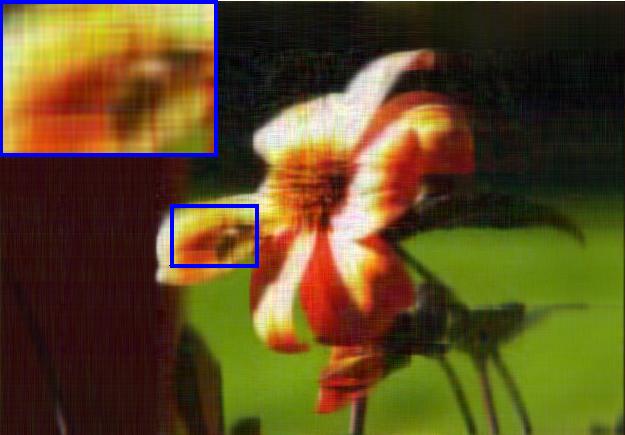}&
\includegraphics[width=0.585in, height=0.48in]{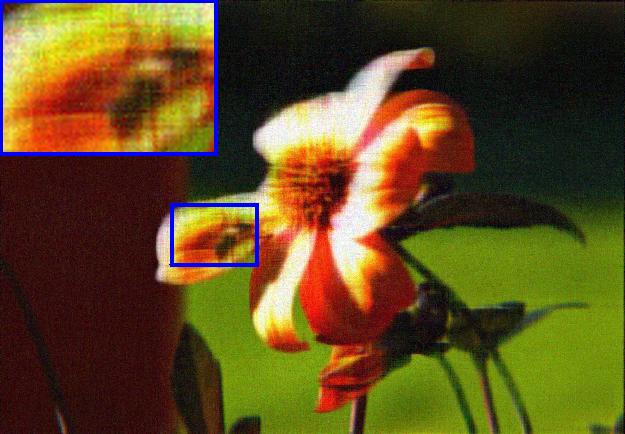}&
\includegraphics[width=0.585in, height=0.48in]{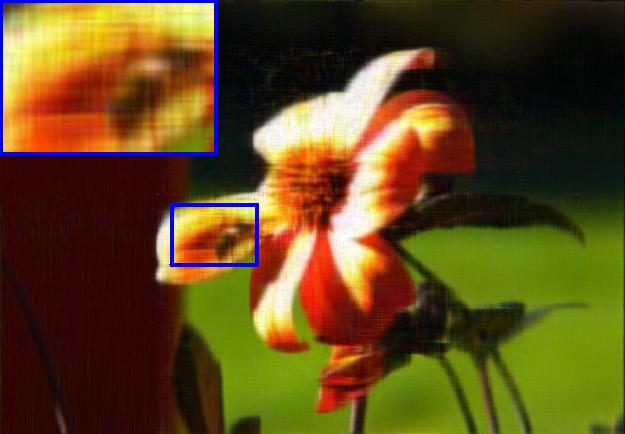}&
\includegraphics[width=0.585in, height=0.48in]{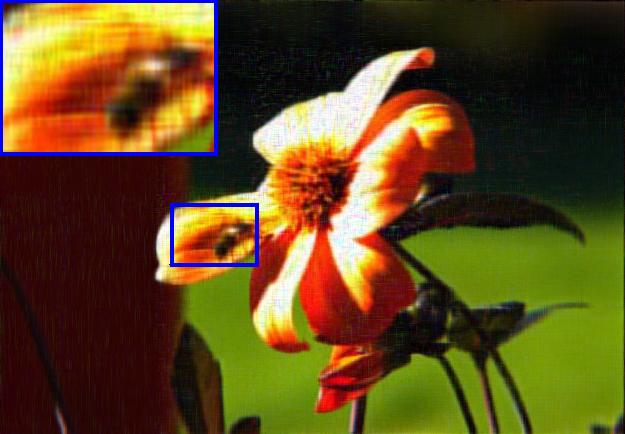}&
\includegraphics[width=0.585in, height=0.48in]{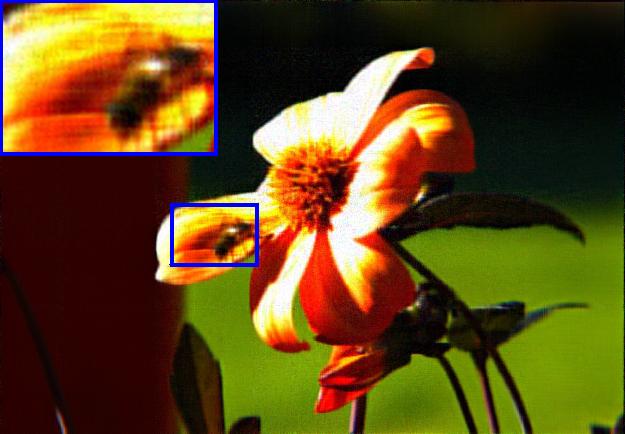}&
\includegraphics[width=0.585in, height=0.48in]{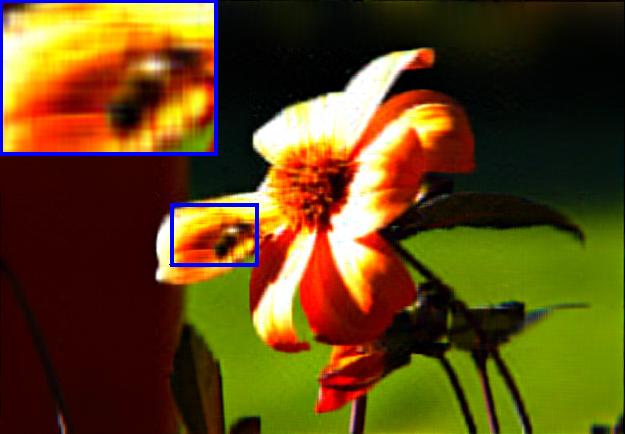}&
\includegraphics[width=0.585in, height=0.48in]{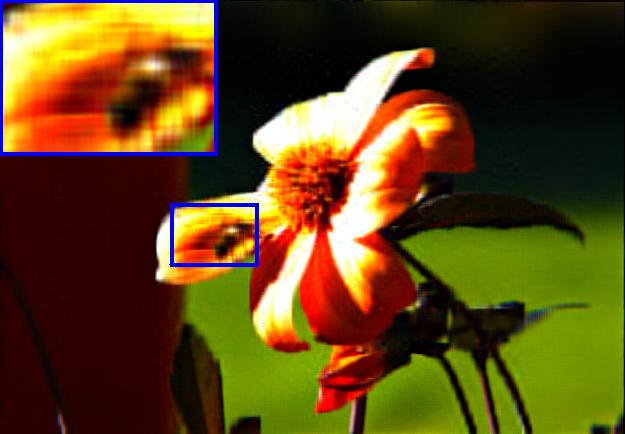}
&
\includegraphics[width=0.585in, height=0.48in,angle=0]{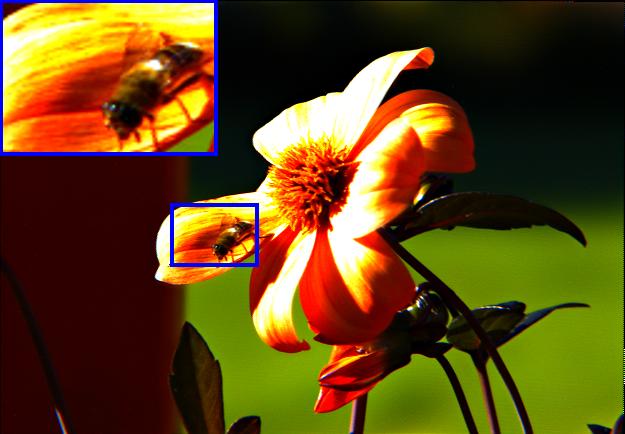}
\\
%

\includegraphics[width=0.585in, height=0.48in,angle=0]{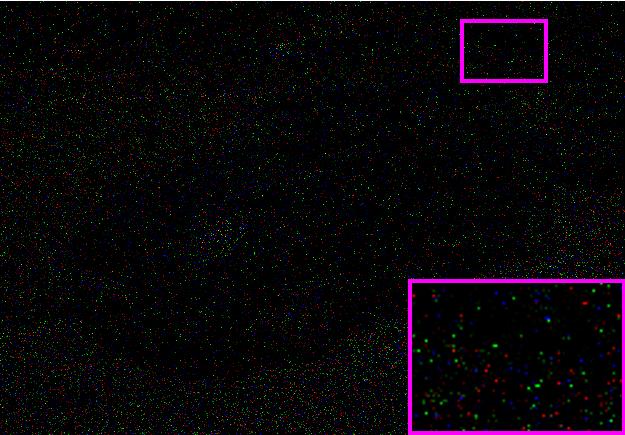}&
\includegraphics[width=0.585in, height=0.48in]{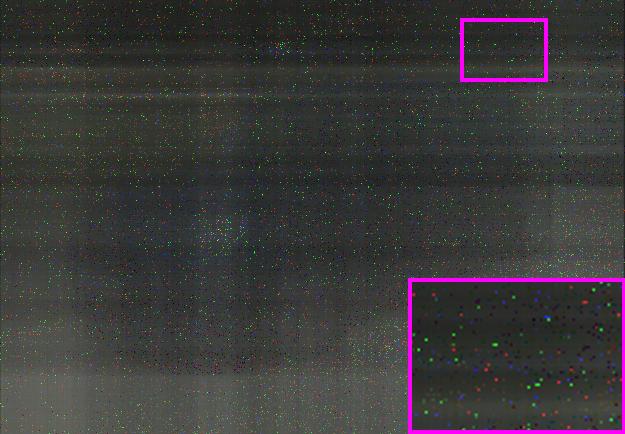}&
\includegraphics[width=0.585in, height=0.48in]{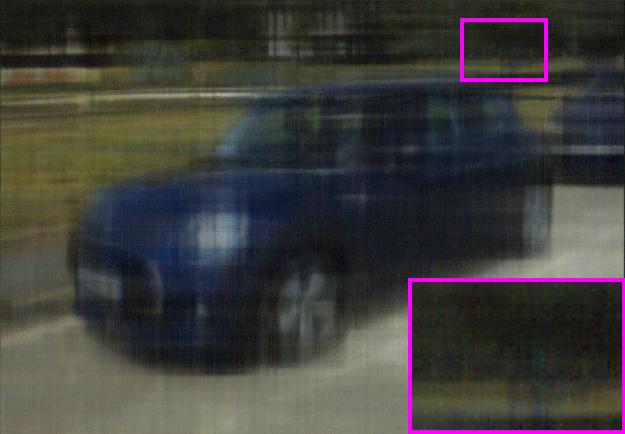}&
\includegraphics[width=0.585in, height=0.48in]{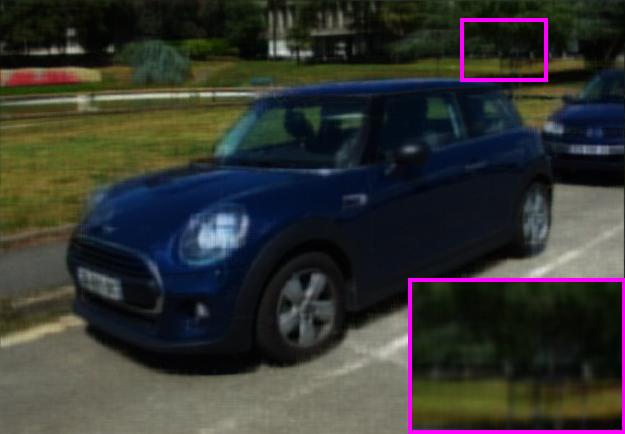}&
\includegraphics[width=0.585in, height=0.48in]{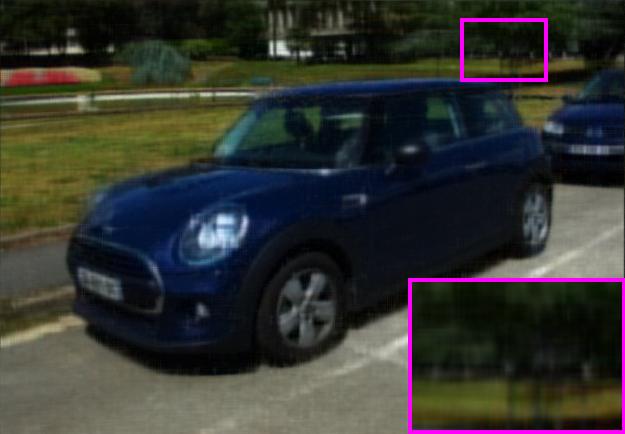}&
\includegraphics[width=0.585in, height=0.48in]{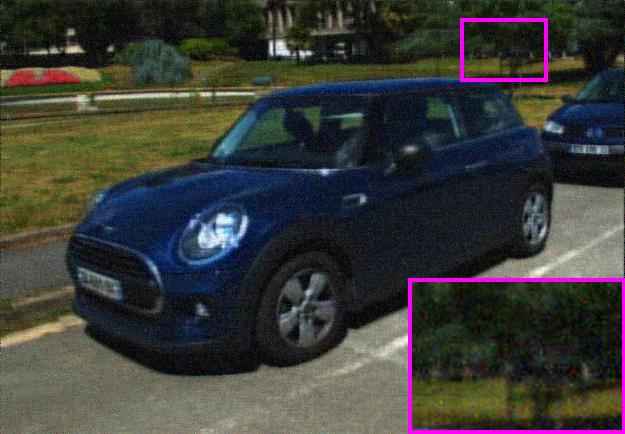}&
\includegraphics[width=0.585in, height=0.48in]{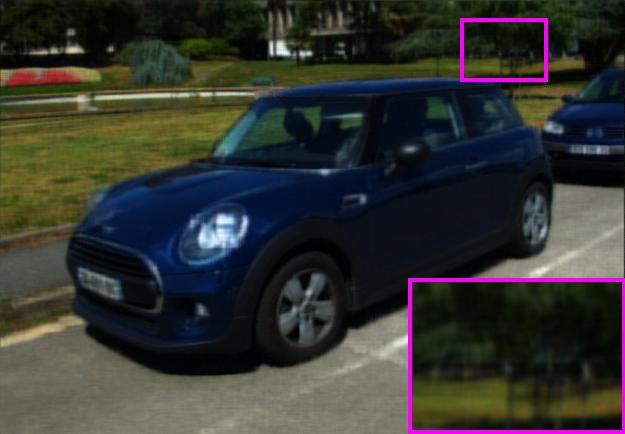}&
\includegraphics[width=0.585in, height=0.48in]{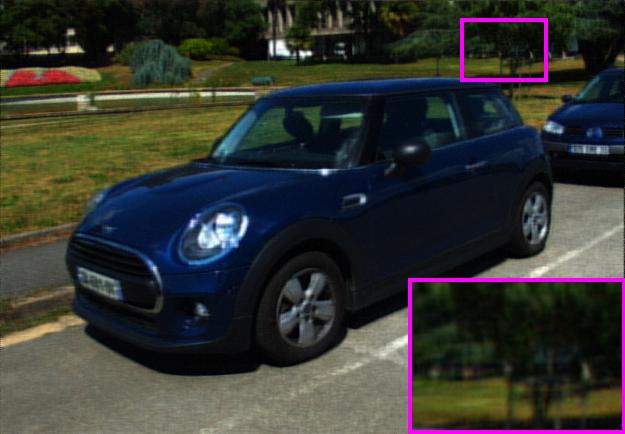}&
\includegraphics[width=0.585in, height=0.48in]{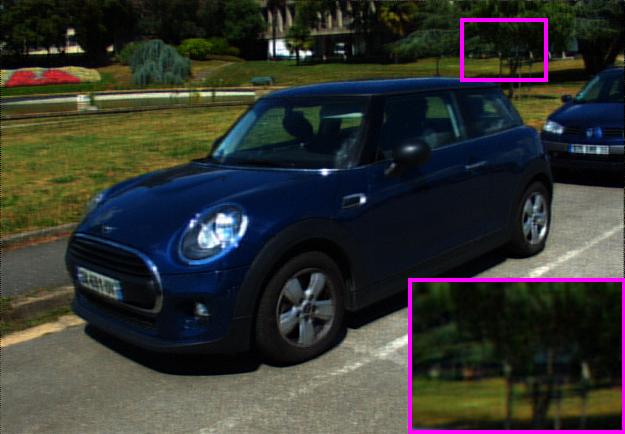}&
\includegraphics[width=0.585in, height=0.48in]{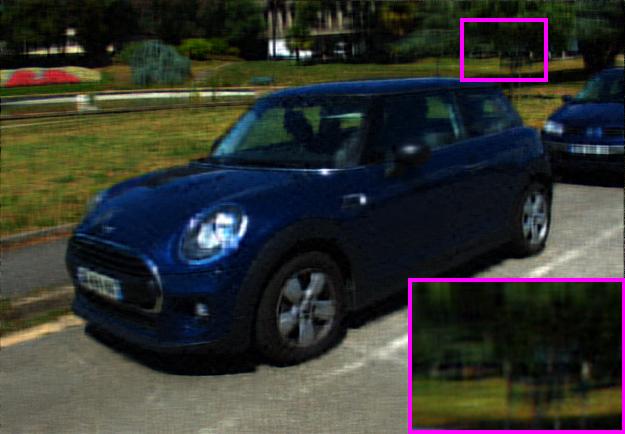}&
\includegraphics[width=0.585in, height=0.48in]{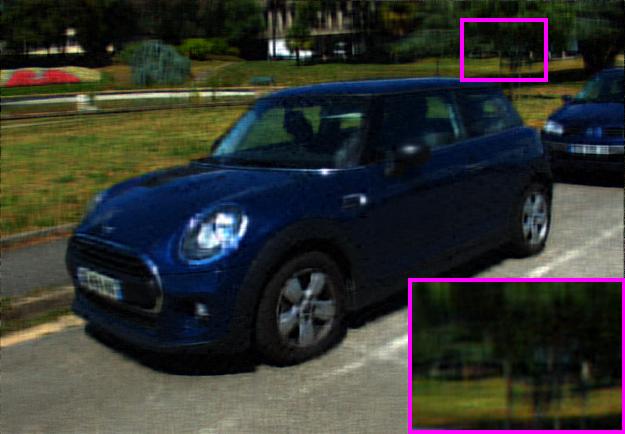}
&
\includegraphics[width=0.585in, height=0.48in,angle=0]{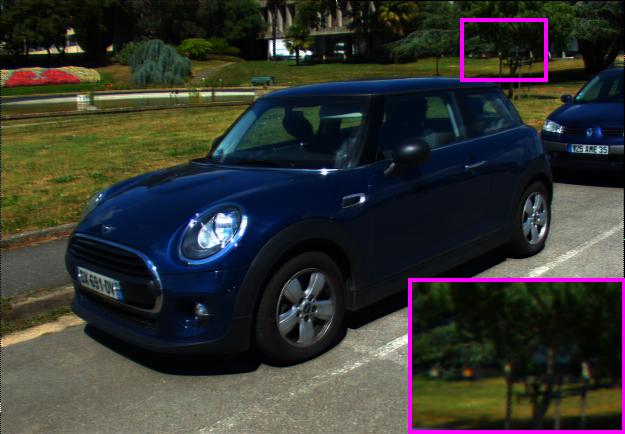}
\\

\includegraphics[width=0.585in, height=0.48in,angle=0]{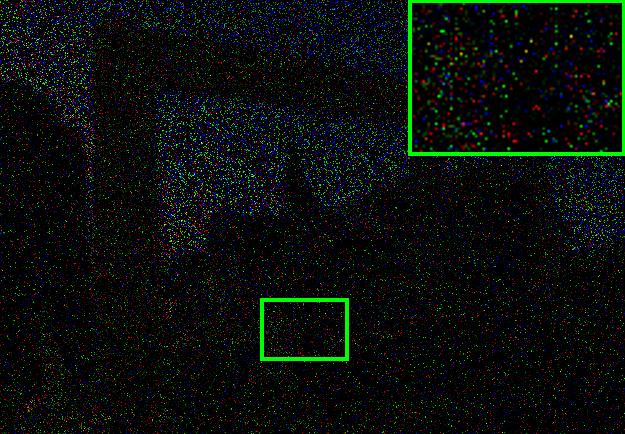}&
\includegraphics[width=0.585in, height=0.48in]{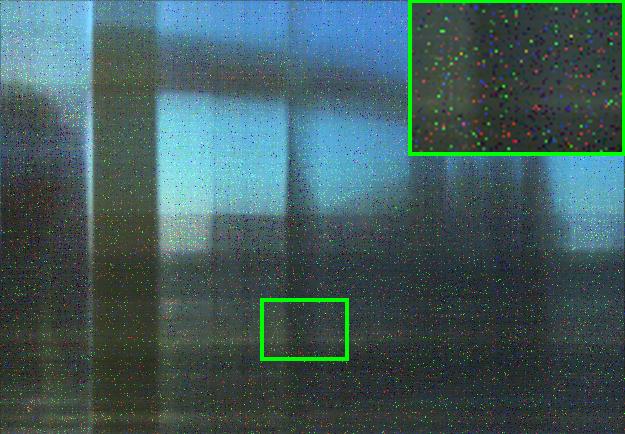}&
\includegraphics[width=0.585in, height=0.48in]{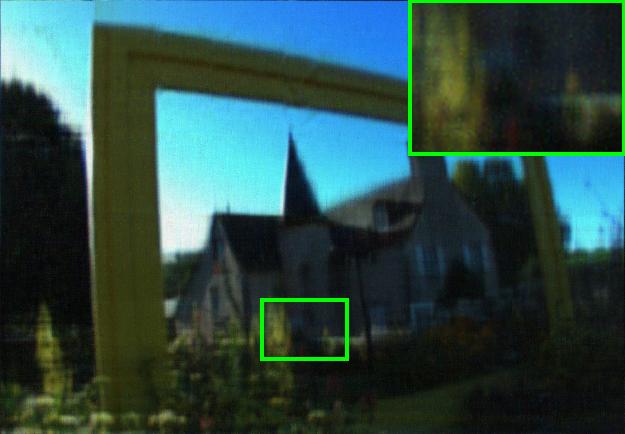}&
\includegraphics[width=0.585in, height=0.48in]{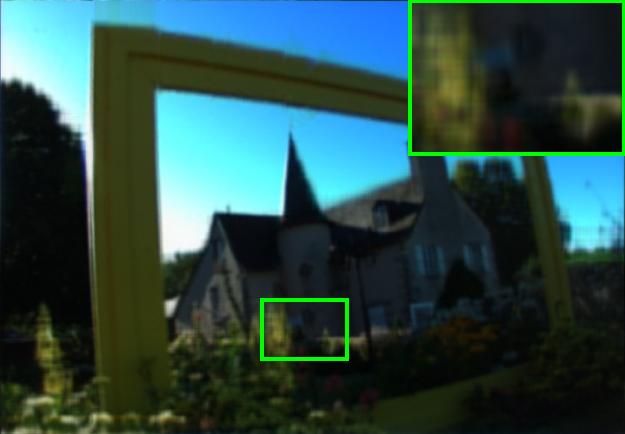}&
\includegraphics[width=0.585in, height=0.48in]{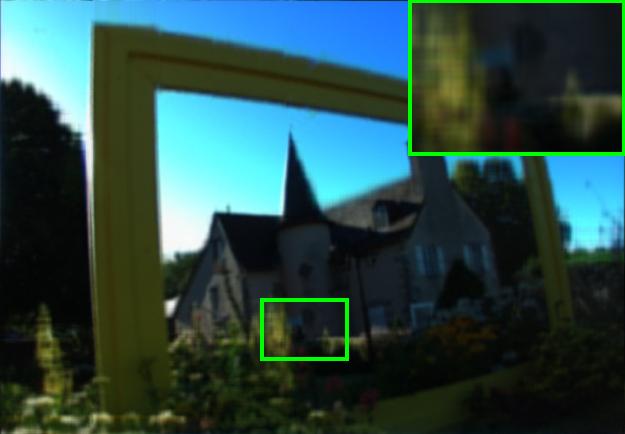}&
\includegraphics[width=0.585in, height=0.48in]{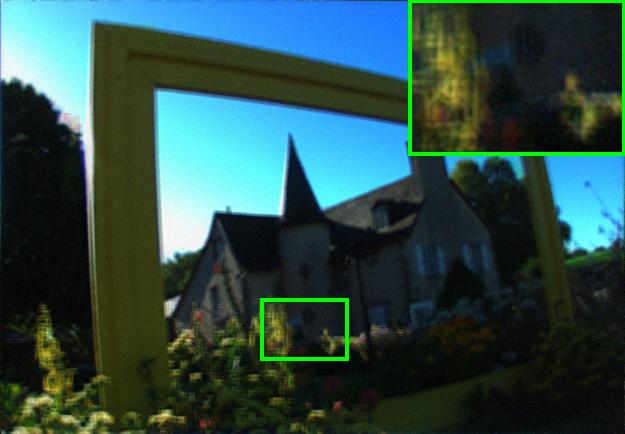}&
\includegraphics[width=0.585in, height=0.48in]{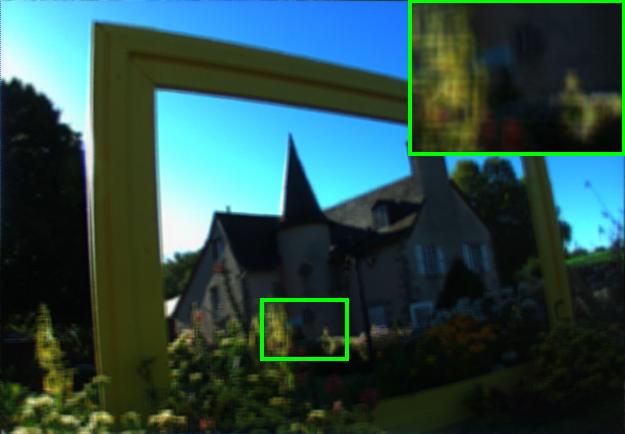}&
\includegraphics[width=0.585in, height=0.48in]{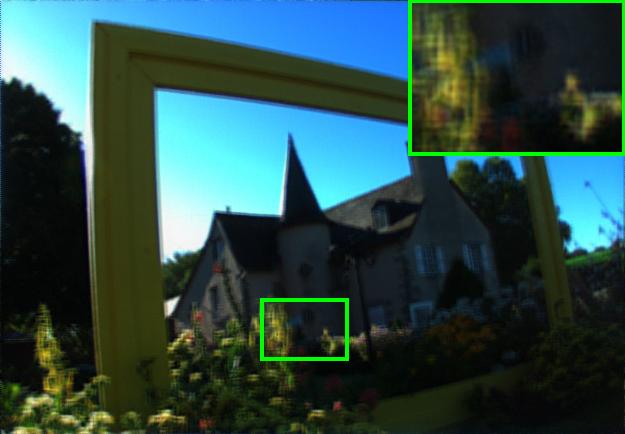}&
\includegraphics[width=0.585in, height=0.48in]{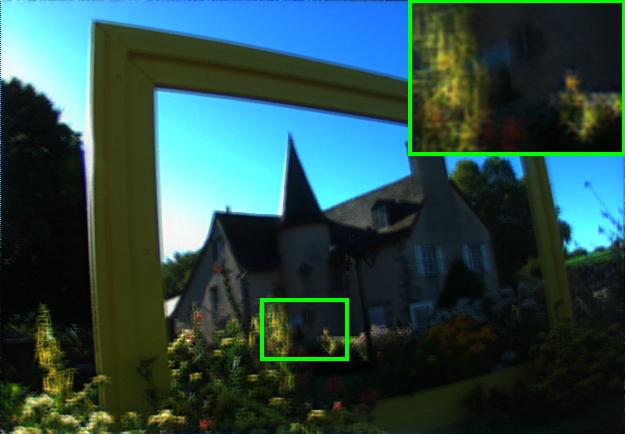}&
\includegraphics[width=0.585in, height=0.48in]{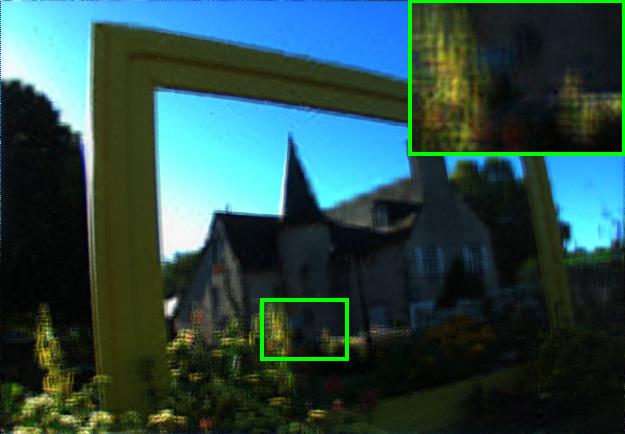}&
\includegraphics[width=0.585in, height=0.48in]{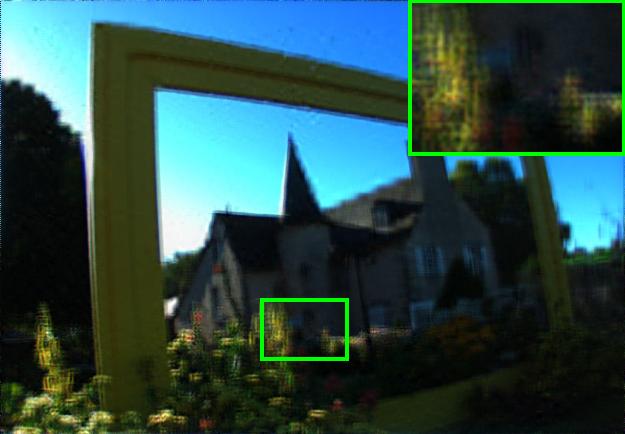}
&
\includegraphics[width=0.585in, height=0.48in,angle=0]{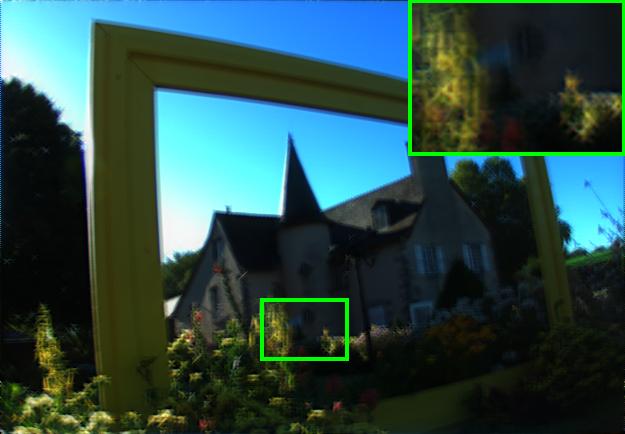}
\\

\textbf{\scriptsize{{Observed}}}  &
  \scriptsize{SNN+$\ell_1$}  &\scriptsize {TRNN+$\ell_1$}
&\scriptsize{TTNN+$\ell_1$} &\scriptsize {TSP-$k$+$\ell_1$}
 &\scriptsize{LNOP}&\scriptsize {NRTRM}&\scriptsize {HWTNN+$\ell_1$}
 & \textbf{\scriptsize {HWTSN+w$\ell_q$}}
  &

 \textbf{\tabincell{c}{ \scriptsize {HWTSN+w$\ell_q$} \\  \scriptsize{(BR)}
     }}

   &
   \textbf{\tabincell{c}{ \scriptsize {HWTSN+w$\ell_q$} \\  \scriptsize{(UR)}
     }}&
     \textbf{\scriptsize{Ground truth}}

%
\end{tabular}
\caption{
Visual comparison of various methods for LFIs recovery.
From top to bottom, the parameter pair $(sr, \tau)$  are 
$(0.05,0.5)$, $(0.1,0.5)$, $(0.05,0.3)$ and $(0.1,0.3)$, respectively.
Top row: 
the $(6,6)$-th frame of Bench.
 %
The second row:  the $(9,9)$-th frame of  Bee-1.
The third row: the $(12,12)$-th frame of  Mini.
Bottom row: the $(15,15)$-th frame of Framed.
}
\vspace{-0.15cm}
\label{fig_visual_lfi}
\end{figure*}
 \begin{table*}[htbp]
  \caption{
  The PSNR, SSIM values and CPU time
  obtained by  various RLRTC methods for different fifth-order LFIs.
  The best  and the second-best results are highlighted in
blue and red, respectively.
  }
  \label{lfi_index}

  \centering
\scriptsize
\renewcommand{\arraystretch}{0.88}
\setlength\tabcolsep{4.5pt}

\begin{tabular}{c c ccccccc  cccc   cccc |c }

     \hline
      \multicolumn{1}{c}{LFI-Name}
     &  \multicolumn{4}{|c|}{Bench} &\multicolumn{4}{c|}{Bee-1} &\multicolumn{4}{c|}{Framed} &\multicolumn{4}{c|}{Mini}
   & \multirow{3}{*}{
   \tabincell{c}{Average\\Time (s)}
   }\\
     \cline{1-1}
     \cline{2-5}
     \cline{6-9}
      \cline{10-13}
      \cline{14-17}
   \multicolumn{1}{c}{$sr$}
     &\multicolumn{2}{|c|}{$5\%$}   &\multicolumn{2}{c|}{$10\%$}  &
     \multicolumn{2}{c|}{$5\%$}&\multicolumn{2}{c|}{$10\%$}&
     \multicolumn{2}{c|}{$5\%$}&\multicolumn{2}{c|}{$10\%$}&
     \multicolumn{2}{c|}{$5\%$}&\multicolumn{2}{c|}{$10\%$}\\
     \cline{1-1}
     \cline{2-5}
     \cline{6-9}
      \cline{10-13}
      \cline{14-17}
   \multicolumn{1}{c}{$\tau$}
    &
    \multicolumn{1}{|c|}{$30\%$}&\multicolumn{1}{c|}{$50\%$}  &
    \multicolumn{1}{c|}{$30\%$}&\multicolumn{1}{c|}{$50\%$}  &
    \multicolumn{1}{c|}{$30\%$}&\multicolumn{1}{c|}{$50\%$}  &
    \multicolumn{1}{c|}{$30\%$}&\multicolumn{1}{c|}{$50\%$}
    &
    \multicolumn{1}{c|}{$30\%$}&\multicolumn{1}{c|}{$50\%$}  &
    \multicolumn{1}{c|}{$30\%$}&\multicolumn{1}{c|}{$50\%$}  &
    \multicolumn{1}{c|}{$30\%$}&\multicolumn{1}{c|}{$50\%$}  &
    \multicolumn{1}{c|}{$30\%$}&\multicolumn{1}{c|}{$50\%$}
    \\
    \hline
     \hline
     \multicolumn{1}{c}{SNN+$\ell_1$}
     &

\tabincell{c}{ 14.29      \\ 0.176  }   &    \tabincell{c}{11.94\\  0.091   }&
\tabincell{c}{  14.81 \\    0.267 }   &    \tabincell{c}{ 13.69\\ 0.168}&

\tabincell{c}{12.72  \\ 0.137  }   &    \tabincell{c}{    11.07   \\ 0.081  }&
\tabincell{c}{  13.39 \\0.179   }   &    \tabincell{c}{ 11.74\\  0.128}&

\tabincell{c}{ 12.76    \\0.203  }   &    \tabincell{c}{ 11.47 \\0.112   }&
\tabincell{c}{ 13.99  \\ 0.269    }   &    \tabincell{c}{11.67\\  0.179}&

\tabincell{c}{  14.29  \\ 0.195   }   &    \tabincell{c}{ 12.92  \\0.132     }&
\tabincell{c}{15.99  \\0.245  }   &    \tabincell{c}{15.37\\ 0.197 }

&

10871
\\

%
%
%
%
%
%
%
%
%
%
%
%
%
%

      \hline
      \multicolumn{1}{c}{TRNN+$\ell_1$}
    &


   \tabincell{c}{ 18.16    \\0.446     }   &
\tabincell{c}{ 14.89   \\   0.312}   &    \tabincell{c}{21.27  \\  0.574    }& \tabincell{c}{ 17.29\\  0.337}&

\tabincell{c}{ 18.04      \\0.356    }   &
\tabincell{c}{  13.71  \\    0.224}   &    \tabincell{c}{22.74  \\0.474    }&\tabincell{c}{ 16.66\\  0.316}&

\tabincell{c}{ 19.07    \\ 0.511     }   &
\tabincell{c}{ 14.02   \\  0.355  }   &   \tabincell{c}{23.07 \\  0.643 }&  \tabincell{c}{17.32\\  0.469 }&

\tabincell{c}{ 20.83  \\0.483    }   &
\tabincell{c}{  16.16   \\0.394    }   &     \tabincell{c}{ 25.25  \\0.658     }& \tabincell{c}{19.72\\0.408 }
 &
    9111
\\

%
%
%
%
%
%
%
%
%
 \hline
    \multicolumn{1}{c}{TTNN+$\ell_1$}
   &

   \tabincell{c}{ 21.65   \\0.704     }   &
\tabincell{c}{   16.48      \\  0.434     }   &    \tabincell{c}{23.04      \\ 0.746      }&\tabincell{c}{ 20.72\\0.608 }&

\tabincell{c}{ 22.44  \\ 0.776   }   &
\tabincell{c}{   16.24      \\0.353    }   &  \tabincell{c}{     24.74     \\   0.831      }&  \tabincell{c}{20.75\\ 0.464 }&

\tabincell{c}{ 23.33      \\0.797    }   &
\tabincell{c}{  16.95     \\ 0.516      }   &   \tabincell{c}{ 25.37      \\  0.857      }& \tabincell{c}{  22.13\\0.667 }&

\tabincell{c}{25.67         \\ 0.775     }   &
\tabincell{c}{  17.51   \\  0.486     }   &    \tabincell{c}{ 27.45     \\ 0.818    }& \tabincell{c}{   22.92\\  0.638 }
&  4708
\\
%
%
%
%
%
%
 \hline
   \multicolumn{1}{c}{TSP-$k$+$\ell_1$}
   &

   \tabincell{c}{22.33    \\0.687   }   &
\tabincell{c}{17.43    \\0.415   }   &    \tabincell{c}{24.02   \\ 0.785     }& \tabincell{c}{21.21  \\0.611 }&
\tabincell{c}{ 22.88\\0.734   }   &
\tabincell{c}{ 17.08\\0.333    }   &   \tabincell{c}{25.62 \\  0.843    }&
  \tabincell{c}{21.32    \\ 0.456}&

\tabincell{c}{ 23.43\\0.772     }   &
\tabincell{c}{ 17.81   \\0.491   }   &
 \tabincell{c}{ 25.67   \\ 0.863     }&  \tabincell{c}{22.24\\  0.661}&

\tabincell{c}{26.36  \\0.757   }   &
\tabincell{c}{ 18.04 \\0.489  }   &
 \tabincell{c}{ 29.02  \\ 0.857     }&
 \tabincell{c}{ 23.93\\   0.663}
&6798
\\
%
%
%
%
%
%

     \hline
     \multicolumn{1}{c}{LNOP}
     &


   \tabincell{c}{ 21.87        \\0.628    }   &
\tabincell{c}{ 18.35   \\  0.436      }   &   \tabincell{c}{  27.02       \\   0.813     }& \tabincell{c}{     21.84\\0.649 }&

\tabincell{c}{   22.46       \\0.658  }   &
\tabincell{c}{   18.08     \\  0.348     }   &   \tabincell{c}{28.37     \\ 0.879     }& \tabincell{c}{  21.87\\  0.512}&

\tabincell{c}{   22.88   \\0.674     }   &
\tabincell{c}{  19.05       \\   0.493      }   & \tabincell{c}{   28.83      \\0.839    }&   \tabincell{c}{22.79\\  0.637}&

\tabincell{c}{ 24.58     \\   0.658     }   &
\tabincell{c}{   20.97      \\ 0.529    }   &     \tabincell{c}{  30.89     \\ 0.853       }& \tabincell{c}{ 25.73\\   0.725}

&
%
 6465
\\

 \hline
     \multicolumn{1}{c}{NRTRM}
     &

   \tabincell{c}{ 23.67   \\ 0.764 }   &    \tabincell{c}{  19.41  \\ 0.458}&
\tabincell{c}{ 26.41 \\ 0.838}   &    \tabincell{c}{22.63 \\ 0.703}&

\tabincell{c}{  25.46  \\  0.834   }   &    \tabincell{c}{19.01\\ 0.362}&
\tabincell{c}{ 28.89 \\ 0.891}   &    \tabincell{c}{ 23.73 \\0.599 }&

\tabincell{c}{ 26.05  \\ 0.864  }   &    \tabincell{c}{20.53\\  0.541 }&
\tabincell{c}{ 29.11\\  0.927}   &    \tabincell{c}{24.84\\  0.789 }&

\tabincell{c}{ 28.30  \\0.829  }   &    \tabincell{c}{20.31\\ 0.518}&
\tabincell{c}{ 31.92 \\  0.913 }   &    \tabincell{c}{  26.59\\0.773 }
&6267
\\

%
%
%
%
%
%
%
%
%
%
%

     \hline
     \multicolumn{1}{c}{HWTNN+$\ell_1$}
     &

     \tabincell{c}{  {\textcolor[rgb]{1.00,0.00,0.00} {25.63}}             \\   {\textcolor[rgb]{1.00,0.00,0.00}{0.808}}          }   & \tabincell{c}{   21.94     \\      0.543   }
     &\tabincell{c}{     {\textcolor[rgb]{1.00,0.00,0.00}{27.77}}    \\   {\textcolor[rgb]{1.00,0.00,0.00}{0.857}}}   & \tabincell{c}{     23.95 \\     0.705  }   &

\tabincell{c}{      28.24            \\    0.839              }   & \tabincell{c}{    22.29   \\    0.407    }
&\tabincell{c}{     {\textcolor[rgb]{1.00,0.00,0.00}{30.53}}    \\    {\textcolor[rgb]{1.00,0.00,0.00}{0.896}}
 }   & \tabincell{c}{    25.25   \\    0.564    }   &

\tabincell{c}{ {\textcolor[rgb]{1.00,0.00,0.00}{28.45}}          \\
        {\textcolor[rgb]{1.00,0.00,0.00}{0.893}}         }   & \tabincell{c}{    23.03      \\   0.583       }
 &\tabincell{c}{     {\textcolor[rgb]{1.00,0.00,0.00}{31.08}}   \\    {\textcolor[rgb]{1.00,0.00,0.00}{0.931}}}   & \tabincell{c}{      26.39 \\   0.767    }   &

\tabincell{c}{    {\textcolor[rgb]{1.00,0.00,0.00}{30.89}}            \\    {\textcolor[rgb]{1.00,0.00,0.00}{0.872}}        }   & \tabincell{c}{   25.41      \\    0.629     }
 &\tabincell{c}{     {\textcolor[rgb]{1.00,0.00,0.00}{33.29}}   \\   {\textcolor[rgb]{1.00,0.00,0.00}{0.919}} }   & \tabincell{c}{       28.44 \\ 0.776     }   &
      5754
%
%
%
\\

  \hline
  \multicolumn{1}{c}{\textbf{HWTSN+w$\ell_q$}}
   &

\tabincell{c}{{\textcolor[rgb]{0,0,1}{26.85}}          \\ \textcolor[rgb]{0,0,1}{0.834}            }
&    \tabincell{c}{  \textcolor[rgb]{0,0,1}  {23.44} \\ \textcolor[rgb]{0,0,1} {0.671}           }&
\tabincell{c}{ \textcolor[rgb]{0,0,1}{30.13}     \\ \textcolor[rgb]{0,0,1}{0.892}   }   &    \tabincell{c}{  \textcolor[rgb]{0,0,1}{26.29}\\  \textcolor[rgb]{0,0,1}{0.808}   }&

\tabincell{c}{  \textcolor[rgb]{0,0,1}{{29.56}}             \\ \textcolor[rgb]{0,0,1}{0.889}           }   &    \tabincell{c}{ \textcolor[rgb]{0,0,1}{25.28}    \\  \textcolor[rgb]{0,0,1}{0.651}   }&
\tabincell{c}{ \textcolor[rgb]{0,0,1}{33.03}  \\ \textcolor[rgb]{0,0,1}{0.938}    }   &    \tabincell{c}{  {27.69}\\ {0.723}  }&

\tabincell{c}{ \textcolor[rgb]{0,0,1}{29.95}        \\  \textcolor[rgb]{0,0,1}{0.916}       }   &    \tabincell{c}{ \textcolor[rgb]{0,0,1}{25.55}  \\  \textcolor[rgb]{0,0,1}{0.743}   }&
\tabincell{c}{  \textcolor[rgb]{0,0,1}{33.78}    \\ \textcolor[rgb]{0,0,1}{0.961}   }   &    \tabincell{c}{\textcolor[rgb]{0,0,1}{29.89}\\  \textcolor[rgb]{0,0,1}{0.915}  }&

\tabincell{c}{ \textcolor[rgb]{0,0,1}{32.25}           \\\textcolor[rgb]{0,0,1}{0.902}         }   &    \tabincell{c}{ \textcolor[rgb]{0,0,1}{28.99}   \\  \textcolor[rgb]{0,0,1}{0.766}    }&
\tabincell{c}{  \textcolor[rgb]{0,0,1}{35.82}    \\ \textcolor[rgb]{0,0,1}{0.952}    }   &    \tabincell{c}{ \textcolor[rgb]{0,0,1}{32.78} \\ \textcolor[rgb]{0,0,1}{0.898}}

&

       7080
\\

%
%
%

\hline
\multicolumn{1}{c}{
   \tabincell{c}{\textbf{HWTSN+w$\ell_q$(BR)}}}

     &

\tabincell{c}{24.83       \\0.743      }   &    \tabincell{c}{ 22.86   \\  0.645   }&
\tabincell{c}{ 25.36  \\0.746   }   &    \tabincell{c}{24.62\\ 0.721}&

\tabincell{c}{ 28.16    \\ 0.846       }   &    \tabincell{c}{   24.85 \\ 0.661  }&
\tabincell{c}{28.69   \\0.874   }   &    \tabincell{c}{ {\textcolor[rgb]{1.00,0.00,0.00}{27.79}}\\ 
{\textcolor[rgb]{1.00,0.00,0.00}{0.847}}
}&

\tabincell{c}{28.15     \\0.863    }   &    \tabincell{c}{ 25.25  \\ 0.735   }&
\tabincell{c}{29.01  \\0.884  }   &    \tabincell{c}{27.98\\ 0.868}&

\tabincell{c}{   29.39    \\0.808       }   &    \tabincell{c}{ 27.29 \\0.723   }&
\tabincell{c}{  29.55  \\0.817  }   &    \tabincell{c}{29.19\\  0.811}
&

\textcolor[rgb]{0.00,0.00,1.00}{2869}

\\

%
%
%
%
%
%
\hline
\multicolumn{1}{c}{
   \tabincell{c}{\textbf{HWTSN+w$\ell_q$(UR)}}}
&

\tabincell{c}{ 24.92     \\0.741      }   &    \tabincell{c}{  {\textcolor[rgb]{1.00,0.00,0.00}{22.96}}  \\  {\textcolor[rgb]{1.00,0.00,0.00}{0.649}} }&
\tabincell{c}{25.46   \\0.749 }   &    \tabincell{c}{  {\textcolor[rgb]{1.00,0.00,0.00}{24.78}} \\ {\textcolor[rgb]{1.00,0.00,0.00}{0.728}}}&

\tabincell{c}{ {\textcolor[rgb]{1.00,0.00,0.00}{28.74}}    \\ {\textcolor[rgb]{1.00,0.00,0.00}{0.846}}      }   &    \tabincell{c}{ {\textcolor[rgb]{1.00,0.00,0.00}{24.94}}   \\ 
{\textcolor[rgb]{1.00,0.00,0.00}{0.663}}
  }&
\tabincell{c}{28.76  \\ 0.874 }   &    \tabincell{c}{ {\textcolor[rgb]{0,0,1}{27.88}}\\
{\textcolor[rgb]{0,0,1}{0.848}}
}&

\tabincell{c}{28.23   \\ 0.864    }   &    \tabincell{c}{ {\textcolor[rgb]{1.00,0.00,0.00}{25.28}}  \\  {\textcolor[rgb]{1.00,0.00,0.00}{0.736}}}&
\tabincell{c}{29.19   \\ 0.886 }   &    \tabincell{c}{ {\textcolor[rgb]{1.00,0.00,0.00}{28.29}}\\  {\textcolor[rgb]{1.00,0.00,0.00}{0.869}}}&

\tabincell{c}{29.46   \\  0.809      }   &    \tabincell{c}{ {\textcolor[rgb]{1.00,0.00,0.00}{27.38}}   \\ {\textcolor[rgb]{1.00,0.00,0.00}{0.725}}}&
\tabincell{c}{ 29.67    \\0.816}   &    \tabincell{c}{\textcolor[rgb]{1.00,0.00,0.00}{29.25}\\\textcolor[rgb]{1.00,0.00,0.00}{0.812}}
&
 \textcolor[rgb]{1.00,0.00,0.00}{2944}

\\

%
%
%
%

     \hline

\end{tabular}
\begin{tablenotes}
\item[**]
In each RLRTC method,
the top represents the PSNR values   while 
the bottom denotes the SSIM values.
\end{tablenotes}
\vspace{-0.45cm}
\end{table*}

%
\textbf{Parameters Setting:} 
The transform
in  TTNN+$\ell_1$, NRTRM,  HWTNN+$\ell_1$, and {HWTSN+w$\ell_q$}
 is set as the FFT for consistency.
The parameter {\bf{$\lambda$}} of \textbf{SNN+$\ell_1$} 
is set as 
$\textbf{$\lambda$}\in \{[10,10,1,1]*\alpha,  [10,10,1,1,1]*\alpha\}$,
$\alpha \in \{1, 3,5,8,10\}$.
For \textbf{TRNN+$\ell_1$}, we set 
$\lambda \in \{0.018, 0.015, 0.012,0.01, 0.009,0.007\}$.
 For \textbf{TTNN+$\ell_1$},
 we set
$\lambda=\varsigma/ (\max{(n_1,n_2)} \cdot \prod_{i=3}^{d} n_i)^{1/2} $,
$\varsigma \in \{1, 1.2, 1.5, 1.8, 2\}$.
For \textbf{TSP-$k$+$\ell_1$},  
we set
 $k\in \{3,4,5\},  \lambda \in \{100,200,300,400\}$.
For \textbf{LNOP},
the parameter $p$ of $\ell_p$-ball projection 
is set to be 
 $0.7$, $\epsilon=500, \lambda=10^{7}$.
For \textbf{NRTRM}, the
minimax concave penalty (MCP)
function is utilized  in both 
regularizers $G_1, G_2$,
the parameter $\eta$ of MCP is chosen as $\max(n_1,n_2)/\alpha^{k}$ for $G_1$ and $1/\beta^{k}$ for $G_2$, respectively;
$c \in \{0.7,0.9,1.4\}$,
and
$\lambda={\kappa}  / 
(\max{(n_1,n_2)}
\cdot \prod_{i=3}^{d} n_i )^{1/2}
  $, 
  $\kappa \in \{1.2, 1.5, 1.8, 2,2.2, 2.5\}$.
 For \textbf{HWTNN+$\ell_1$},
 we set
$c=\max(n_1,n_2), \epsilon=10^{-16}$ and
$\lambda=\theta/ (\max{(n_1,n_2)} \cdot \prod_{i=3}^{d} n_i)^{1/2} $,
$\theta \in \{25,30, 35, 40, 45, 50, 55, 60,65,70\}$.
For \textbf{our algorithms},
 we set
%
 $t=1, \vartheta=1.15,
  \beta^0=10^{-3}, \beta^{\max}=10^{8},
  \varpi=10^{-4},\epsilon_{1}=\epsilon_{2}=10^{-16}$,
$c_1= \omega \cdot \max(n_1,n_2)$, $\omega \in  \{0.5,1,2,5,10,15,20,25,30\}$,
$c_2=1$,
$\lambda=\xi/ (\max{(n_1,n_2)} \cdot  \prod_{i=3}^{d} n_i)^{1/2} $, 
$\xi\in \{1, 3,5,6, 8, 10,12,15\}$,
$k=100, b=20$ for 
 multitemporal remote sensing images
($k=50, b=10$ for 
color videos and light field images).
%
The
adjustable parameters
$p$ and $q$ 
are 
set to be inversely proportional to
the constant $c_1$, respectively.
In other words, as $p$ and $q$  go up, $\omega$ goes down.

\subsubsection{\textbf{Application in Light Field Images Recovery}}
 In this experiment,
 we choose four 
 fifth-order light field images
 (LFIs)
including Bench, Bee-1, Framed and Mini 
to 
showcase the superiority and effectiveness of the proposed algorithms.
These LFIs with the size of $434\times 625\times 3\times 15\times15$ can be downloaded from the lytro illum light field dataset
website
\footnote{\url{https://www.irisa.fr/temics/demos/IllumDatasetLF/index.html}}.

\begin{figure*}[!htbp]
\renewcommand{\arraystretch}{0.4}
\setlength\tabcolsep{0.4pt}
\centering
\begin{tabular}{ccc ccc  cccccc}
\centering

\includegraphics[width=0.585in, height=0.48in,angle=0]{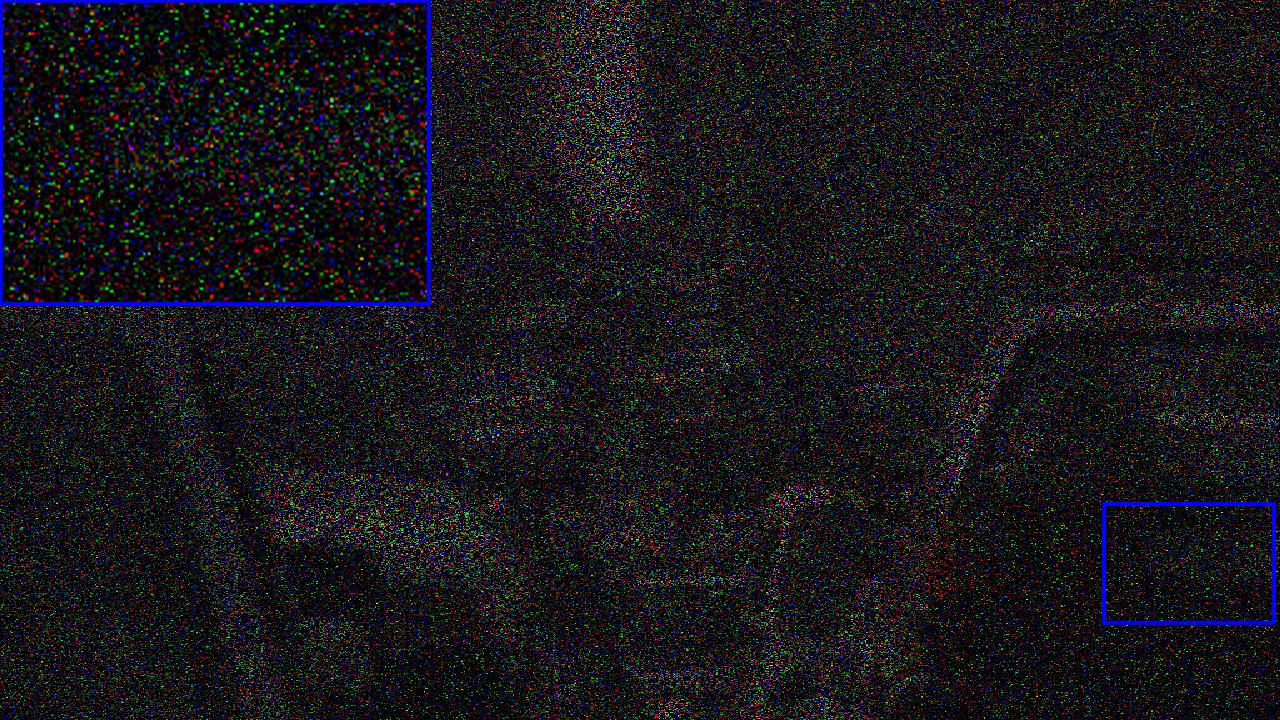}&
\includegraphics[width=0.585in, height=0.48in]{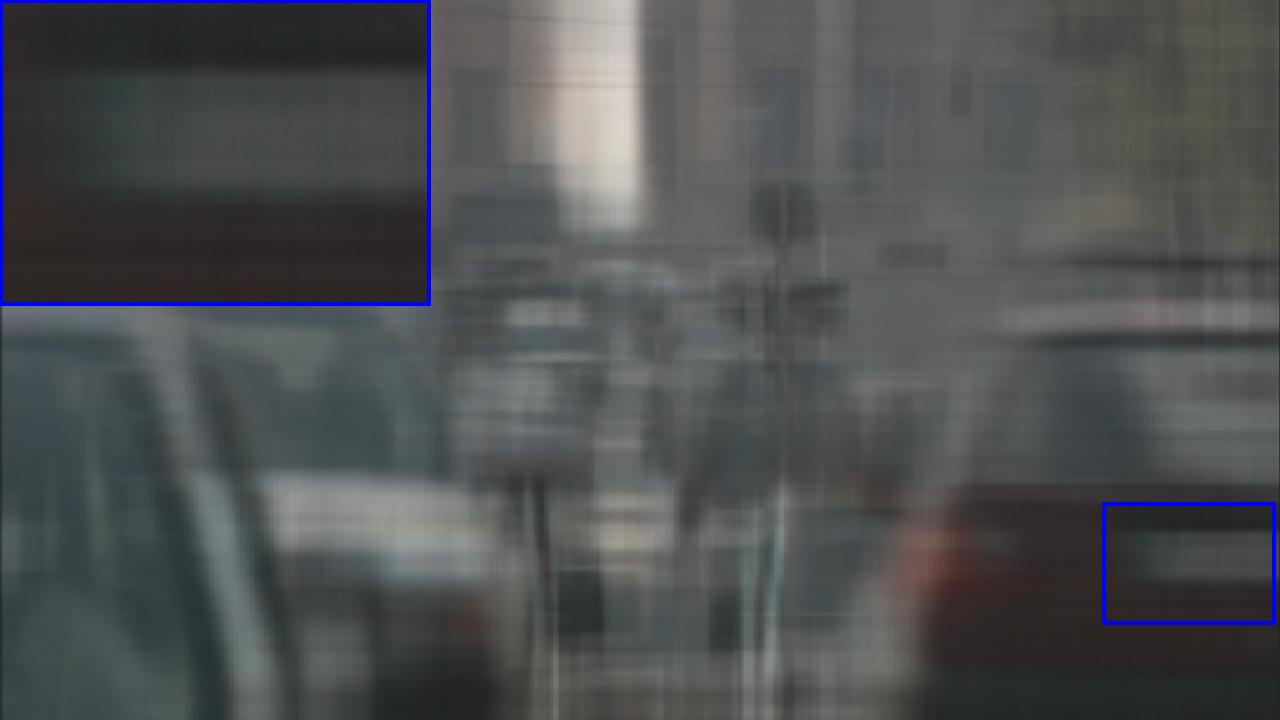}&
\includegraphics[width=0.585in, height=0.48in]{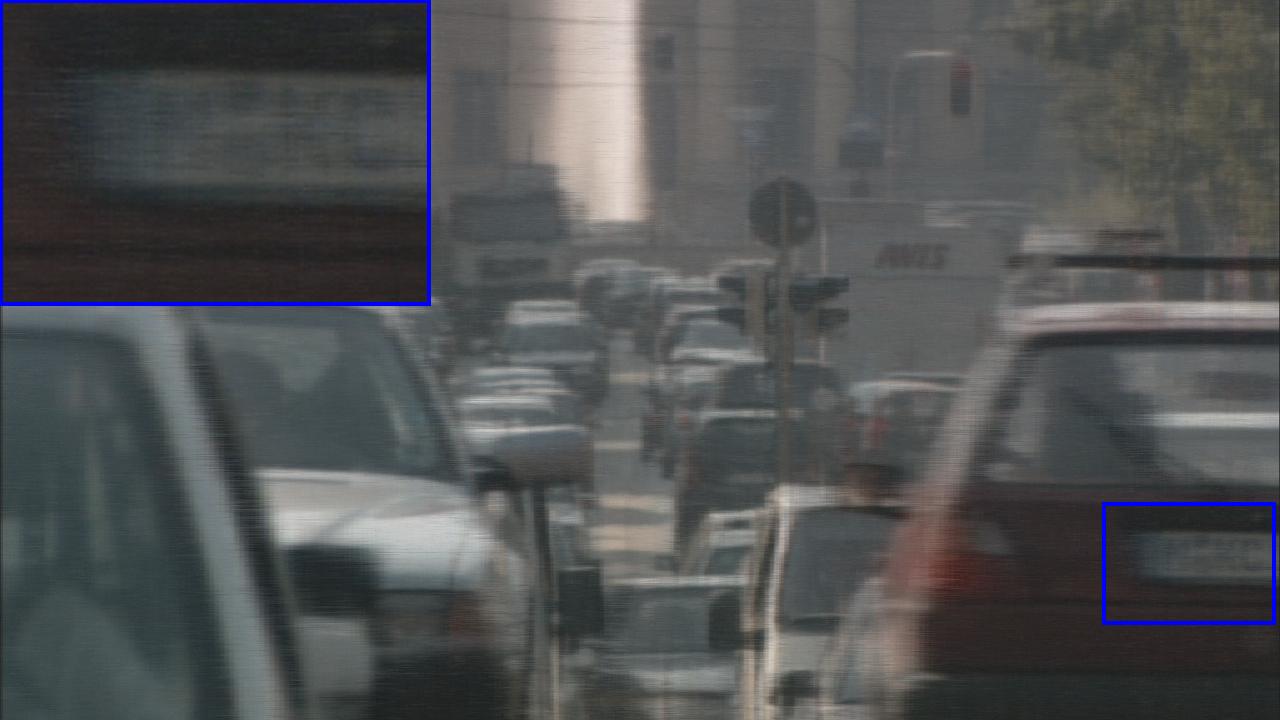}&
\includegraphics[width=0.585in, height=0.48in]{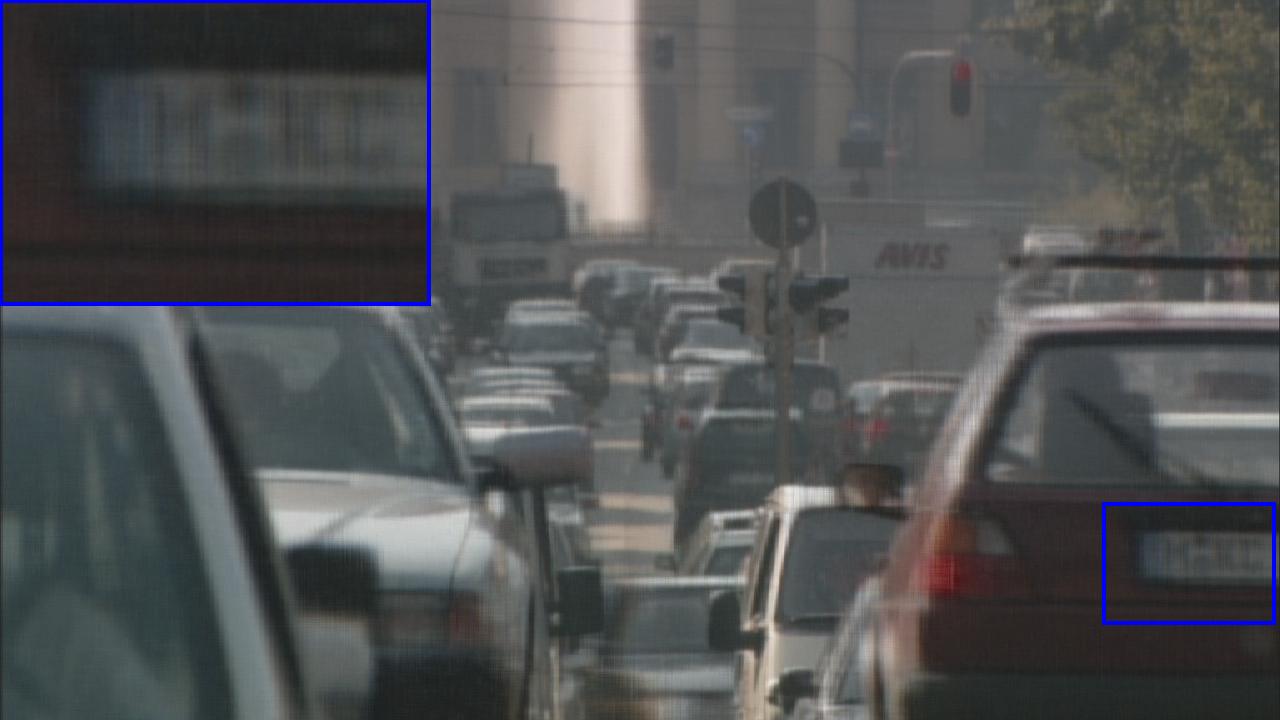}&
\includegraphics[width=0.585in, height=0.48in]{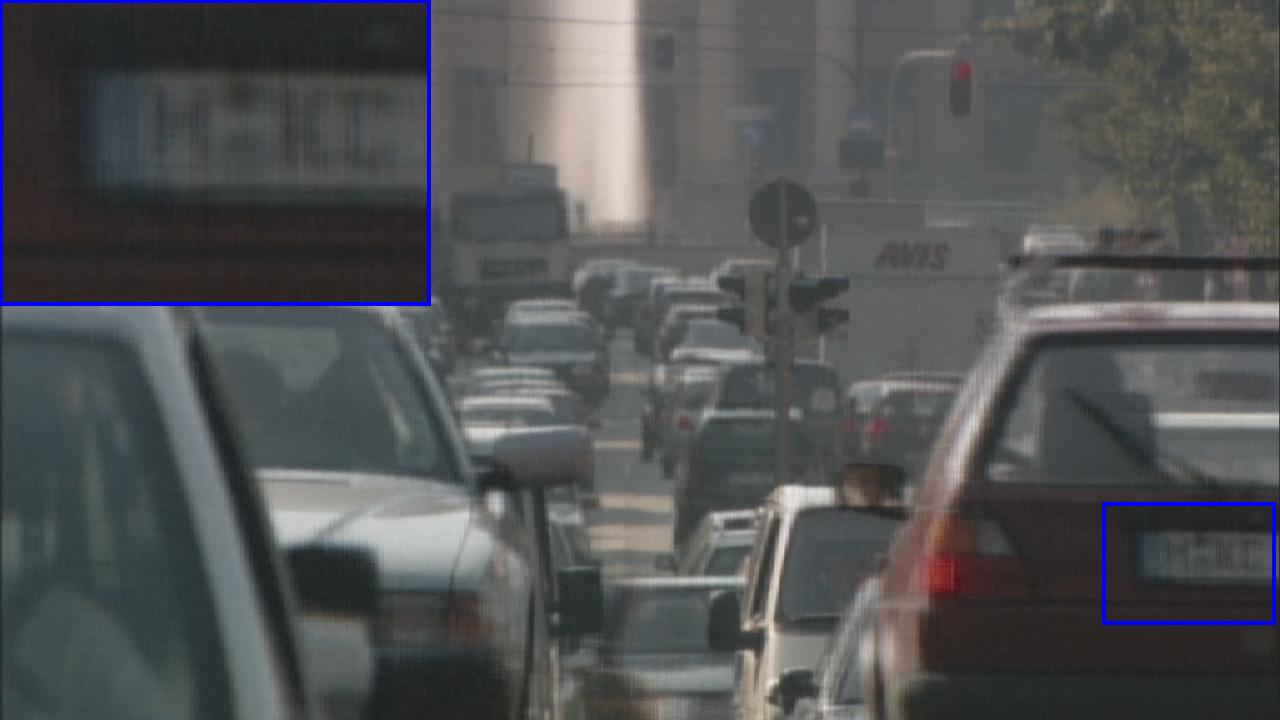}&
\includegraphics[width=0.585in, height=0.48in]{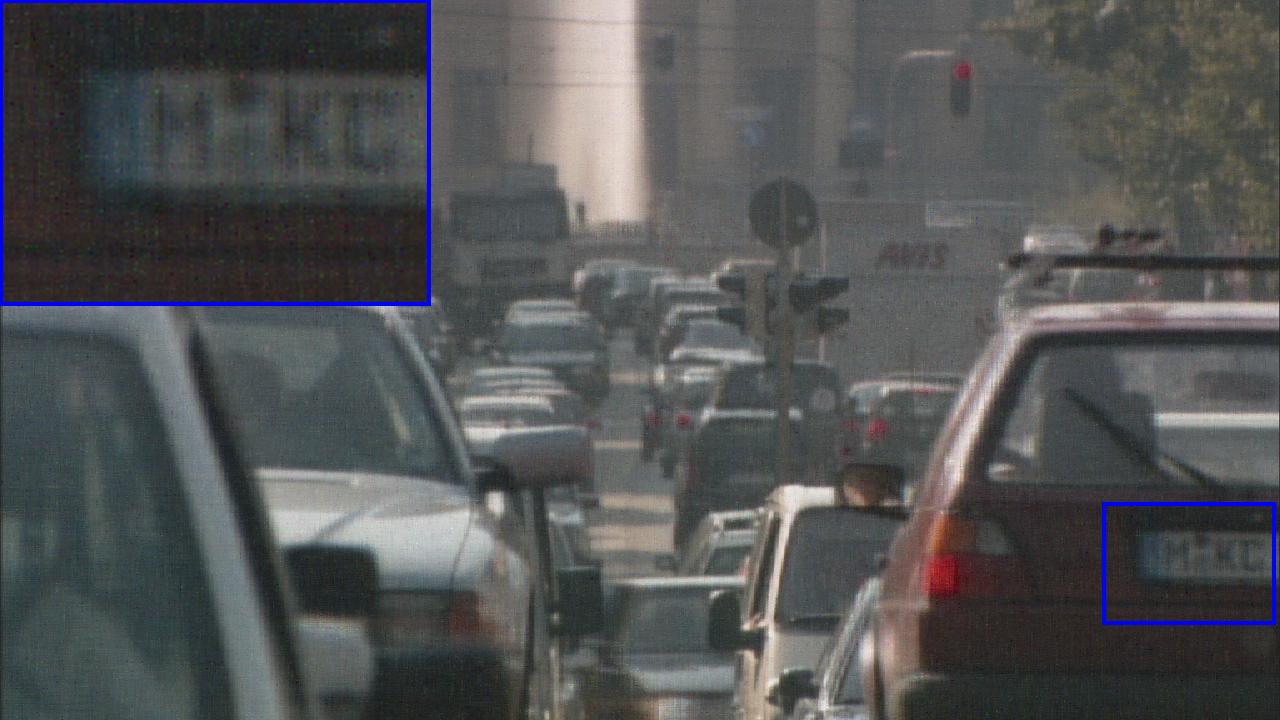}&
\includegraphics[width=0.585in, height=0.48in]{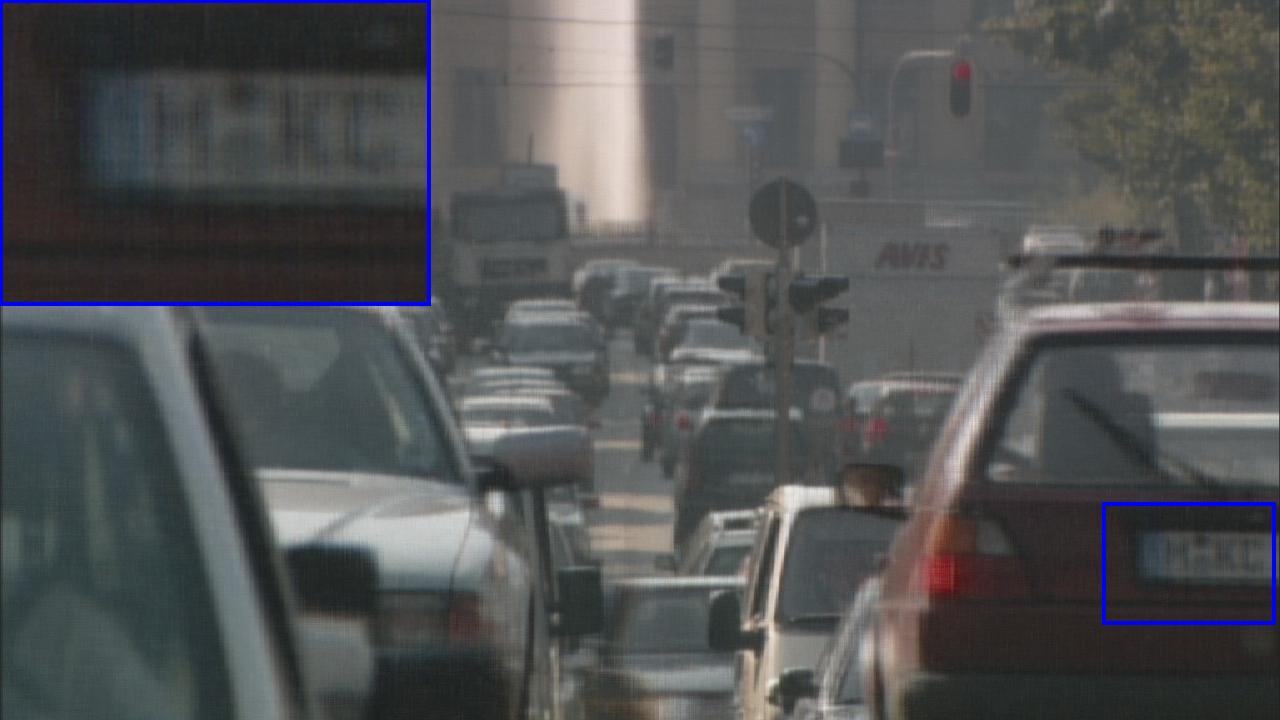}&
\includegraphics[width=0.585in, height=0.48in]{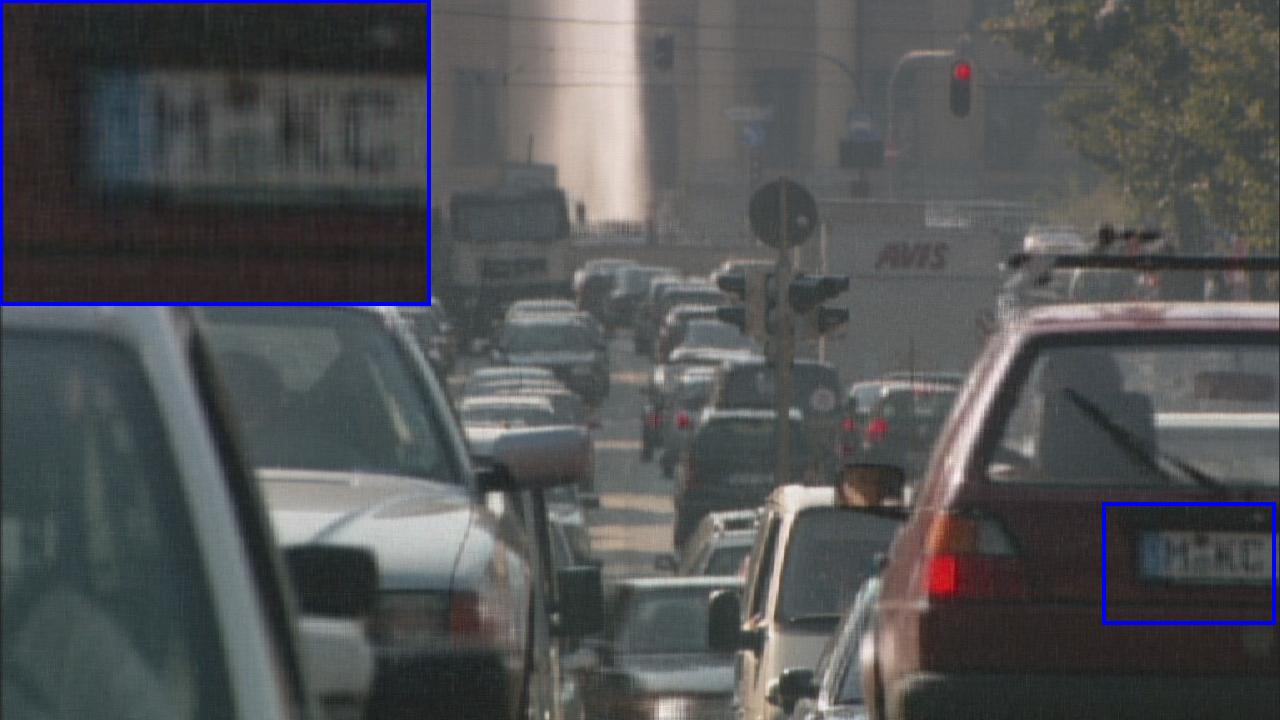}&
\includegraphics[width=0.585in, height=0.48in]{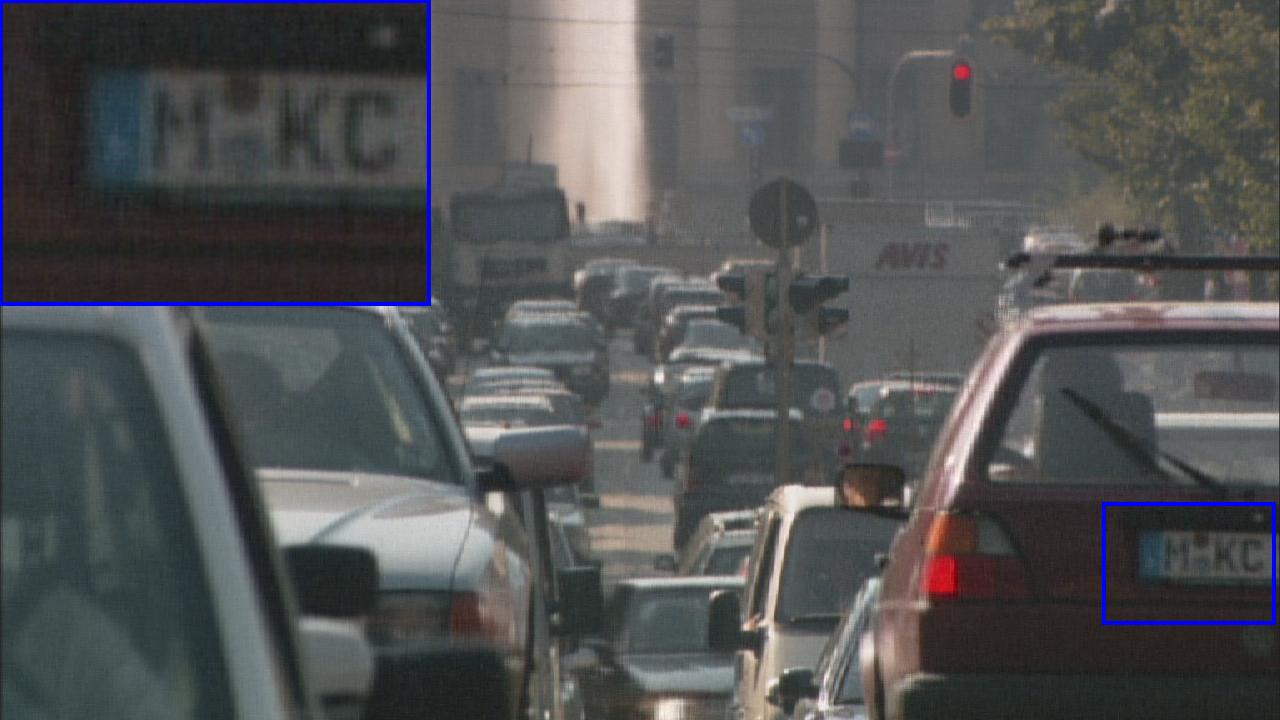}&
\includegraphics[width=0.585in, height=0.48in]{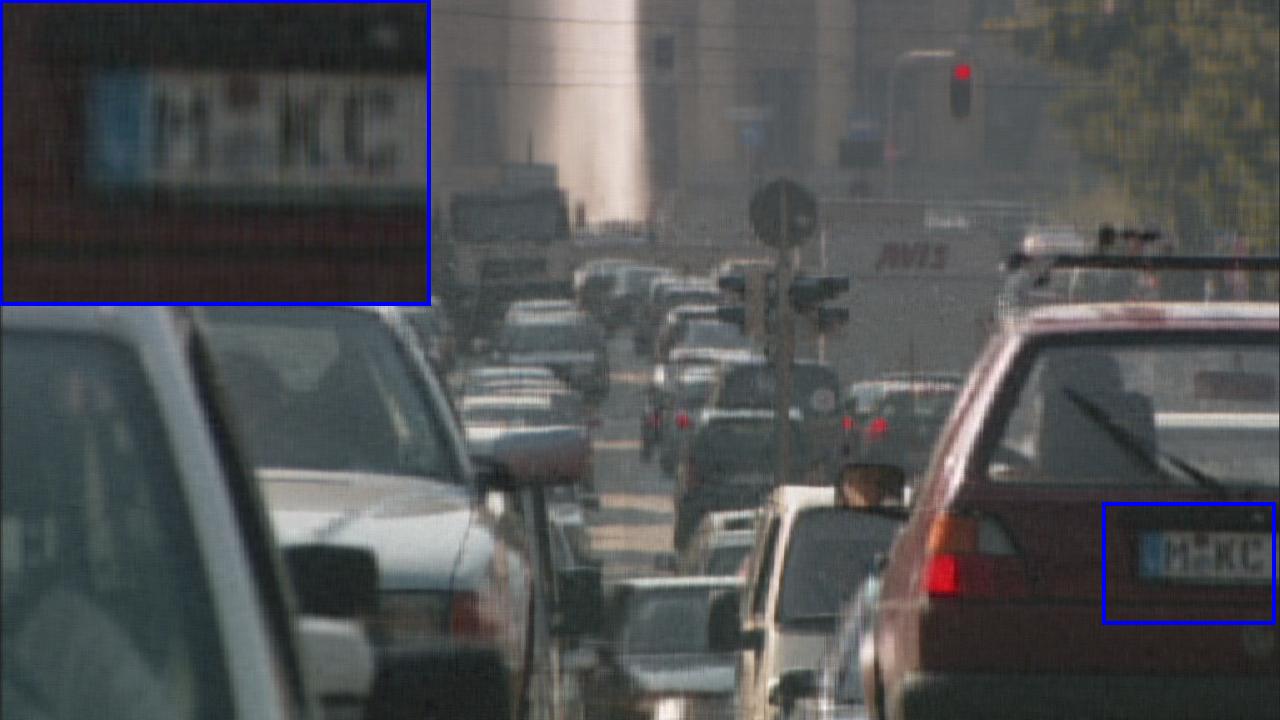}&
\includegraphics[width=0.585in, height=0.48in]{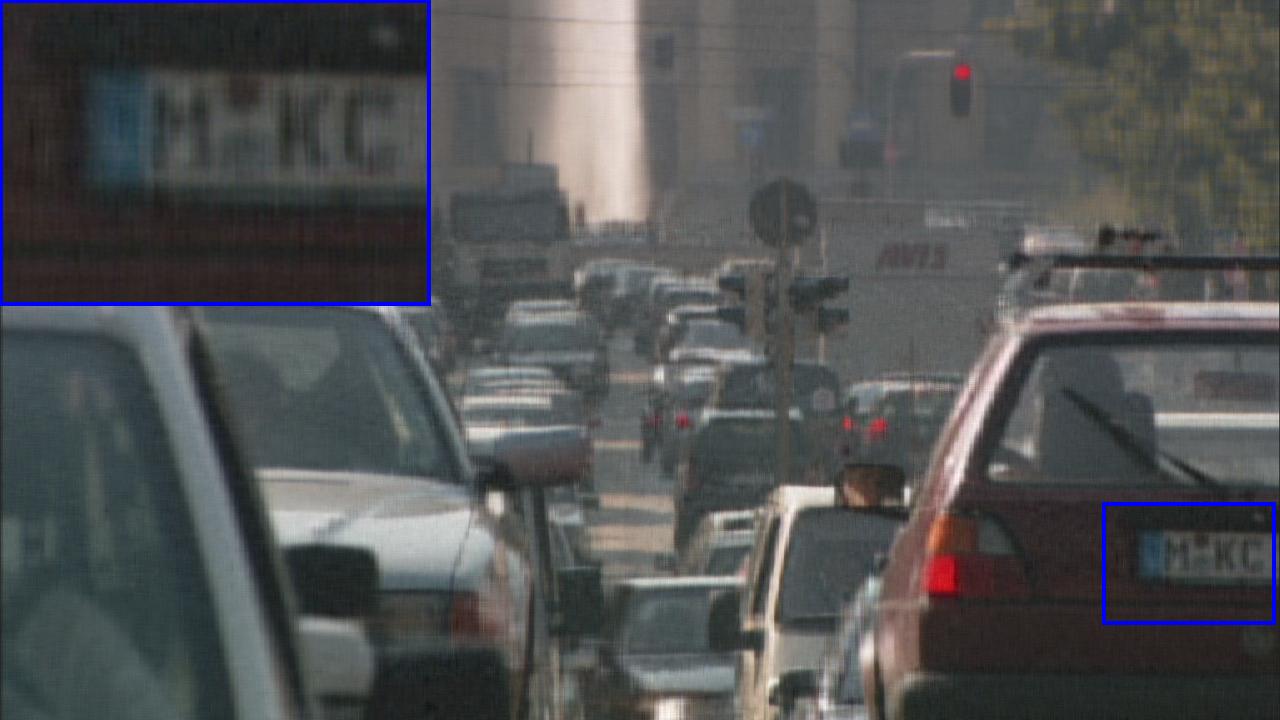}
&
\includegraphics[width=0.585in, height=0.48in,angle=0]{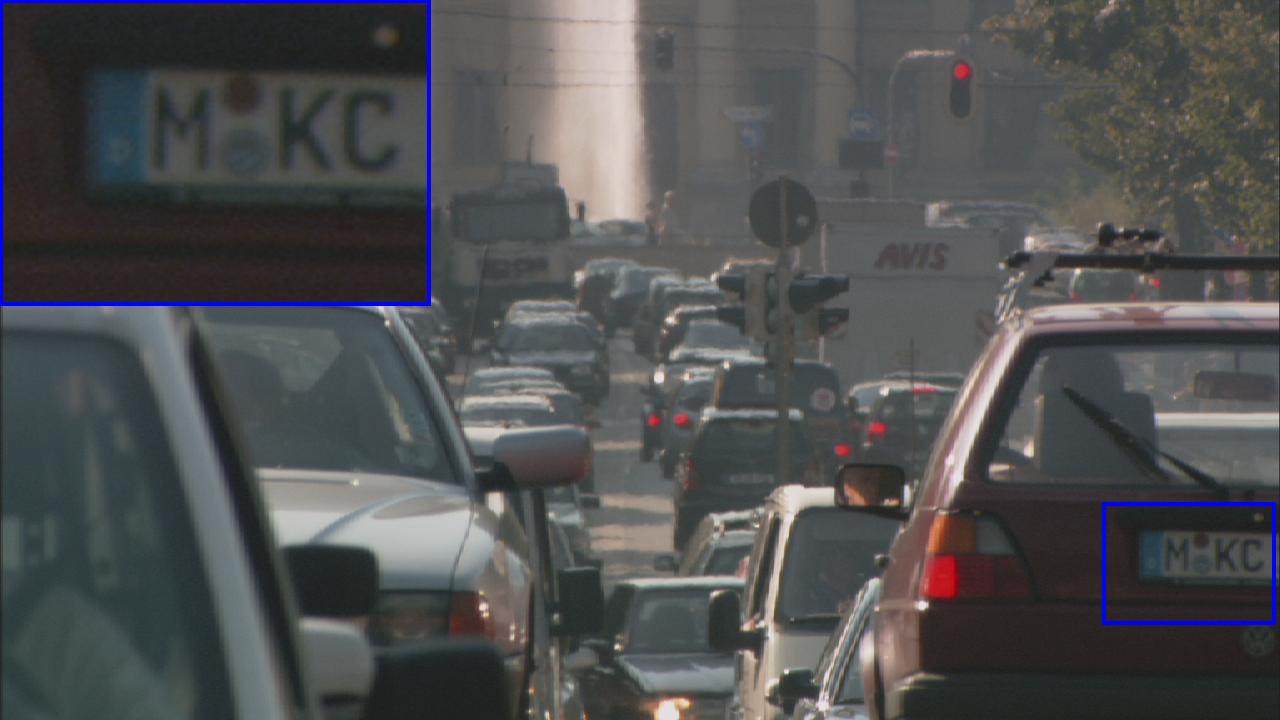}
\\

\includegraphics[width=0.585in, height=0.48in,angle=0]{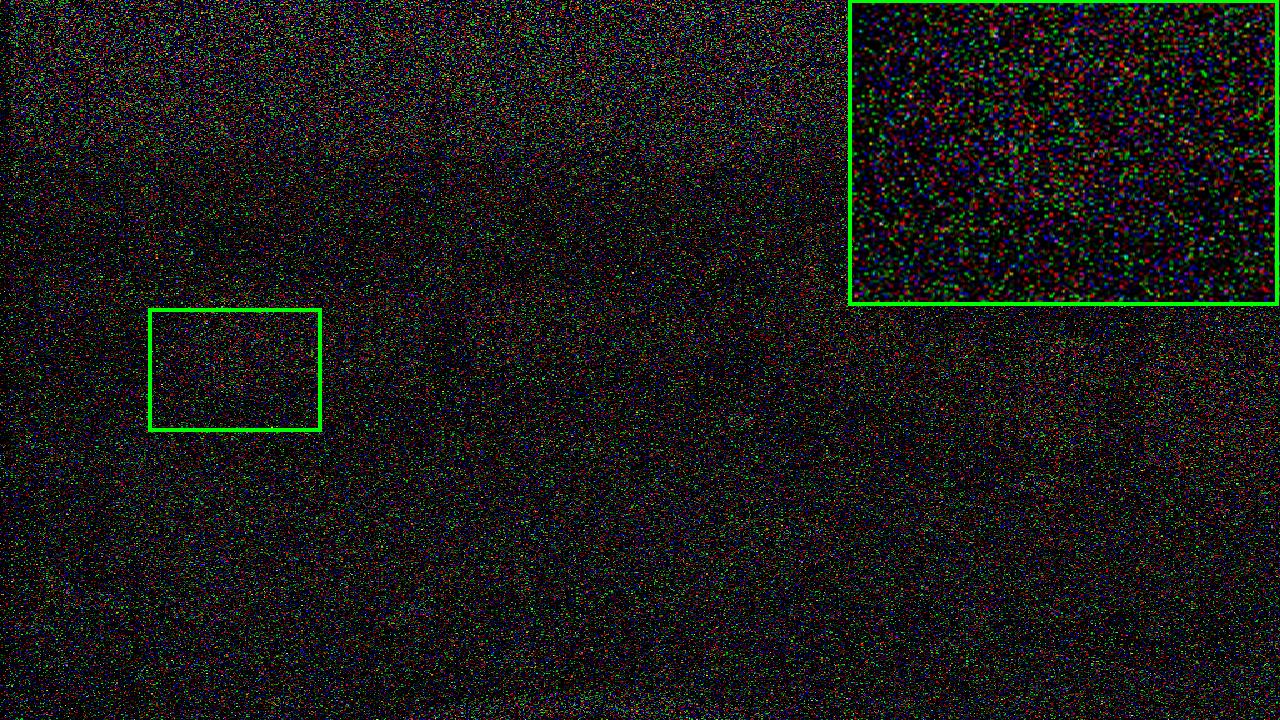}&
\includegraphics[width=0.585in, height=0.48in]{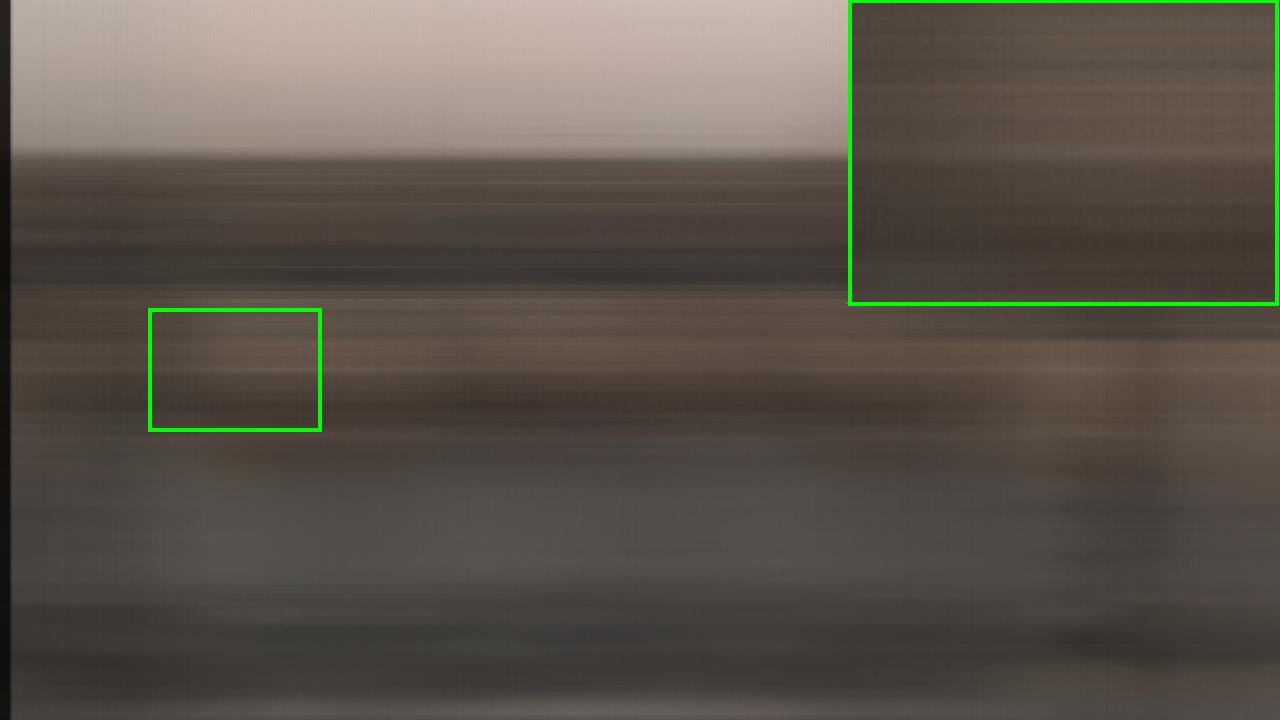}&
\includegraphics[width=0.585in, height=0.48in]{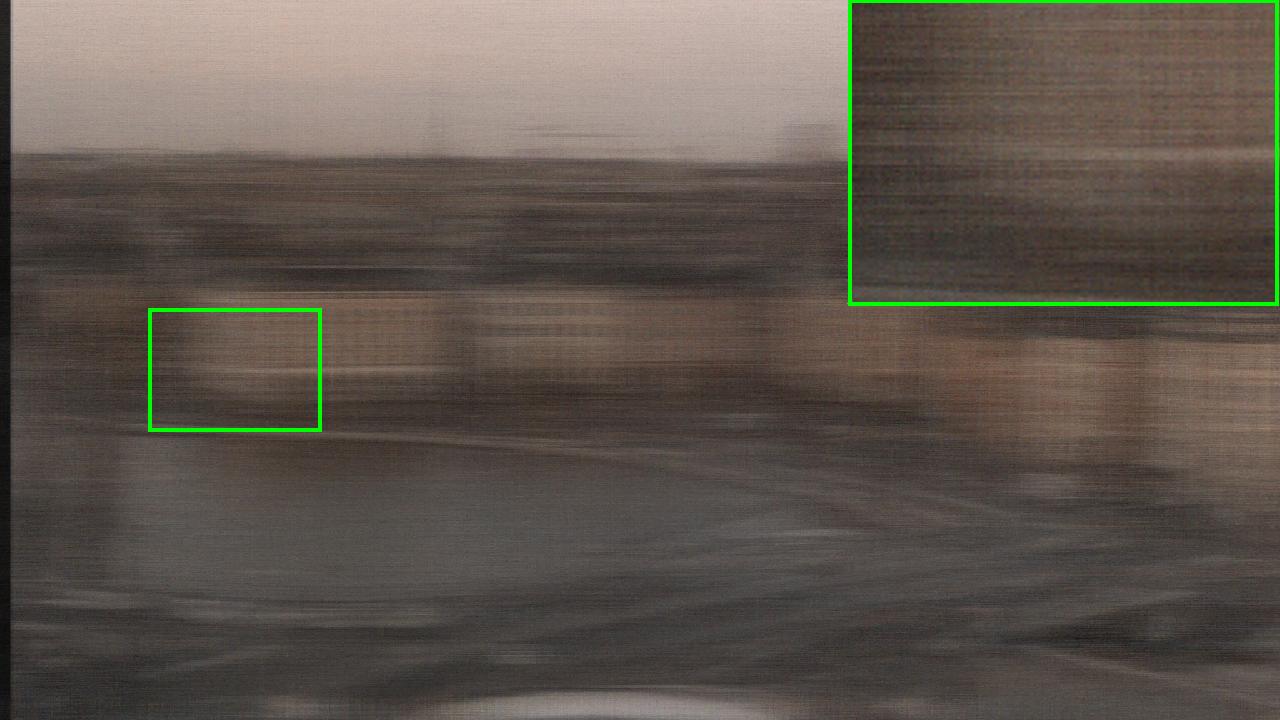}&
\includegraphics[width=0.585in, height=0.48in]{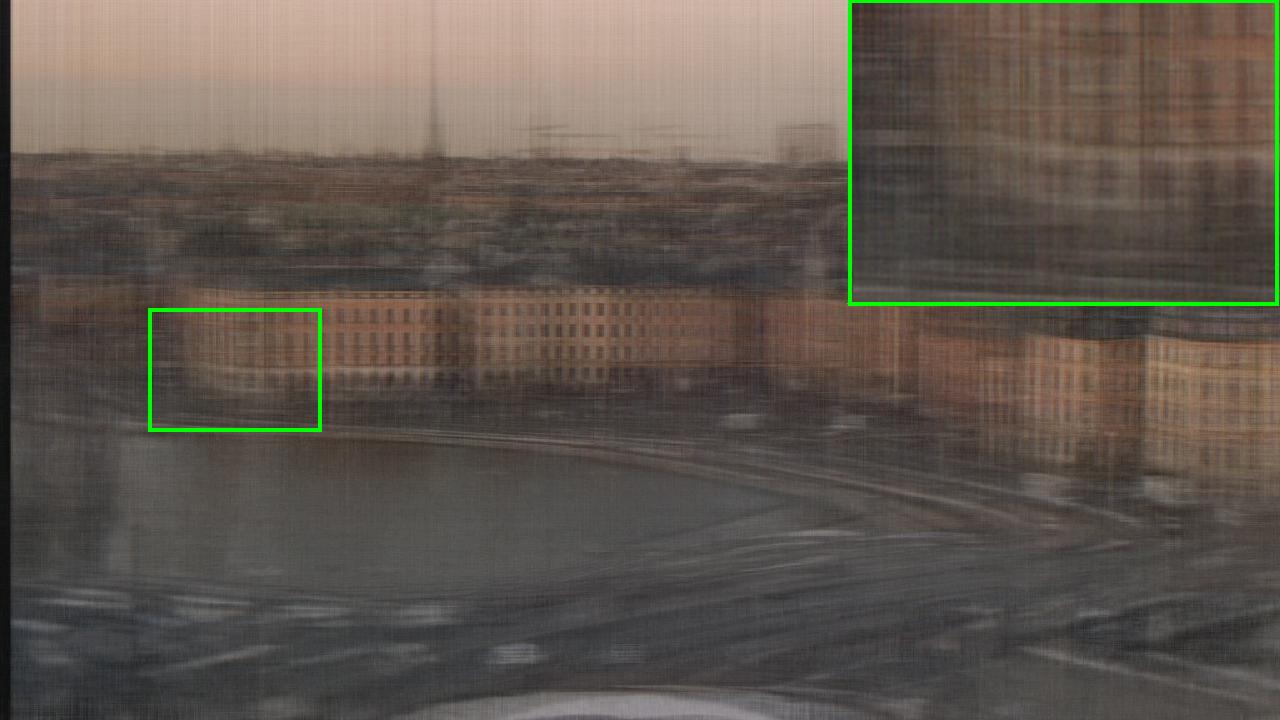}&
\includegraphics[width=0.585in, height=0.48in]{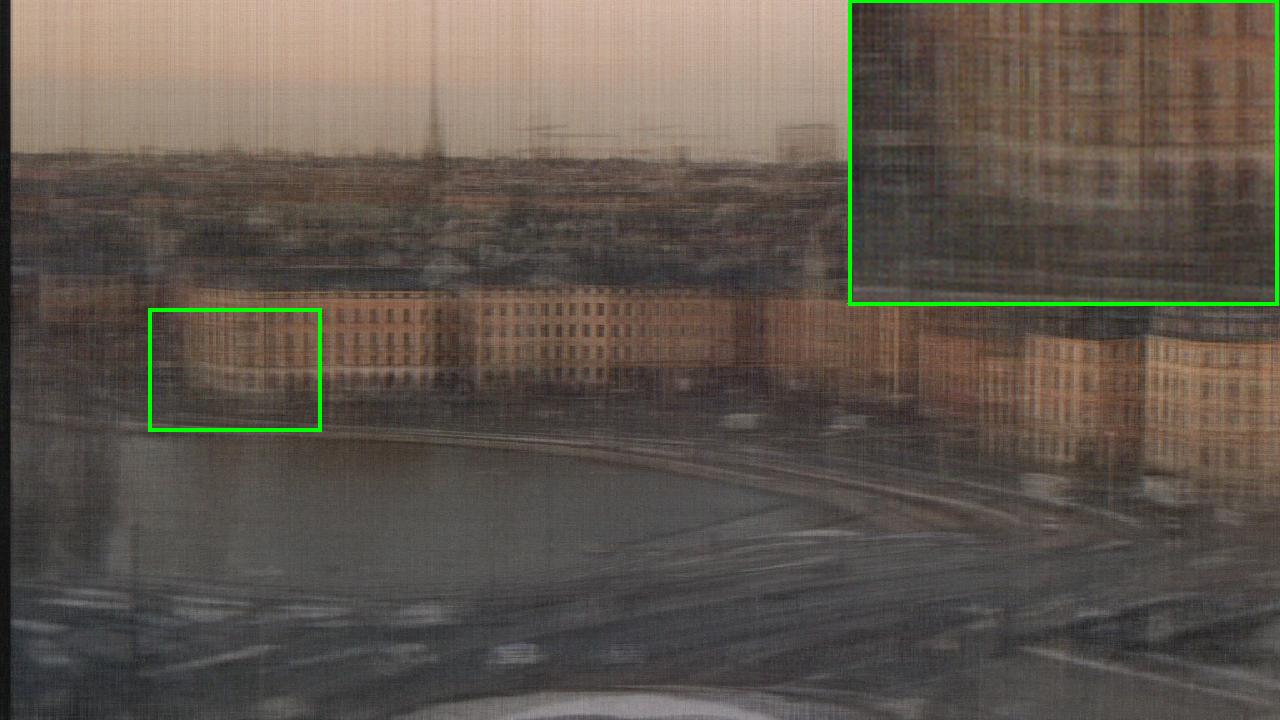}&
\includegraphics[width=0.585in, height=0.48in]{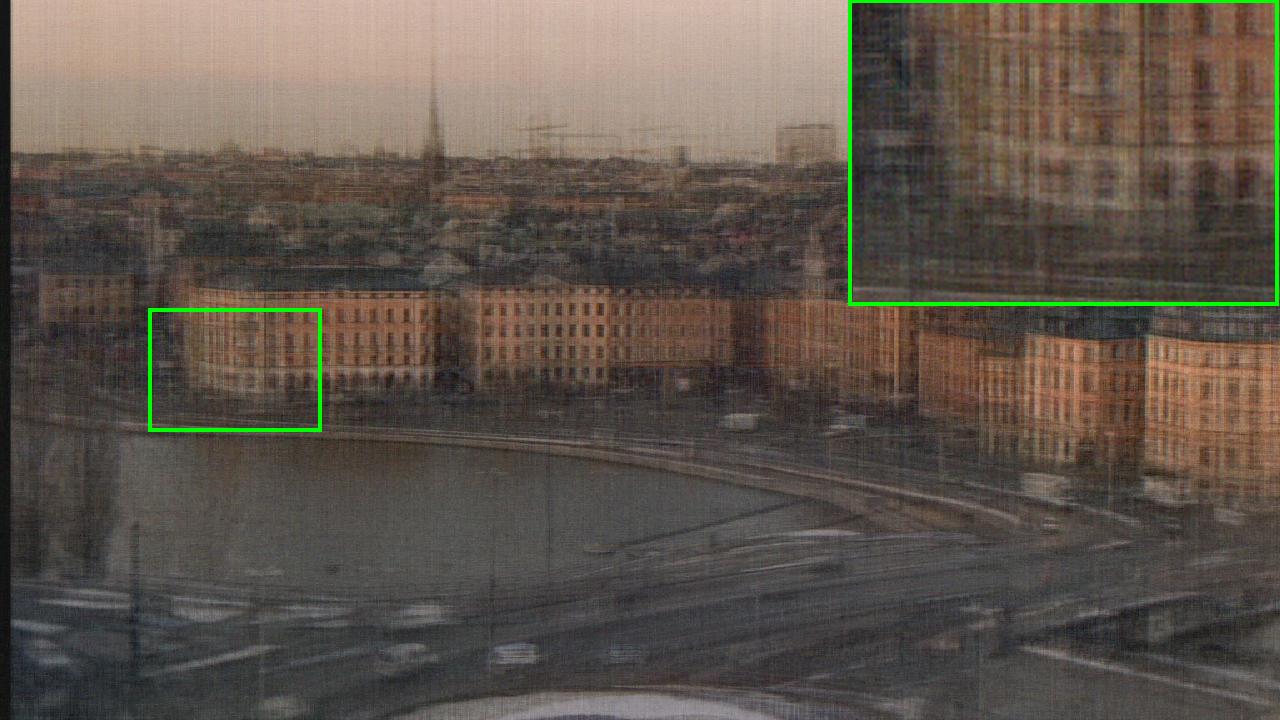}&
\includegraphics[width=0.585in, height=0.48in]{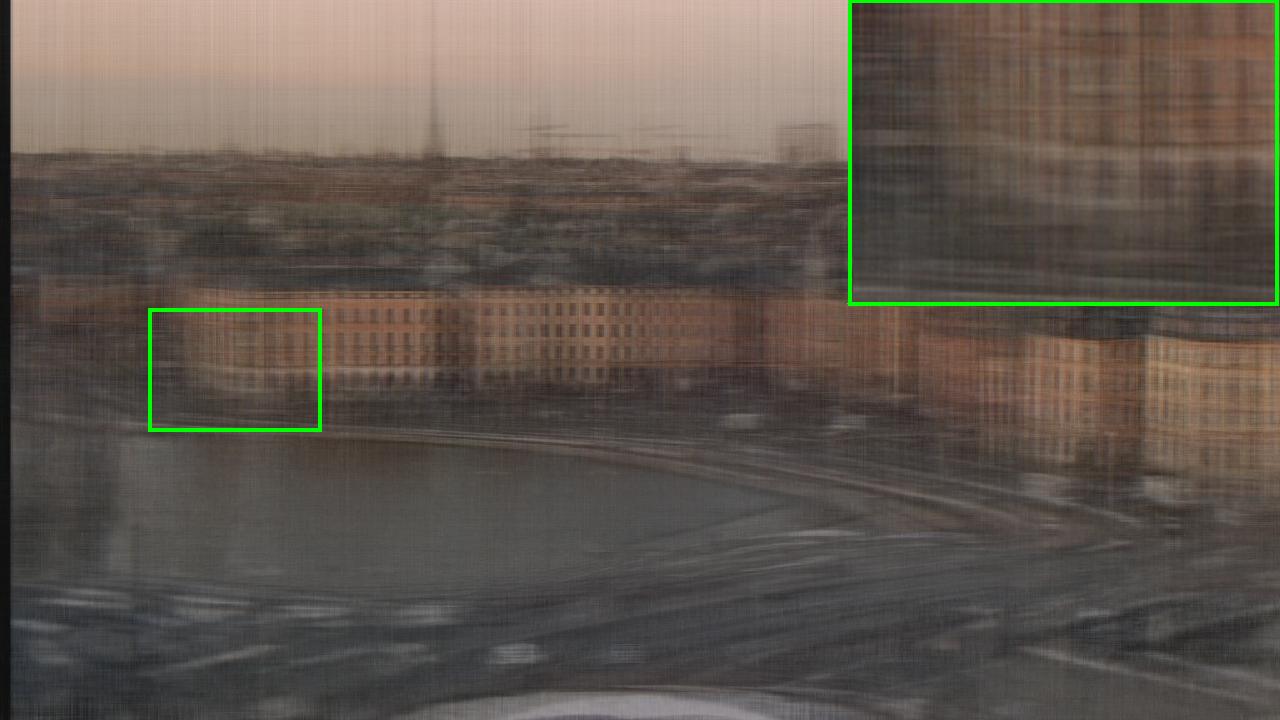}&
\includegraphics[width=0.585in, height=0.48in]{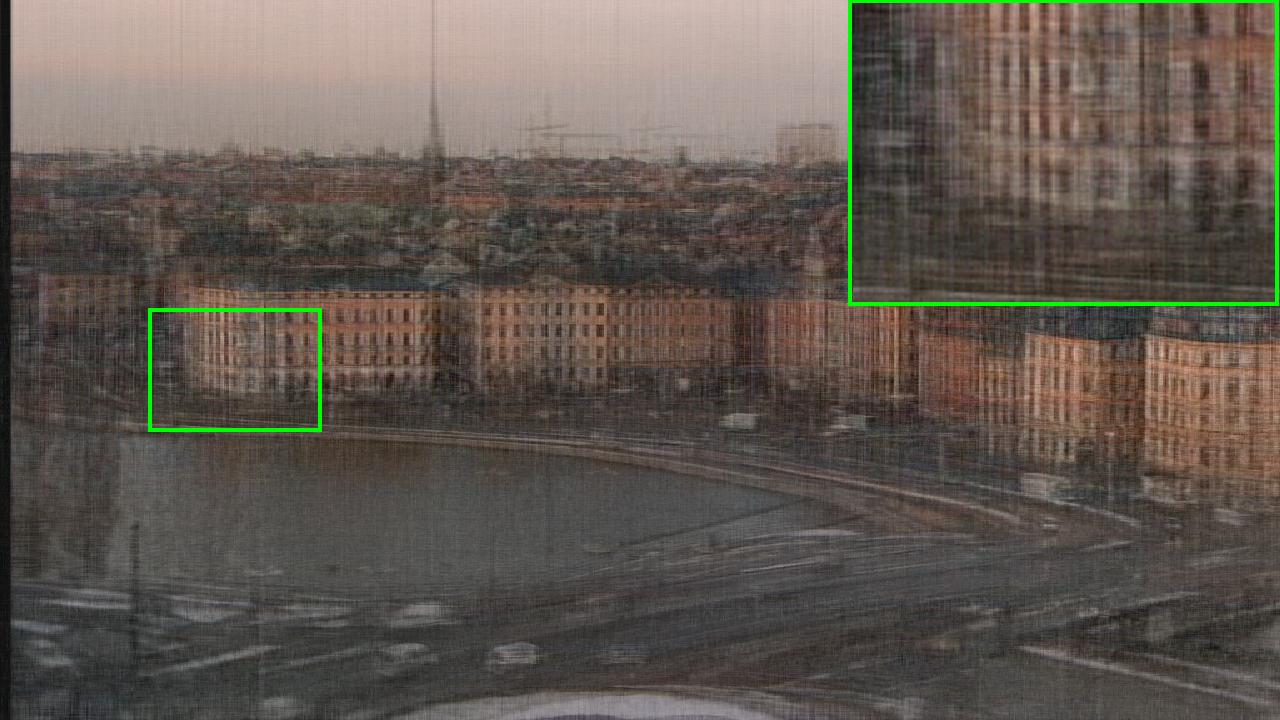}&
\includegraphics[width=0.585in, height=0.48in]{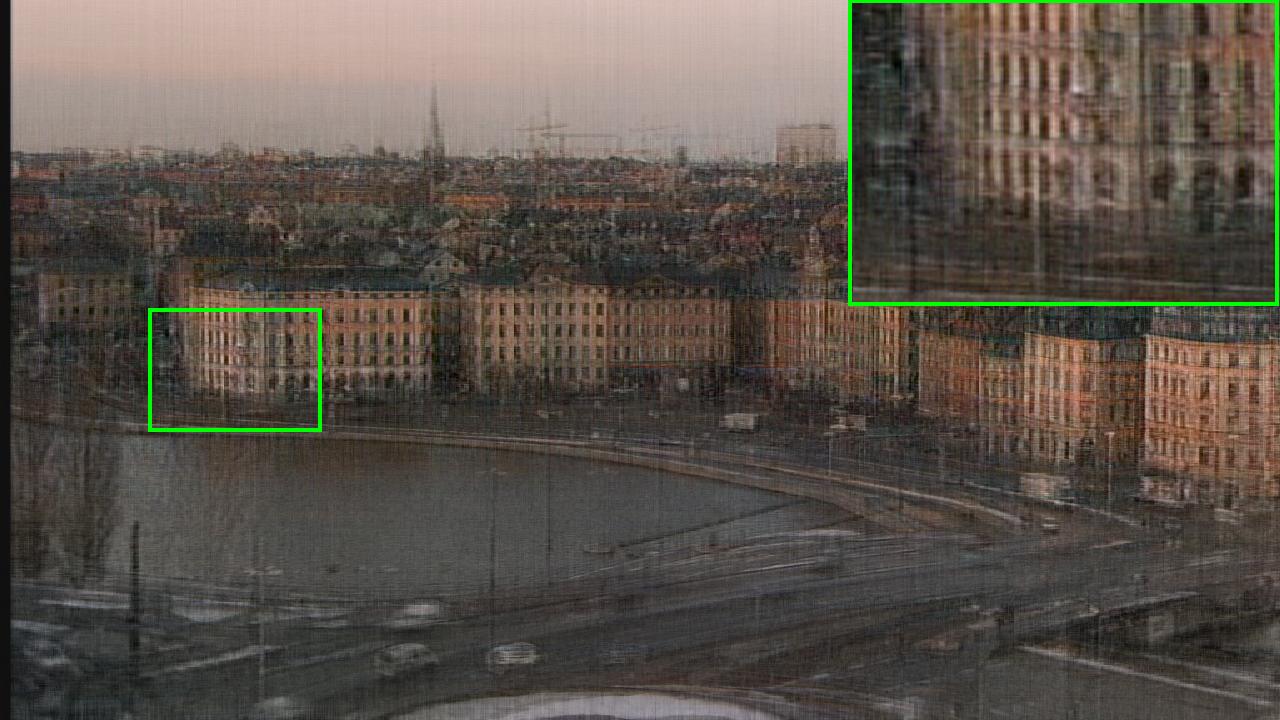}&
\includegraphics[width=0.585in, height=0.48in]{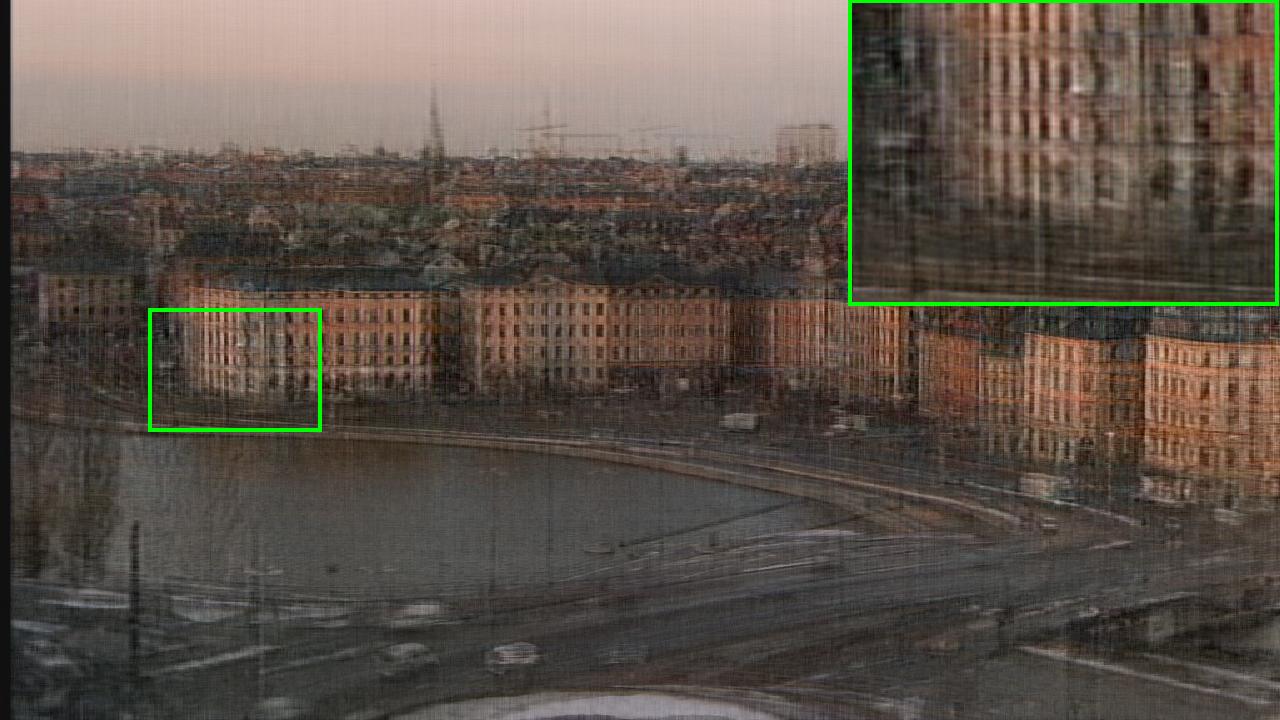}&
\includegraphics[width=0.585in, height=0.48in]{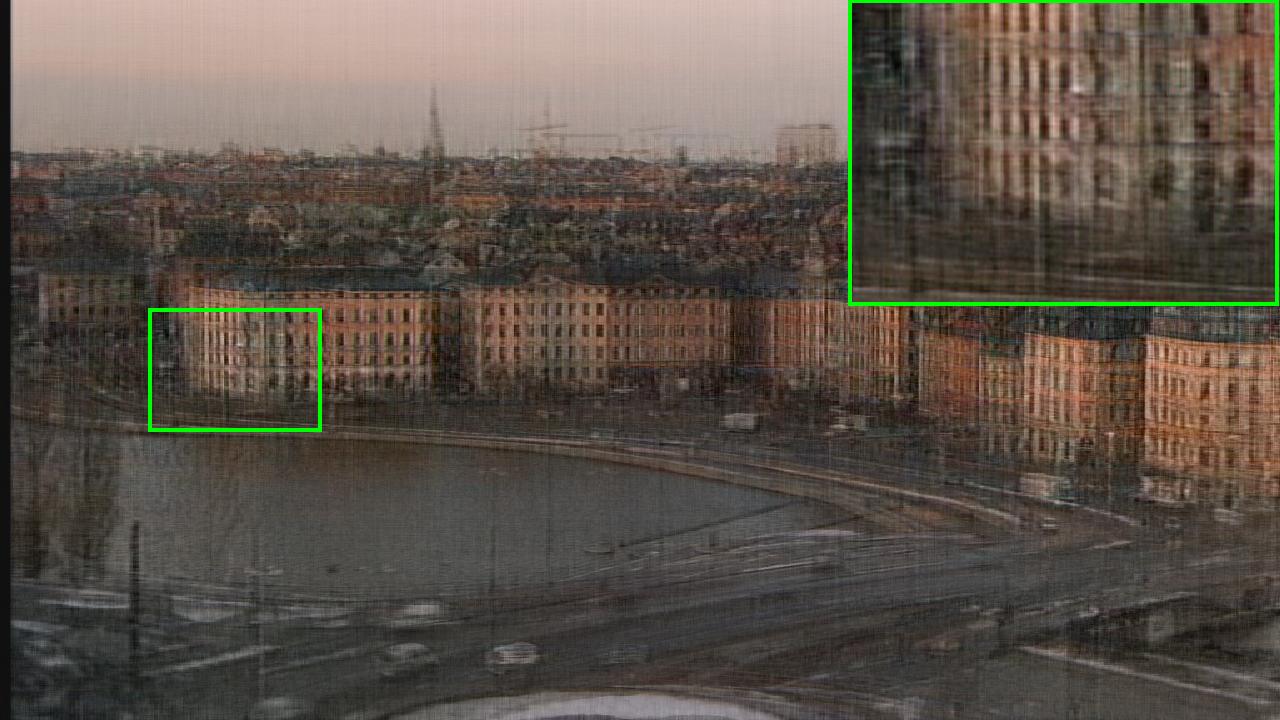}
&
\includegraphics[width=0.585in, height=0.48in,angle=0]{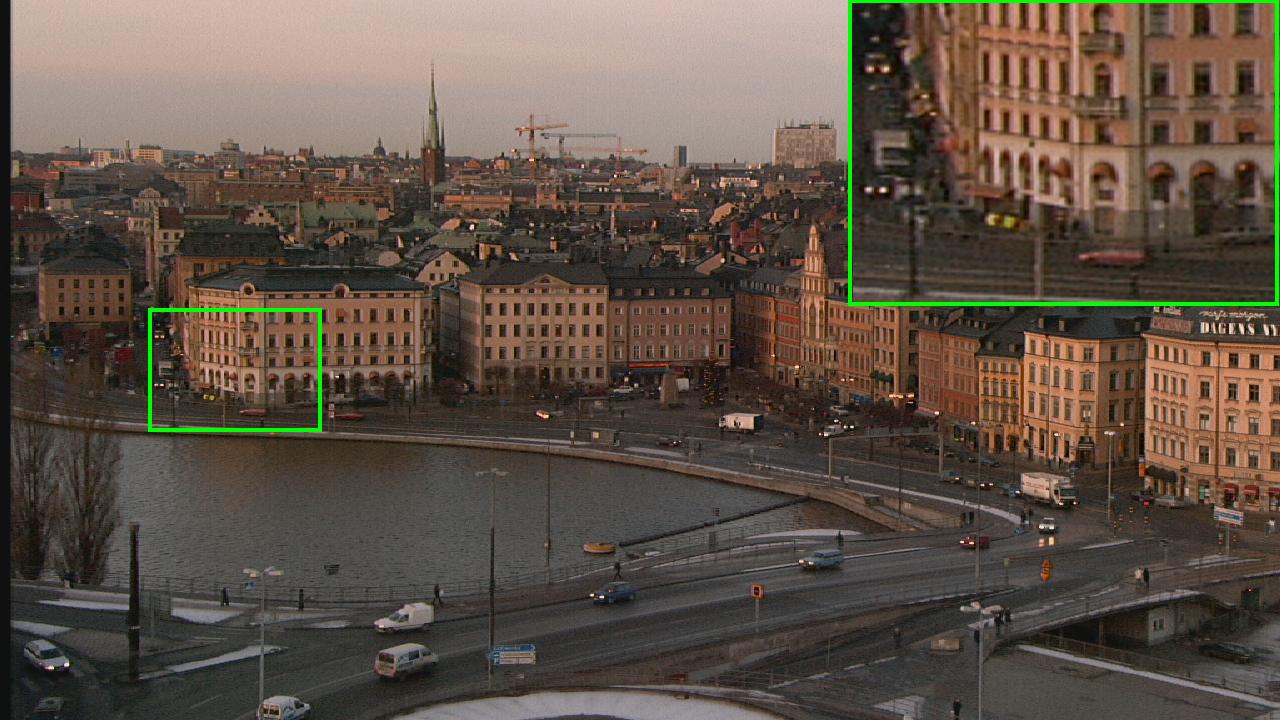}
\\

\includegraphics[width=0.585in, height=0.48in,angle=0]{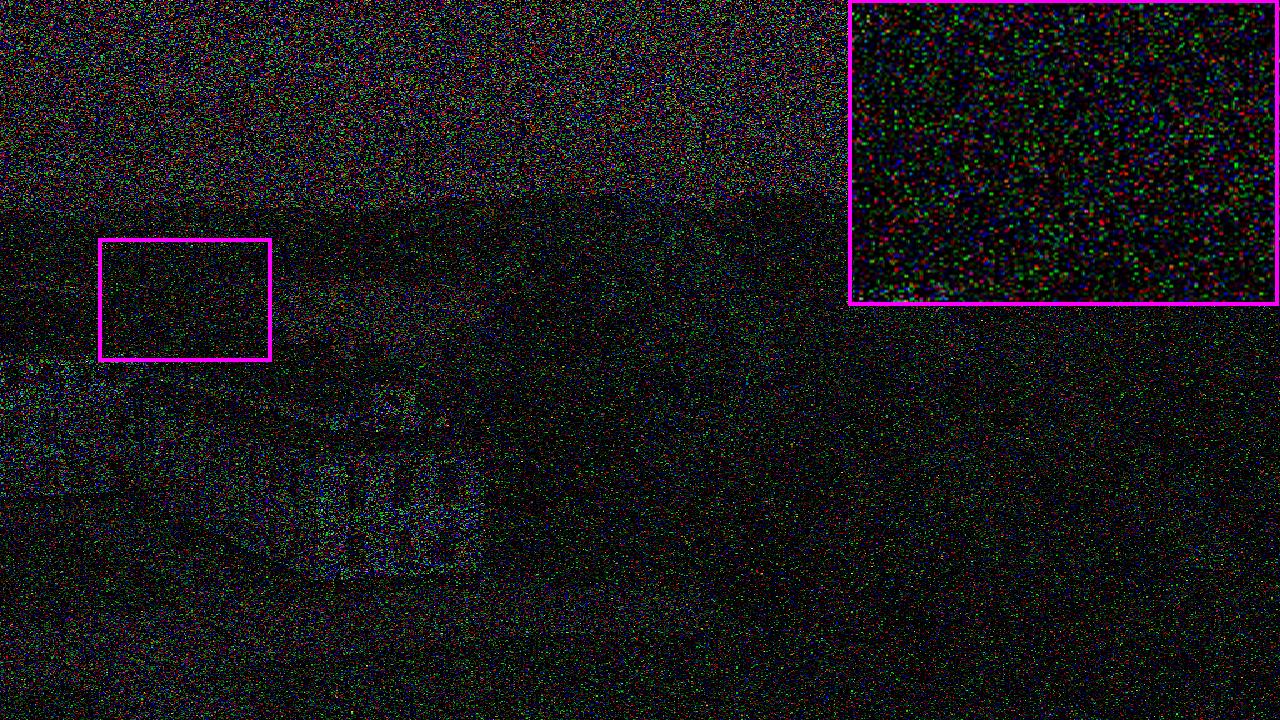}&
\includegraphics[width=0.585in, height=0.48in]{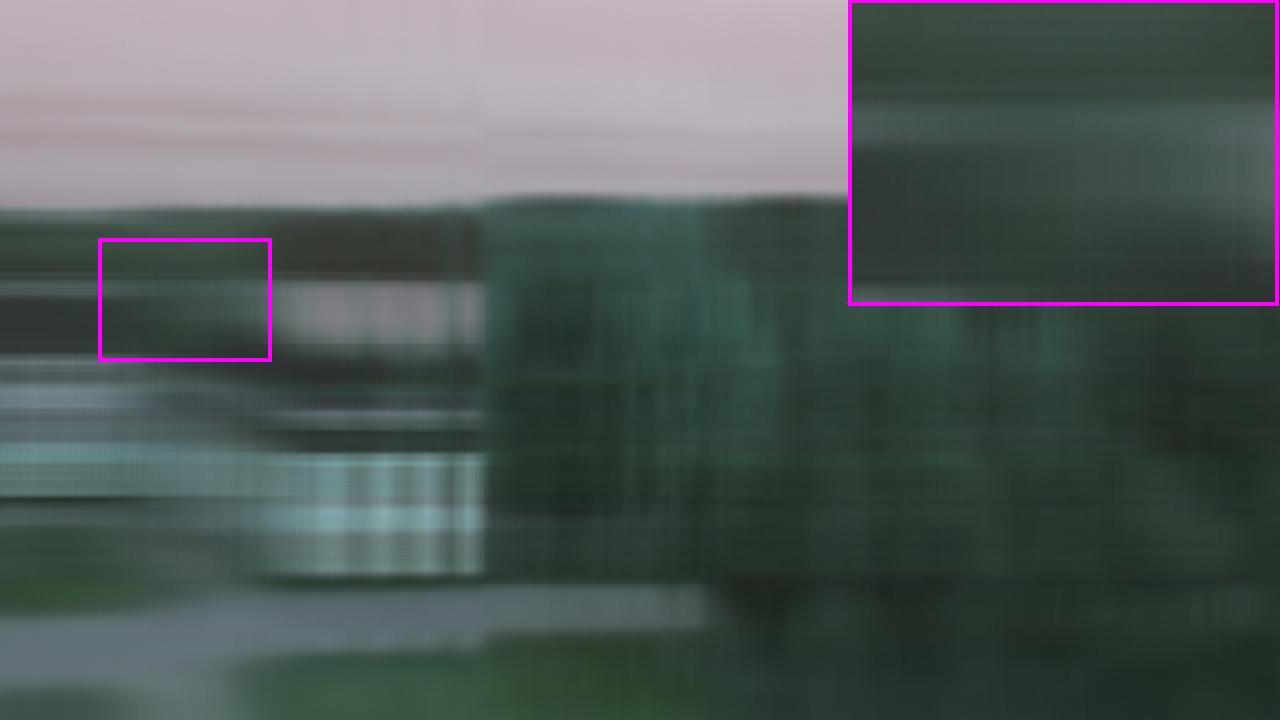}&
\includegraphics[width=0.585in, height=0.48in]{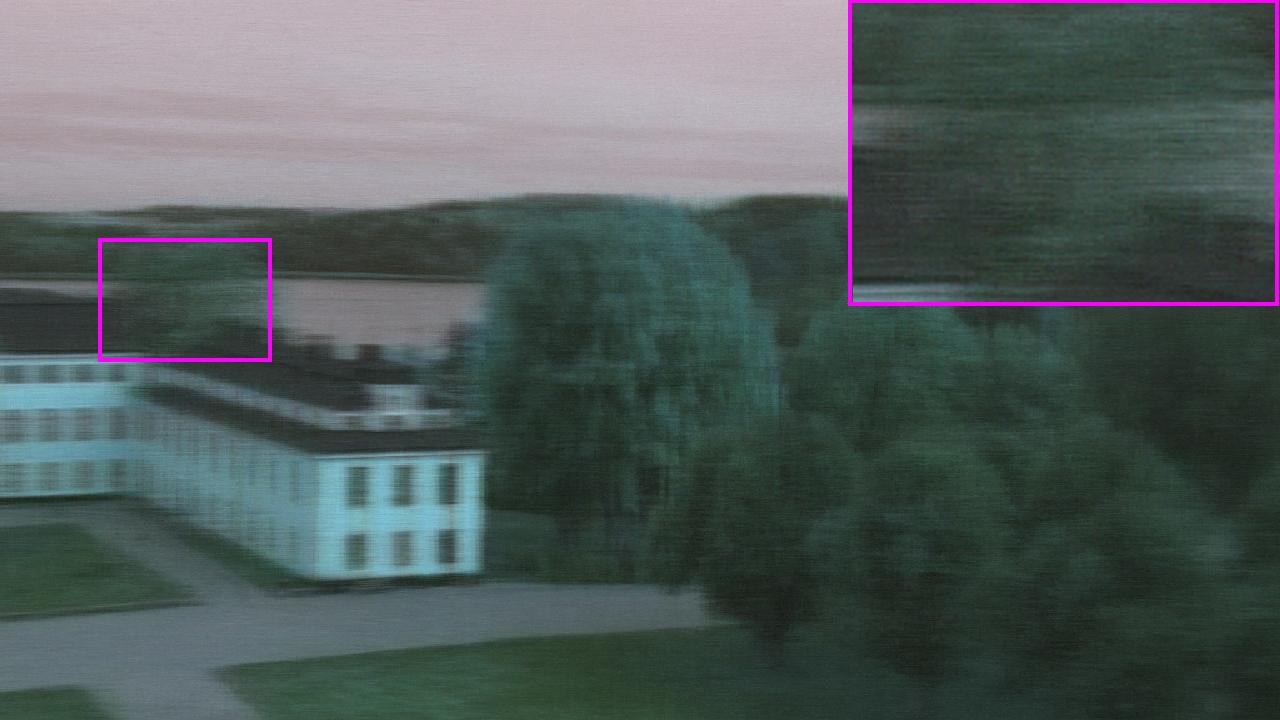}&
\includegraphics[width=0.585in, height=0.48in]{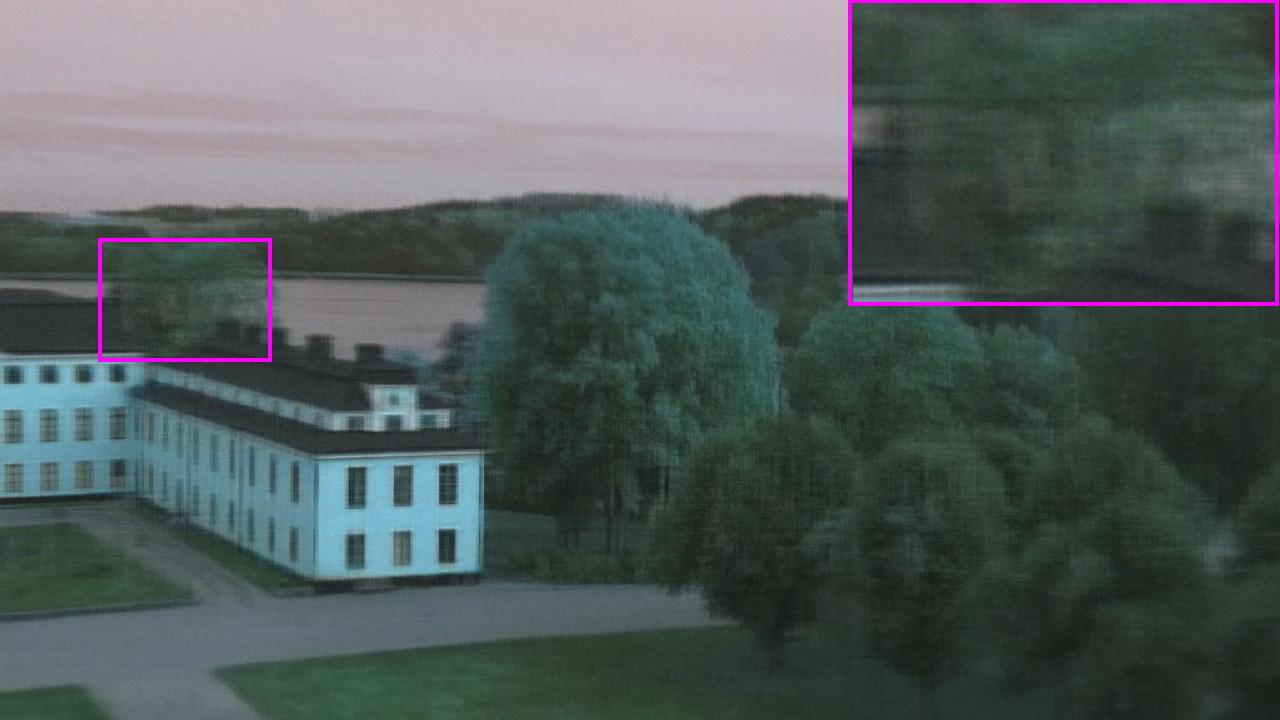}&
\includegraphics[width=0.585in, height=0.48in]{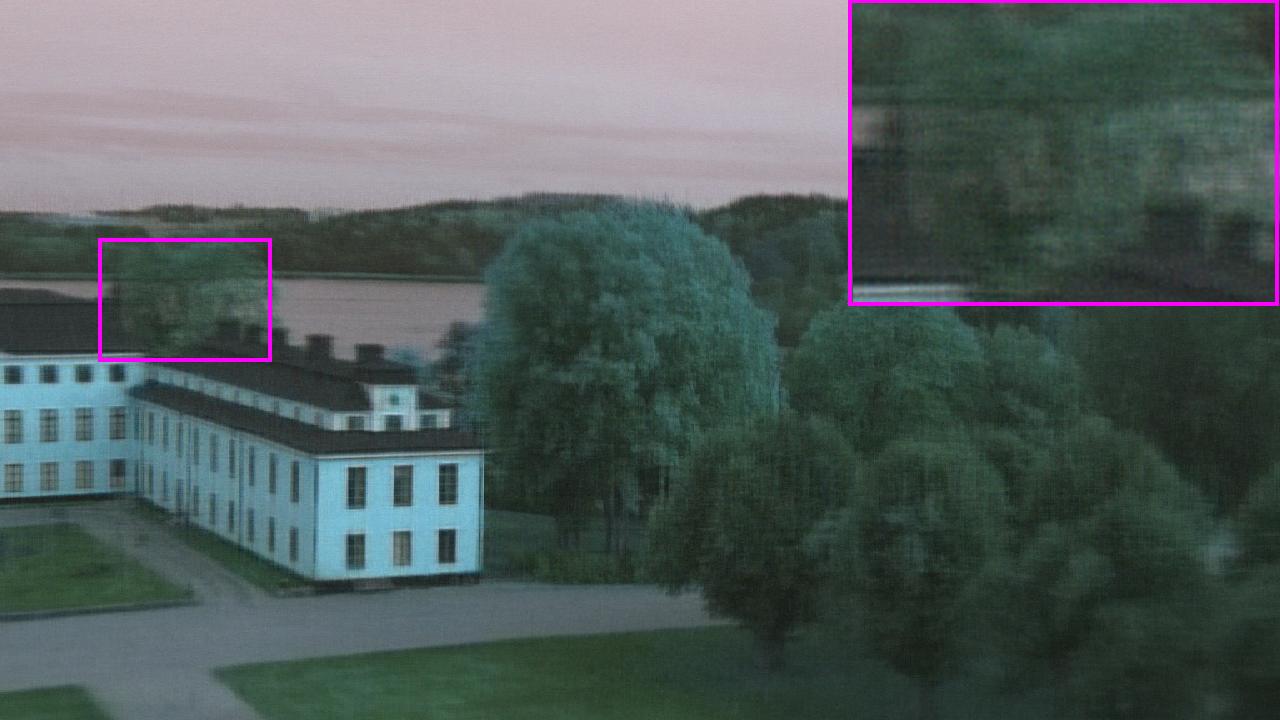}&
\includegraphics[width=0.585in, height=0.48in]{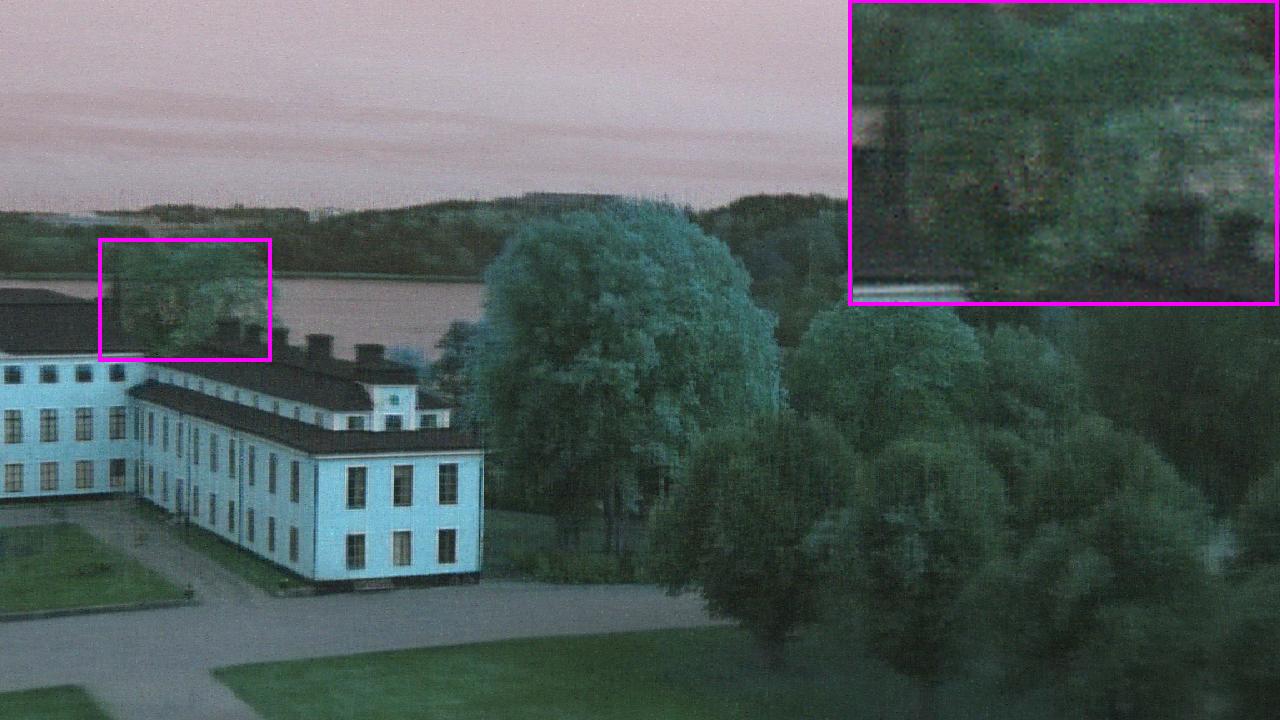}&
\includegraphics[width=0.585in, height=0.48in]{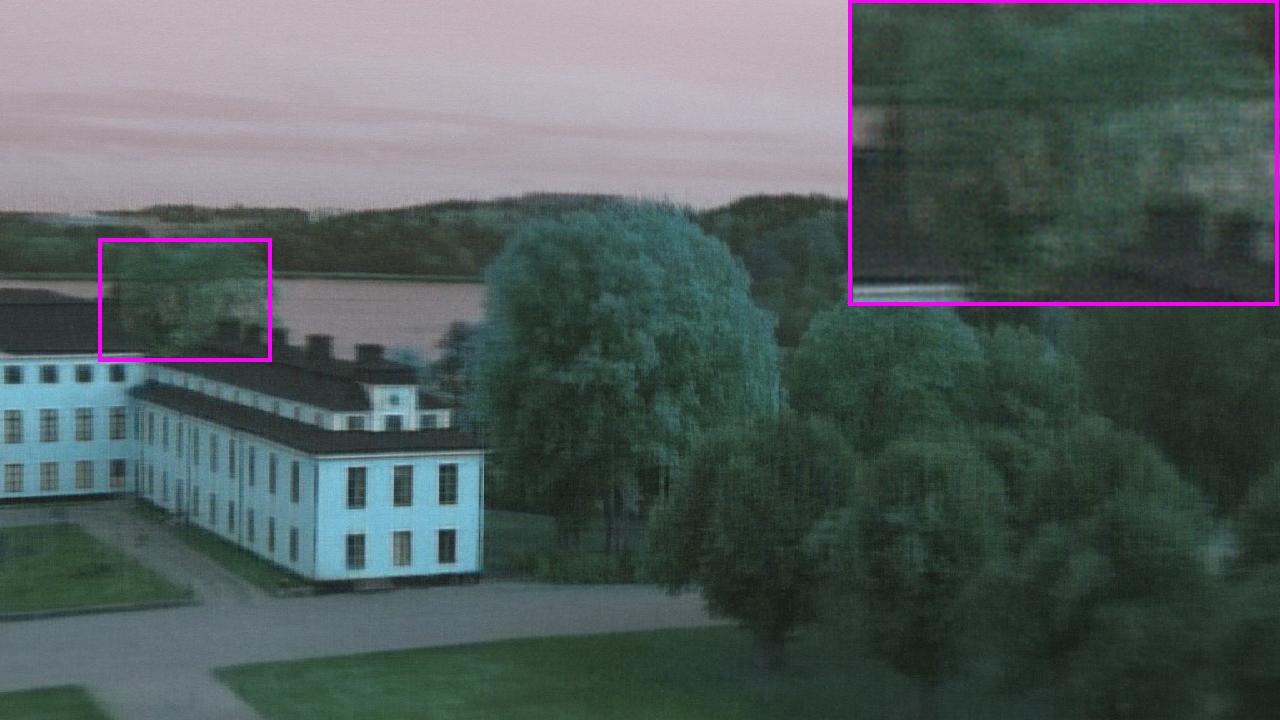}&
\includegraphics[width=0.585in, height=0.48in]{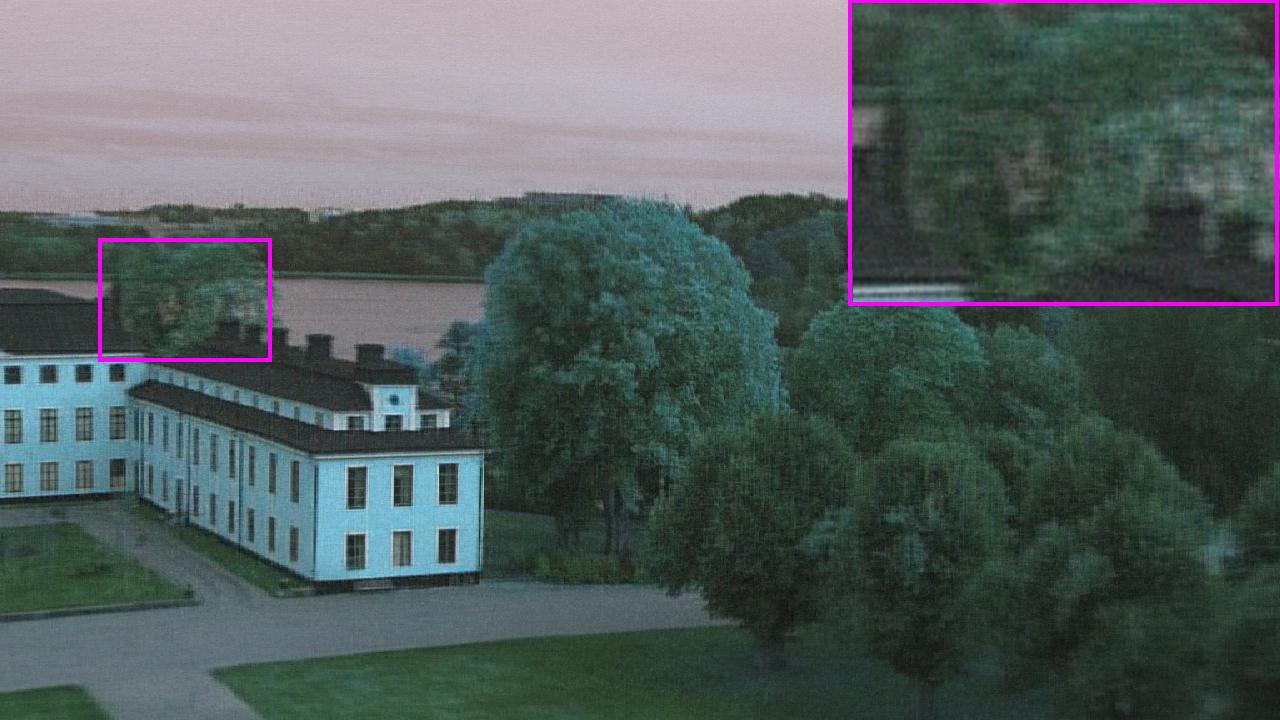}&
\includegraphics[width=0.585in, height=0.48in]{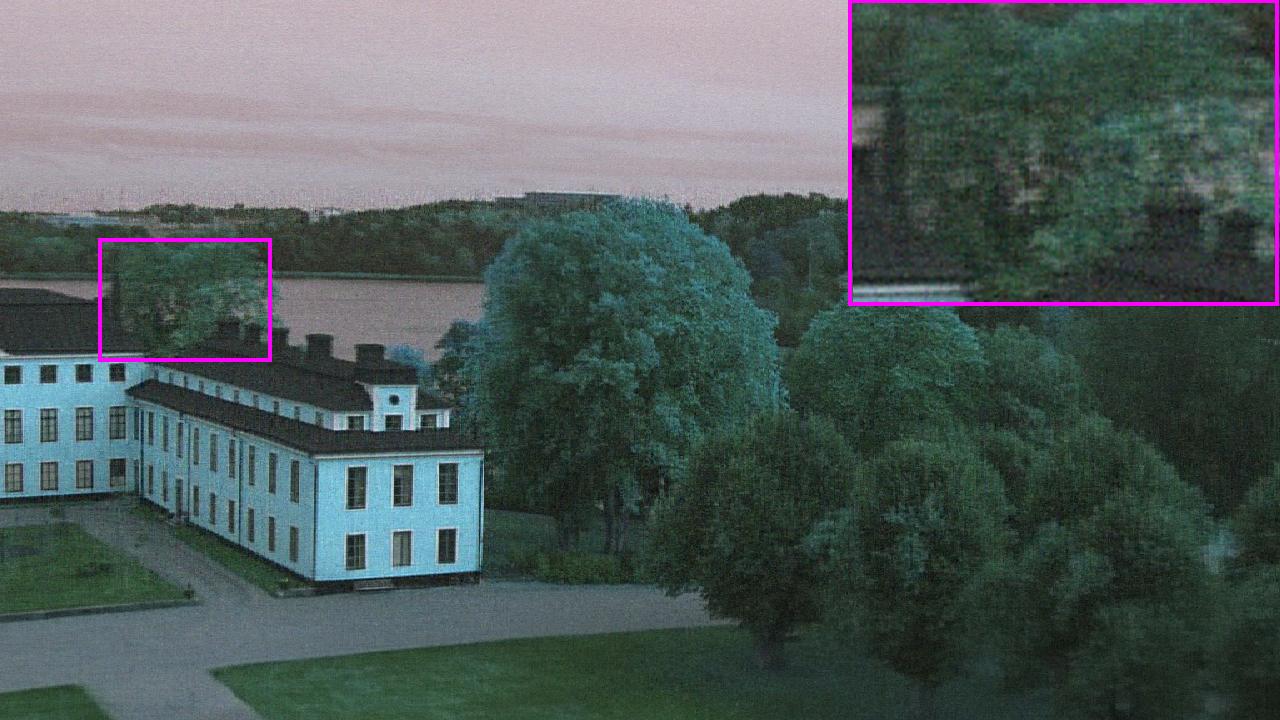}&
\includegraphics[width=0.585in, height=0.48in]{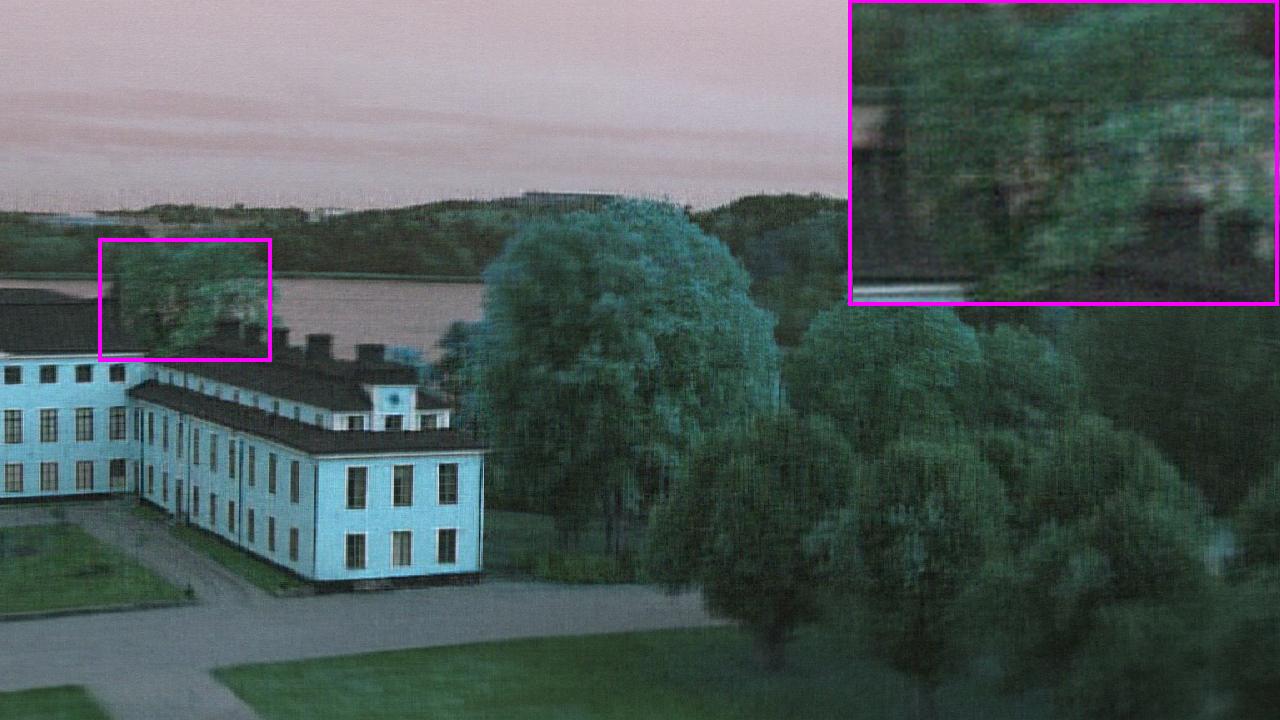}&
\includegraphics[width=0.585in, height=0.48in]{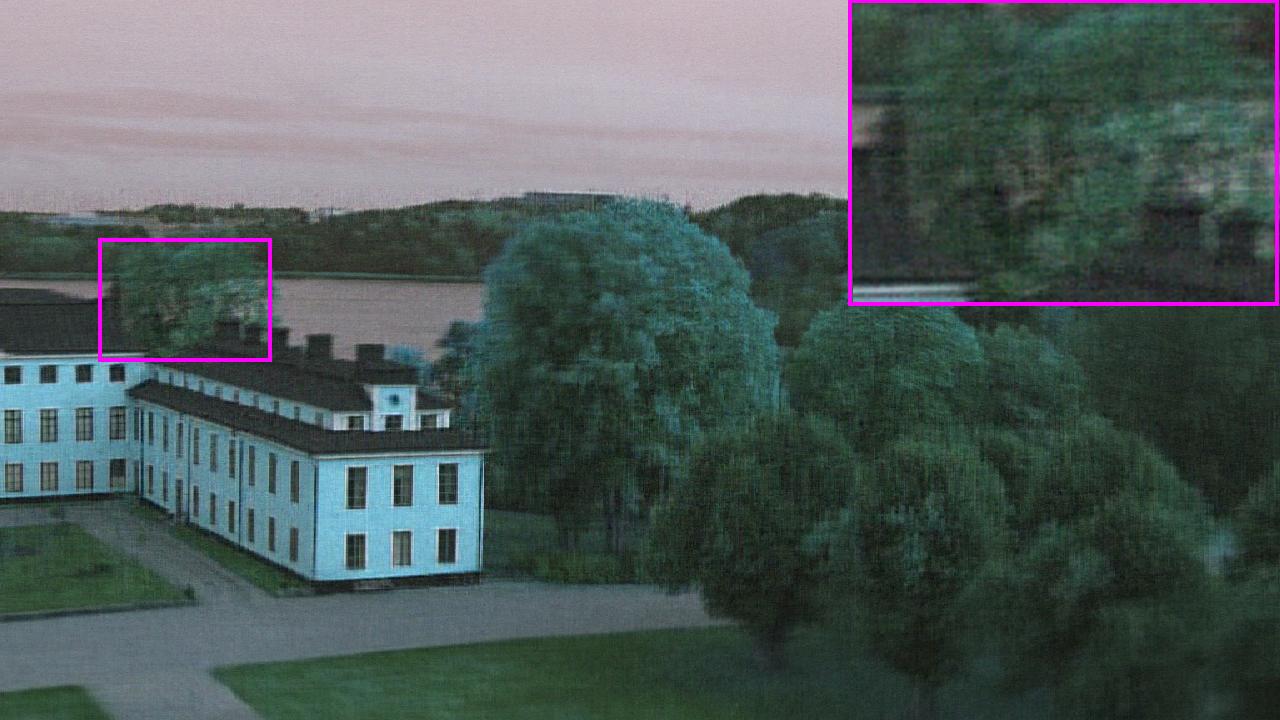}
&
\includegraphics[width=0.585in, height=0.48in,angle=0]{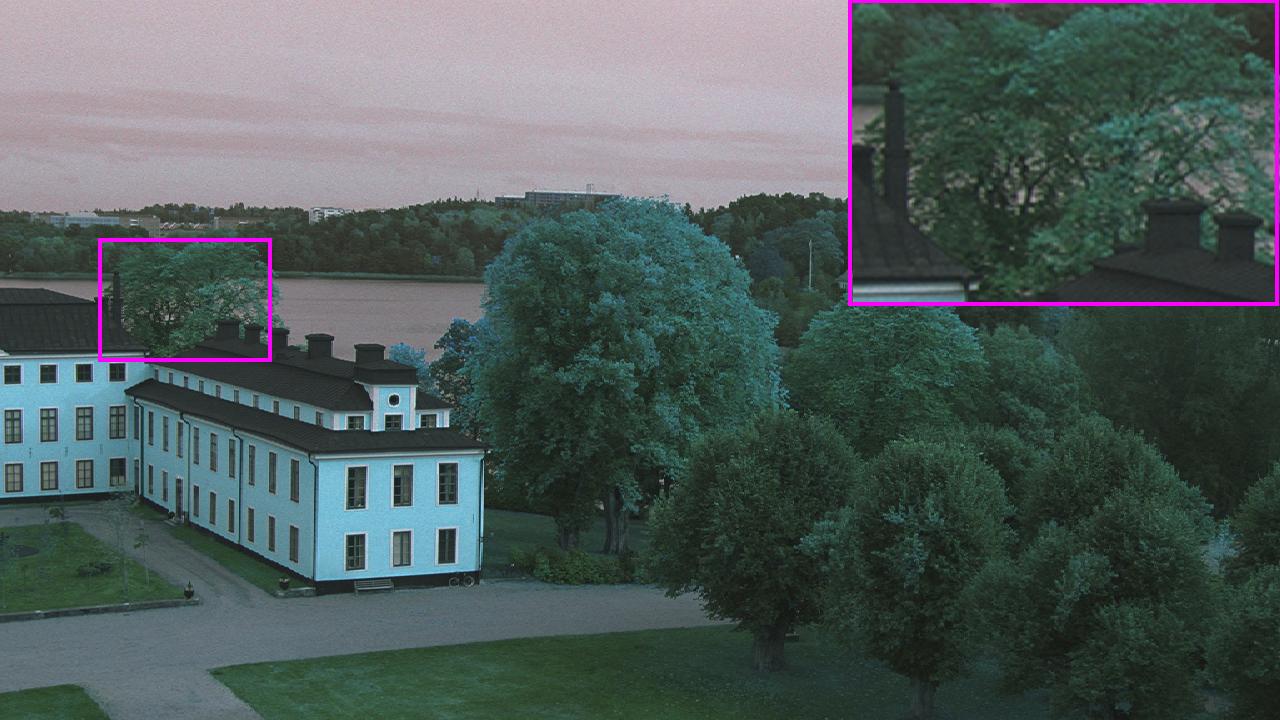}
\\

%
%
%

\includegraphics[width=0.585in, height=0.48in,angle=0]{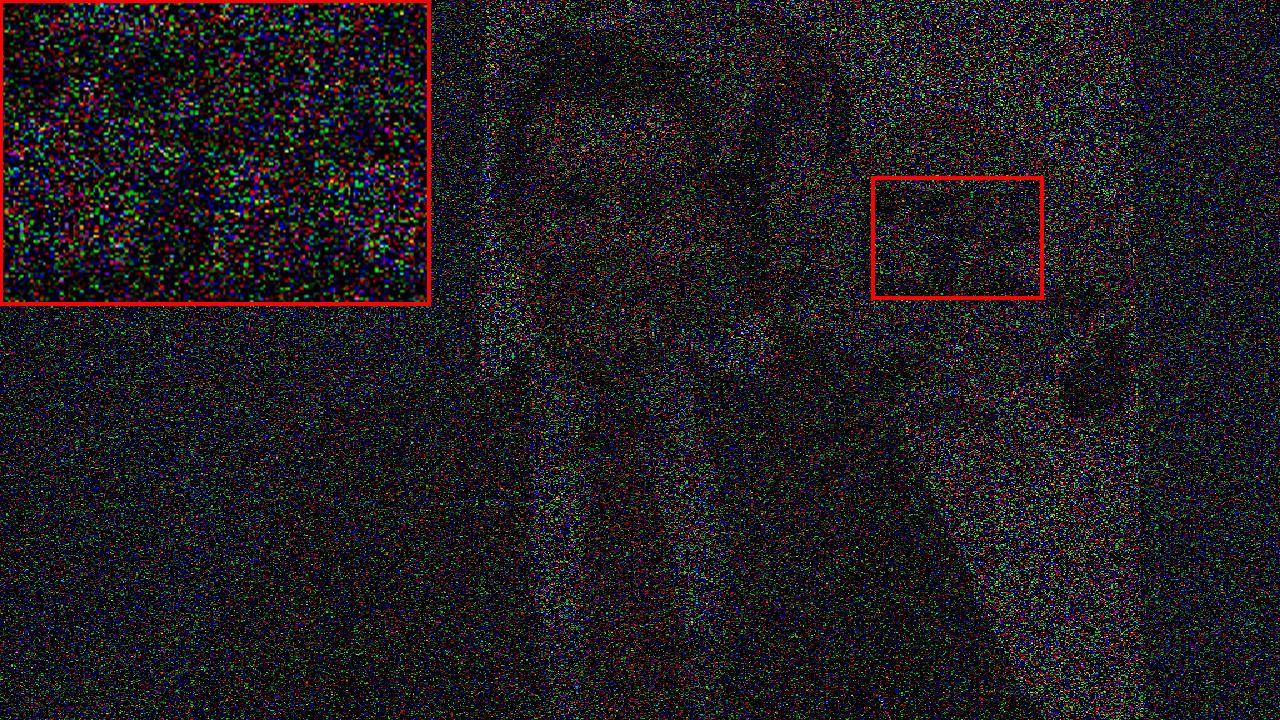}&
\includegraphics[width=0.585in, height=0.48in]{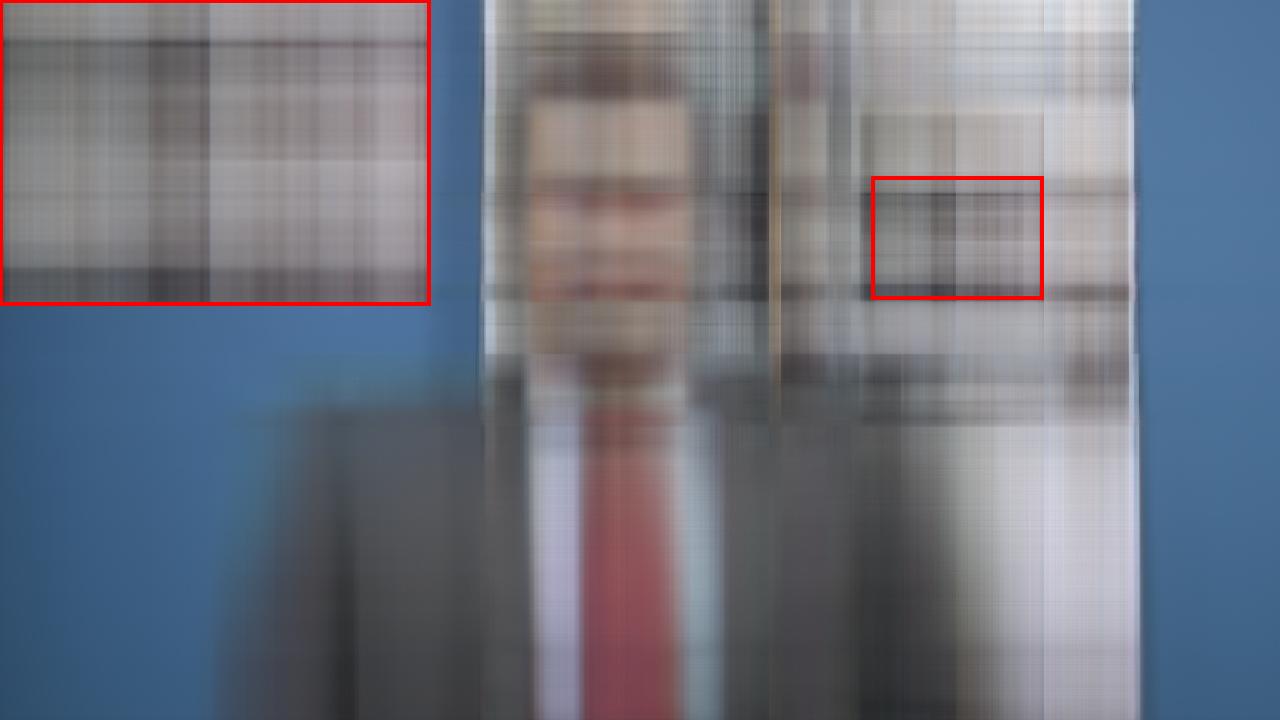}&
\includegraphics[width=0.585in, height=0.48in]{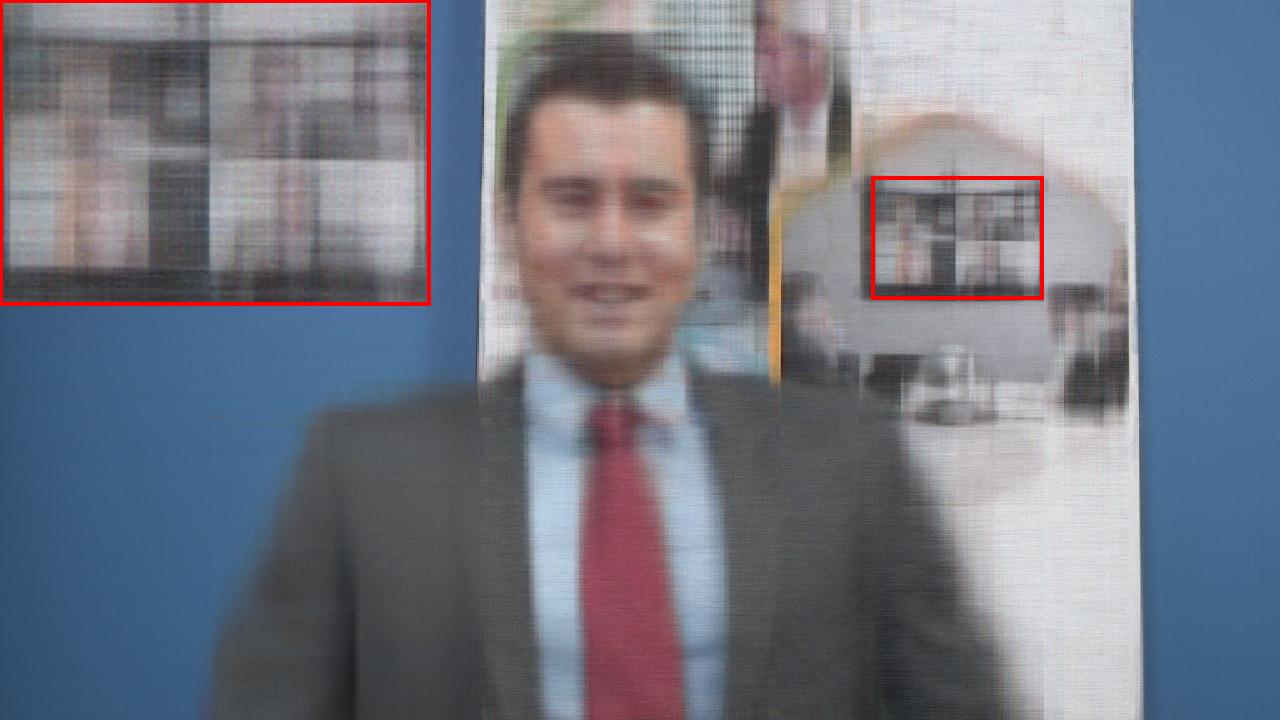}&
\includegraphics[width=0.585in, height=0.48in]{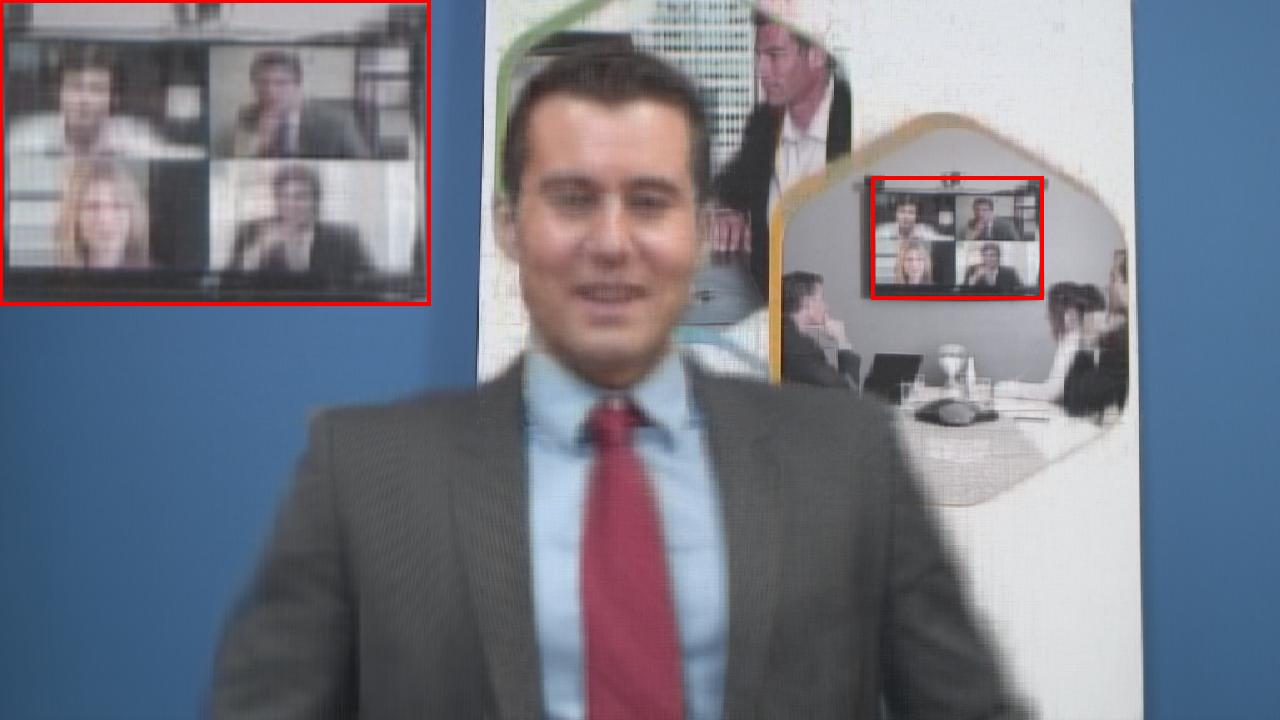}&
\includegraphics[width=0.585in, height=0.48in]{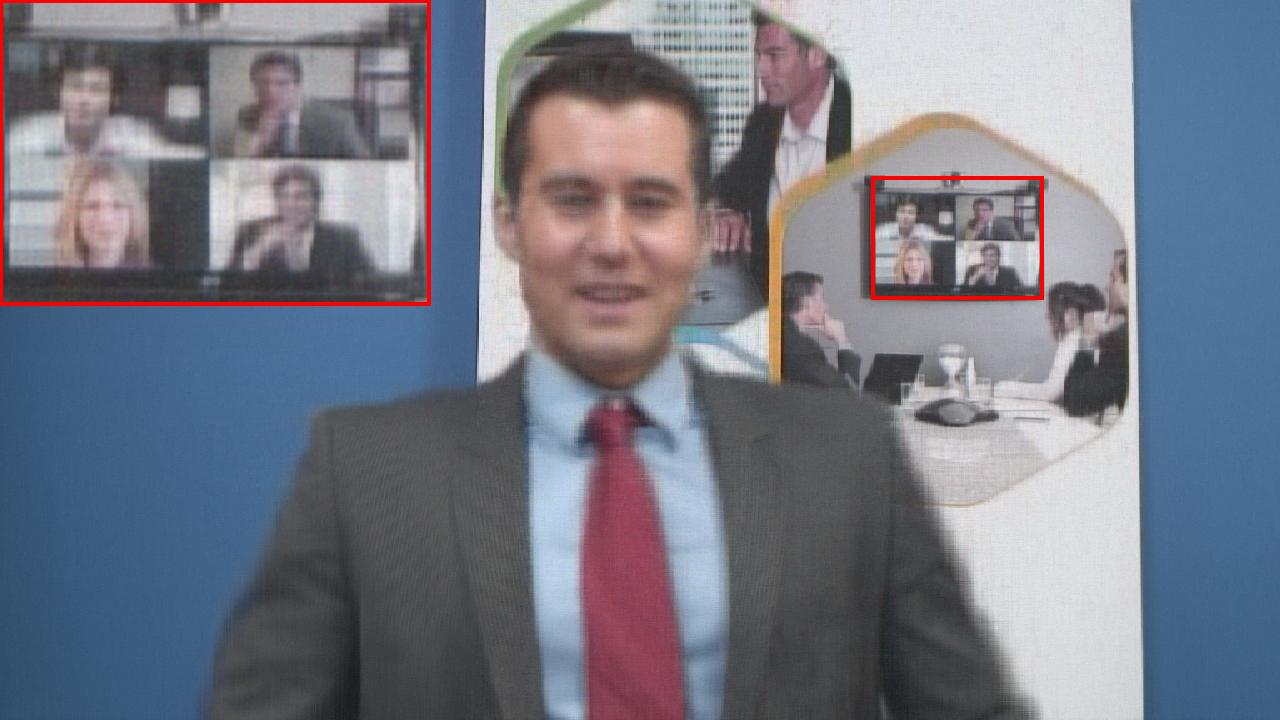}&
\includegraphics[width=0.585in, height=0.48in]{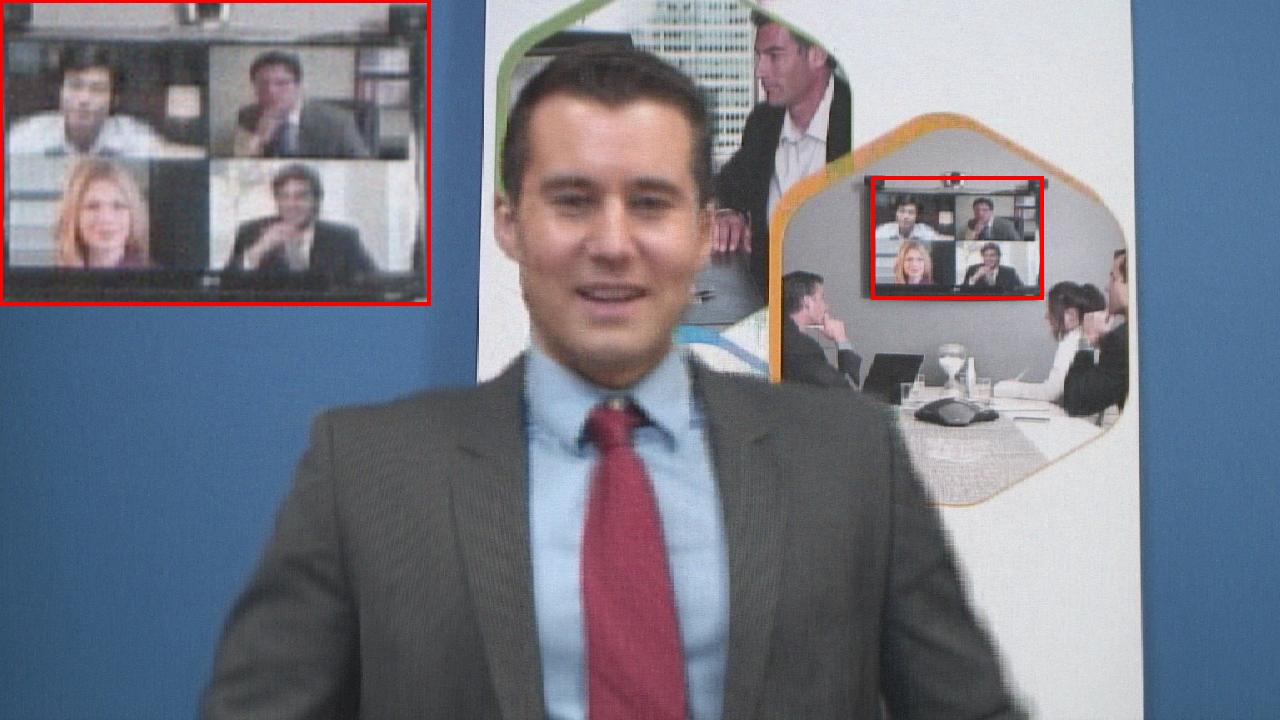}&
\includegraphics[width=0.585in, height=0.48in]{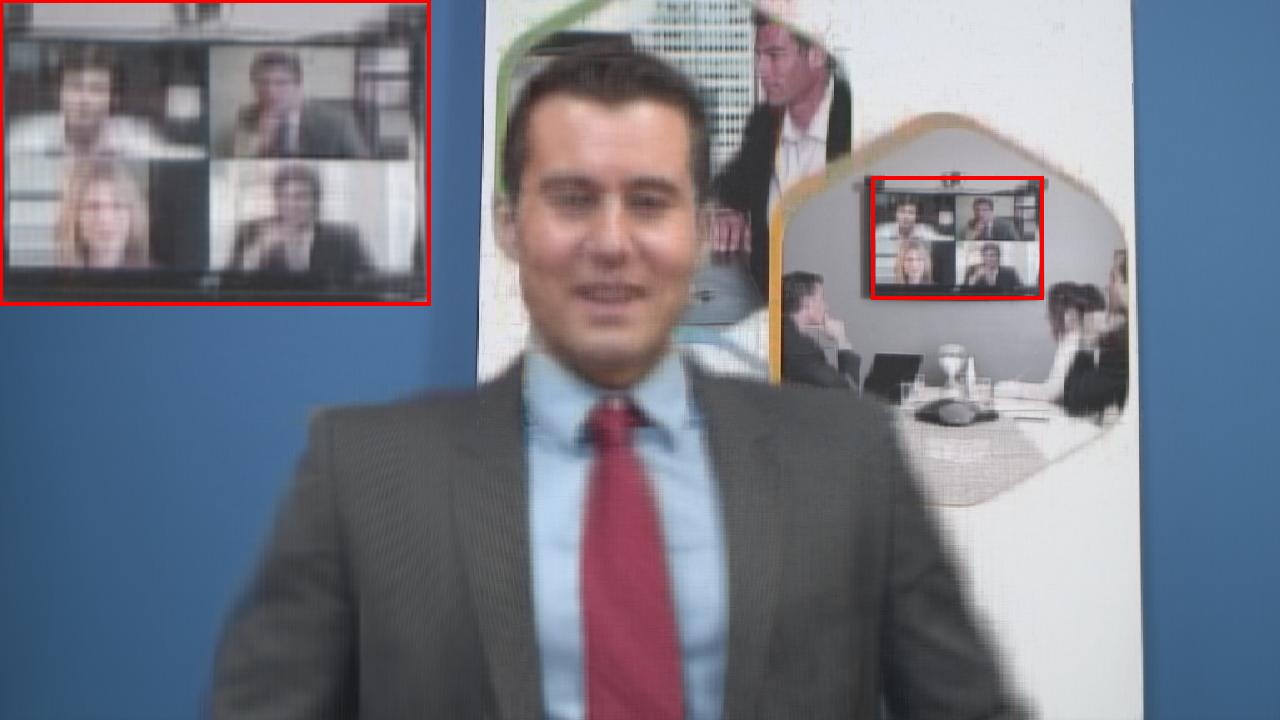}&
\includegraphics[width=0.585in, height=0.48in]{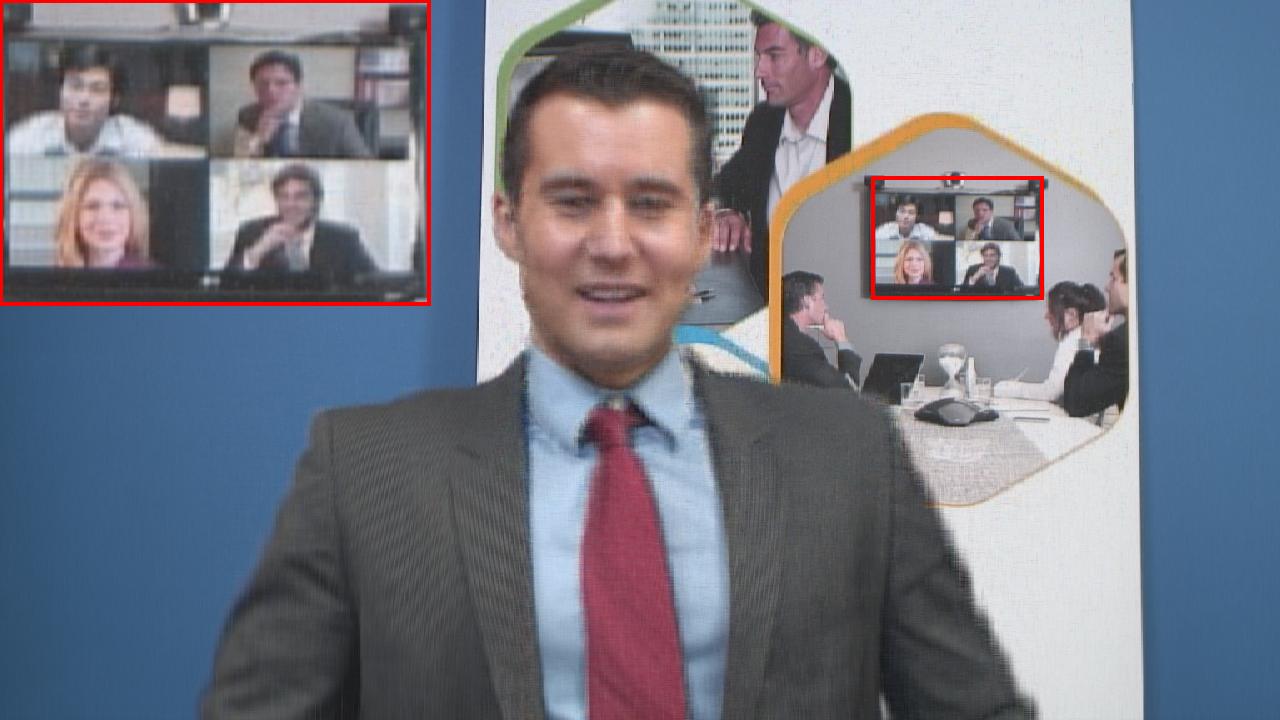}&
\includegraphics[width=0.585in, height=0.48in]{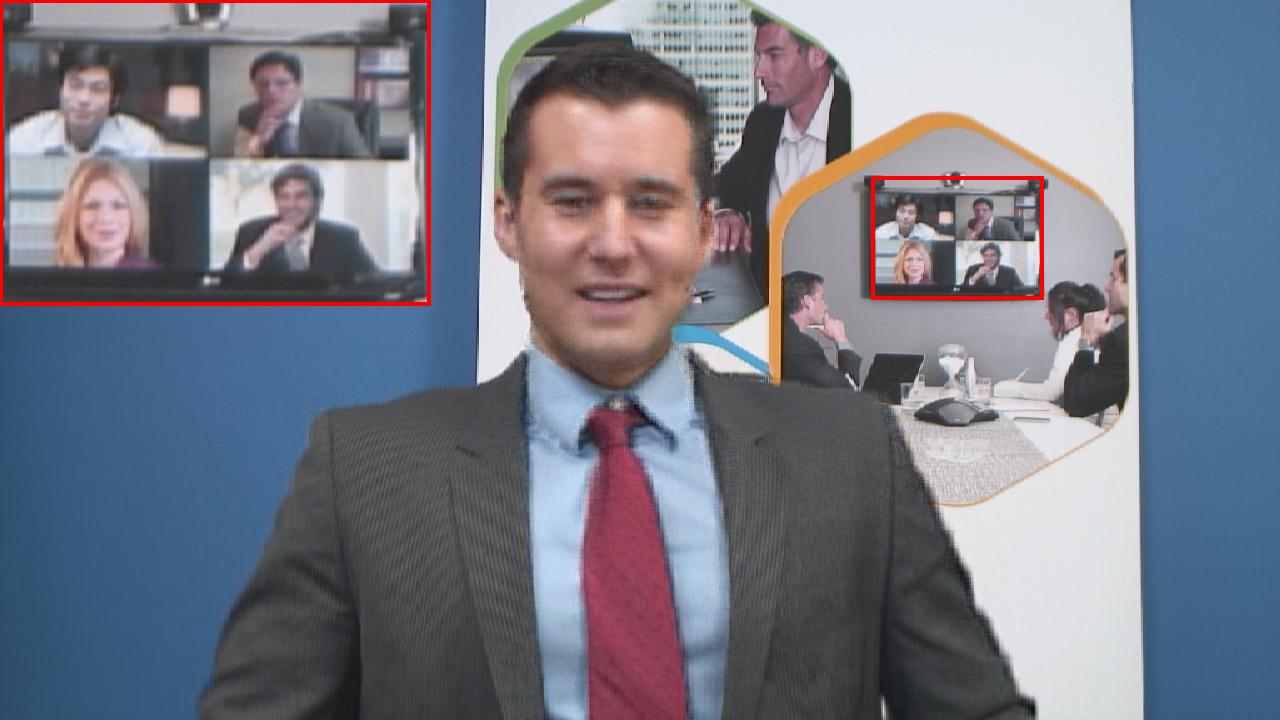}&
\includegraphics[width=0.585in, height=0.48in]{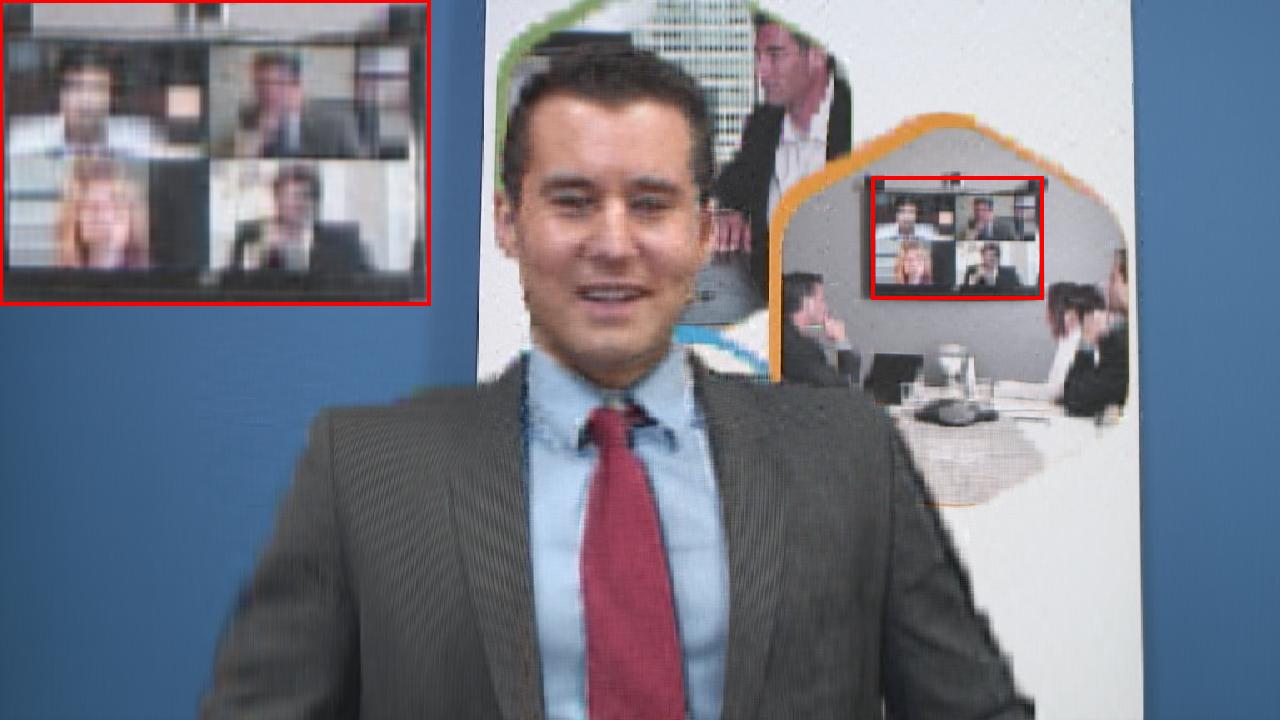}&
\includegraphics[width=0.585in, height=0.48in]{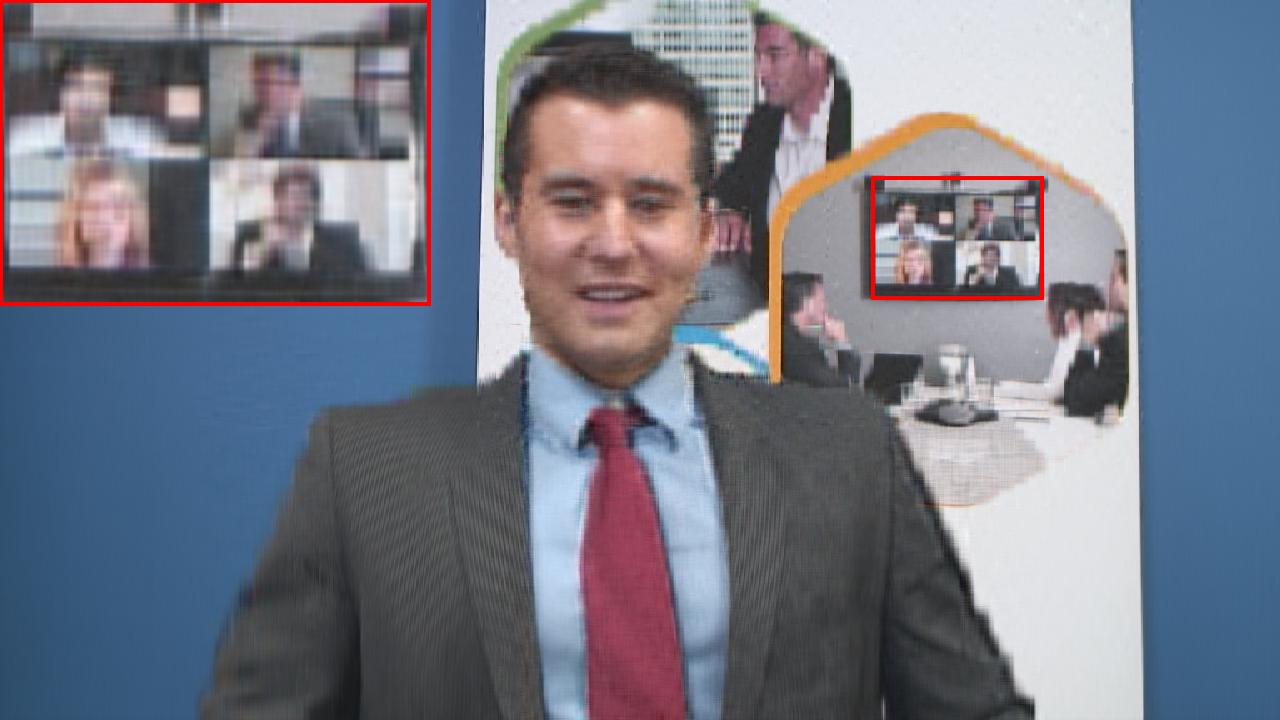}
&
\includegraphics[width=0.585in, height=0.48in,angle=0]{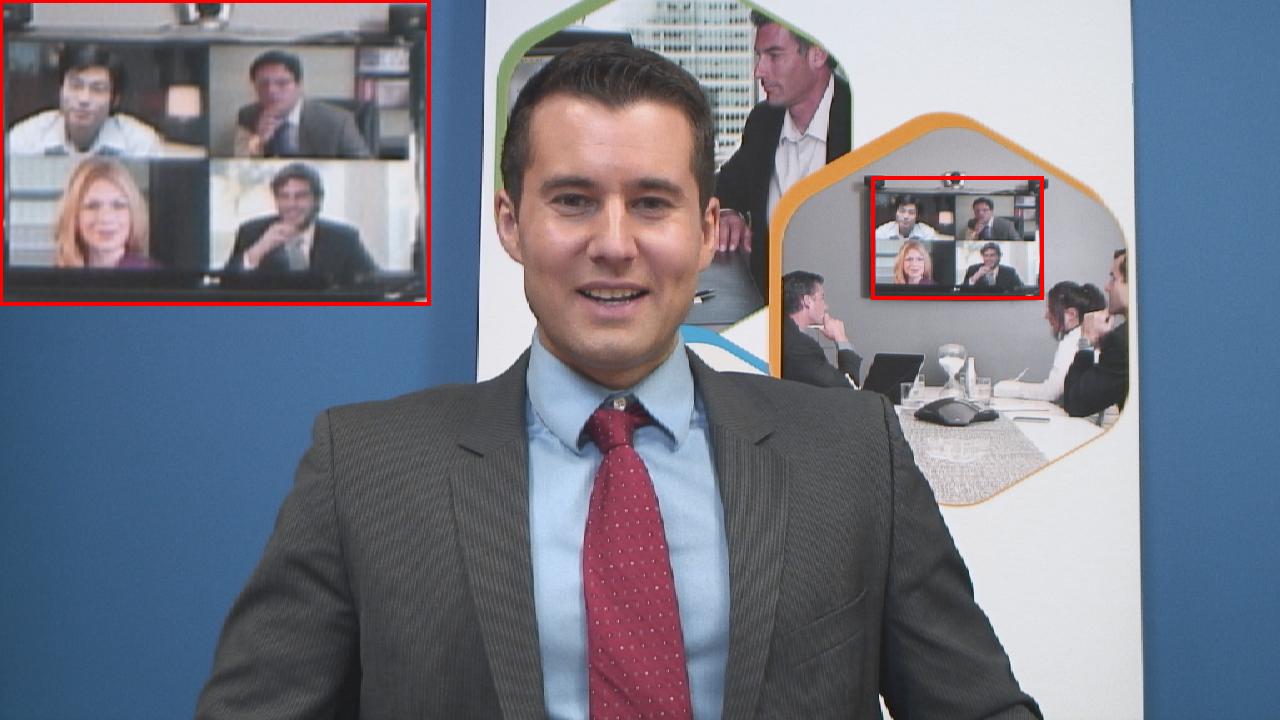}
\\


\textbf{\scriptsize{{Observed}}}  &
  \scriptsize{SNN+$\ell_1$}  &\scriptsize {TRNN+$\ell_1$}
&\scriptsize{TTNN+$\ell_1$} &\scriptsize {TSP-$k$+$\ell_1$}
 &\scriptsize{LNOP}&\scriptsize {NRTRM}&\scriptsize {HWTNN+$\ell_1$}
 & \textbf{\scriptsize {HWTSN+w$\ell_q$}}
  &

 \textbf{\tabincell{c}{ \scriptsize {HWTSN+w$\ell_q$} \\  \scriptsize{(BR)}
     }}

   &
   \textbf{\tabincell{c}{ \scriptsize {HWTSN+w$\ell_q$} \\  \scriptsize{(UR)}
     }}&
     \textbf{\scriptsize{Ground truth}}

%
\end{tabular}
\caption{
Visual comparison of various methods for 
CVs restoration.
From top to bottom, the parameter pair $(sr, \tau)$  are 
$(0.1,0.3)$, $(0.1,0.5)$, $(0.2,0.3)$ and $(0.2,0.5)$, respectively.
Top row: 
 the  $60$-th frame of  Rush-hour.
The second row: the  $50$-th frame of  Stockholm.
%
%
The third row: the  $34$-th frame of Intotree. 
Bottom row: 
the $10$-th frame of Johnny.
}
\vspace{-0.15cm}
\label{fig_visual_cv}
\end{figure*}
 \begin{table*}[htbp]
  \caption{
 The PSNR, SSIM values and CPU time
  obtained by  various RLRTC methods for different fourth-order CVs.
   The best  and the second-best results are highlighted in
blue and red, respectively.
  }
  \label{cv_index}

  \centering
\scriptsize
\renewcommand{\arraystretch}{0.880}
\setlength\tabcolsep{4.5pt}

\begin{tabular}{c c ccccccc  cccc   cccc | c}

     \hline
     {CV-Name}
     &  \multicolumn{4}{|c|}{Rush-hour
     } &\multicolumn{4}{c|}{Johnny} &\multicolumn{4}{c|}{Stockholm} &\multicolumn{4}{c|}{Intotree}  
     & \multirow{3}{*}{
   \tabincell{c}{Average\\Time (s)}
   }\\
     \cline{1-1}
     \cline{2-5}
     \cline{6-9}
      \cline{10-13}
      \cline{14-17}
     $sr$
     &\multicolumn{2}{|c|}{$10\%$}   &\multicolumn{2}{c|}{$20\%$}  &
     \multicolumn{2}{c|}{$10\%$}&\multicolumn{2}{c|}{$20\%$}&
     \multicolumn{2}{c|}{$10\%$}&\multicolumn{2}{c|}{$20\%$}&
     \multicolumn{2}{c|}{$10\%$}&\multicolumn{2}{c|}{$20\%$}\\
     \cline{1-1}
     \cline{2-5}
     \cline{6-9}
      \cline{10-13}
      \cline{14-17}
    $\tau$
    &
    \multicolumn{1}{|c|}{$30\%$}&\multicolumn{1}{c|}{$50\%$}  &
    \multicolumn{1}{c|}{$30\%$}&\multicolumn{1}{c|}{$50\%$}  &
    \multicolumn{1}{c|}{$30\%$}&\multicolumn{1}{c|}{$50\%$}  &
    \multicolumn{1}{c|}{$30\%$}&\multicolumn{1}{c|}{$50\%$}
    &
    \multicolumn{1}{c|}{$30\%$}&\multicolumn{1}{c|}{$50\%$}  &
    \multicolumn{1}{c|}{$30\%$}&\multicolumn{1}{c|}{$50\%$}  &
    \multicolumn{1}{c|}{$30\%$}&\multicolumn{1}{c|}{$50\%$}  &
    \multicolumn{1}{c|}{$30\%$}&\multicolumn{1}{c|}{$50\%$}
    \\
    \hline
     \hline
    SNN+$\ell_1$
     &

\tabincell{c}{18.37   \\0.711    }   &    \tabincell{c}{ 17.26  \\0.559  }&
\tabincell{c}{ 21.78   \\ 0.771    }   &    \tabincell{c}{19.57\\0.689 }&

\tabincell{c}{  18.03   \\0.758    }   &    \tabincell{c}{16.61 \\0.734   }&
\tabincell{c}{  21.80  \\0.803    }   &    \tabincell{c}{ 18.61\\ 0.745}&

\tabincell{c}{   19.96   \\0.501   }   &    \tabincell{c}{18.96  \\ 0.484  }&
\tabincell{c}{21.42   \\  0.537    }   &    \tabincell{c}{ 20.53\\0.502 }&

\tabincell{c}{ 20.81   \\0.591   }   &    \tabincell{c}{18.88  \\  0.512   }&
\tabincell{c}{23.08   \\0.622   }   &    \tabincell{c}{ 21.85\\ 0.539 }
& 12968
\\


%
%
%


%
%
%
%

      \hline
    TRNN+$\ell_1$ &

   \tabincell{c}{ 23.79    \\ 0.733   }   &    \tabincell{c}{20.47  \\  0.701 }&
\tabincell{c}{ 26.28  \\0.811    }   &    \tabincell{c}{ 23.02\\ 0.718}&

\tabincell{c}{  24.59  \\ 0.765  }   &    \tabincell{c}{ 19.84  \\  0.739  }&
\tabincell{c}{27.18   \\0.847    }   &    \tabincell{c}{23.06\\0.755 }&

\tabincell{c}{ 22.19    \\0.499  }   &    \tabincell{c}{20.77  \\   0.488 }&
\tabincell{c}{ 23.48 \\0.594    }   &    \tabincell{c}{ 21.64\\ 0.512}&

\tabincell{c}{  24.54  \\ 0.599   }   &    \tabincell{c}{ 22.28\\ 0.567}&
\tabincell{c}{ 25.97    \\ 0.631    }   &    \tabincell{c}{23.64\\  0.572}&6968
\\

%
%
%
%
%
%
%
%
%
 \hline
   TTNN+$\ell_1$
   &

   \tabincell{c}{24.77   \\ 0.758   }   &    \tabincell{c}{ 22.39  \\  0.734 }&
\tabincell{c}{ 28.13  \\ 0.833   }   &    \tabincell{c}{ 24.94\\0.753 }&

\tabincell{c}{  27.29  \\0.868   }   &    \tabincell{c}{ 24.19 \\ 0.782   }&
\tabincell{c}{ 30.57  \\0.911   }   &    \tabincell{c}{ 26.93\\  0.791}&

\tabincell{c}{23.02    \\0.609   }   &    \tabincell{c}{21.32 \\  0.494 }&
\tabincell{c}{ 25.82   \\ 0.649  }   &    \tabincell{c}{ 22.82\\  0.567 }&

\tabincell{c}{25.51   \\0.605   }   &    \tabincell{c}{23.64   \\  0.577 }&
\tabincell{c}{ 27.54   \\0.632   }   &    \tabincell{c}{24.99\\  0.602}&5843
\\
%
%
%
%
%
%
%
 \hline
   TSP-$k$+$\ell_1$
   &

   \tabincell{c}{26.59    \\0.819    }   &    \tabincell{c}{23.46  \\0.741  }&
\tabincell{c}{  28.68   \\0.845   }   &    \tabincell{c}{25.46\\  0.808}&

\tabincell{c}{ 29.42   \\0.886    }   &    \tabincell{c}{24.98 \\ 0.813 }&
\tabincell{c}{  31.55  \\0.912    }   &    \tabincell{c}{ 27.48\\ 0.835}&

\tabincell{c}{ 24.39   \\0.643    }   &    \tabincell{c}{21.81 \\ 0.526  }&
\tabincell{c}{  26.08   \\0.686    }   &    \tabincell{c}{23.24\\0.593 }&

\tabincell{c}{ 26.58   \\0.633    }   &    \tabincell{c}{23.64 \\ 0.578 }&
\tabincell{c}{ 27.85   \\0.637    }   &    \tabincell{c}{24.85\\ 0.617}&9924
\\
%
%
%
%
%
%
%
     \hline
     LNOP
     &

%
%
%

 \tabincell{c}{ 26.67      \\0.830    }   &    \tabincell{c}{ 23.31  \\ 0.748  }&
\tabincell{c}{29.13      \\0.857        }   &    \tabincell{c}{  25.91\\  0.829}&

\tabincell{c}{  28.03      \\ 0.768  }   &    \tabincell{c}{ 23.69 \\ 0.742      }&
\tabincell{c}{ 29.55      \\0.861   }   &    \tabincell{c}{ 28.39\\   0.832}&

\tabincell{c}{ 24.66       \\ 0.638     }   &    \tabincell{c}{22.39 \\ 0.533  }&
\tabincell{c}{ 26.85      \\0.694    }   &    \tabincell{c}{ 24.43\\  0.603  }&

\tabincell{c}{ 26.24      \\ 0.609   }   &    \tabincell{c}{ 23.78   \\ 0.603   }&
\tabincell{c}{27.70       \\0.634    }   &    \tabincell{c}{ 25.95\\0.623 }&9138
\\

  \hline
     NRTRM
     &

   \tabincell{c}{ 26.52      \\  0.831     }   &    \tabincell{c}{ 23.94 \\ 0.772  }&
\tabincell{c}{ 29.06       \\0.855     }   &    \tabincell{c}{ 25.16\\  0.776}&

\tabincell{c}{ 28.91       \\ 0.891    }   &    \tabincell{c}{ 25.96  \\  0.838  }&
\tabincell{c}{ 31.61       \\0.919     }   &    \tabincell{c}{ 27.27\\ 0.868 }&

\tabincell{c}{ 24.33      \\ 0.642    }   &    \tabincell{c}{ 22.19 \\  0.538    }&
\tabincell{c}{ 26.21      \\0.706      }   &    \tabincell{c}{ 22.89\\ 0.605 }&

\tabincell{c}{ 26.49       \\ 0.639     }   &    \tabincell{c}{ 24.46  \\ 0.611  }&
\tabincell{c}{27.86       \\ 0.661     }   &    \tabincell{c}{ 25.19\\  0.632}
&7834
\\

%
%
%
%
%
%
%
%
%
%

     \hline
HWTNN+$\ell_1$
     &

   \tabincell{c}{  29.23        \\0.839      }   &    \tabincell{c}{26.09   \\ 0.777  }&
\tabincell{c}{ \textcolor[rgb]{1.00,0.00,0.00}{32.22}      \\ \textcolor[rgb]{1.00,0.00,0.00}{0.876}}   &    \tabincell{c}{ 28.22\\ 0.838}&

\tabincell{c}{    \textcolor[rgb]{1.00,0.00,0.00}{31.25}       \\ \textcolor[rgb]{1.00,0.00,0.00}{0.893}}   &    \tabincell{c}{28.16   \\ 0.843  }&
\tabincell{c}{ \textcolor[rgb]{1.00,0.00,0.00}{34.24}       \\\textcolor[rgb]{1.00,0.00,0.00}{0.923}}   &    \tabincell{c}{ \textcolor[rgb]{1.00,0.00,0.00}{30.43}\\\textcolor[rgb]{1.00,0.00,0.00}{0.876}}&

\tabincell{c}{26.16     \\ 0.663     }   &    \tabincell{c}{   23.34   \\ 0.551 }&
\tabincell{c}{  28.38      \\ 0.718     }   &    \tabincell{c}{ 24.66\\  0.627 }&

\tabincell{c}{  27.49       \\ 0.647   }   &    \tabincell{c}{ 25.56   \\   0.619   }&
\tabincell{c}{\textcolor[rgb]{1.00,0.00,0.00}{28.72}       \\\textcolor[rgb]{1.00,0.00,0.00}{0.687}}   &    \tabincell{c}{ 26.27\\   0.643}
& 6737 
\\
     \hline
    \textbf{HWTSN+w$\ell_q$}
   &

   \tabincell{c}{\textcolor[rgb]{0.00,0.00,1.00}{30.27}        \\\textcolor[rgb]{0.00,0.00,1.00}{0.851}}   &    \tabincell{c}{\textcolor[rgb]{0.00,0.00,1.00}{27.38} \\  \textcolor[rgb]{0.00,0.00,1.00}{0.794}}&
\tabincell{c}{\textcolor[rgb]{0.00,0.00,1.00}{33.68}         \\ \textcolor[rgb]{0.00,0.00,1.00}{0.885}}   &    \tabincell{c}{\textcolor[rgb]{0.00,0.00,1.00}{30.02}\\   \textcolor[rgb]{0.00,0.00,1.00}{0.867}}&

\tabincell{c}{ \textcolor[rgb]{0.00,0.00,1.00}{32.36}      \\  \textcolor[rgb]{0.00,0.00,1.00}{0.905}}   &    \tabincell{c}{  \textcolor[rgb]{0.00,0.00,1.00}{29.45}  \\ \textcolor[rgb]{0.00,0.00,1.00}{0.868}}&
\tabincell{c}{\textcolor[rgb]{0.00,0.00,1.00}{35.33}       \\ \textcolor[rgb]{0.00,0.00,1.00}{0.931}}   &    \tabincell{c}{ \textcolor[rgb]{0.00,0.00,1.00}{32.07}\\  \textcolor[rgb]{0.00,0.00,1.00}{0.907}}&

\tabincell{c}{ \textcolor[rgb]{0.00,0.00,1.00}{26.86}      \\ 0.668      }   &    \tabincell{c}{ 24.01  \\0.554   }&
\tabincell{c}{  28.62     \\0.741     }   &    \tabincell{c}{  \textcolor[rgb]{0.00,0.00,1.00}{26.89}\\  \textcolor[rgb]{0.00,0.00,1.00}{0.707}}&

\tabincell{c}{\textcolor[rgb]{0.00,0.00,1.00}{27.92}      \\  \textcolor[rgb]{0.00,0.00,1.00}{0.659}}   &    \tabincell{c}{ \textcolor[rgb]{0.00,0.00,1.00}{26.53}   \\  \textcolor[rgb]{0.00,0.00,1.00}{0.638} }&
\tabincell{c}{\textcolor[rgb]{0.00,0.00,1.00}{28.98}        \\\textcolor[rgb]{0.00,0.00,1.00}{0.695}}   &    \tabincell{c}{\textcolor[rgb]{0.00,0.00,1.00}{27.94}\\\textcolor[rgb]{0.00,0.00,1.00}{0.667}}
& 8760
\\

  \hline

       \tabincell{c}{ \textbf {HWTSN+w$\ell_q$(BR)}
    }
     &

   \tabincell{c}{ 29.15   \\0.841 }   &    \tabincell{c}{26.66  \\ 0.784 }&
\tabincell{c}{  31.27  \\0.868}   &    \tabincell{c}{ 28.89\\     0.850 }&

\tabincell{c}{  29.85  \\0.868}   &    \tabincell{c}{ 28.06   \\0.844 }&
\tabincell{c}{30.66  \\0.879}   &    \tabincell{c}{ 29.46 \\  0.869}&

\tabincell{c}{26.50   \\\textcolor[rgb]{1.00,0.00,0.00}{0.686}}   &    \tabincell{c}{ \textcolor[rgb]{1.00,0.00,0.00}{24.19}  \\ \textcolor[rgb]{1.00,0.00,0.00}{0.579}}&
\tabincell{c}{ \textcolor[rgb]{1.00,0.00,0.00}{29.21}   \\\textcolor[rgb]{1.00,0.00,0.00}{0.755}}   &    \tabincell{c}{ 26.19\\   0.688}&

\tabincell{c}{27.60     \\ 0.649    }   &    \tabincell{c}{ 26.16   \\ 0.623  }&
\tabincell{c}{ 28.43    \\0.679}   &    \tabincell{c}{27.38\\  0.658}&\textcolor[rgb]{0.00,0.00,1.00}{4243}
\\

     \hline
 \tabincell{c}{\textbf{{HWTSN+w$\ell_q$(UR)}}
    }
&

   \tabincell{c}{\textcolor[rgb]{1.00,0.00,0.00}{29.28}  \\ \textcolor[rgb]{1.00,0.00,0.00}{0.842}}   &    \tabincell{c}{ \textcolor[rgb]{1.00,0.00,0.00}{26.77}   \\ \textcolor[rgb]{1.00,0.00,0.00}{0.789}}&
\tabincell{c}{31.41    \\ 0.865    }   &    \tabincell{c}{ \textcolor[rgb]{1.00,0.00,0.00}{29.03}\\\textcolor[rgb]{1.00,0.00,0.00}{0.852}}&

\tabincell{c}{  29.93  \\ 0.869    }   &    \tabincell{c}{\textcolor[rgb]{1.00,0.00,0.00}{28.19}  \\ \textcolor[rgb]{1.00,0.00,0.00}{0.844}}&
\tabincell{c}{  30.76   \\  0.878   }   &    \tabincell{c}{29.53\\  0.870}&

\tabincell{c}{ \textcolor[rgb]{1.00,0.00,0.00}{26.60}  \\\textcolor[rgb]{0.00,0.00,1.00}{0.688}}   &    \tabincell{c}{ \textcolor[rgb]{0.00,0.00,1.00}{24.27} \\ \textcolor[rgb]{0.00,0.00,1.00}{0.582}}&
\tabincell{c}{  \textcolor[rgb]{0.00,0.00,1.00}{29.27}  \\\textcolor[rgb]{0.00,0.00,1.00}{0.758}}   &    \tabincell{c}{ \textcolor[rgb]{1.00,0.00,0.00}{26.30}\\  \textcolor[rgb]{1.00,0.00,0.00}{0.692}}&

\tabincell{c}{\textcolor[rgb]{1.00,0.00,0.00}{27.68}    \\\textcolor[rgb]{1.00,0.00,0.00}{0.651}}   &    \tabincell{c}{ \textcolor[rgb]{1.00,0.00,0.00}{26.21}  \\\textcolor[rgb]{1.00,0.00,0.00}{0.625}}&
\tabincell{c}{  28.45  \\0.684    }   &    \tabincell{c}{ \textcolor[rgb]{1.00,0.00,0.00}{27.48}\\ \textcolor[rgb]{1.00,0.00,0.00}{0.659}}
&\textcolor[rgb]{1.00,0.00,0.00}{4374}
\\

     \hline

\end{tabular}
\begin{tablenotes}
\item[**]
In each RLRTC method,
the top represents the PSNR values   while 
the bottom denotes the SSIM values.
\end{tablenotes}
\vspace{-0.45cm}
\end{table*}

Figure \ref{fig_visual_lfi} shows the recovered LFIs and corresponding zoomed regions  acquired by 
different RLRTC methods
at extremely low sampling rates.
From the enlarged areas, 
we can observed that
the LFIs restored  by our method 
 preserves more details than those achieved by 
 other competitive algorithms. 
The PSNR, SSIM values and average CPU time of various RLRTC methods for   four   LFIs with different  noise levels and 
observation ratios
are displayed in Table \ref{lfi_index}.
The following conclusions can be drawn from above quantitative metrics. 
\textbf{1)}
Classical 
methods induced by the Tucker and TR format, i.e., SNN+$\ell_1$ and TRNN+$\ell_1$, perform
relative poorer in term of recovery quality. 
\textbf{2)}
 Among those
 algorithms 
 induced by third-order T-SVD,
although the ones employing 
nonconvex  schemes (i.e., LNOP and NRTRM) require more running time  over the ones utilizing  convex methods
(i.e., TTNN+$\ell_1$ and TSP-$k$+$\ell_1$),
they achieve higher   PSNR and SSIM values in most cases.
This phenomenon also exists in the 
 methods based on high-order T-SVD.
\textbf{3)}
Compared with other methods,
the \textbf{HWTSN+w$\ell_q$} 
achieves  an approximately 
$2$$\sim$$5$ dB
 gain in the mean PSNR values, and
its accelerated versions obtains 
an about $40\%$$\sim$$60\%$ percent drop in the average CPU time. 
\textbf{4)} In our 
%
%
algorithms,
the versions fused randomization ideas
shorten  the
running time by about  $55\%$ percent over the  deterministic version
 with a slight reduction
 of psnr and ssim values.

 \subsubsection {\textbf{Application in Color Videos Restoration}}  
 In this experiment, color videos (CVs) are used to evaluate the performance of the proposed algorithm.
We   download four large-scale CVs 
from  the derf website \footnote{\url{https://media.xiph.org/video/derf/}}
for this test.
Only the first $100$ frames of each video sequence are selected as
the test data owing  to the computational limitation,
in which  each frame has the  size $720 \times 1280 \times 3$.
For each CV 
 with $100$ frames, it can be formulated as an
$720 \times 1280 \times 3 \times 100$
 fourth-order tensor.

Figure \ref{fig_visual_cv} displays the  visual comparison of 
the proposed and competitive RLRTC algorithms
for  various 
CVs restoration.
From the zoomed regions, we can see that the \textbf{HWTSN+w$\ell_q$}  exhibits tangibly better 
restortation  quality
over other
 comparative methods 
according to the color, brightness,  and outline. 
In Table \ref{cv_index}, we report
the PSNR, SSIM values and  CPU time of ten RLRTC methods for four CVs,
where $sr=0.1,0.2$ and  $\tau=0.3,0.5$.
These results show that the PSNR  and SSIM metrics acquired by \textbf{HWTSN+w$\ell_q$} are higher than those obtained by
the baseline method, i.e., {HWTNN+$\ell_1$}.
%
In contrast to the competitive  
non-convex 
methods (i.e., LNOP and NRTRM),
the improvements of proposed non-convex algorithm (i.e.,\textbf{HWTSN+w$\ell_q$}) are around $3$ dB  
 in term of PSNR index
while the reductions  of its randomized version are about  $52\%$  according to  the CPU time. 
Furthermore, under the comprehensive balance of PSNR, SSIM, and CPU time,
the proposed randomized RHTC method
is always superior to other popular algorithms.
Other findings are similar to the case of LFIs recovery.

\subsubsection{\textbf{Application in Multitemporal 
Remote Sensing Images Inpainting}}
This  experiment mainly 
tests three 
fourth-order multi-temporal remote sensing 
images (MRSIs), which are named  SPOT-5
\footnote{\url{{https://take5.theia.cnes.fr/atdistrib/take5/client/\#/home
% https://take5.theia.cnes.fr/atdistrib/take5/client/#/home
}}} ($2000 \times 2000 \times   4 \times 13$),
Landsat-7
($4500 \times 4500 \times   6 \times 11$),
 and
T22LGN
\footnote{\url{{https://theia.cnes.fr/atdistrib/rocket/\#/home
% https://take5.theia.cnes.fr/atdistrib/take5/client/#/home
}}} ($5001\times 5001 \times   4 \times 7$), respectively.
To speed up the calculation process,
 the spatial size of 
these
MRSIs
  is downsampled (resized) to 
  $2000\times 2000$.

%
\begin{figure*}[!htbp]
\renewcommand{\arraystretch}{0.528}
\setlength\tabcolsep{0.4pt}
\centering
\begin{tabular}{ccc ccc  cccccc}
\centering

\includegraphics[width=0.585in, height=0.585in,angle=0]{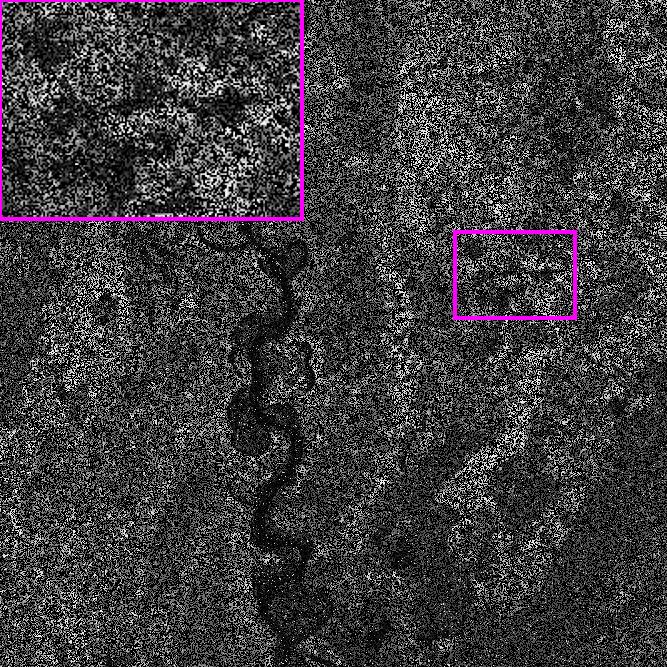}&
\includegraphics[width=0.585in, height=0.585in]{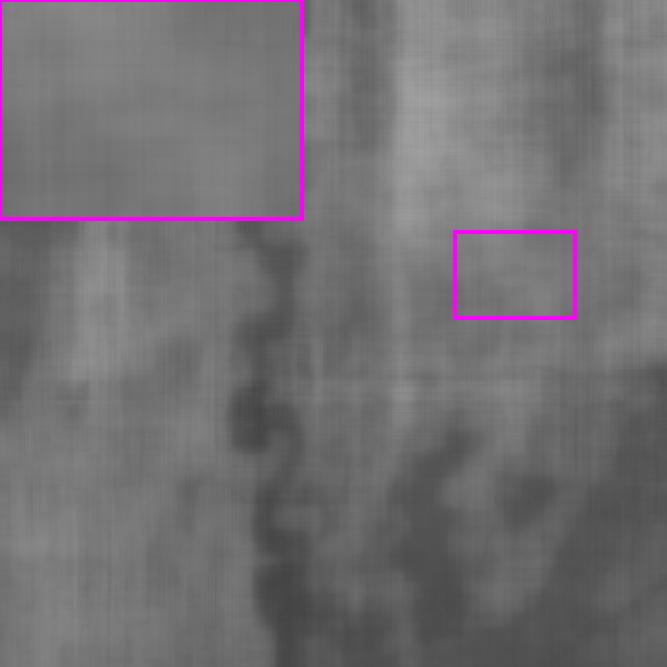}&
\includegraphics[width=0.585in, height=0.585in]{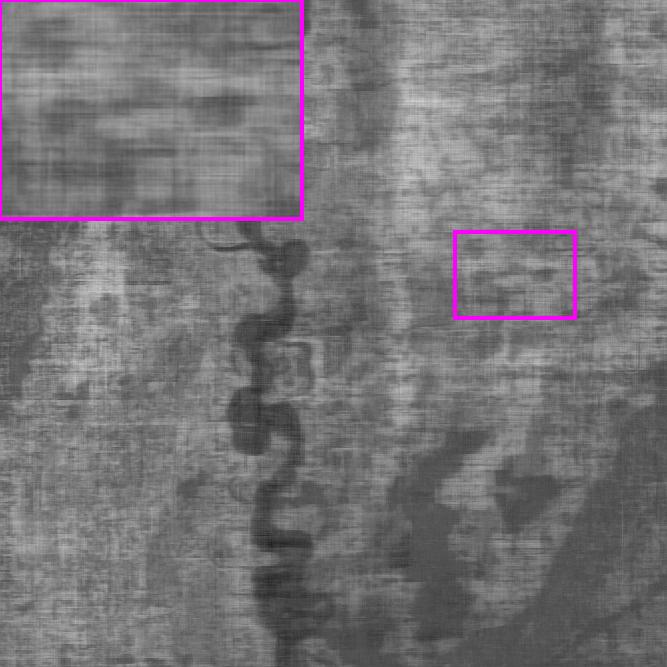}&
\includegraphics[width=0.585in, height=0.585in]{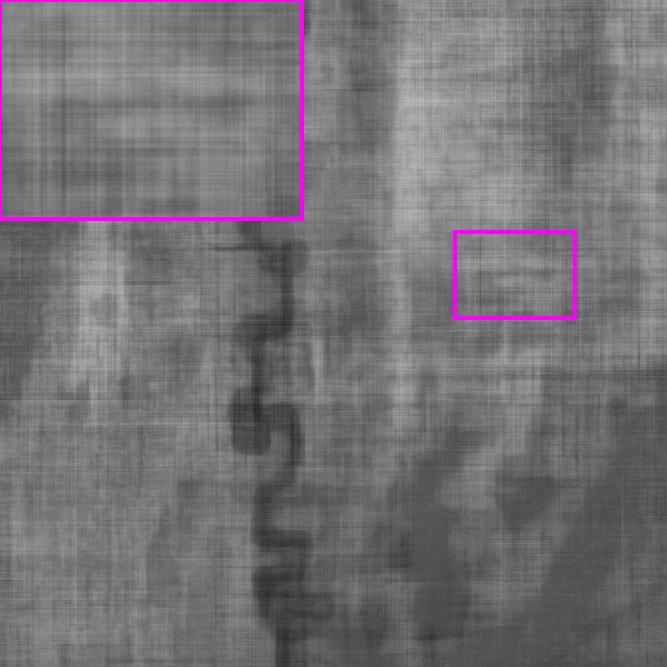}&
\includegraphics[width=0.585in, height=0.585in]{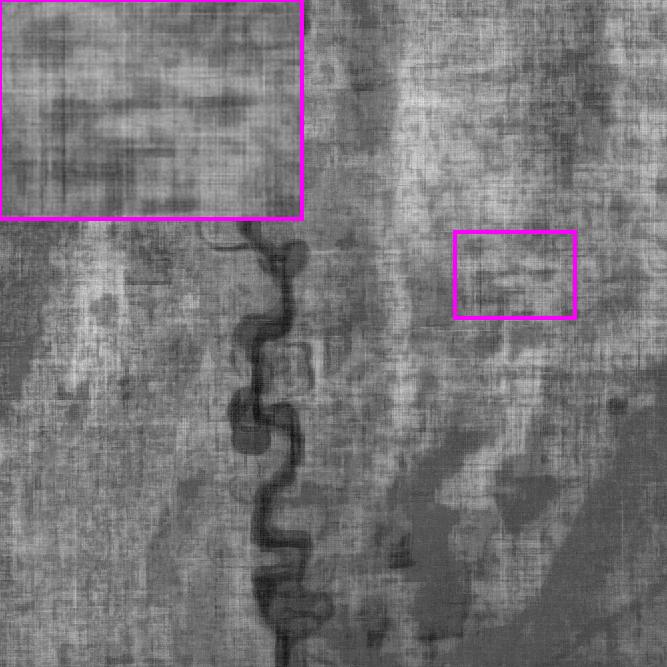}&
\includegraphics[width=0.585in, height=0.585in]{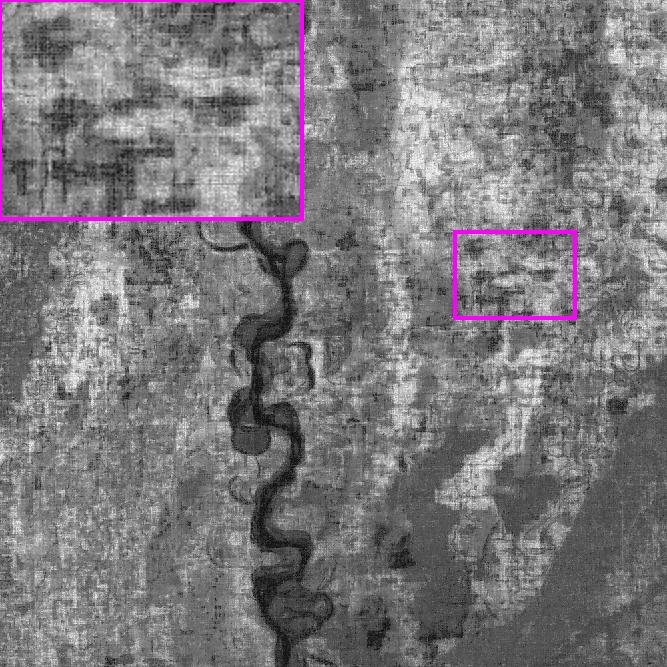}&
\includegraphics[width=0.585in, height=0.585in]{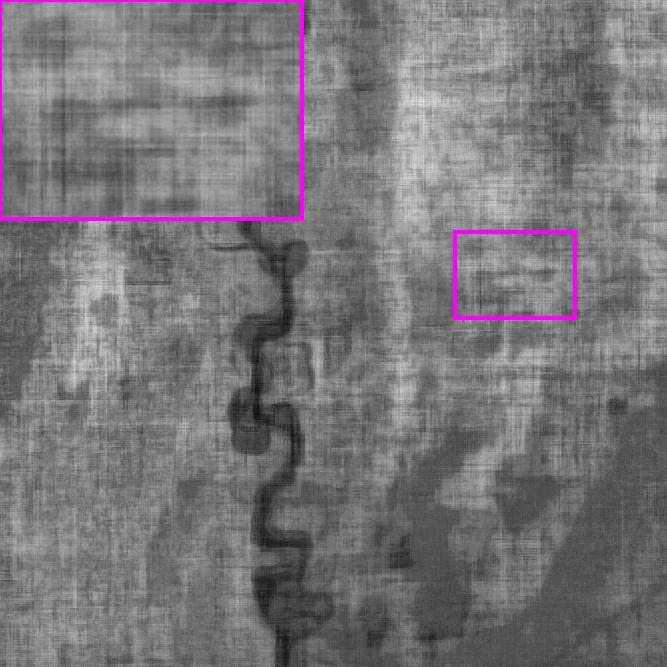}&
\includegraphics[width=0.585in, height=0.585in]{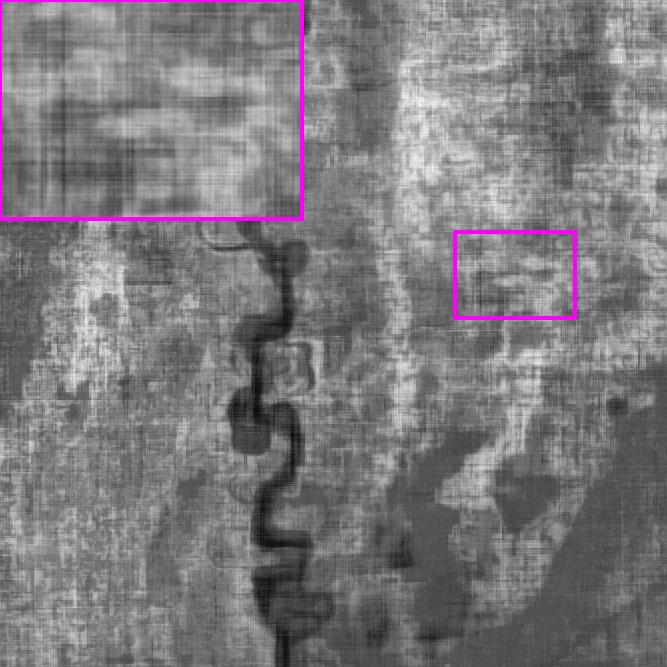}&
\includegraphics[width=0.585in, height=0.585in]{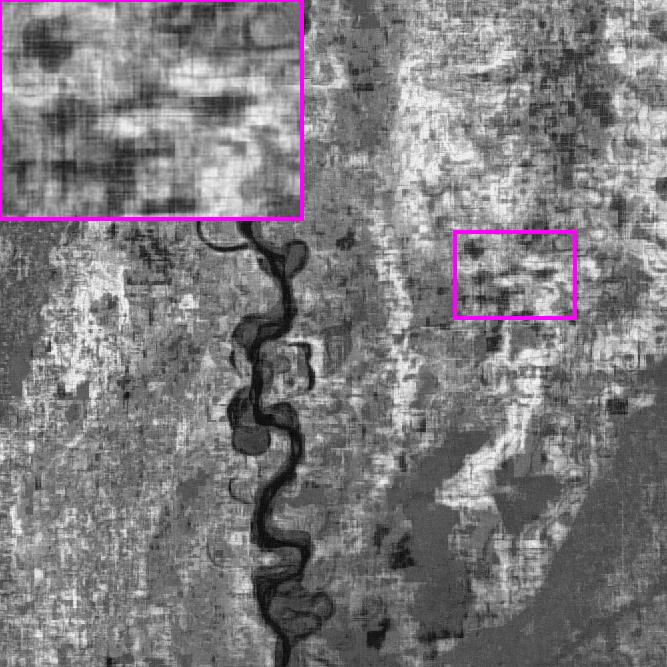}&
\includegraphics[width=0.585in, height=0.585in]{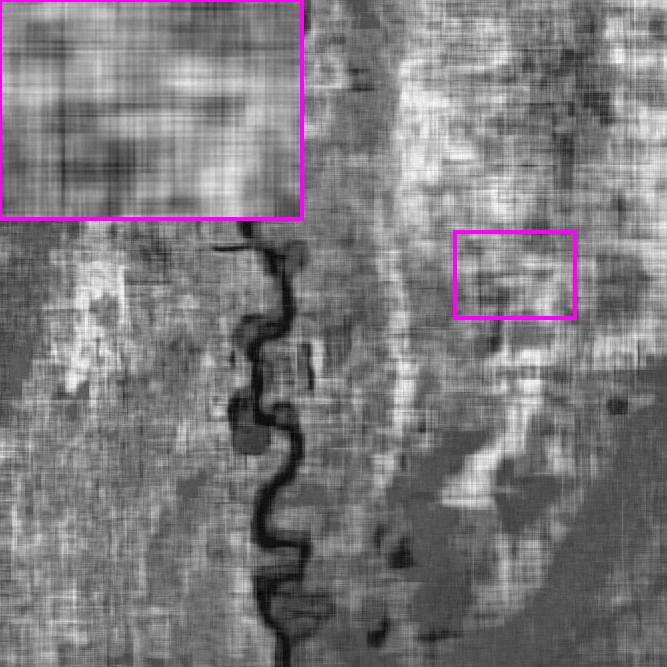}&
\includegraphics[width=0.585in, height=0.585in]{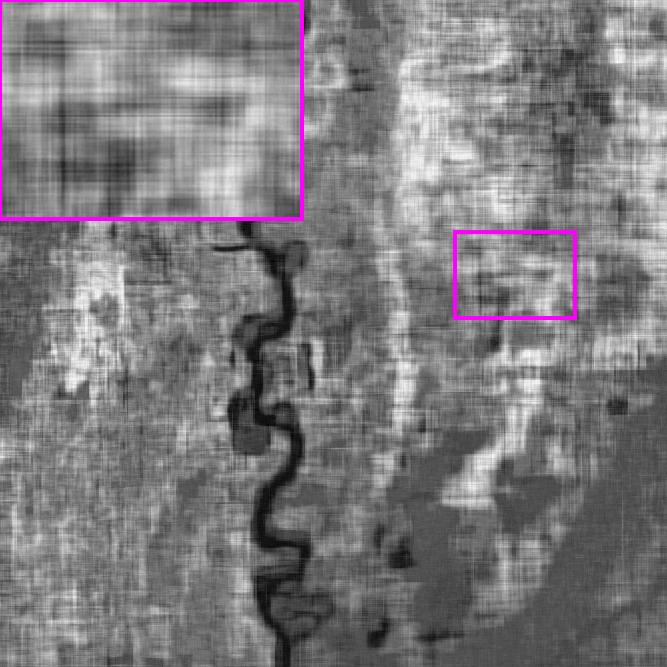}
&
\includegraphics[width=0.585in, height=0.585in,angle=0]{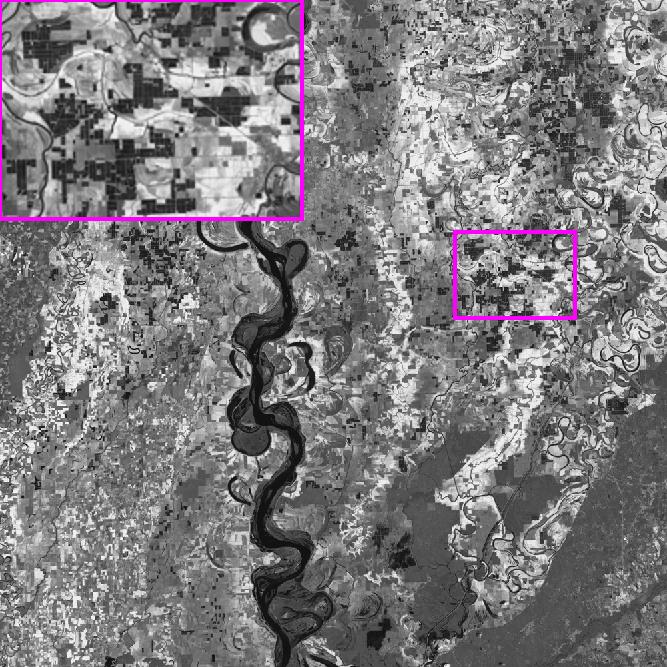}
\\

\includegraphics[width=0.585in, height=0.585in,angle=0]{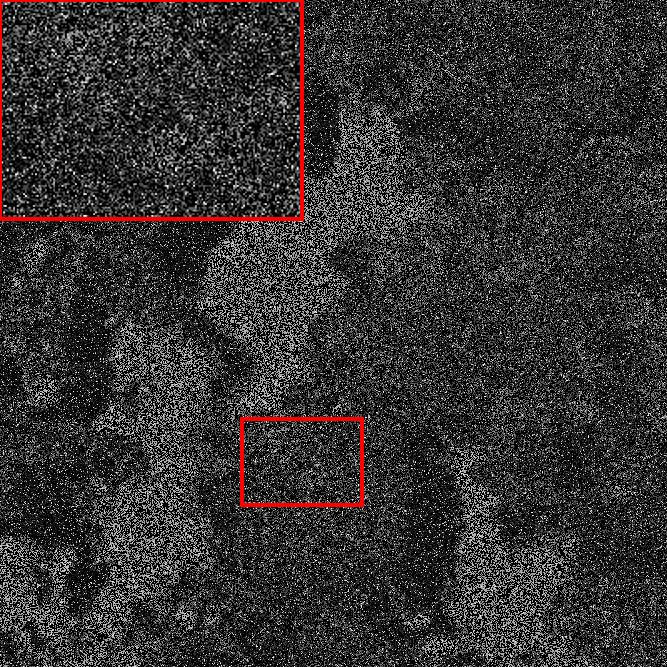}&
\includegraphics[width=0.585in, height=0.585in]{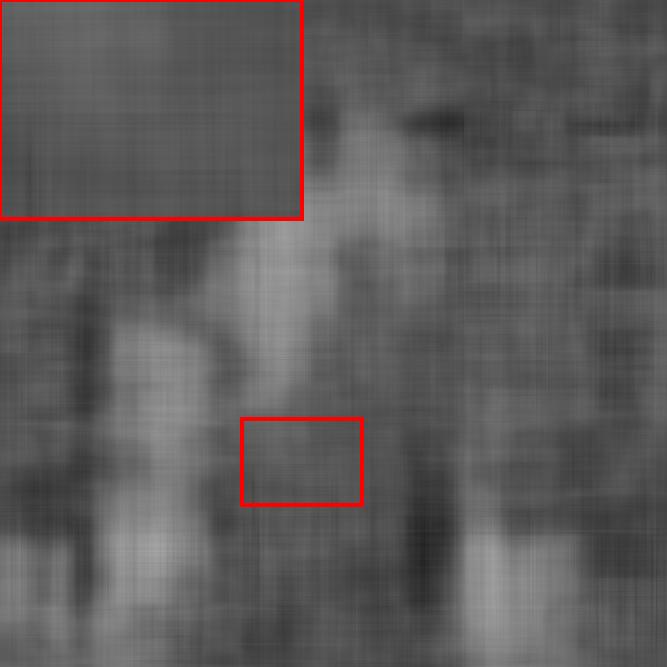}&
\includegraphics[width=0.585in, height=0.585in]{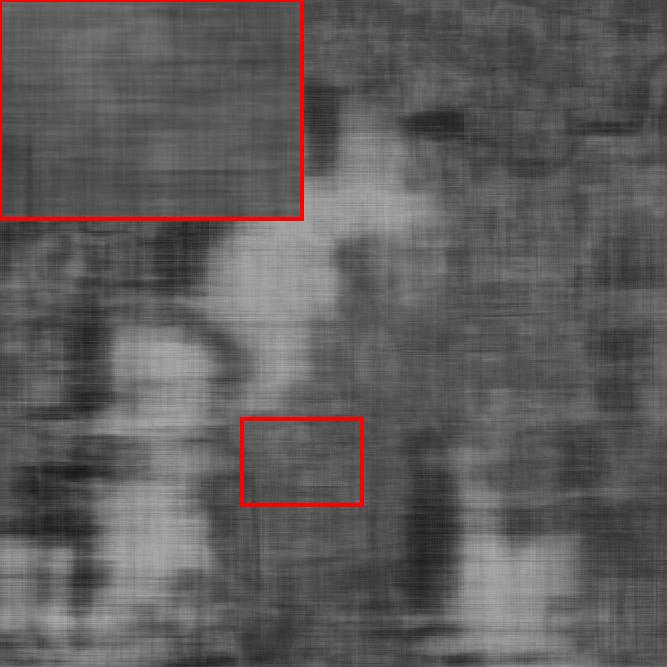}&
\includegraphics[width=0.585in, height=0.585in]{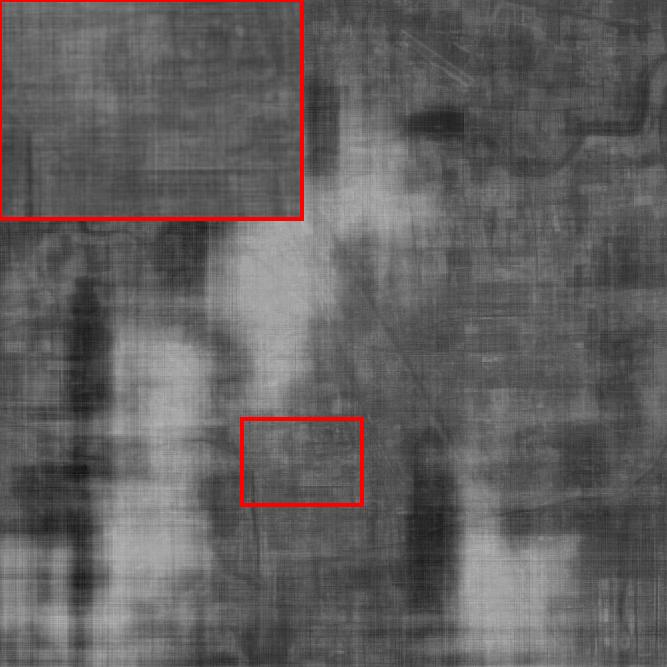}&
\includegraphics[width=0.585in, height=0.585in]{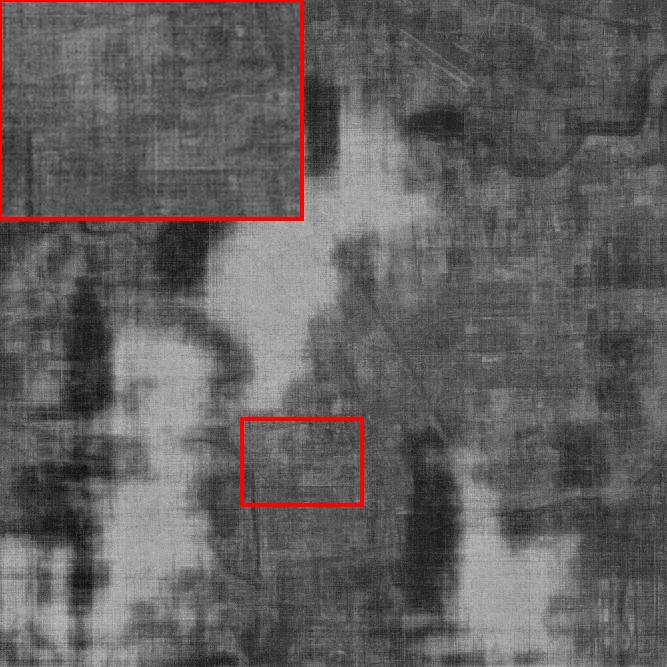}&
\includegraphics[width=0.585in, height=0.585in]{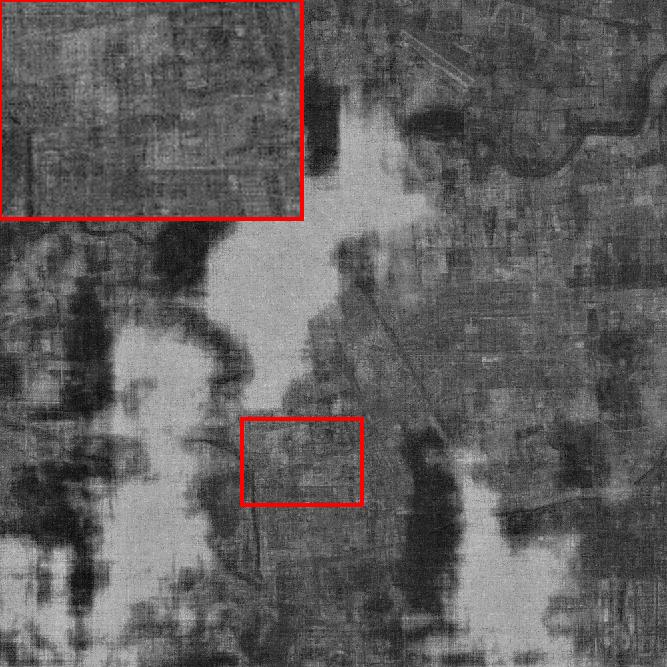}&
\includegraphics[width=0.585in, height=0.585in]{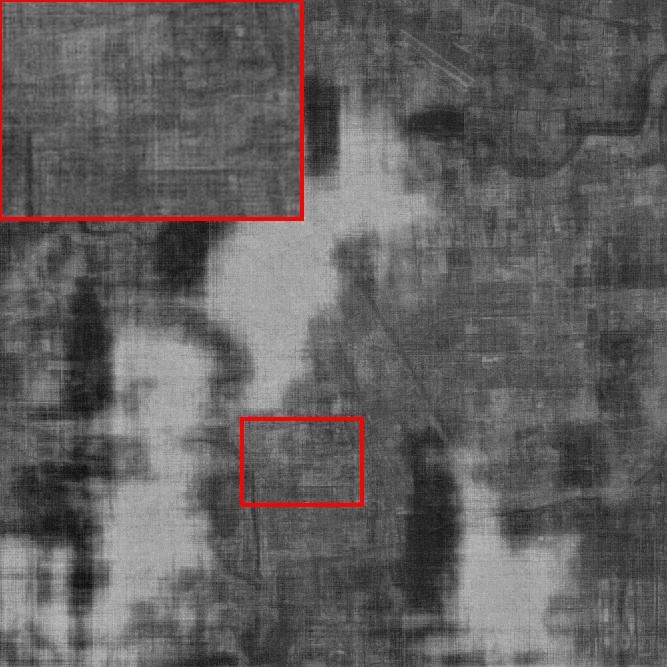}&
\includegraphics[width=0.585in, height=0.585in]{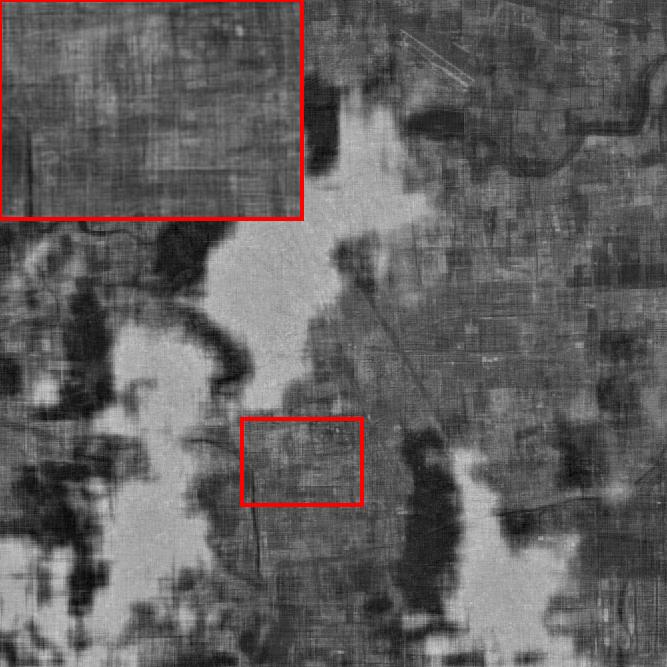}&
\includegraphics[width=0.585in, height=0.585in]{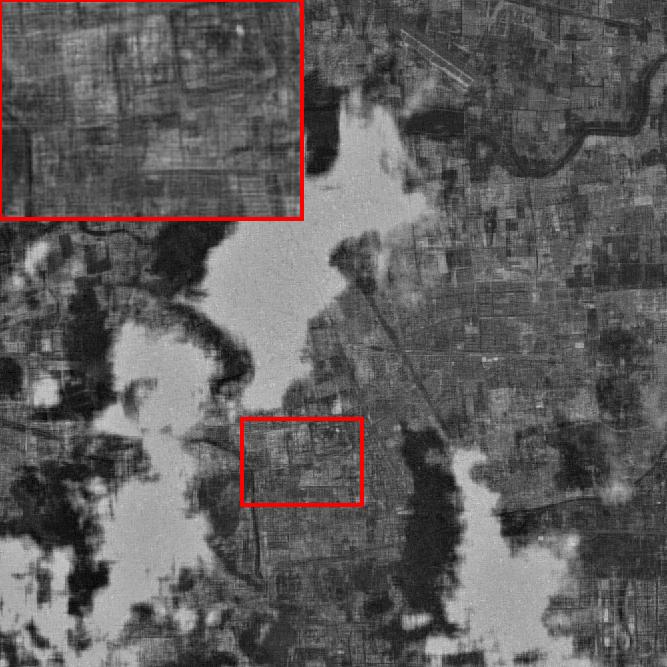}&
\includegraphics[width=0.585in, height=0.585in]{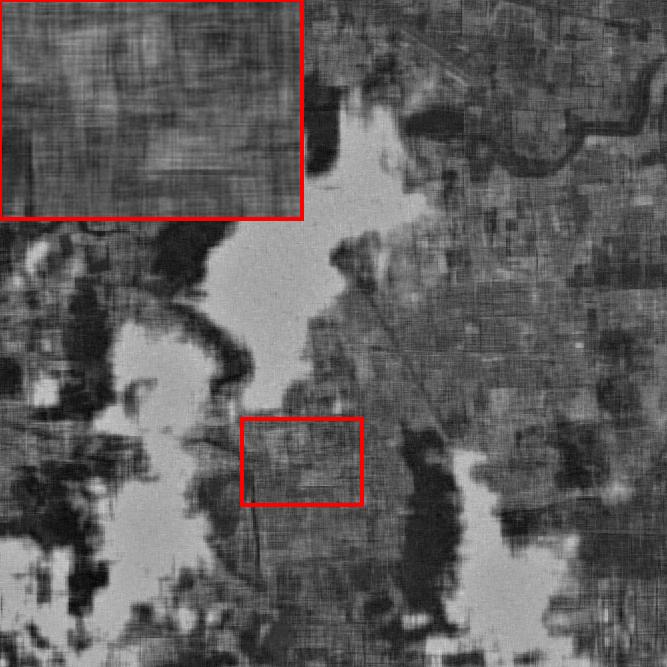}&
\includegraphics[width=0.585in, height=0.585in]{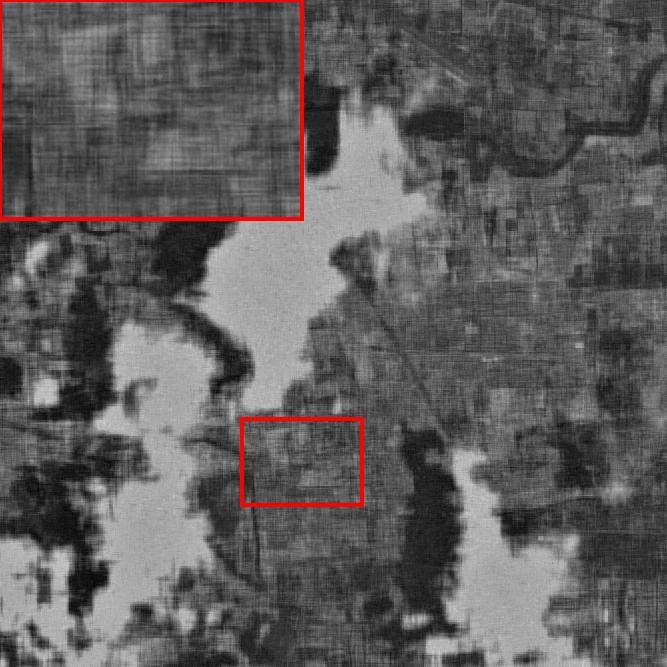}
&
\includegraphics[width=0.585in, height=0.585in,angle=0]{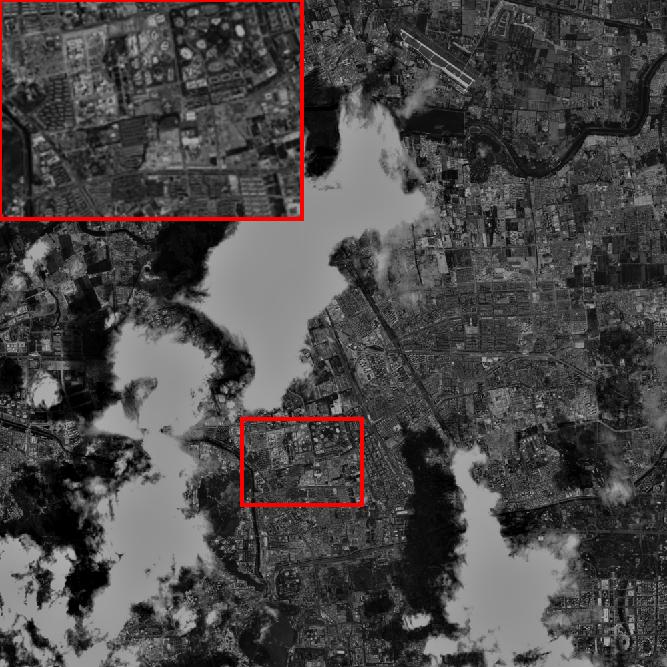}
\\

\includegraphics[width=0.585in, height=0.585in,angle=0]{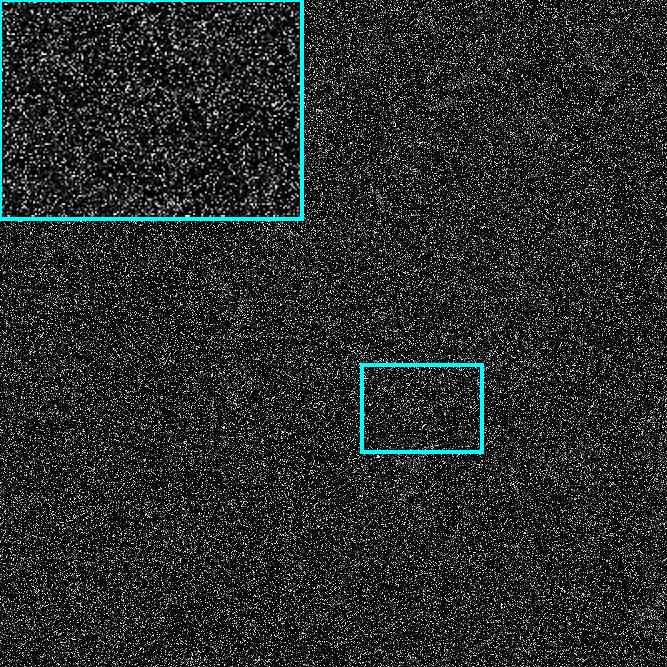}&
\includegraphics[width=0.585in, height=0.585in]{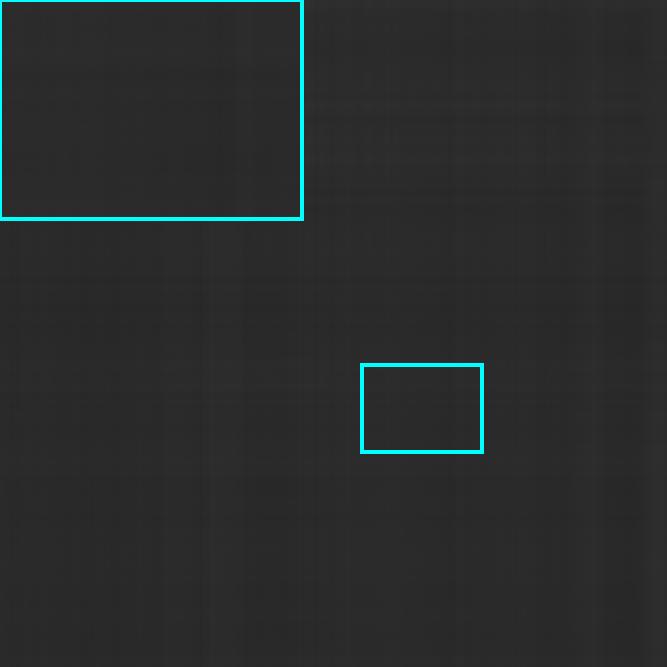}&
\includegraphics[width=0.585in, height=0.585in]{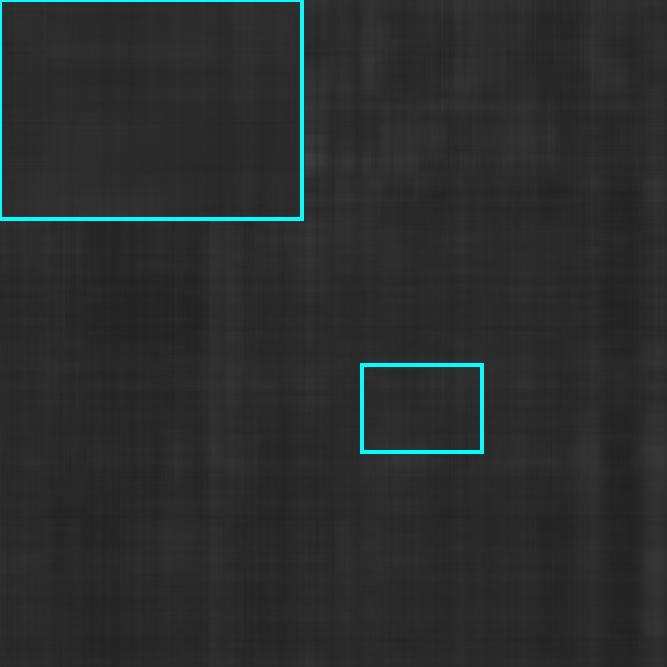}&
\includegraphics[width=0.585in, height=0.585in]{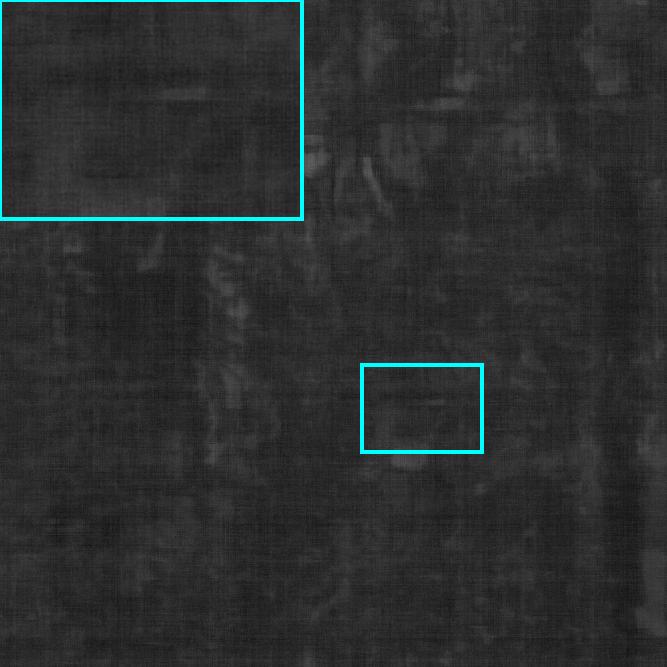}&
\includegraphics[width=0.585in, height=0.585in]{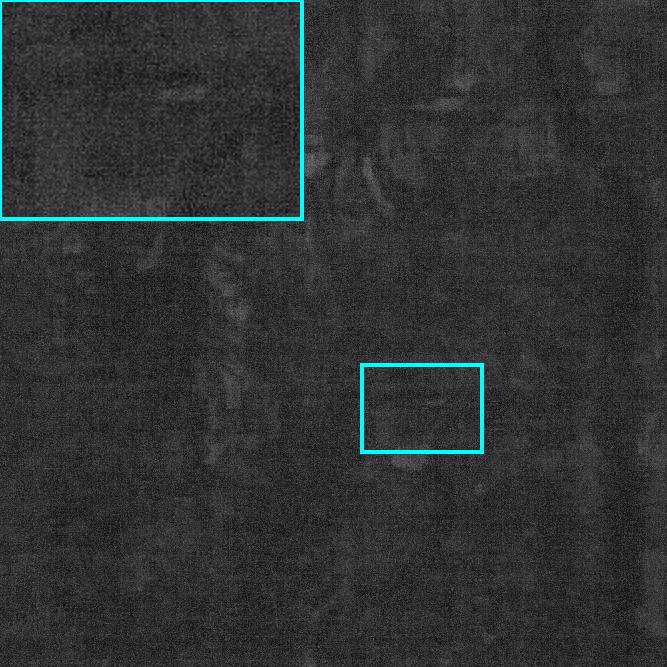}&
\includegraphics[width=0.585in, height=0.585in]{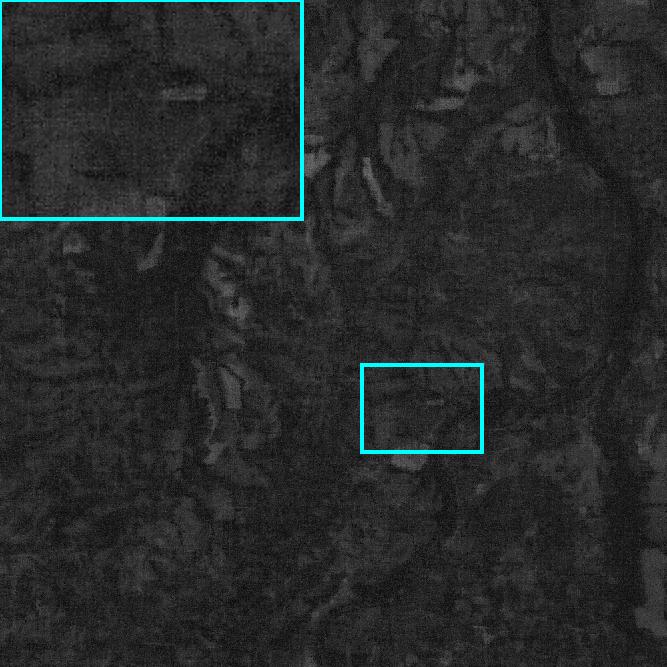}&
\includegraphics[width=0.585in, height=0.585in]{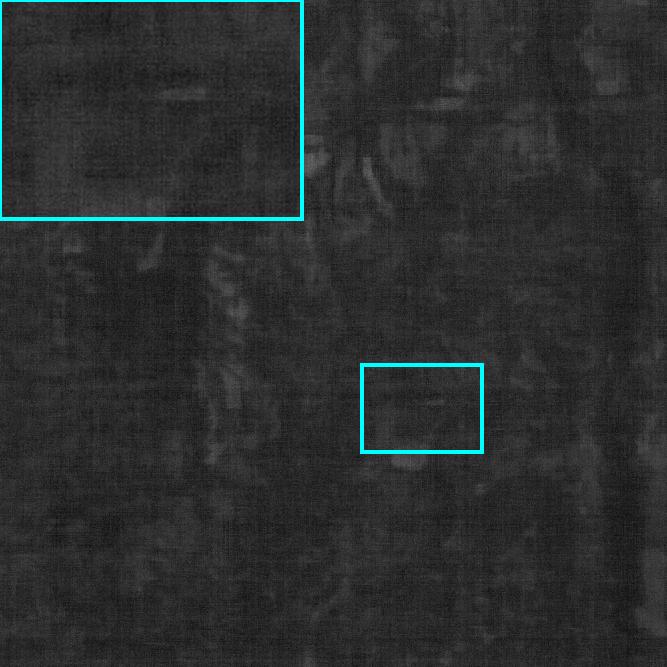}&
\includegraphics[width=0.585in, height=0.585in]{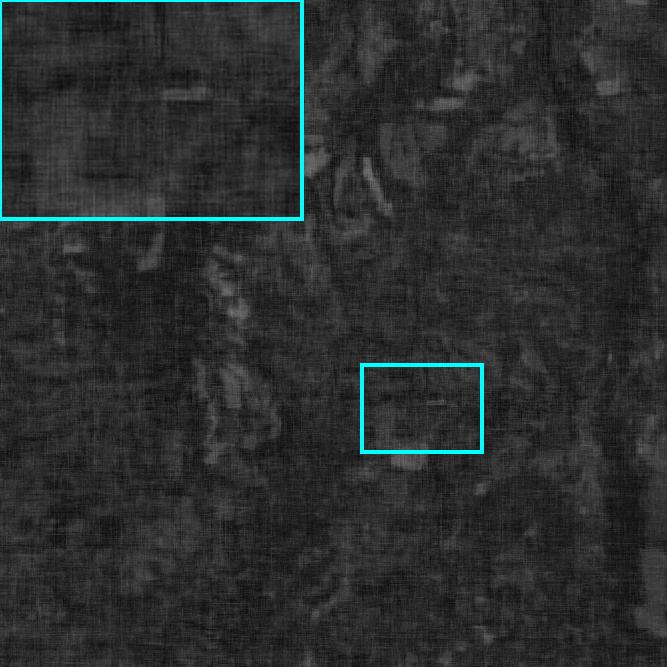}&
\includegraphics[width=0.585in, height=0.585in]{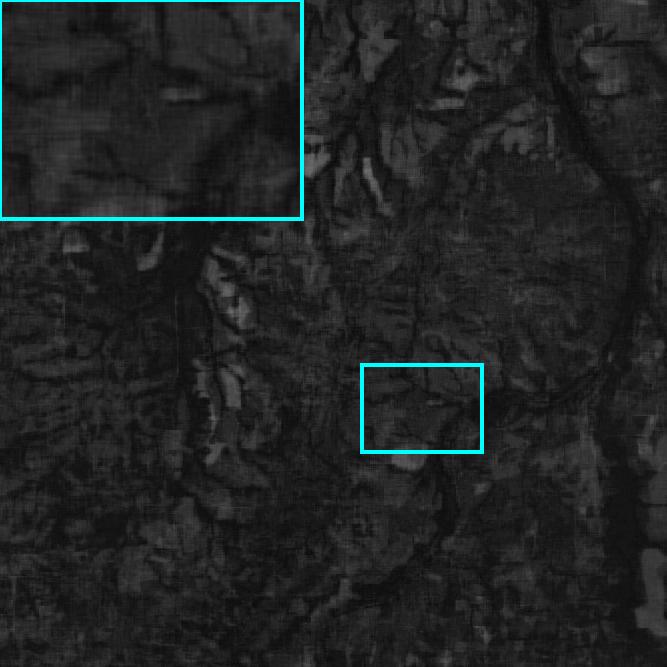}&
\includegraphics[width=0.585in, height=0.585in]{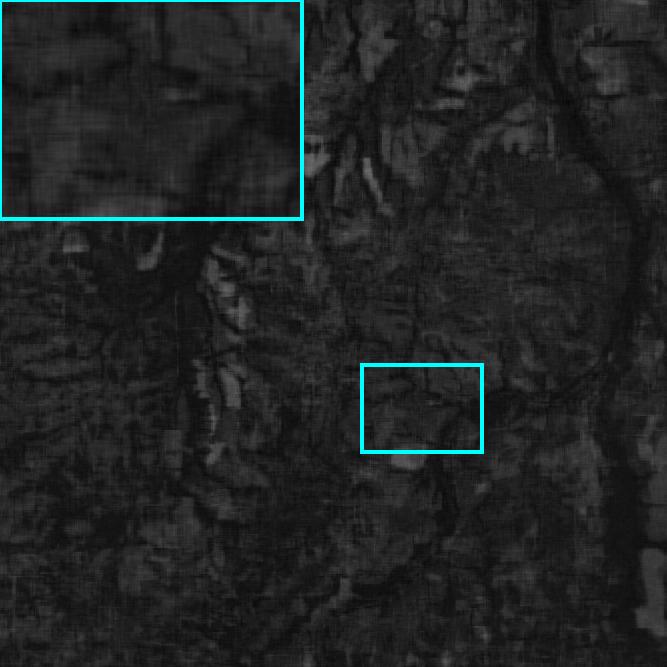}&
\includegraphics[width=0.585in, height=0.585in]{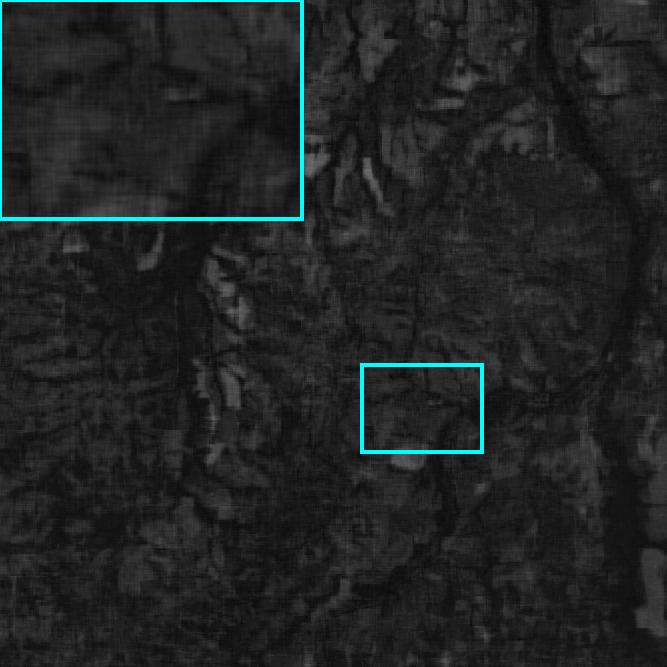}
&
\includegraphics[width=0.585in, height=0.585in,angle=0]{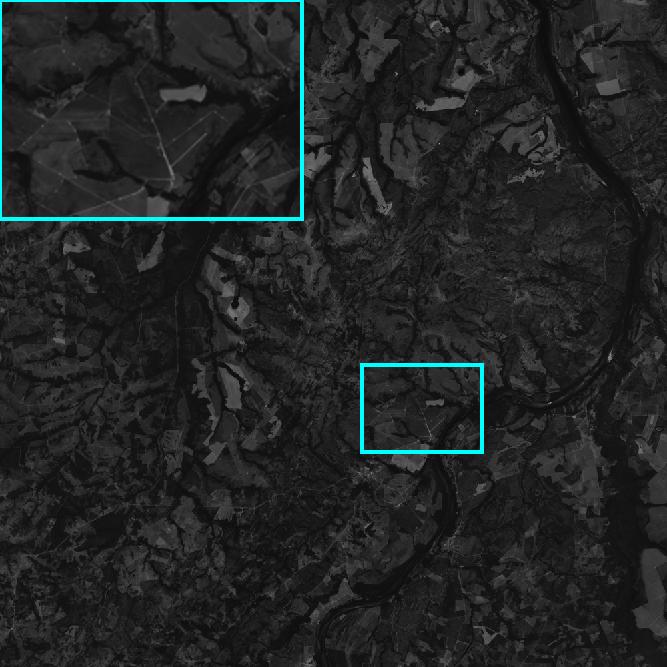}
\\

\textbf{\scriptsize{{Observed}}}  &
  \scriptsize{SNN+$\ell_1$}  &\scriptsize {TRNN+$\ell_1$}
&\scriptsize{TTNN+$\ell_1$} &\scriptsize {TSP-$k$+$\ell_1$}
 &\scriptsize{LNOP}&\scriptsize {NRTRM}&\scriptsize {HWTNN+$\ell_1$}
 & \textbf{\scriptsize {HWTSN+w$\ell_q$}}
  &

 \textbf{\tabincell{c}{ \scriptsize {HWTSN+w$\ell_q$} \\  \scriptsize{(BR)}
     }}

   &
   \textbf{\tabincell{c}{ \scriptsize {HWTSN+w$\ell_q$} \\  \scriptsize{(UR)}
     }}&
     \textbf{\scriptsize{Ground truth}}

%
\end{tabular}
\caption{
Visual comparison of various methods for 
MRSIs inpainting.
From top to bottom, the parameter pair $(sr, \tau)$  are 
$(0.4,0.1)$, $(0.4,0.3)$,  and $(0.4,0.5)$, respectively.
Top row: 
 the  $(5,1)$-th frame of  Landsat-7.
Middle row: the  $(2,6)$-th frame of SPOT-5.
Bottom row: the  $(3,5)$-th frame of T22LGN.
}
\vspace{-0.2cm}
\label{fig_visual_rs}
\end{figure*}

%
 \begin{table*}[htbp]
  \caption{
  The PSNR, SSIM values and CPU time
  obtained by  various RLRTC methods for different fourth-order MRSIs. 
   The best  and the second-best results are highlighted in
blue and red, respectively.
  }
  \label{mrsi_index}

  \centering
\scriptsize
\renewcommand{\arraystretch}{0.88}
\setlength\tabcolsep{3.0pt}

\begin{tabular}{c cccccc cccccc cccccc  |c}

     \hline
     \multicolumn{1}{c} {MRSI-Name}
     &  \multicolumn{6}{|c|} {Landsat-7} &\multicolumn{6}{c|}{T22LGN} &\multicolumn{6}{c|} {SPOT-5}  
      & \multirow{3}{*}{
   \tabincell{c}{Average\\Time (s)}
   }\\
%
     \cline{1-1}
     \cline{2-7}
     \cline{8-13}
      \cline{14-19}
     $sr$
     &\multicolumn{3}{|c|}{$20\%$}   &\multicolumn{3}{c|}{$40\%$}  &
     \multicolumn{3}{c|}{$20\%$}&\multicolumn{3}{c|}{$40\%$}&
     \multicolumn{3}{c|}{$20\%$}&\multicolumn{3}{c|}{$40\%$}\\
     \cline{1-1}
    \cline{2-4}
     \cline{5-7}
      \cline{8-10}
      \cline{11-13}
      \cline{14-16}
      \cline{17-19}
    $\tau$
    &
    \multicolumn{1}{|c|}{$10\%$}&\multicolumn{1}{c|}{$30\%$}  &\multicolumn{1}{c|}{$50\%$}&
    \multicolumn{1}{c|}{$10\%$}&\multicolumn{1}{c|}{$30\%$}  &\multicolumn{1}{c|}{$50\%$}&
    \multicolumn{1}{c|}{$10\%$}&\multicolumn{1}{c|}{$30\%$}  &\multicolumn{1}{c|}{$50\%$}&
    \multicolumn{1}{c|}{$10\%$}&\multicolumn{1}{c|}{$30\%$}  &\multicolumn{1}{c|}{$50\%$}&
    \multicolumn{1}{c|}{$10\%$}&\multicolumn{1}{c|}{$30\%$}  &\multicolumn{1}{c|}{$50\%$}&
   \multicolumn{1}{c|}{$10\%$}&\multicolumn{1}{c|}{$30\%$}  &\multicolumn{1}{c|}{$50\%$}
    \\
    \hline
     \hline
    SNN+$\ell_1$
     &

\tabincell{c}{21.45   \\0.537}   &    \tabincell{c}{20.86   \\0.453}&
\tabincell{c}{20.13 \\ 0.386}   &    \tabincell{c}{23.12   \\ 0.576}&
\tabincell{c}{ 22.49  \\ 0.498}   &    \tabincell{c}{ 21.44\\     0.477}&

\tabincell{c}{  25.94   \\0.637}   &    \tabincell{c}{25.91   \\     0.605}&
\tabincell{c}{ 25.69\\    0.592}   &    \tabincell{c}{27.01\\ 0.718}&
\tabincell{c}{    26.68  \\    0.666}   &    \tabincell{c}{ 25.76\\     0.596}&

\tabincell{c}{22.38   \\0.573}   &    \tabincell{c}{ 21.58  \\     0.509}&
\tabincell{c}{  20.91\\    0.395}   &    \tabincell{c}{  25.63   \\ 0.631}&
\tabincell{c}{ 24.19  \\    0.563}   &    \tabincell{c}{ 22.55\\     0.492}&23817
\\

      \hline
    TRNN+$\ell_1$ &

\tabincell{c}{23.96   \\ 0.671}   &    \tabincell{c}{22.32 \\ 0.589}&
\tabincell{c}{   20.70\\    0.565}   &    \tabincell{c}{ 24.61   \\  0.699}&
\tabincell{c}{ 22.67  \\    0.691}   &    \tabincell{c}{ 21.30\\     0.614}&

\tabincell{c}{ 28.57
   \\0.716
}   &    \tabincell{c}{27.22 \\ 0.696}&
\tabincell{c}{ 25.02  \\   0.586}   &    \tabincell{c}{ 29.15 \\ 0.797}&
\tabincell{c}{   27.52  \\   0.723}   &    \tabincell{c}{ 25.89\\     0.666}&

\tabincell{c}{ 27.06  \\ 0.681}   &    \tabincell{c}{24.55   \\ 0.568}&
\tabincell{c}{ 22.16\\    0.467}   &    \tabincell{c}{28.19   \\ 0.716}&
\tabincell{c}{ 25.07  \\    0.584}   &    \tabincell{c}{ 23.11\\     0.523}&  22414 
\\

 \hline
   TTNN+$\ell_1$
   &

\tabincell{c}{ 23.17       \\ 0.652}   &    \tabincell{c}{21.89       \\       0.543}&
\tabincell{c}{ 20.19\\     0.458}   &    \tabincell{c}{  24.05 \\ 0.739}&
\tabincell{c}{        22.61   \\      0.691}   &    \tabincell{c}{    21.15\\        0.575}&

\tabincell{c}{29.88  \\ 0.793}   &    \tabincell{c}{      28.34   \\  0.778}&
\tabincell{c}{     24.58\\      0.497}   &    \tabincell{c}{ 31.26        \\  0.817}&
\tabincell{c}{  29.37       \\     0.797}   &    \tabincell{c}{26.29\\       0.692}&

\tabincell{c}{26.49    \\0.684}   &    \tabincell{c}{    24.38 \\0.537}&
\tabincell{c}{       22.22\\      0.431}   &    \tabincell{c}{ 28.17       \\ 0.701}&
\tabincell{c}{ 25.65       \\      0.596}   &    \tabincell{c}{23.49\\       0.519}&10107
\\

 \hline
   TSP-$k$+$\ell_1$
   &

%
%
%

   \tabincell{c}{ 24.07
             \\ 0.673}   &    \tabincell{c}{22.41     \\0.588}&
\tabincell{c}{   20.36\\     0.461}   &    \tabincell{c}{25.82 \\  0.742}&
\tabincell{c}{ 23.62      \\     0.628}   &    \tabincell{c}{ 21.14\\      0.472}&

\tabincell{c}{ 29.12      \\0.791}   &    \tabincell{c}{ 25.22      \\       0.511}&
\tabincell{c}{   22.54\\     0.409}   &    \tabincell{c}{31.14     \\  0.836}&
\tabincell{c}{   26.03      \\    0.545}   &    \tabincell{c}{ 22.59\\        0.419}&

\tabincell{c}{ 27.07
       \\0.657}   &    \tabincell{c}{24.57       \\ 0.571}&
\tabincell{c}{ 22.29\\     0.413}   &    \tabincell{c}{29.74  \\ 0.775}&
\tabincell{c}{      26.36      \\      0.601}   &    \tabincell{c}{ 23.34\\       0.506}&27412
\\

     \hline
    LNOP
     &

   \tabincell{c}{24.61   \\0.679}   &    \tabincell{c}{22.72\\     0.658}&
\tabincell{c}{20.83 \\   0.589}   &    \tabincell{c}{26.27\\  0.763}&
\tabincell{c}{23.35   \\   0.688}   &    \tabincell{c}{21.42\\     0.592}&

\tabincell{c}{30.98\\0.841}   &    \tabincell{c}{28.61\\ 0.736}&
\tabincell{c}{26.43 \\    0.703}   &    \tabincell{c}{33.63\\ 0.919}&
\tabincell{c}{29.12   \\   0.825}   &    \tabincell{c}{26.89 \\     0.706}&

\tabincell{c}{27.66   \\0.643}   &    \tabincell{c}{25.55   \\ 0.593}&
\tabincell{c}{23.45\\0.484}   &    \tabincell{c}{29.88\\ 0.781}&
\tabincell{c}{26.63  \\ 0.632}   &    \tabincell{c}{24.66\\ 0.572}&18418
\\

\hline

     NRTRM
     &

 \tabincell{c}{ 23.76
       \\0.669}   &    \tabincell{c}{22.40\\ 0.626}&
\tabincell{c}{20.45 \\     0.546}   &    \tabincell{c}{25.72\\ 0.757}&
\tabincell{c}{23.84 \\      0.639}   &    \tabincell{c}{21.33\\        0.577}&

\tabincell{c}{
       30.34
        \\0.833}   &    \tabincell{c}{   28.67 \\  0.782}&
\tabincell{c}{       25.68\\      0.607}   &    \tabincell{c}{33.12 \\ {0.893}}&
\tabincell{c}{       30.41        \\     {0.844}}   &    \tabincell{c}{25.87\\       0.621}&

\tabincell{c}{    27.08
       \\{0.682}}   &    \tabincell{c}{24.84       \\      0.589}&
\tabincell{c}{ 22.55\\       0.432}   &    \tabincell{c}{29.95 \\  \textcolor[rgb]{0.00,0.00,0.00}{0.799}}&
\tabincell{c}{      26.84  \\      0.645}   &    \tabincell{c}{      23.73\\       0.569}&18560
\\

     \hline
HWTNN+$\ell_1$
     &

  \tabincell{c}{\textcolor[rgb]{1.00,0.00,0.00}{25.18}
       \\\textcolor[rgb]{1.00,0.00,0.00}{0.692}}   &    \tabincell{c}{23.48      \\      0.659}&
\tabincell{c}{  21.59\\     0.630}   &    \tabincell{c}{\textcolor[rgb]{1.00,0.00,0.00}{ 26.33}       \\ \textcolor[rgb]{1.00,0.00,0.00}{0.768}}&
\tabincell{c}{ 24.98       \\     0.693}   &    \tabincell{c}{22.89\\       0.665}&

\tabincell{c}{  \textcolor[rgb]{1.00,0.00,0.00}{32.09}
       \\ \textcolor[rgb]{1.00,0.00,0.00}{0.854}}   &    \tabincell{c}{ 29.19       \\  0.758}&
\tabincell{c}{ 25.53\\      0.639}   &    \tabincell{c}{\textcolor[rgb]{1.00,0.00,0.00}{34.24}      \\\textcolor[rgb]{1.00,0.00,0.00}{0.923}}&
\tabincell{c}{   \textcolor[rgb]{1.00,0.00,0.00}{31.46}      \\      \textcolor[rgb]{1.00,0.00,0.00}{0.852}}   &    \tabincell{c}{ 26.46\\       0.706}&

\tabincell{c}{  \textcolor[rgb]{1.00,0.00,0.00}{28.45}
       \\ \textcolor[rgb]{1.00,0.00,0.00}{0.687}}   &    \tabincell{c}{ {26.24}   \\  0.596}&
\tabincell{c}{     23.61\\      0.490}   &    \tabincell{c}{ \textcolor[rgb]{1.00,0.00,0.00}{30.47}       \\ \textcolor[rgb]{1.00,0.00,0.00}{0.808}}&
\tabincell{c}{ \textcolor[rgb]{1.00,0.00,0.00}{28.36}       \\      \textcolor[rgb]{1.00,0.00,0.00}{0.699}}   &    \tabincell{c}{ 25.49\\       0.581}&14835
\\

\hline
   %
   \tabincell{c}{               \textbf{HWTSN+w$\ell_q$}
    }
     &

%
%

  \tabincell{c}{ \textcolor[rgb]{0,0,1}{25.99}
   \\\textcolor[rgb]{0,0,1}{0.722}}   &    \tabincell{c}{   \textcolor[rgb]{0,0,1}{24.71}  \\\textcolor[rgb]{0,0,1}{0.683}}&
\tabincell{c}{  \textcolor[rgb]{0,0,1}{23.21}\\   \textcolor[rgb]{0,0,1}{0.669}}   &    \tabincell{c}{\textcolor[rgb]{0,0,1}{28.11}\\ \textcolor[rgb]{0,0,1}{0.777}}&
\tabincell{c}{\textcolor[rgb]{0,0,1}{25.99}   \\\textcolor[rgb]{0,0,1}{0.724}}   &    \tabincell{c}{\textcolor[rgb]{0,0,1}{23.61}\\ \textcolor[rgb]{0,0,1}{0.677}}&

\tabincell{c}{\textcolor[rgb]{0,0,1}{32.46}
    \\\textcolor[rgb]{0,0,1}{0.888}}   &    \tabincell{c}{ \textcolor[rgb]{0,0,1}{30.16}   \\ \textcolor[rgb]{0,0,1}{0.833}}&
\tabincell{c}{\textcolor[rgb]{0,0,1}{27.98} \\    \textcolor[rgb]{0,0,1}{0.746}}   &    \tabincell{c}{\textcolor[rgb]{0,0,1}{34.98}   \\ \textcolor[rgb]{0,0,1}{0.944}}&
\tabincell{c}{ \textcolor[rgb]{0,0,1}{31.92}\\   \textcolor[rgb]{0,0,1}{0.896}}   &    \tabincell{c}{\textcolor[rgb]{0,0,1}{28.64}\\    \textcolor[rgb]{0,0,1}{0.794}}&

\tabincell{c}{\textcolor[rgb]{0,0,1}{29.58}   \\ \textcolor[rgb]{0,0,1}{0.713}}   &    \tabincell{c}{\textcolor[rgb]{0,0,1}{27.79}\\ \textcolor[rgb]{0,0,1}{0.615}}&
\tabincell{c}{\textcolor[rgb]{0,0,1}{25.82} \\    \textcolor[rgb]{0,0,1}{0.565}}   &    \tabincell{c}{ \textcolor[rgb]{0,0,1}{32.47}\\  \textcolor[rgb]{0,0,1}{0.836}}&
\tabincell{c}{ \textcolor[rgb]{0,0,1}{30.07} \\    \textcolor[rgb]{0,0,1}{0.756}}   &    \tabincell{c}{\textcolor[rgb]{0,0,1}{26.77} \\     \textcolor[rgb]{0,0,1}{0.644}}&  18974
\\

  \hline
%
   \tabincell{c}{\textbf{HWTSN+w$\ell_q$(BR)}
    }

 &

%
%
%
%
%
%

\tabincell{c}{ 24.94
       \\0.682}   &    \tabincell{c}{      24.30 \\     0.662}&
\tabincell{c}{       22.88\\     0.641}   &    \tabincell{c}{25.55 \\ 0.729}&
\tabincell{c}{       24.94     \\     0.711}   &    \tabincell{c}{  23.13\\       0.674}&

\tabincell{c}{  30.54 \\0.809}   &    \tabincell{c}{   29.32 \\       0.791}&
\tabincell{c}{     27.79 \\     0.738}   &    \tabincell{c}{30.94 \\0.819}&
\tabincell{c}{       30.02 \\     0.814}   &    \tabincell{c}{      28.39\\      0.777}&

\tabincell{c}{   27.94    \\ 0.624}   &    \tabincell{c}{   27.19\\       {0.599}}&
\tabincell{c}{        25.48\\      0.551}   &    \tabincell{c}{28.72       \\ 0.695}&
\tabincell{c}{ 28.19      \\     0.675}   &    \tabincell{c}{ 26.37\\       0.614}&5412
\\

\hline

   \tabincell{c}{\textbf{HWTSN+w$\ell_q$(UR)}
    }

 &

%
%

  \tabincell{c}{25.02
    \\ 0.683}   &    \tabincell{c}{  \textcolor[rgb]{1.00,0.00,0.00}{24.37}   \\ \textcolor[rgb]{1.00,0.00,0.00}{{0.676}}}&
\tabincell{c}{ \textcolor[rgb]{1.00,0.00,0.00}{22.91}\\    \textcolor[rgb]{1.00,0.00,0.00}{0.645}}   &    \tabincell{c}{25.69\\   0.729}&
\tabincell{c}{    \textcolor[rgb]{1.00,0.00,0.00}{25.05}  \\  \textcolor[rgb]{1.00,0.00,0.00}{0.712}}   &    \tabincell{c}{ \textcolor[rgb]{1.00,0.00,0.00}{23.15}\\     \textcolor[rgb]{1.00,0.00,0.00}{0.674}}&

\tabincell{c}{ 30.65  \\0.811}   &    \tabincell{c}{ \textcolor[rgb]{1.00,0.00,0.00}{29.39} \\    \textcolor[rgb]{1.00,0.00,0.00}{0.792}}&
\tabincell{c}{\textcolor[rgb]{1.00,0.00,0.00}{27.81} \\    \textcolor[rgb]{1.00,0.00,0.00}{0.739}}   &    \tabincell{c}{ 31.06\\ 0.822}&
\tabincell{c}{    30.17  \\   0.816}   &    \tabincell{c}{ \textcolor[rgb]{1.00,0.00,0.00}{28.45}\\     \textcolor[rgb]{1.00,0.00,0.00}{0.779}}&

\tabincell{c}{27.99
 \\0.624}   &    \tabincell{c}{   \textcolor[rgb]{1.00,0.00,0.00}{27.21} \\ \textcolor[rgb]{1.00,0.00,0.00}{0.605}}&
\tabincell{c}{   \textcolor[rgb]{1.00,0.00,0.00}{25.51}\\    \textcolor[rgb]{1.00,0.00,0.00}{0.557}}   &    \tabincell{c}{ 28.82 \\ 0.694}&
\tabincell{c}{   28.26  \\    0.686}   &    \tabincell{c}{  \textcolor[rgb]{1.00,0.00,0.00}{26.40}\\      \textcolor[rgb]{1.00,0.00,0.00}{0.617}}&5746
\\

   \hline

\end{tabular}
\begin{tablenotes}
\item[**]
In each RLRTC method, the top represents the PSNR values while  the bottom denotes the SSIM values.
\end{tablenotes}
\vspace{-0.47cm}
\end{table*}

Table \ref{mrsi_index}  presents
the PSNR, SSIM values and  CPU time provided  by  ten RLRTC methods on three  large-scale MRSIs with
 $sr \in \{0.4,0.2\}$, $\tau \in \{0.1, 0.3,0.5\}$.
 Accordingly,
 some visual examples are illustrated in Figure \ref{fig_visual_rs}, which indicates that
 the proposed method retains more details and textures over the other state-of-the-art approaches.
 Strikingly,
 in comparison with the  deterministic RLRTC algorithms,
 our  RHTC algorithms  incorporating with randomized technology
 can decrease the computational time by about $70\%$
 with little or no loss of PSNR and SSIM. 
 This demonstrates the effectiveness of randomized  
 approach for processing  large-scale  tensor data.
Other
conclusions achieved from the quantitative results are similar to those obtained from the  tasks of  CVs restoration and
 LFIs recovery.
%
%
%
%
Overall,   the  proposed randomized  RHTC method     can dramatically  shorten the CPU running time 
 while still achieving reasonable recovery 
 precision
 over other popular approaches, 
  especially for large-scale inpainting tasks. 

\begin{figure}[!htbp]
\centering
\renewcommand{\arraystretch}{0.0}
\setlength\tabcolsep{0.0pt}
\begin{tabular}
{ccc }
\centering
\includegraphics[width=1.15in, height=0.9in]{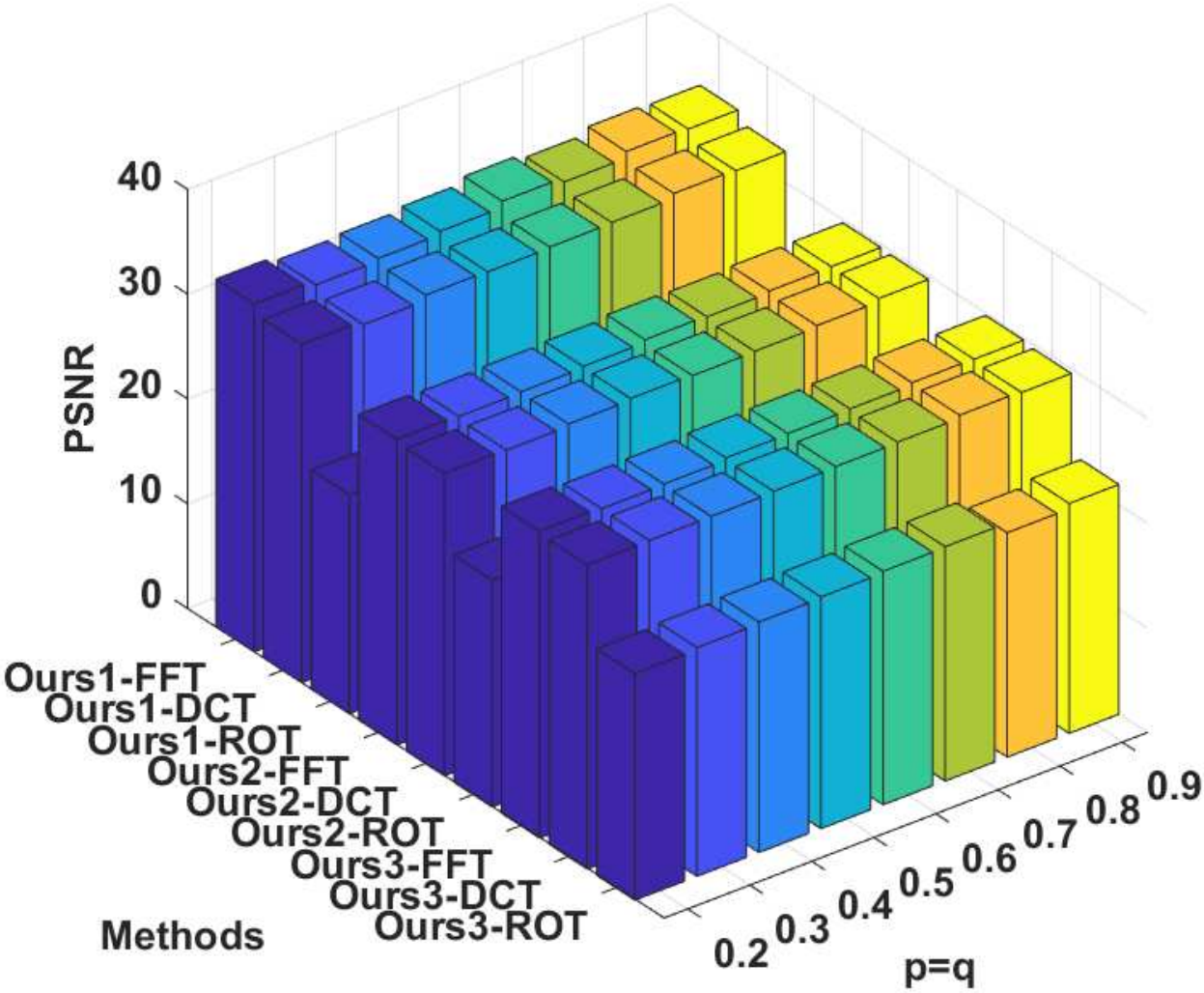} & 
\includegraphics[width=1.15in, height=0.9in]{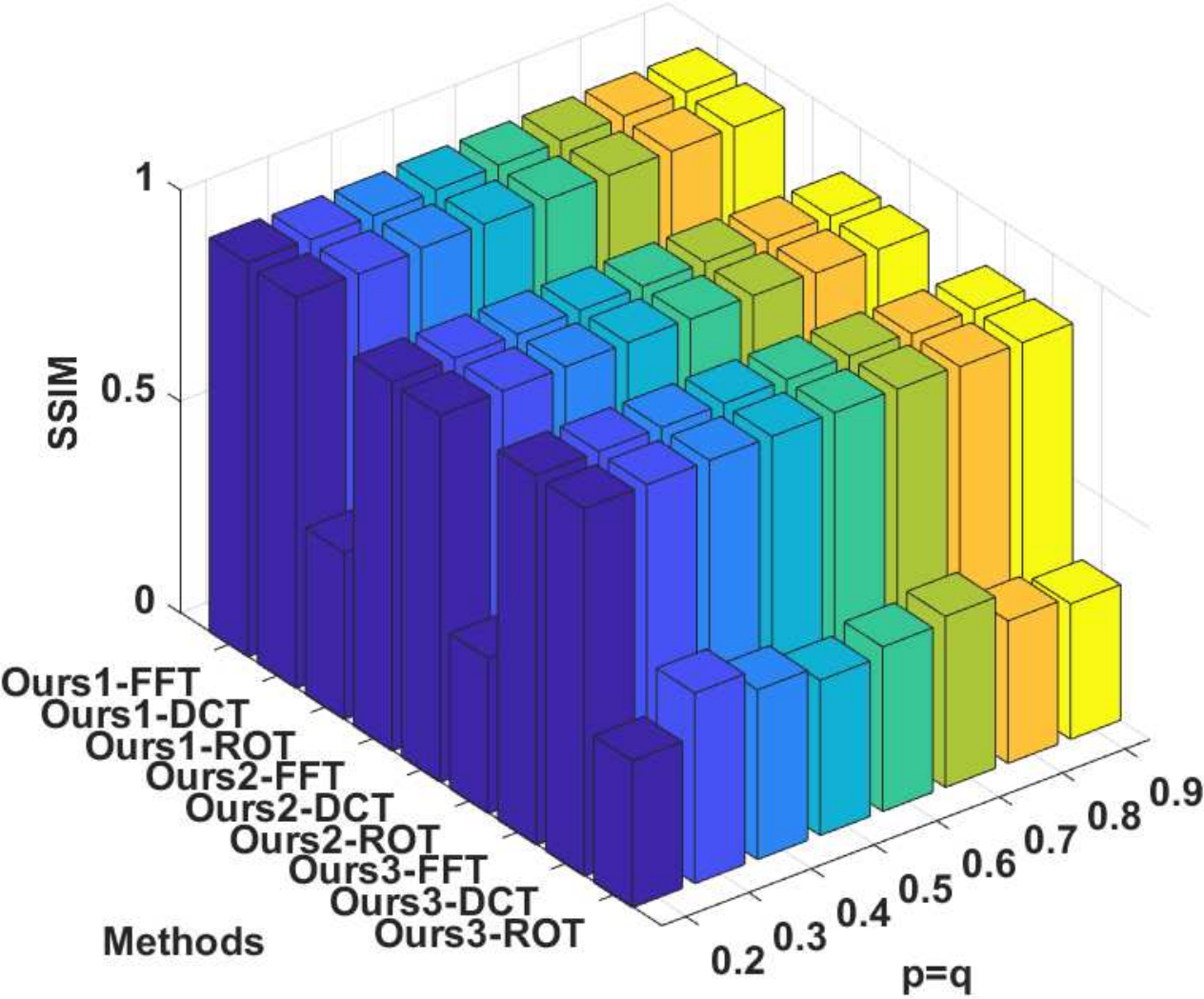}& 
\includegraphics[width=1.15in, height=0.9in]{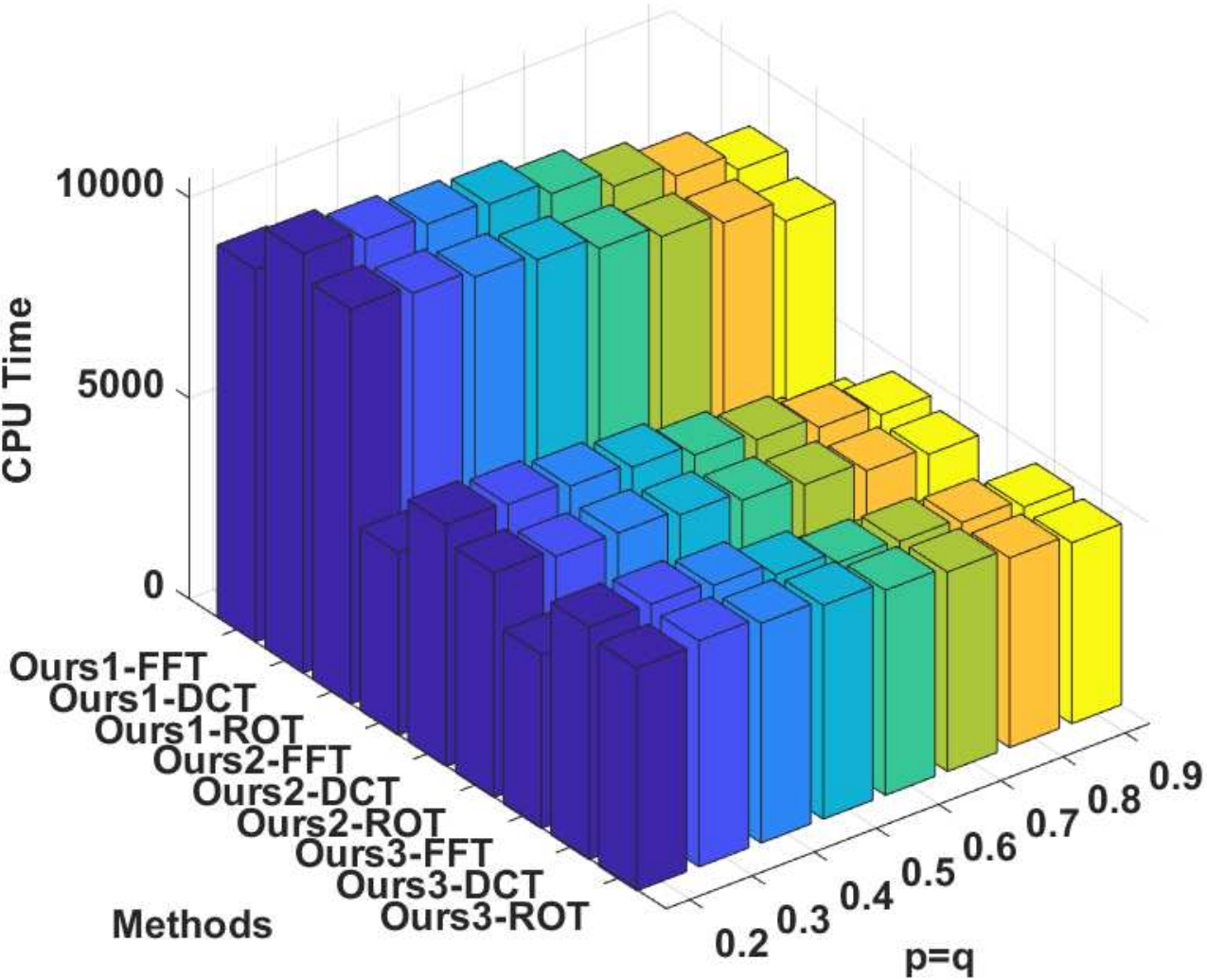}\\
\includegraphics[width=1.15in, height=0.9in]{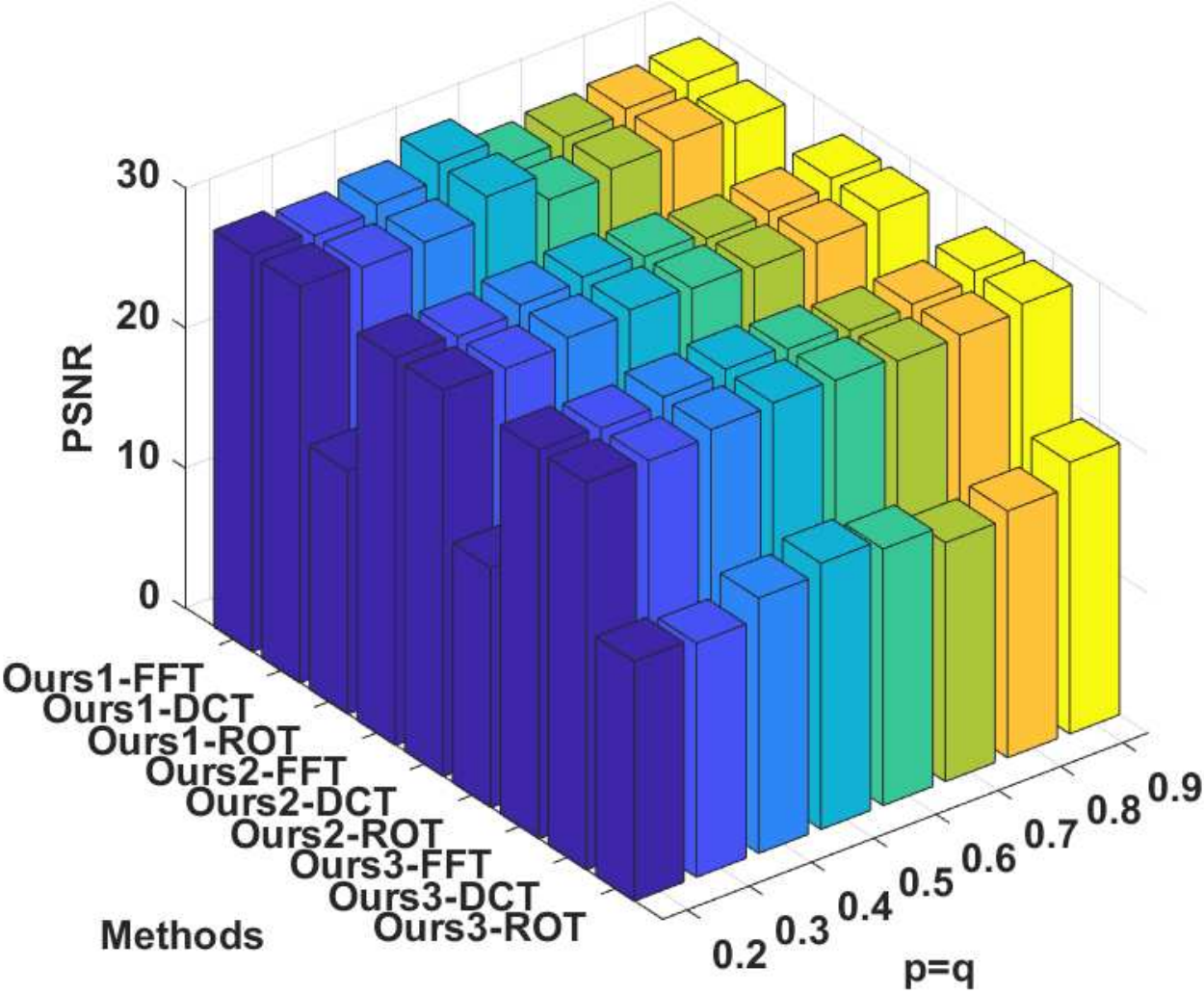} & 
\includegraphics[width=1.15in, height=0.9in]{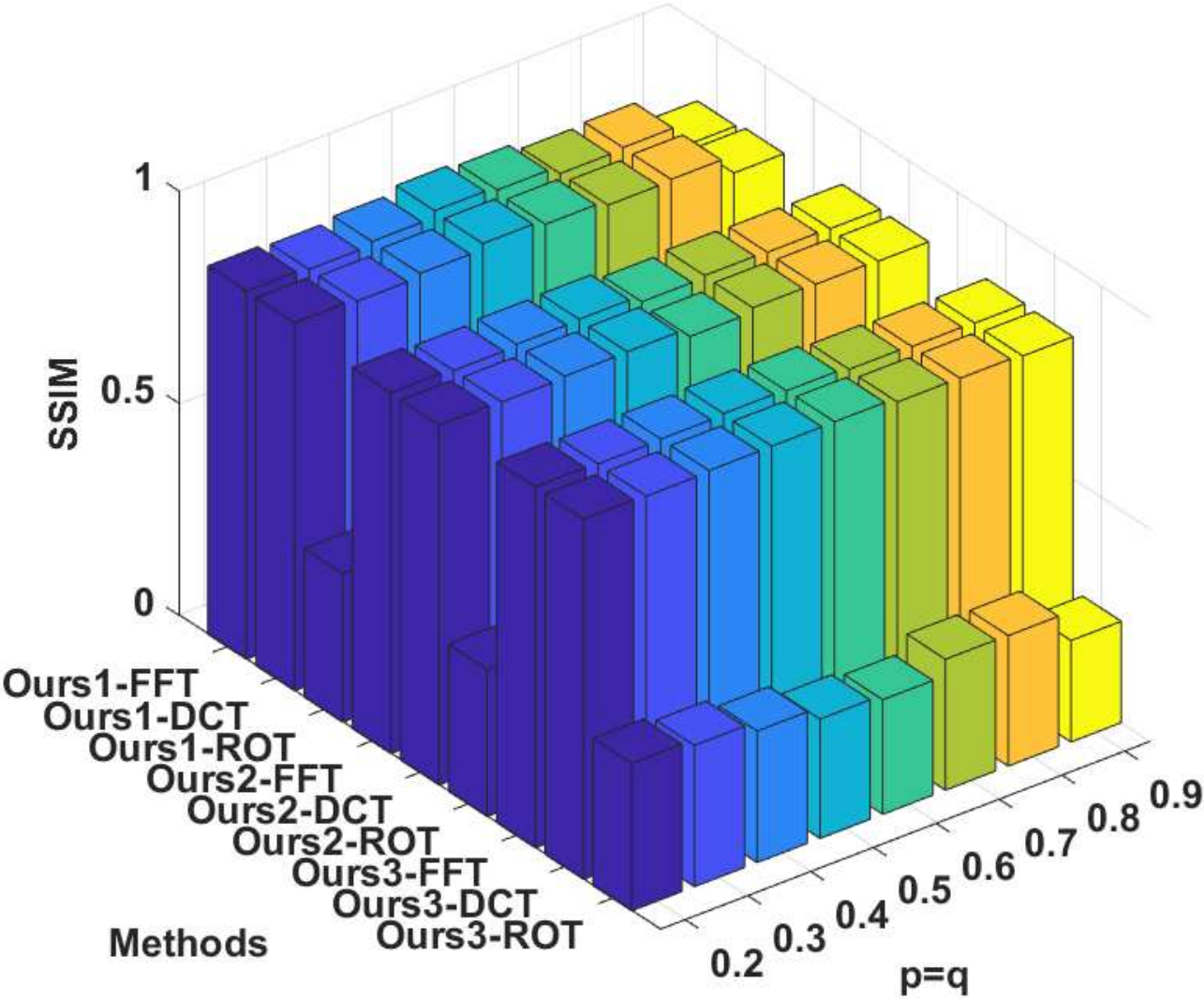}& 
\includegraphics[width=1.15in, height=0.9in]{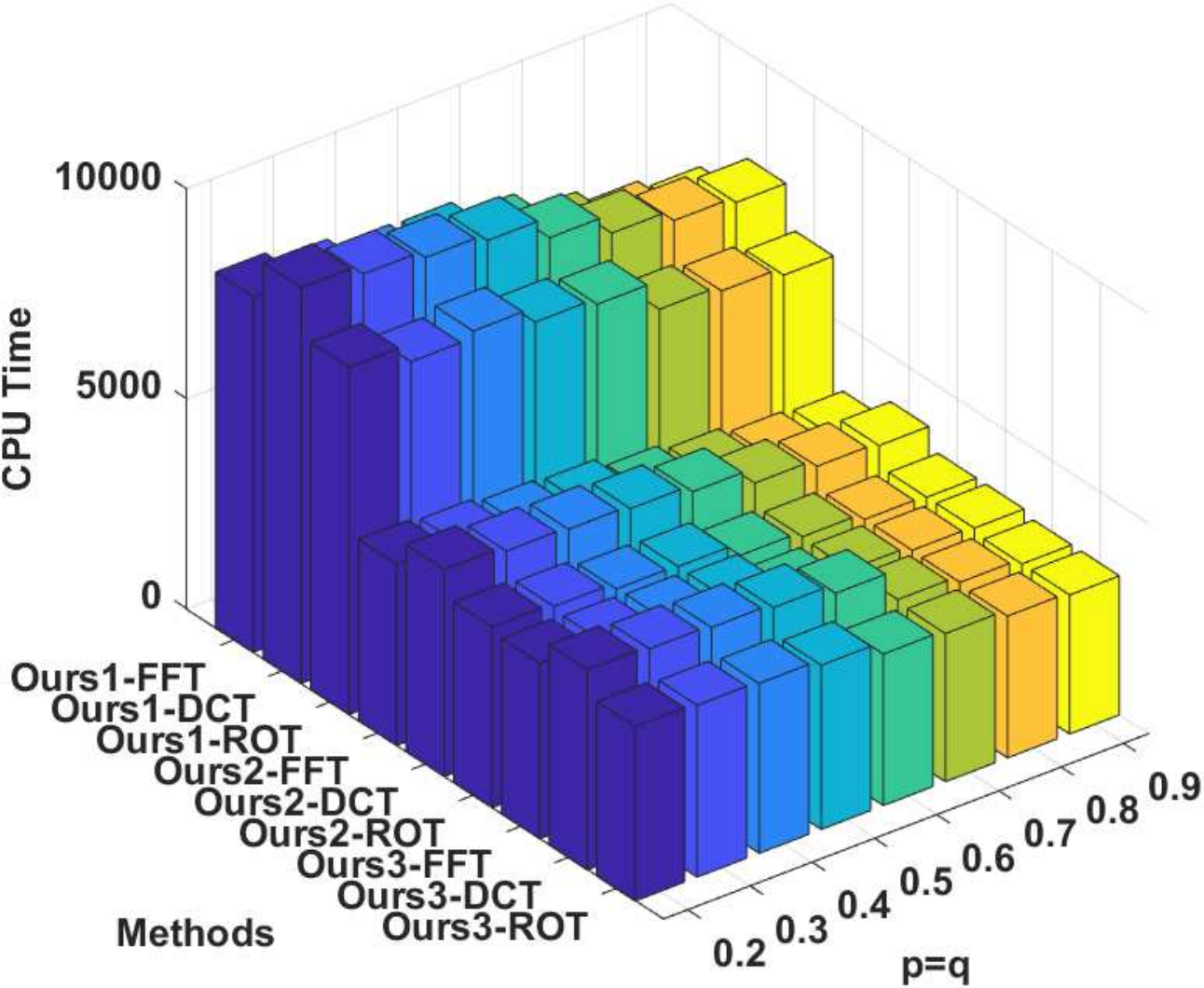} 
\end{tabular}
\caption{
The influence of adjustable  parameters $(p,q)$, 
and
 changing 
linear transforms $\mathfrak{L}$
upon  LFIs recovery.
%
Top row: $sr=0.1, \tau=0.3$,
Bottom row: $sr=0.1, \tau=0.5$.
}
\label{disscu1-lfi}
\vspace{-0.3cm}
\end{figure}

\begin{figure}[!htbp]
\centering
\renewcommand{\arraystretch}{0.0}
\setlength\tabcolsep{0.0pt}
\begin{tabular}
{ccc}
\centering
\includegraphics[width=1.15in, height=0.9in]{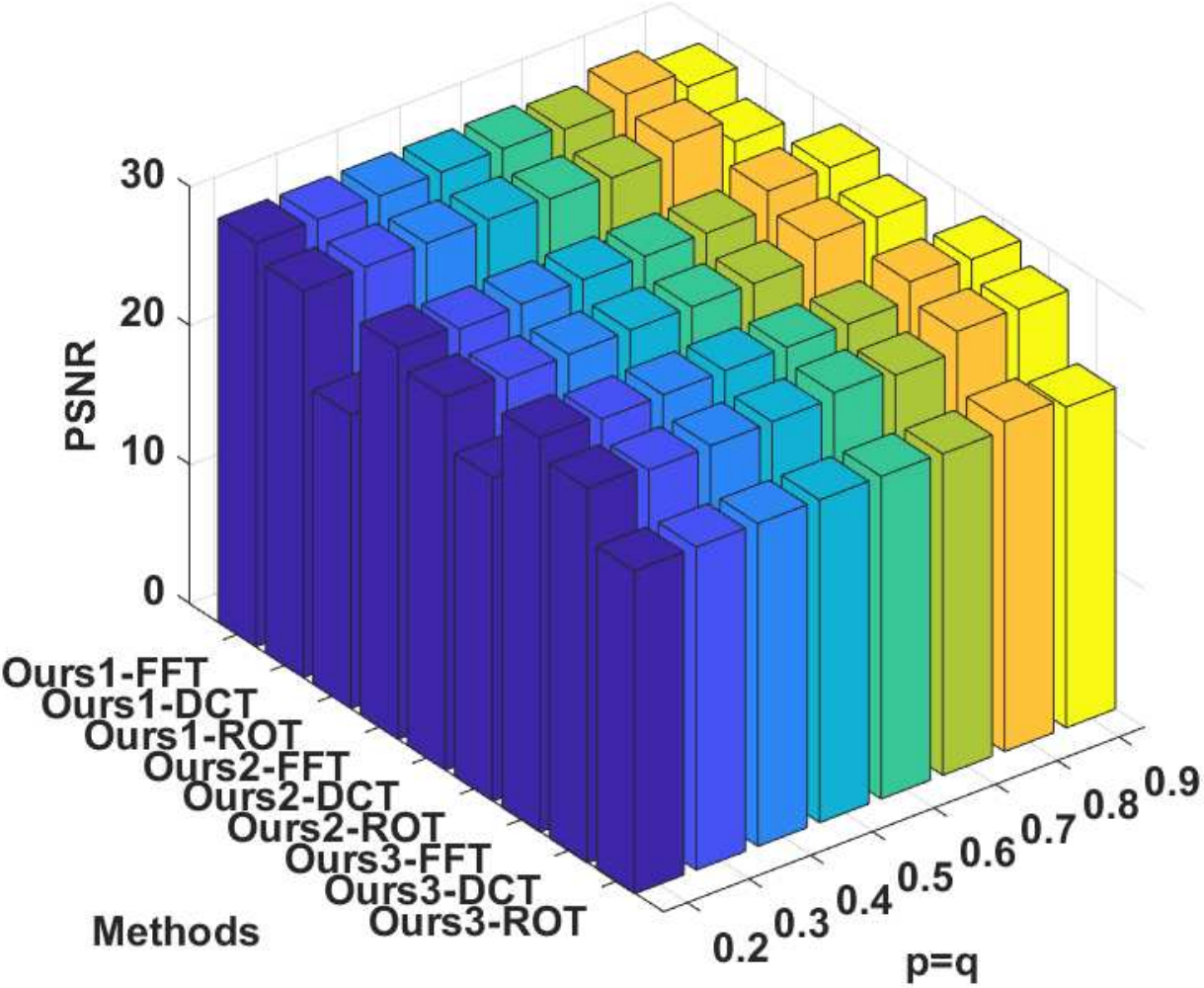} &
\includegraphics[width=1.15in, height=0.9in]{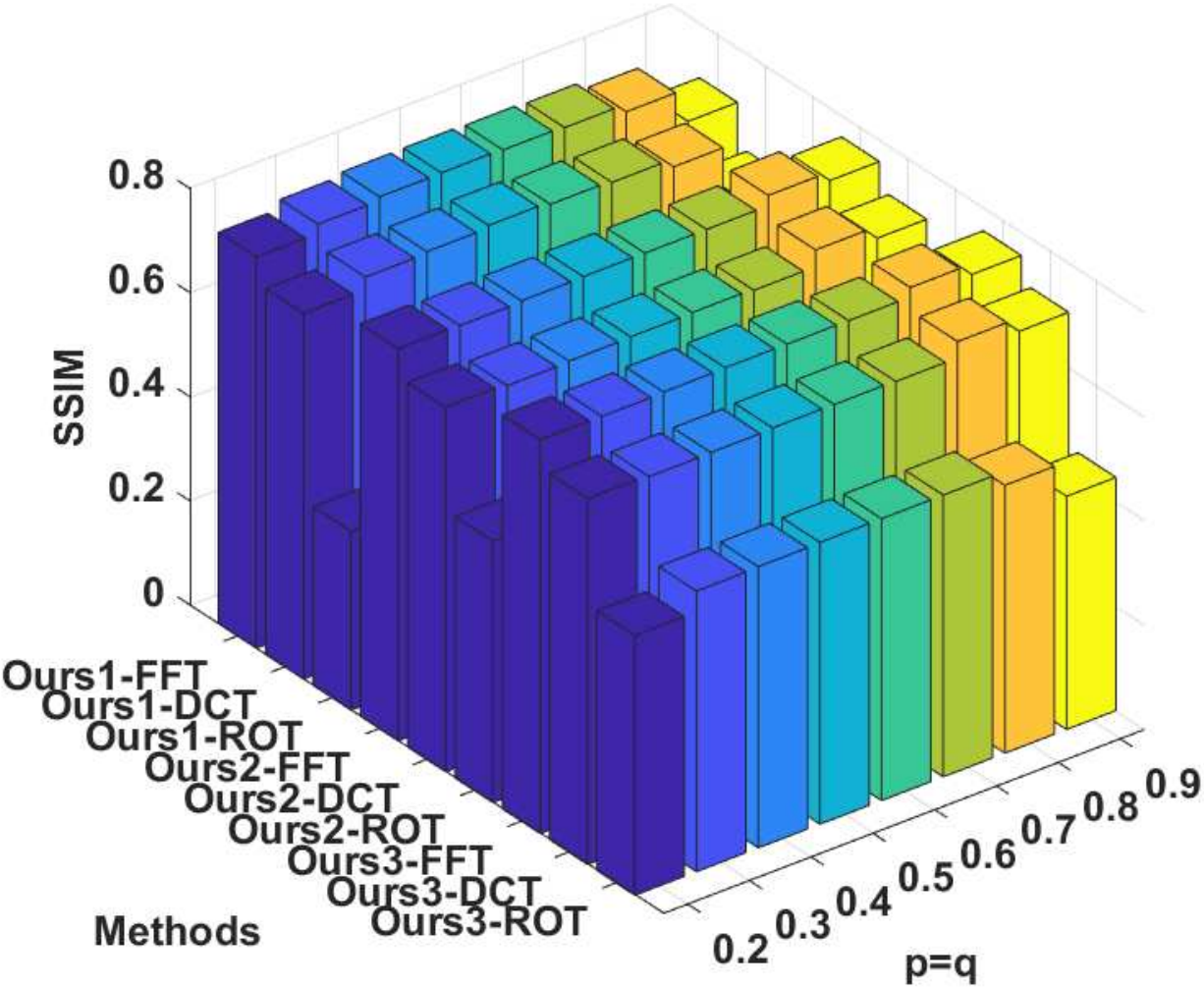}&
\includegraphics[width=1.15in, height=0.9in]{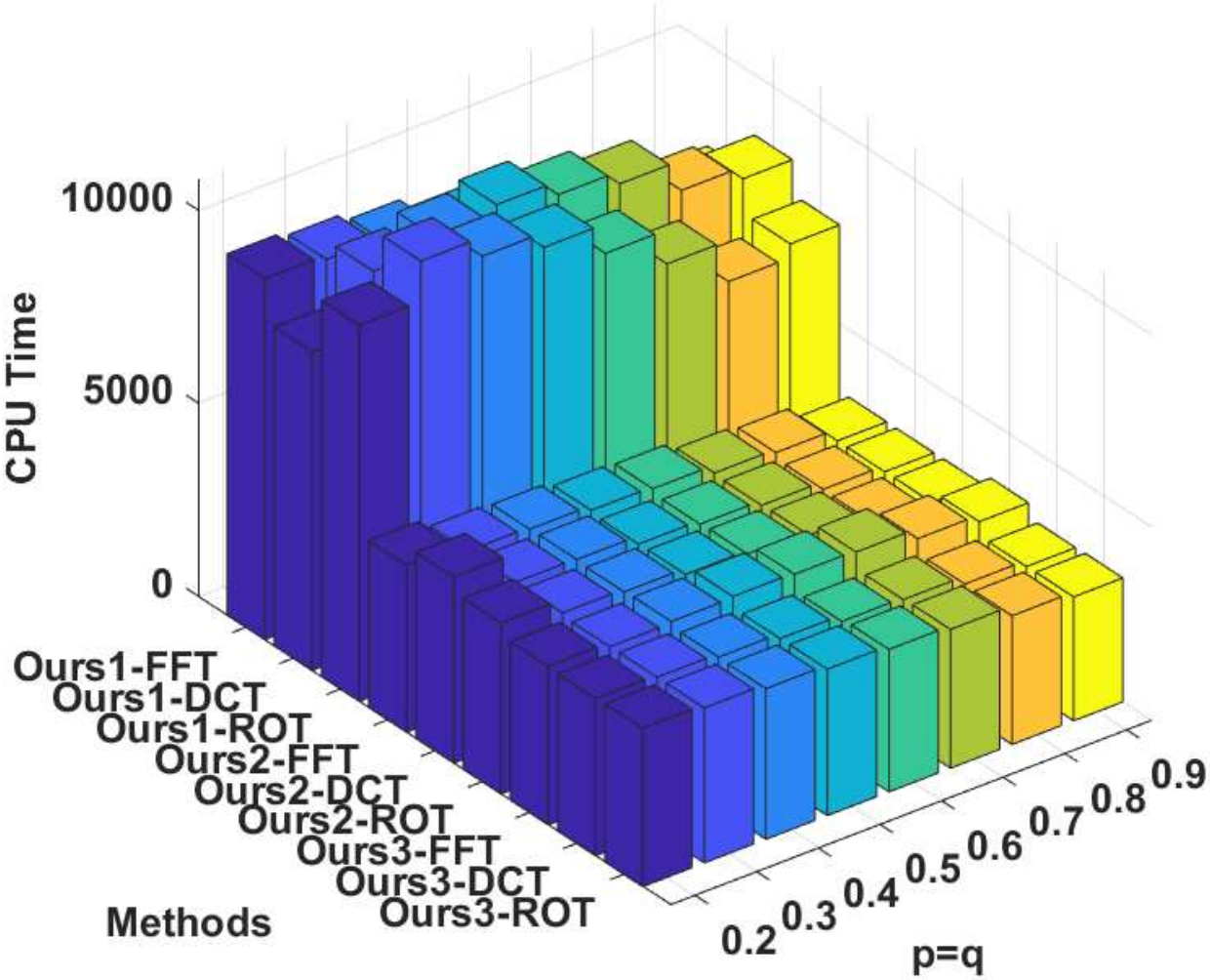}
 \\
\includegraphics[width=1.15in, height=0.9in]{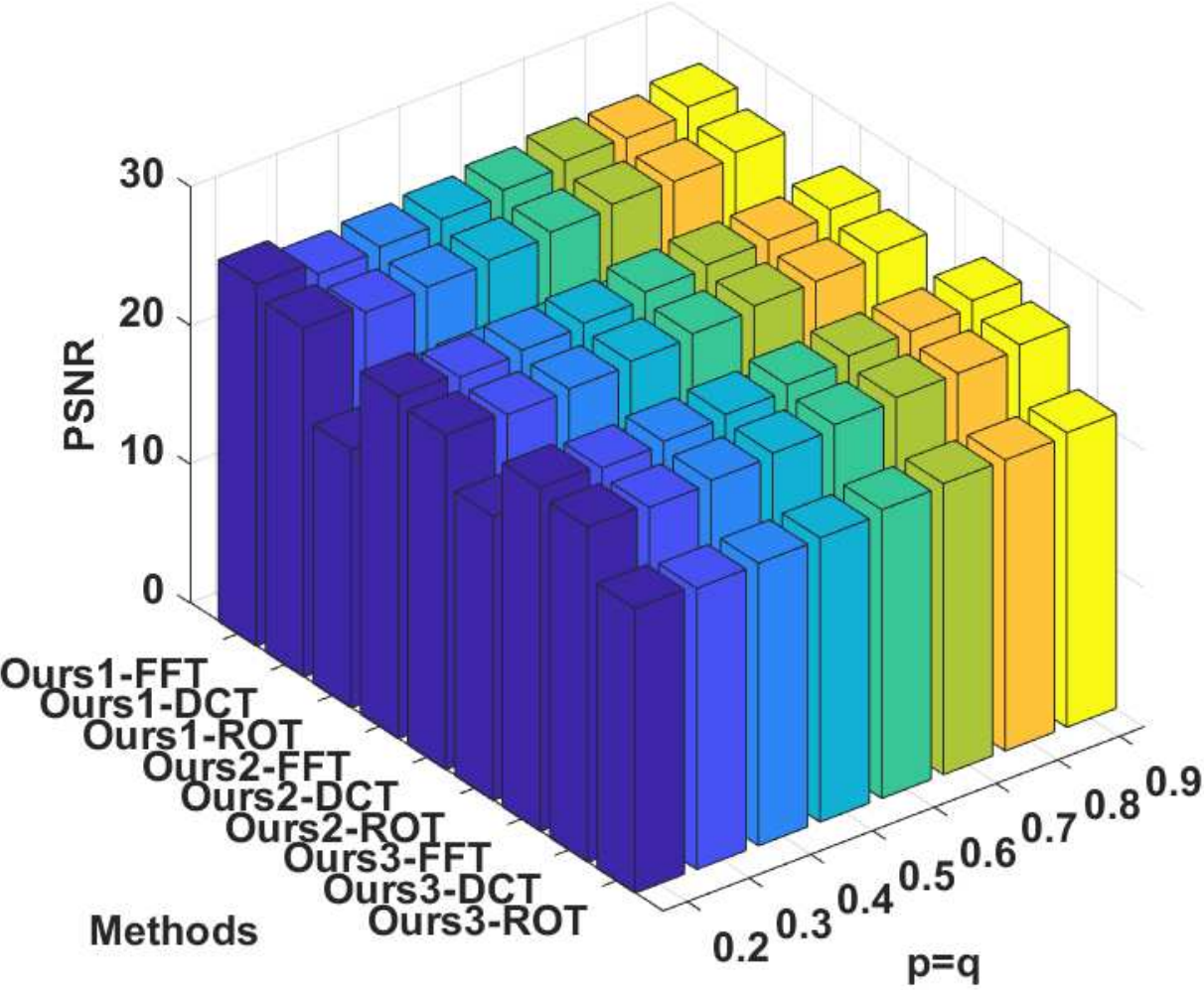} &
\includegraphics[width=1.15in, height=0.9in]{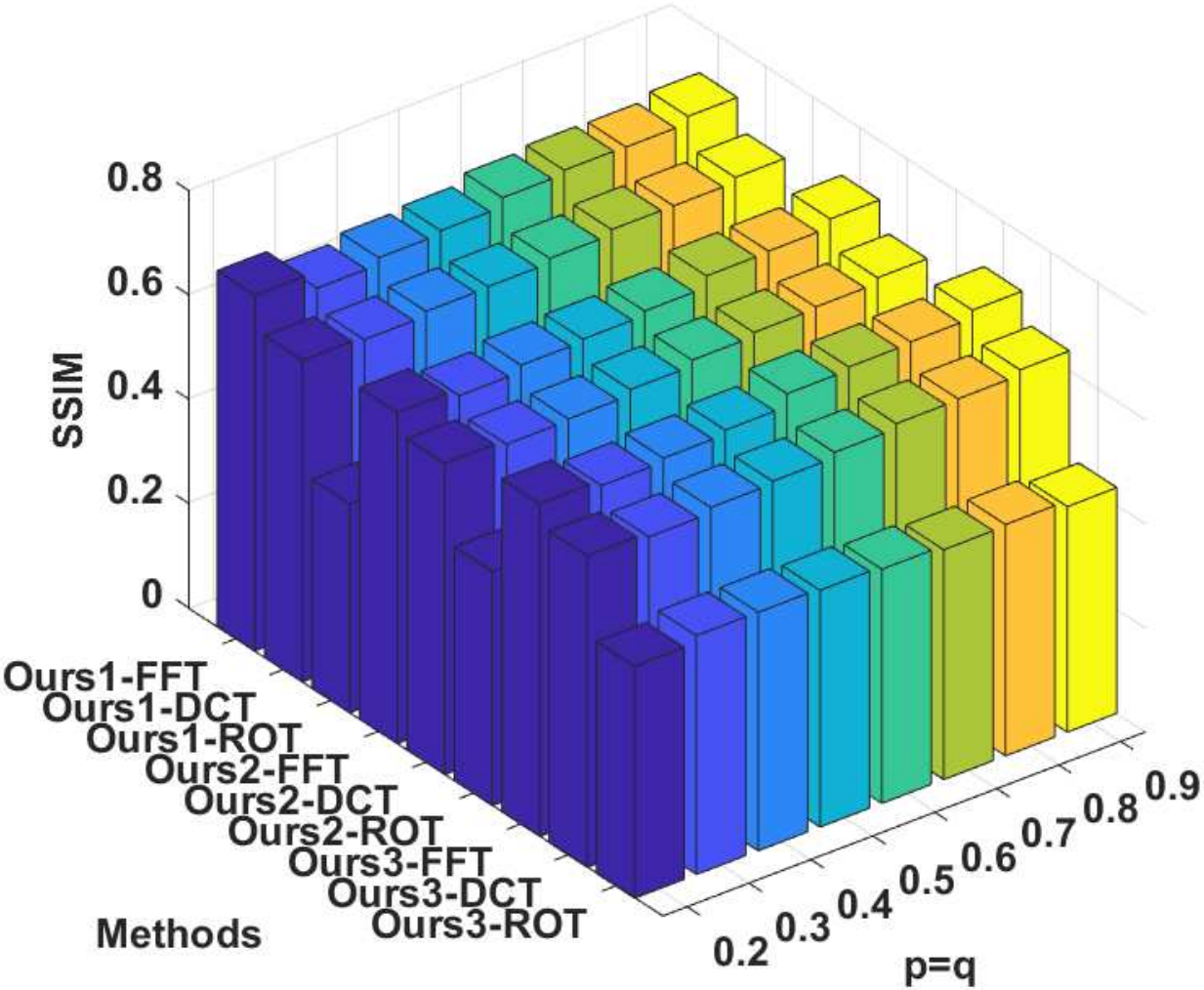}&
\includegraphics[width=1.15in, height=0.9in]{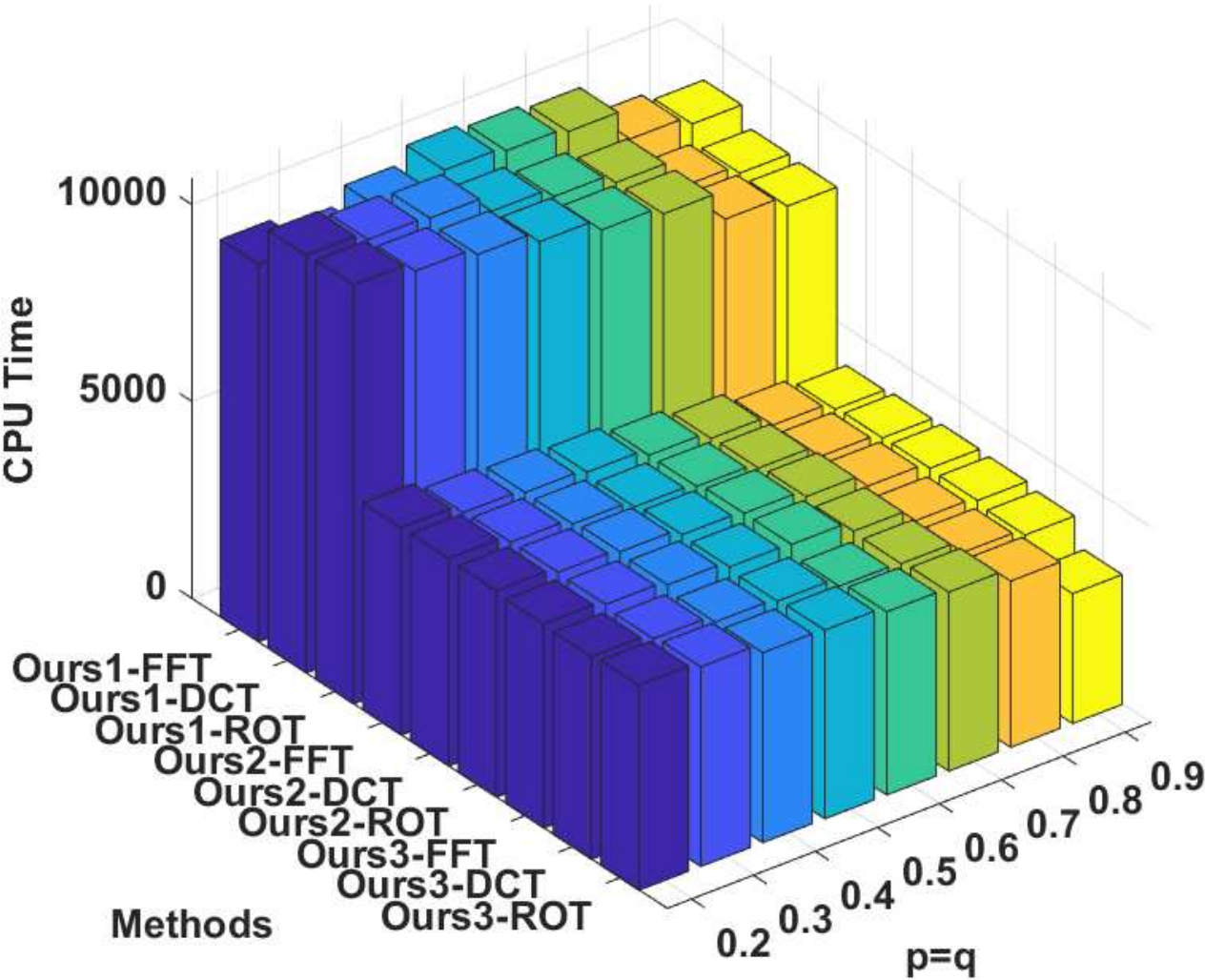}
\end{tabular}
\caption{
The influence of adjustable  parameters $(p,q)$, 
and
 changing 
linear transforms $\mathfrak{L}$
upon CVs restoration.
%
Top row: $sr=0.2, \tau=0.3$,
Bottom row: $sr=0.2, \tau=0.5$.
%
}
\label{disscu1-cv}
\vspace{-0.3cm}
\end{figure}

\begin{figure}[!htbp]
\centering
\renewcommand{\arraystretch}{0.0}
\setlength\tabcolsep{0.0pt}
\begin{tabular}
{ccc}
\centering
\includegraphics[width=1.15in, height=0.9in]{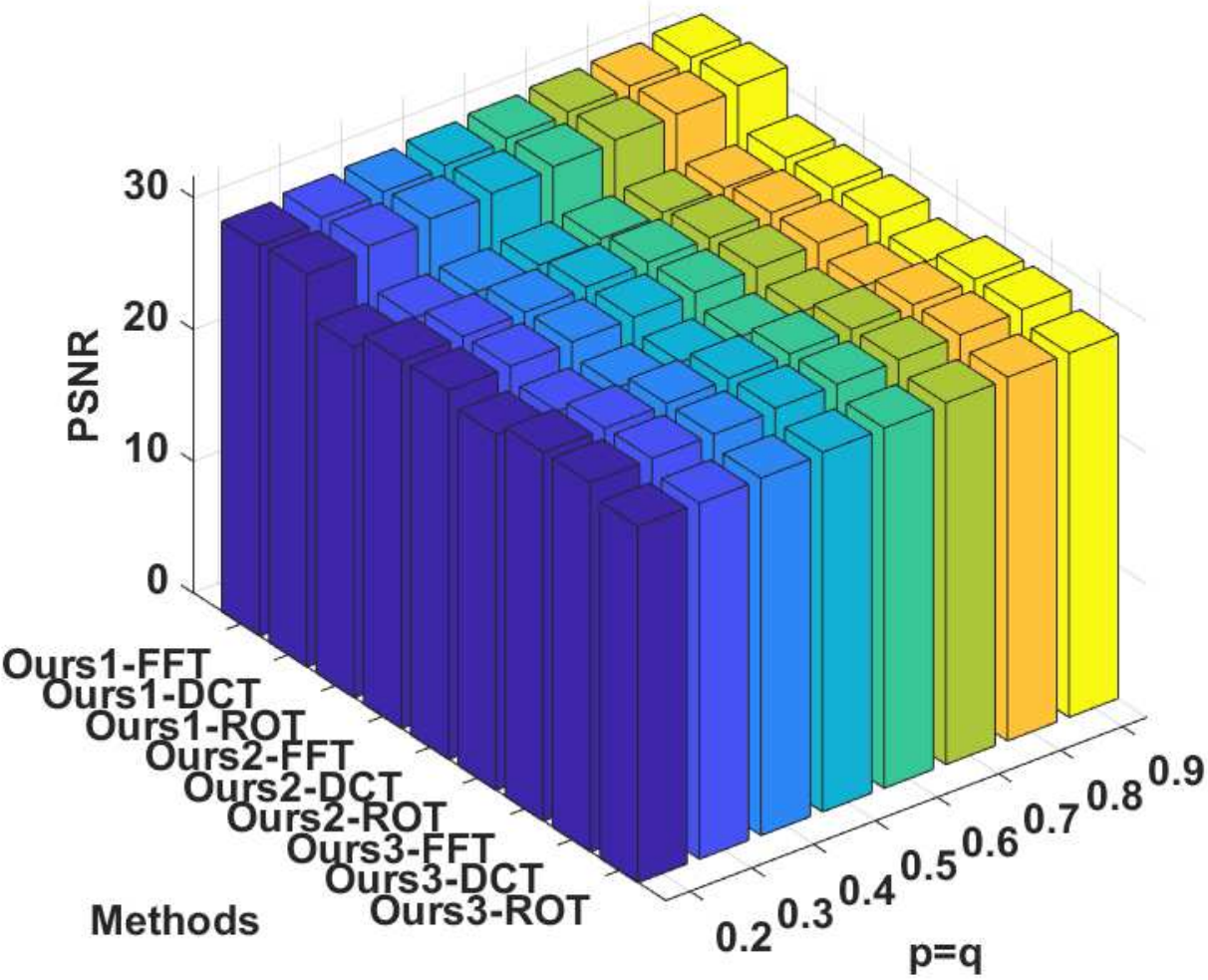} &
\includegraphics[width=1.15in, height=0.9in]{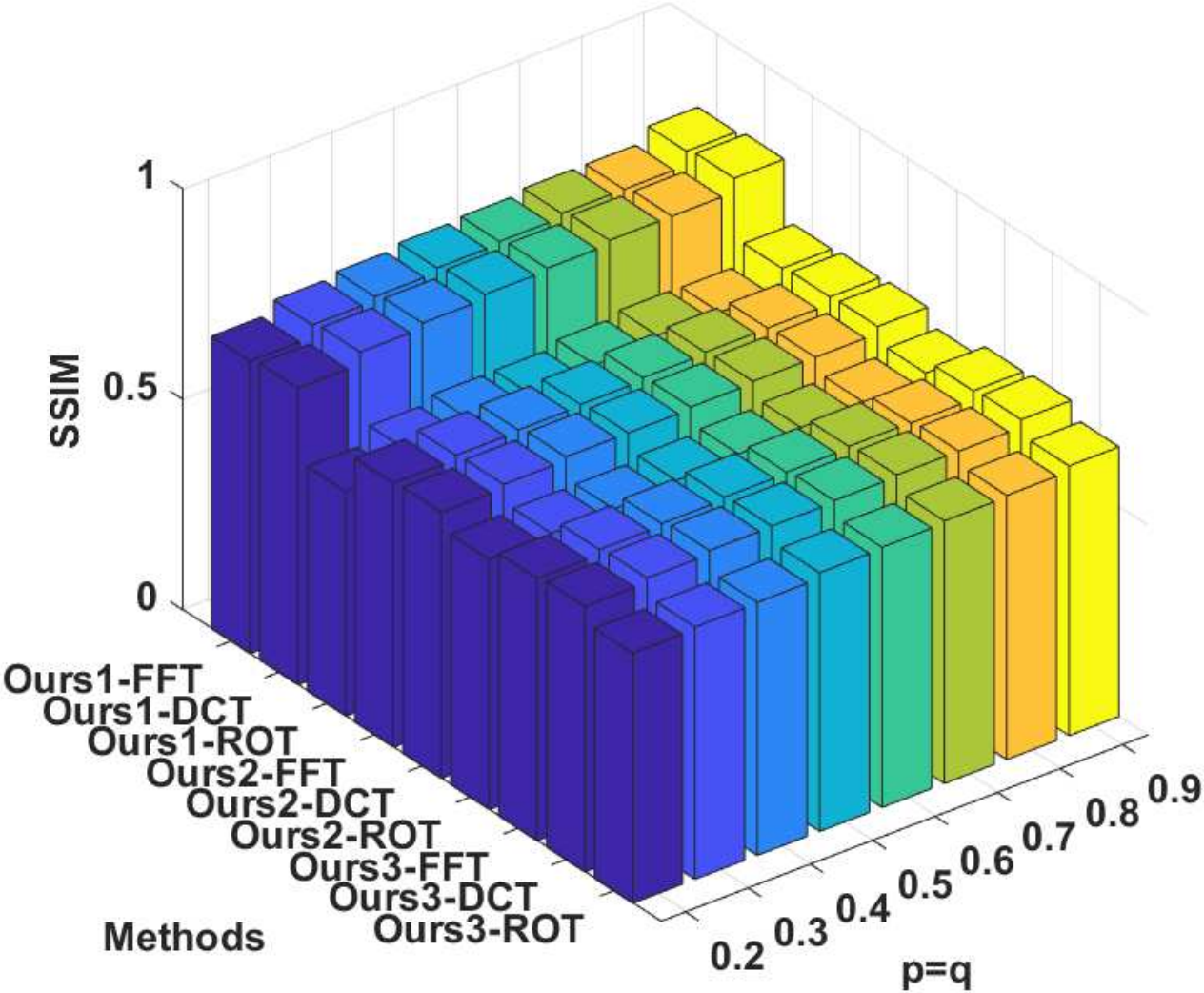}&
\includegraphics[width=1.15in, height=0.9in]{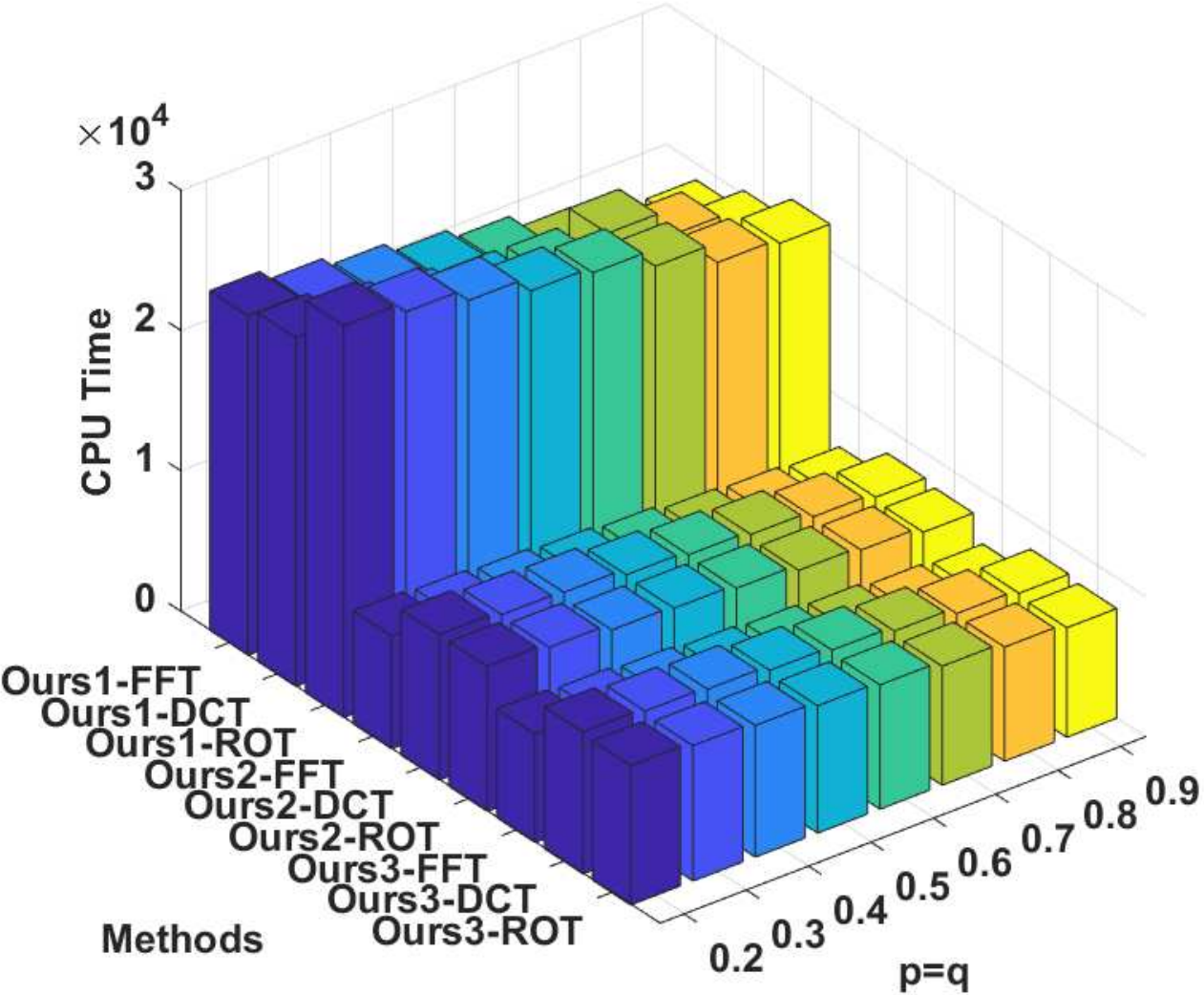} \\
\includegraphics[width=1.15in, height=0.9in]{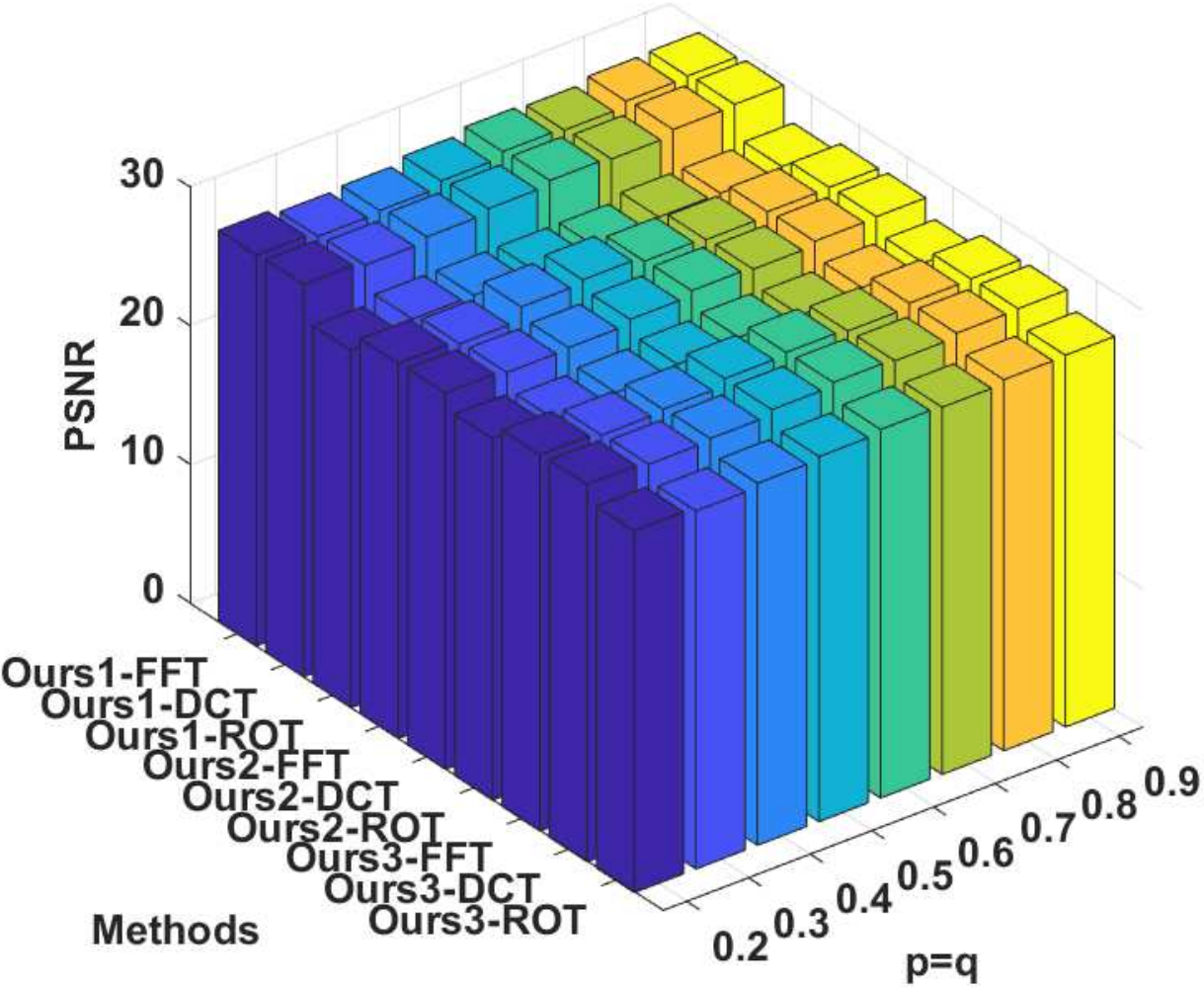} &
\includegraphics[width=1.15in, height=0.9in]{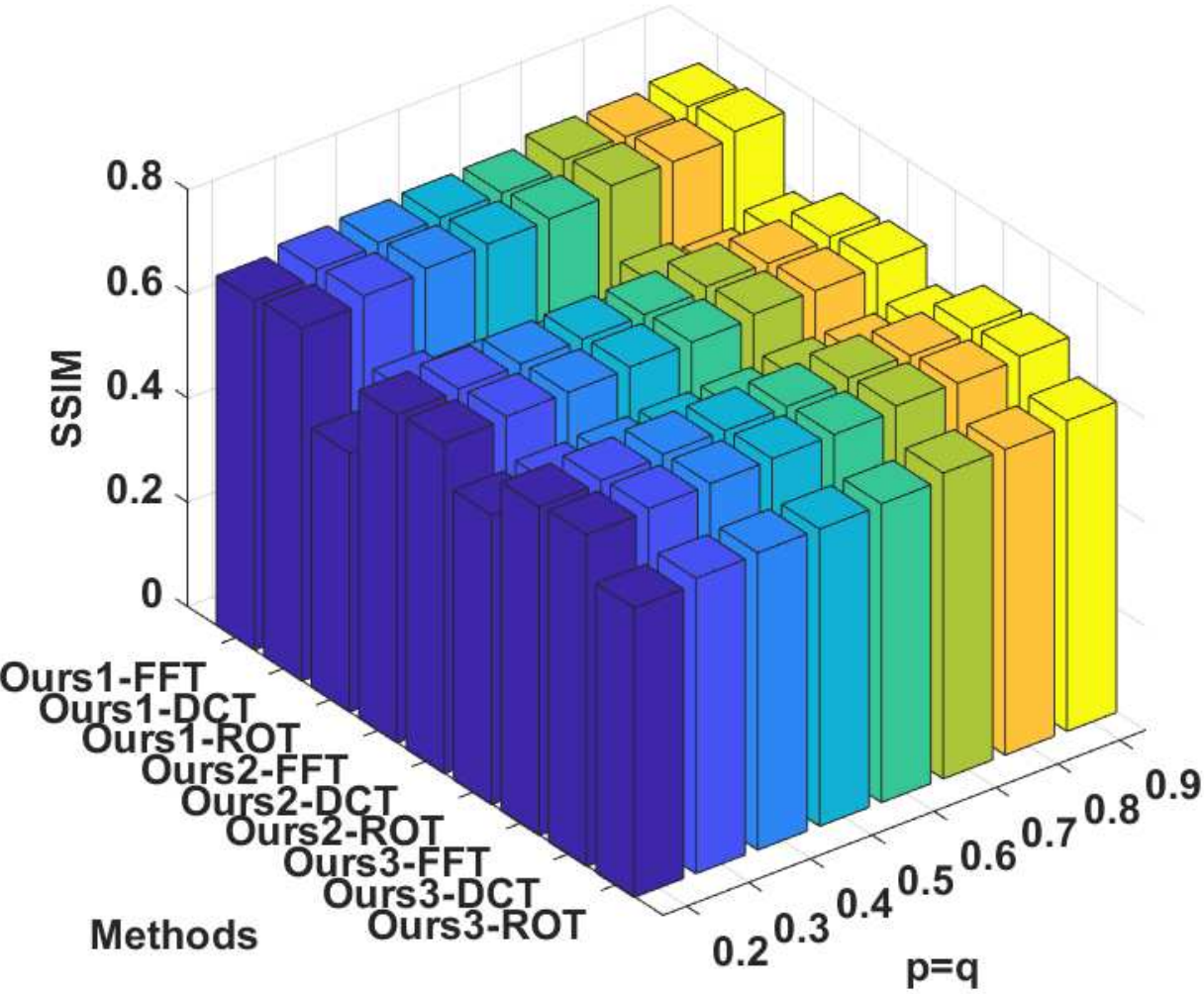}&
\includegraphics[width=1.15in, height=0.9in]{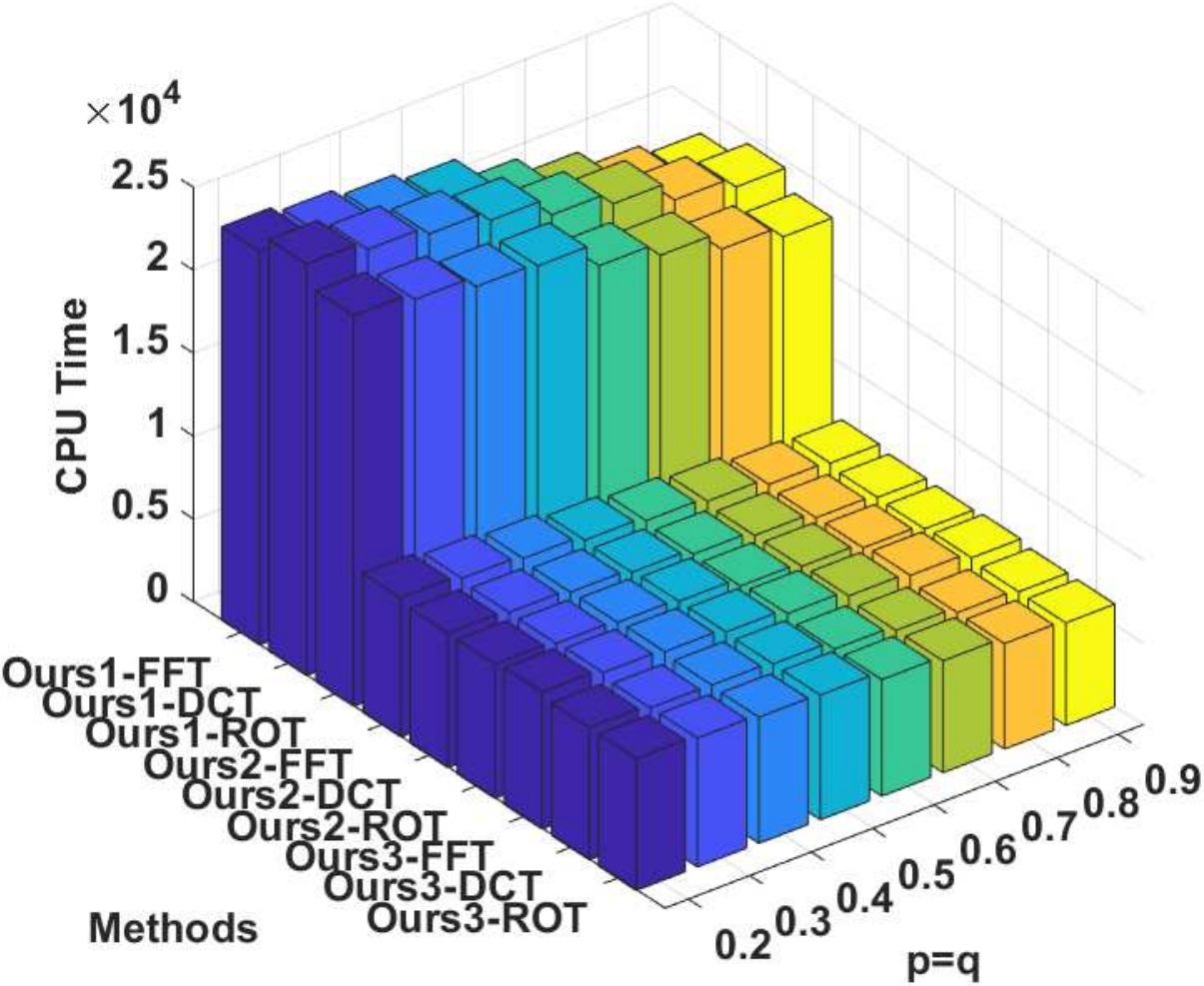}
\end{tabular}
\caption{
The influence of adjustable  parameters $(p,q)$, 
and
 changing 
linear transforms $\mathfrak{L}$
upon MRSIs inpainting.
Top row: $sr=0.4, \tau=0.1$,
Bottom row: $sr=0.4, \tau=0.3$.
%
}
\label{disscu1-rs}
\vspace{-0.3cm}
\end{figure}



\subsubsection{\textbf{Discussion}}
%
In the previous real applications,
the proposed methods only employ 
 the linear transform: FFT,
and only 
set the adjustable parameters $(p,q)$  to be $(0.9,0.9)$.
In this subsection, we  additionally 
 utilize 
other
linear transforms (e.g.,  DCT, ROT)  
and 
adjustable
parameters $(p,q)$
to perform   related    
experiments                         
on the proposed algorithm and 
its two accelerated versions.
Our goal is to investigate 
the influence of adjustable  parameters $(p,q)$, 
and
 changing invertible
linear transforms $\mathfrak{L}$
upon restoration results of various 
 tensor data
with  different noise levels and observed ratios 
in both 
deterministic and randomized approximation 
patterns.
%
In our experiments, 
the  values  $p, q (p=q)$ are set from $0.2$ to $0.9$ with an interval of $0.1$.
%
%
For brevity,  
the proposed  algorithm:
``\textbf{HWTSN+w$\ell_q$}",
and its accelerated versions:  ``\textbf{HWTSN+w$\ell_q$(UR)}" and
``\textbf{HWTSN+w$\ell_q$(BR)}"
are  abbreviated as Ours1, Ours2, and Ours3, respectively.

The
corresponding
 experimental results of the above  investigation are shown in Figure \ref{disscu1-lfi},\ref{disscu1-cv},\ref{disscu1-rs},
from which
some instructive conclusions  and guidelines
can be drawn.
\textbf{(I)}
 The PSNR or   SSIM value obtained by ROT 
 is always worse than
 that
 achieved by FFT and DCT 
  under the same parameters $p$ and
 $q$.
This implies that ROT may not be a good choice for 
 the restoration of 
 real-world high-order  
 tensors.
\textbf{(II)}
 For the recovery of  three types of tensors, 
  with the increase of adjustable parameters $p$ and $q$, the  PSNR and SSIM values obtained by various methods gradually increase 
  whereas
    the corresponding running time gradually decreases in most cases.
    This suggests that 
   selecting relatively large $p$ and  $q$
     may yield better recovery performance 
    for different 
    types of high-order 
     tensors. 
\textbf{(III)}
Just as we expected,
in comparison with
the deterministic version (i.e., Ours1),
the 
randomized 
methods (i.e., Ours2 and Ours3)
greatly boost the computational efficiency at the premise of compromising a little PSNR and SSIM 
for various inpainting tasks.
\textbf{(IV)}
 In our  randomized versions,
there is remarkably  little difference between Ours2 and Ours3
in CPU running time for different recovery tasks.
This indicates that it is very likely that only
 for very large-scale 
tensors, the   computational cost 
of Ours3 (i.e., the version fusing 
blocked randomized scheme) is significantly lower  
than that of Ours2.


\section{\textbf{Conclusions and  Future Work}}\label{conclusion}
%
In this article,
we first develop two
efficient
low-rank 
tensor approximation methods 
%
fusing
 random
 projection
schemes,  
  based on which
we further 
study
%
 the
effective model and algorithm
for RHTC. 
%
The model construction, algorithm design and theoretical analysis are all based on
the algebraic framework of high-order T-SVD.
Extensive experiments  on both synthetic and real-world tensor data
have verified the effectiveness and superiority of the proposed approximation and completion 
approaches.
This work will
lay the foundation for 
many
tensor-based 
data analysis
tasks
such as 
 high-order tensor 
 clustering,
regression,  classification, etc.
%

 In the future,
 %
  under the  
  T-SVD framework,
 we first intend to  explore the
  effective
 randomized  algorithm
  for the fixed-precision low-rank tensor approximation
 by devising novel  mode-wise projection  strategy
 that differs from the literature \cite{haselby2023modewise}. 
On this basis, we further investigate 
the
high-order tensor recovery 
from the perspective of model, algorithm and theory.
%
Secondly,
we will exploit the
fast high-order tensor 
clustering, regression and classification approaches
in virtue of 
some popular
randomized  sketching
techniques (e.g., 
random projection/sampling
and count-sketch).
Finally,
we plan to
extend the above
batch-based randomized methods 
to the online versions,
which can deal with 
large-scale streaming tensors incrementally
in online mode, and even 
with dynamically changing tensors. 




\ifCLASSOPTIONcaptionsoff
  \newpage
\fi

\ifCLASSOPTIONcaptionsoff
  \newpage
\fi


\bibliographystyle{IEEEtran}
\bibliography{hyperspectral,various_application,orderp_tensor}

\end{document}